\documentclass[12pt,twoside,openright]{ut_thesis}  
\pdfoutput=1

\newcommand{\thesistitletext}{Predicting Performance of a Face Recognition System Based on Image Quality}
\newcommand{\authornametext}{Abhishek Dutta}

\usepackage{xspace} 
\usepackage{url} 
\usepackage{hyperref}
\usepackage{color}           
\usepackage{array}           
\usepackage{multirow}        
\usepackage{rotating}        
\usepackage{array} 
\usepackage{amssymb}
\usepackage{amsmath}
\usepackage{subfig}
\usepackage{graphicx}
\usepackage{paralist} 
\usepackage{pdflscape} 
\usepackage{multirow} 
\usepackage{fancyhdr} 
\usepackage{time}
\usepackage{colortbl} 
\usepackage{enumitem} 
\usepackage{cleveref} 
\usepackage{lipsum} 

\hypersetup{
    pdffitwindow=false,     
    pdfstartview={FitH},    
    pdftitle={PhD Thesis - Abhishek Dutta},    
    pdfauthor={Abhishek Dutta},     
    pdfsubject={PhD thesis publicly defended at the University of Twente (Netherlands) on April 24, 2015 at 12.45.},   
    pdfnewwindow=true,      
    colorlinks=true,       
    linkcolor=black,          
    citecolor=black,        
    filecolor=black,      
    urlcolor=black           
}

\renewcommand{\tablename}{Table~} 
\renewcommand{\figurename}{Fig.~} 
 
\newcommand{\eg}{e.\,g.\xspace}
\newcommand{\ie}{i.\,e.\xspace}
\newcommand{\etc}{etc~}
\renewcommand{\figurename}{Figure~}
\newcommand{\argmin}{\operatornamewithlimits{arg\ min}}

\title{\thesistitletext}   
\author{\authornametext}   

\newcommand{\ResearchQuestionOne}{Which type of performance predictor features, score-based or quality-based, are suitable for predicting the performance of a face recognition system?}
\newcommand{\ResearchQuestionTwo}{Given a set of measurable performance predictor features, how can we predict the performance of a face recognition system?}
\newcommand{\ResearchQuestionThree}{What is the impact of automatic eye detection error on the performance of a face recognition system?}
\newcommand{\ResearchQuestionFour}{In forensic cases involving face recognition, how can we adapt the pose of probe or reference image such that pose variation has minimal impact on the performance of a face recognition system?}

\begin{document}


\setcounter{secnumdepth}{3}
\setcounter{tocdepth}{3}

\maketitle                  
\newpage
\thispagestyle{empty}

\noindent
Composition of the Graduation Committee:

\begin{tabular}{l l l}
  Prof.Dr.Ir. & R.N.J. Veldhuis & University of Twente, Netherlands \\
  Dr.Ir. & L.J. Spreeuwers & University of Twente, Netherlands \\
  Prof.Dr. & D. Meuwly & \begin{tabular}{@{}l@{}}University of Twente, Netherlands \\ Netherlands Forensic Institute, Netherlands\end{tabular} \\
  Prof.Dr.Ir. & C.H. Slump & University of Twente, Netherlands \\  
  Prof.Dr. & Christoph Busch & Gj\o vik University College, Norway \\
  Dr. & Arun Ross & Michigan State University, USA
\end{tabular}

\vspace{2cm}

\noindent
\begin{tabular}{l p{11cm}}
\vspace{0.6cm}
 \multirow{1}{*}{\includegraphics[width=0.15\linewidth]{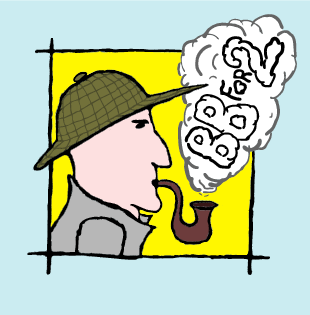}} & The doctoral research of A. Dutta was funded by the BBfor2 project which in turn was funded by the European~Commission as a Marie-Curie ITN-project (FP7-PEOPLE-ITN-2008) under Grant Agreement number 238803. \\

 \multirow{4}{*}{\includegraphics[width=0.2\linewidth]{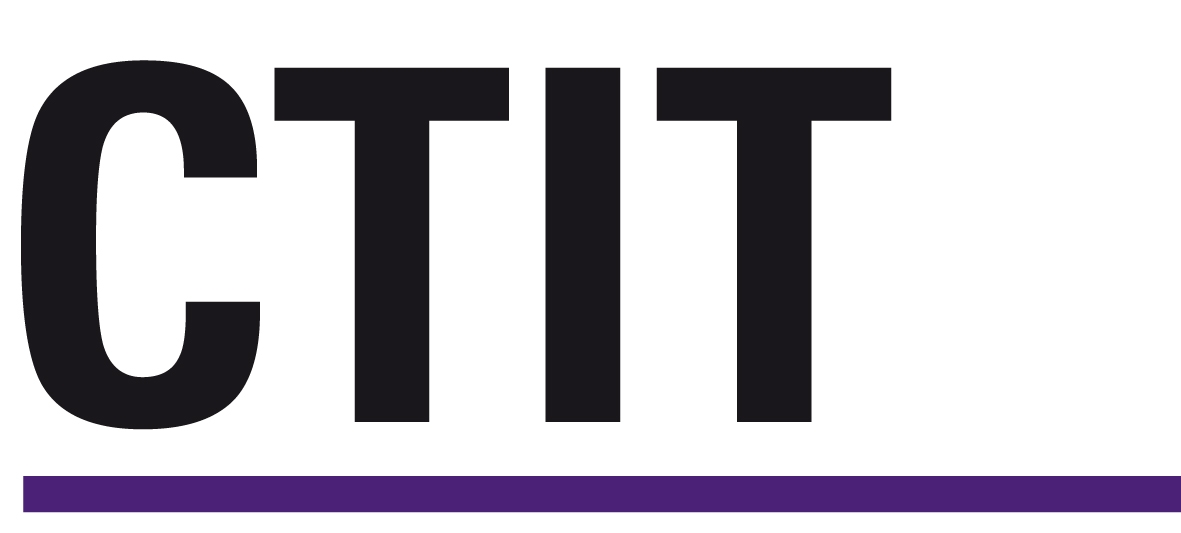}} &  CTIT Ph.D. Thesis Series No. 15-353 \\
 & Centre for Telematics and Information Technology \\
 & P.O. Box 217, 7500 AE \\
 & Enschede, The Netherlands\\
\end{tabular}

\vspace{2cm}

\begin{tabular}{l l}
 ISBN & 978-90-365-3872-5 \\
 ISSN & 1381-3617 \\
 DOI & \url{http://dx.doi.org/10.3990/1.9789036538725} \\
 Code & \url{http://abhishekdutta.org/phd-research/}
\end{tabular}

\vspace{2cm}
{\footnotesize
\noindent Cover: The colourful patches correspond to the visualization of Quality-Performance (QR) space of face recognition systems. The two cartoon characters are inspired from the fictional cardboard box robot character called the Danbo from Yotsuba$\&!$ manga.}

\vspace{2cm}

\noindent Copyright~\textcopyright~2015 Abhishek Dutta

\noindent 
\begin{textnormal}
All rights reserved. No part of this book may be reproduced or transmitted, in any form or by any means, electronic or mechanical, including photocopying, microfilming, and recording, or by any information storage or retrieval system, without the prior written permission of the author.
\end{textnormal}
\newpage
\thispagestyle{empty}

\begin{center}
\vfill
 
  \begin{large} \bfseries
\MakeUppercase{\thesistitletext}\par
  \end{large}
  
\vspace{4cm}

  \begin{large} \bfseries
\MakeUppercase{Dissertation}\par
  \end{large}

\vspace{4cm}
\noindent\mbox{%
\parbox{11.35cm}{\centering {\normalfont to obtain\\the degree of doctor at the University of Twente,\\on the authority of the rector magnificus,\\prof.dr. H. Brinksma,\\on account of the decision of the graduation committee,\\to be publicly defended \\on Friday the $24^{\textnormal{th}}$ of April 2015 at 12.45}}
}

\vspace{2cm}
{by}
\vspace{2cm}

\begin{large}
Abhishek Dutta\par
\end{large}

\vspace{0.2cm}

\begin{textnormal}
born on the $5^{\textnormal{th}}$ of September 1985 \\ in Janakpur, Nepal
\end{textnormal}

\vfill

\end{center}
\clearpage

\newpage
\thispagestyle{empty}

\noindent This dissertation has been approved by:

\vspace{0.15cm}
\begin{tabular}{ l l }
  Promotor & Prof.Dr.Ir. R.N.J. Veldhuis \\
  Co-promotor & Dr.Ir. L.J. Spreeuwers
\end{tabular}

\clearpage

\newpage
\thispagestyle{empty}

\begin{flushright}
to Baba ...
\end{flushright}

\clearpage

\newpage
\thispagestyle{empty}

\begin{figure}[p]
\centering
\includegraphics[width=0.9\linewidth]{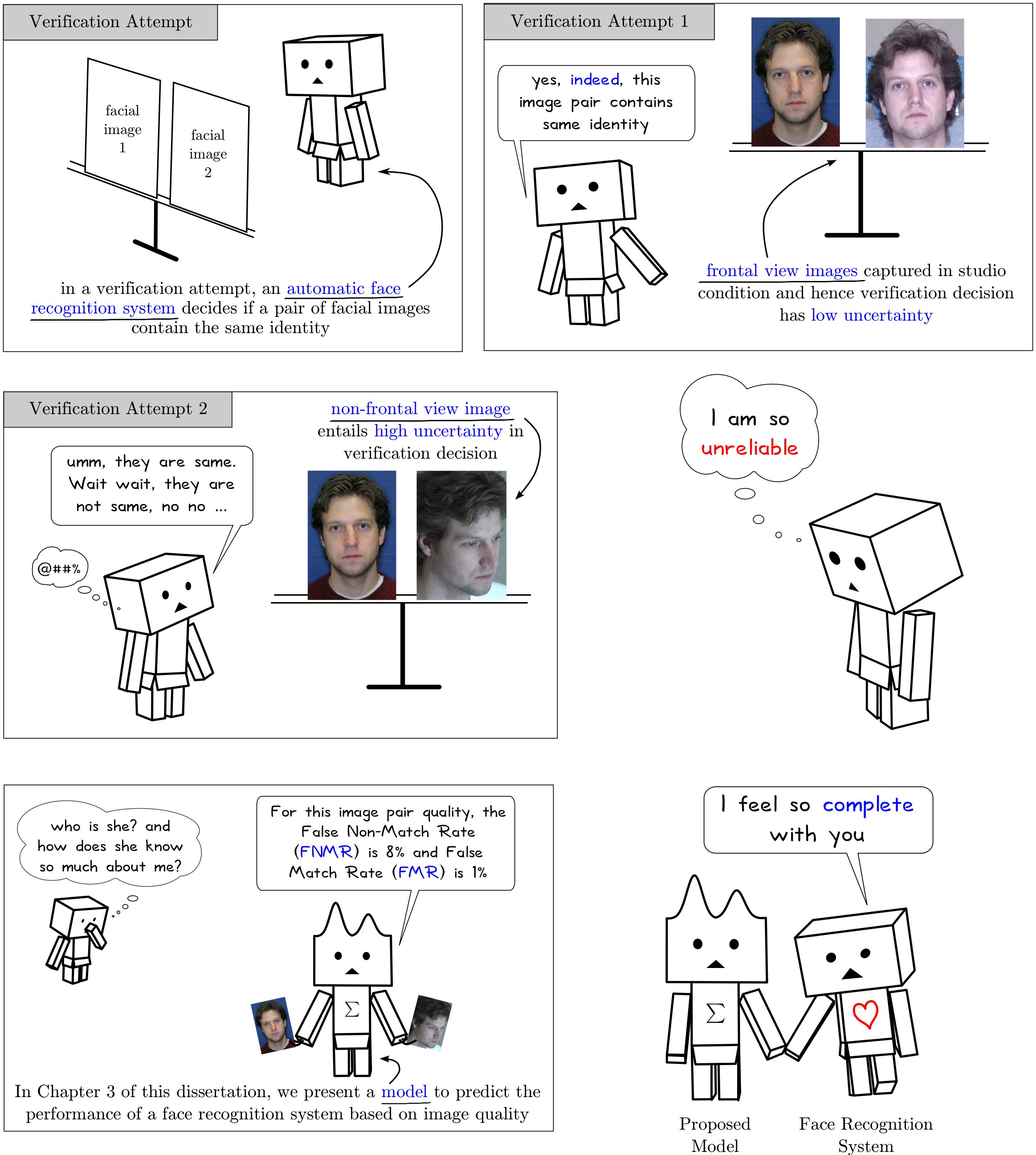}
\end{figure}

\clearpage


\begin{romanpages}          
\tableofcontents            
\listoffigures              

\chapter*{Abstract}

In this dissertation, we focus on several aspects of models that aim to predict performance of a face recognition system.
Performance prediction models are commonly based on the following two types of performance predictor features: a) image quality features; and b) features derived solely from similarity scores.
We first investigate the merit of these two types of performance predictor features.
The evidence from our experiments suggests that the features derived solely from similarity scores are unstable under image quality variations.
On the other hand, image quality features have a proven record of being a reliable predictor of face recognition performance.
Therefore, the performance prediction model proposed in this dissertation is based only on image quality features.
We present a generative model to capture the relation between image quality features $\mathbf{q}$ (\eg pose, illumination, \etc) and face recognition performance $\mathbf{r}$ (\eg FMR and FNMR at operating point).
Since the model is based only on image quality features, the face recognition performance can be predicted even before the actual recognition has taken place thereby facilitating many preemptive action.
A practical limitation of such a data driven generative model is the limited nature of training data set.
To address this limitation, we have developed a Bayesian approach to model the nature of FNMR and FMR distribution based on the number of match and non-match scores in small regions of the quality space.
Random samples drawn from these models provide the initial data essential for training the generative model $P(\mathbf{q},\mathbf{r})$.
Experiment results based on six face recognition systems operating on three independent data sets show that the proposed performance prediction model can accurately predict face recognition performance using an accurate and unbiased Image Quality Assessor (IQA).
Furthermore, variability in the \textit{unaccounted quality space} -- the image quality features not considered by the IQA -- is the major factor causing inaccuracies in predicted performance.

Many automatic face recognition systems use automatically detected eye coordinates for facial image registration.
We investigate the influence of automatic eye detection error on the performance of face recognition systems. 
We simulate the error in automatic eye detection by performing facial image registration based on perturbed manually
annotated eye coordinates. 
Since the image quality of probe images are fixed to frontal pose and ambient illumination, the performance variations are solely due to the impact of facial image registration error on face recognition performance.
This study helps us understand how image quality variations can amplify its influence on recognition performance by having dual impact on both facial image registration and facial feature extraction/comparison stages of a face recognition system.
Our study has shown that, for a face recognition system sensitive to errors in facial image registration, the performance predictor feature set should include some features that can predict the accuracy of automatic eye detector used in the face recognition system.
This is essential to accurately model and predict the performance variations in a practical face recognition system.
So far, existing work has only focused on using features that predict the performance of face recognition algorithms.
Our work has laid the foundation for future work in this direction.

A forensic case involving face recognition commonly contains a surveillance view trace (usually a frame from CCTV footage) and a frontal suspect reference set containing facial images of suspects narrowed down by police and forensic investigation.
If the forensic investigator chooses to use an automatic face recognition system for this task, there are two choices available: a model based approach or a view based approach.
In a model based approach, a frontal view probe image is synthesized based on a 3D model reconstructed from the surveillance view trace.
Most face recognition systems are fine tuned for optimal recognition performance for comparing frontal view images and therefore the model based approach, with synthesized frontal probe and frontal suspect reference images, ensures high recognition performance.
In a view based approach, the reference set is adapted such that it matches the pose of the surveillance view trace.
This approach ensures that a face recognition system always gets to compare facial images under similar pose -- not necessarily the frontal view.
We investigate if it is potentially more useful to apply a view based approach in forensic cases.
The evidence from our experiments suggests that the view based approach should be used if: a) it is possible to exactly match the pose, illumination condition and camera of the suspect reference set to that of the probe image (or, forensic trace acquired from CCTV footage); and b) one uses a face recognition system that is capable of comparing non-frontal view facial images with high accuracy.
A view based approach may not always be practical because matching pose and camera requires cooperative suspects and access to the same camera that captured the trace image.


\end{romanpages}            


\chapter{Introduction}
\label{dutta2014phdthesis_intro}

A face recognition system compares a pair of facial images and decides if the image pair contains same identity.
This comparison is based on facial features extracted from the image pair.
The outcome of this verification process is a verification decision which is either a match or non-match -- match corresponds to an image pair containing same identity while a non-match decision corresponds to different identity.
Such a verification system helps ascertain the validity of claimed identity and therefore has many applications in areas like access control, border security, \etc.

Practical face recognition systems make occasional mistake in their verification decision and therefore many recognition performance measures exist to quantify the error rate of a face recognition system.
Commonly, the verification performance of a face recognition system is measured in terms of False Match Rate - FMR (or False Accept Rate) and False Non-Match Rate - FNMR (or, False Reject Rate).
The FMR denotes the rate at which a verification system misses to correctly spot a non-match identity claim whereas FNMR measures the rate at which the verification system misses to correctly spot a match identity claim.
These two measures collectively define the uncertainty in decision about identity.
In practical applications of a verification system, we are not only interested in the verification decision -- match or non-match -- but also want to know the uncertainty (\eg FMR and FNMR) associated with this decision.

\begin{figure}[t]
\centering
\includegraphics[width=0.8\linewidth]{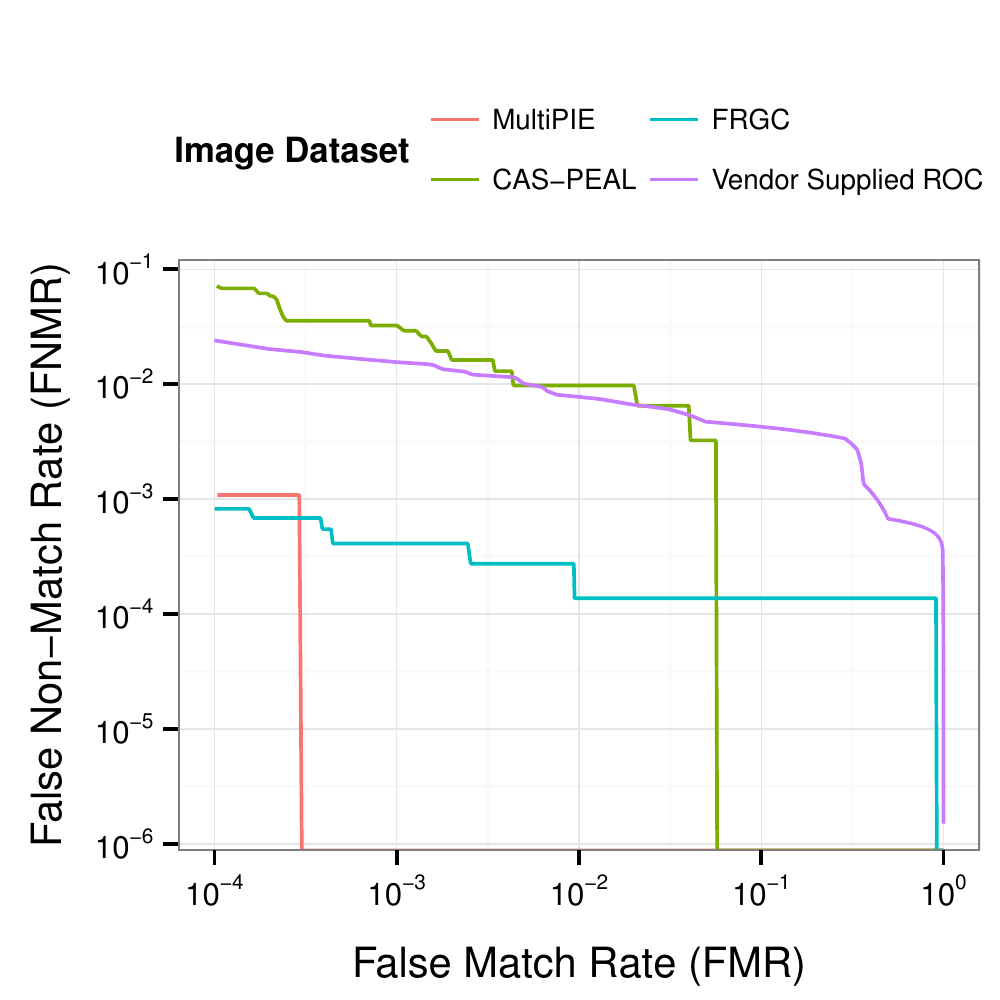}
\caption{Vendor supplied Receiver Operating Characteristic (ROC) and actual ROC curve of a COTS face recognition system~\cite{facevacs2010} operating on frontal pose, illumination, neutral expression subset of three independent data sets (sample facial images are shown in~\figurename~\ref{fig:prb_ref_img_illus.pdf}).}
\label{fig:intro_fv_vendor_roc_vs_true_roc.pdf}
\end{figure}

The vendors of commercial off-the-shelf (COTS) face recognition systems provide Receiver Operating Characteristics (ROC)\footnote{ROC curve is generated by plotting (FMR,FNMR) pairs at several operating point (\ie a decision threshold)} curve which characterizes the uncertainty in decision about identity at several operating points.
As shown in~\figurename\ref{fig:fv_vendor_roc_vs_true_roc.pdf}, the vendor supplied ROC for a COTS face recognition system~\cite{facevacs2010} differs significantly for frontal image subset of three independent but controlled facial image data sets~\cite{gross2008multipie,phillips2005overview,gao2008caspeal} that were captured using different devices and under different setup.
This suggests that practical applications of verification systems cannot rely on the vendor supplied ROC curve to quantify uncertainty in decision about identity on per verification instance basis.
Usually, the vendor supplied ROC represents recognition performance that the face recognition system is expected to deliver under ideal conditions.
In practice, the ideal conditions are rarely met and therefore the actual recognition performance varies as illustrated in~\figurename\ref{fig:intro_fv_vendor_roc_vs_true_roc.pdf}.
Therefore, practical applications of verification systems cannot rely on the vendor supplied ROC curve to quantify uncertainty in decision about identity on per verification instance basis.

The past decade has seen considerable effort being invested in building systems that can predict the uncertainty in verification decision of a face recognition system~\cite{beveridge2008focus,ozay2009improving,aggarwal2012predicting,scheirer2011meta,wang2007modeling} and biometrics systems in general~\cite{tabassi2005novel,wein2005using,shi2008modeling,zuo2010adaptive}.
Such systems have several applications:
\begin{itemize}
\item \textbf{Forensic Face Recognition} : In a forensic case involving face recognition, forensic investigators often have deal with a large volume of CCTV footage from a crime scene.
It is not possible to examine every CCTV frame and therefore investigators have to rank them based on their quality.
Such a ranking helps the forensic investigators focus their resources on a small number of CCTV frames with high evidential value.
A performance prediction system can be used to rank the CCTV frames based on the predicted verification performance of individual frames.

\item \textbf{Enrollment} : When capturing facial images for enrollment (\ie gallery or reference set), we have control over the static and dynamic properties of the subject or acquisition process~\cite{iso_iec_29794-5:2010}.
A performance prediction system can alert the operator whenever a ``poor'' quality facial image sneaks into the enrollment set thereby allowing the operator to take appropriate corrective action.

\item \textbf{Decision Threshold} : Verification decisions are made using a decision threshold score such that any similarity score above (or below) this threshold is assigned as a match (or non-match).
The value of this decision threshold defines the operating point of the face recognition system and is usually supplied by vendor to match the user requirement of certain minimum False Non-Match Rate -- FNMR (or, False Match Rate -- FMR).
With image quality variations, the true FNMR (or FMR) varies and therefore a performance prediction system can be used to dynamically adapt this decision threshold based on the image quality.

\item \textbf{Multi-algorithm Fusion} : The tolerance of face recognition algorithms towards image quality degradation varies.
For example, a face recognition algorithm may be able to maintain high level of performance even under non-frontal illumination while its performance may degrade rapidly for non-frontal pose.
Some other face recognition system may be able to maintain good performance level for small deviation in pose ($\pm 30^{\circ}$) while it may be highly sensitive to illumination variations.
Therefore, recognition results from multiple face recognition algorithms can be fused based on the performance prediction for each individual algorithm corresponding to same facial image.
Such fusion scheme often results in performance better than individual algorithms.
\end{itemize}

Due to a large number of potential application avenues, the research into systems that can predict performance of a face recognition system has received much greater attention in recent years.

Before continuing onto further discussion, we define two key terms used frequently in this dissertation. Throughout this dissertation, we use the term \textit{face recognition system} to refer to a complete biometric system that contains, in addition to other specific components, image preprocessing modules and a \textit{face recognition algorithm} which handles the core task of facial feature extraction and comparison.
Furthermore, we use the term \textit{image quality} to denote all the static or dynamic characteristics of the subject or acquisition process as described in~\cite{iso_iec_29794-5:2010}.

In this dissertation, we focus on several aspects of models that aim to predict performance of a face recognition system.
Chapter~\ref{dutta2015predicting_intro} addresses the following \textit{main research question}:
\begin{center}
\textit{\ResearchQuestionTwo}
\end{center}

Performance prediction models are commonly based on the two types of performance predictor features: 
\begin{inparaenum}[\itshape a\upshape)]
\item image quality features, and
\item features derived solely from similarity scores.
\end{inparaenum}
Image quality features have a proven record of being a predictor of face recognition performance~\cite{beveridge2010quantifying,phillips2013existence}.
For example, facial features can be extracted more accurately from images captured under studio conditions -- frontal pose and illumination, sharp, high resolution \etc.
This contributes to more certainty in the decision about identity and therefore results in better recognition performance.
However, when the studio conditions are not met, facial features may be occluded or obscured causing inaccuracies in the extracted facial features which results in more uncertainty in decision about identity.
Furthermore, features derived solely from similarity scores have also been widely used for predicting recognition performance~\cite{scheirer2011meta,wang2007modeling,klare2012face}.
To investigate the merit of these two types of performance predictor features we investigate, in Chapter~\ref{dutta2014phdthesis_pred-feat-intro}, the merit of these two types of features by addressing the following subordinate research question:
\begin{center}
\textit{\ResearchQuestionOne}
\end{center}

\begin{figure}[h]
\centering
\includegraphics[width=\linewidth]{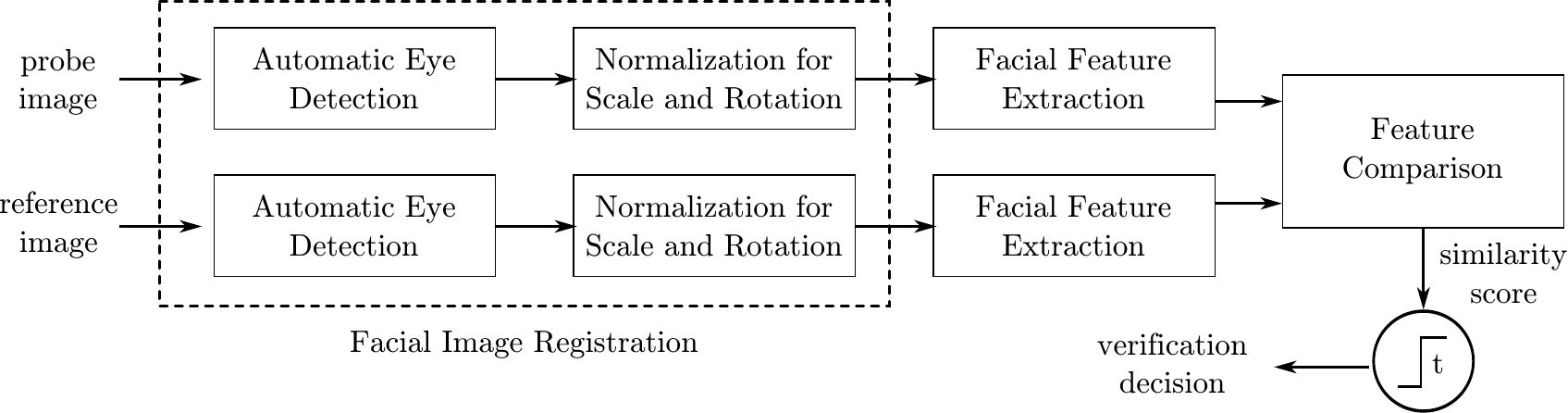}
\caption{Processing stages of a face recognition system.}
\label{fig:face_rec_sys_pipeline.pdf}
\end{figure}

Facial image registration is one of the critical preprocessing stages of most face recognition systems.
It ensures that facial features such as eyes, nose, lips, \etc consistently occupy similar spatial position in all the facial images provided to the facial feature extraction stage.
Face recognition systems commonly employ automatically detected eye coordinates for facial image registration: a preprocessing stage that corrects for variations scale and orientation of facial images as shown in~\figurename~\ref{fig:face_rec_sys_pipeline.pdf}.
Therefore, the performance of such face recognition systems depend not only on capabilities of the facial feature extraction and comparison stages -- core components of a face recognition algorithm -- but also on the accuracy of automatic eye detectors.
The accuracy of automatic eye detector is known to be influenced by image quality variations~\cite{dutta2014automatic}.
Furthermore, image quality variations also influence the accuracy of face recognition algorithms by either occluding or obscuring facial features present in a image~\cite{beveridge2008focus}.
If we wish to accurately model and predict the performance of such face recognition systems, we must take into account this dual impact of image quality variations:
\begin{inparaenum}[\itshape a\upshape)]
\item impact on the accuracy of automatic eye detection; and
\item impact on the accuracy of facial feature extraction and comparison.
\end{inparaenum}

In Chapter~\ref{dutta2015impact_intro}, we investigate the influence of automatic eye detection error on the performance of face recognition systems.
This chapter addresses the following subordinate research question:
\begin{center}
\textit{\ResearchQuestionThree}
\end{center}
This study helps us understand how image quality variations can amplify its influence on recognition performance by having dual impact on both facial image registration and facial feature extraction and comparison stages of a face recognition system.
Note that for all the experiments presented in Section~\ref{dutta2014automatic_intro} and Chapter~\ref{dutta2015predicting_intro}, we use manually annotated eye locations for facial image registration to ensure that performance variations are solely due to the impact of image quality variations on the feature extraction and comparison stages of a face recognition system.
In Chapter~\ref{dutta2015impact_intro}, we keep image quality fixed (frontal pose and ambient illumination) in all the images and therefore the performance variations are solely due to the impact of error in facial image registration.

A forensic case involving face recognition commonly contains a surveillance view trace (usually a frame from CCTV footage) and a frontal suspect reference set containing facial images of suspects narrowed down by police and forensic investigation~\cite{grgic2011scface,burton1999face,klontz2013case}.
When a forensic investigator is tasked to compare the surveillance view trace (or, probe) to the suspect reference set, it is quite common to manually compare these images.
However, if the forensic investigator chooses to use an automatic face recognition system for this task, there are two choices available: a model based approach or a view based approach.
In a model based approach, a frontal view probe image is synthesized based on a 3D model reconstructed from the surveillance view trace.
Most face recognition systems are fine tuned for optimal recognition performance for comparing frontal view images and therefore the model based approach, with synthesized frontal probe and frontal suspect reference images, ensures high recognition performance.
In a view based approach, the reference set is adapted such that it matches the pose of the surveillance view trace.
This approach ensures that a face recognition system always gets to compare facial images under similar pose -- not necessarily the frontal view.
In a forensic face recognition case, prior knowledge about the impact of pose variations on the performance of a face recognition system -- addressed by the main research question -- can be used to decide between the two approaches: view based or model based.
In Chapter~\ref{dutta2014phdthesis_forensic-intro}, we investigate if it is potentially more useful to apply a view based approach in forensic cases.
This chapter addresses the following subordinate research question:
\begin{center}
\textit{\ResearchQuestionFour}
\end{center}

\section{Research Questions}
\label{dutta2014phdthesis_intro-resques}
In this dissertation, we address the following main research questions which in turn results in three subordinate research questions:
\begin{enumerate}
\item \ResearchQuestionTwo
\begin{enumerate}
 \item \ResearchQuestionOne
 \item \ResearchQuestionThree
 \item \ResearchQuestionFour
 \end{enumerate}
\end{enumerate}

\section{Contributions}
\label{dutta2014phdthesis_intro-contrib}
The work presented in this dissertation make the following major contributions:
\begin{description}
\item [A model for performance prediction based on image quality] In Chapter~\ref{dutta2015predicting_intro}, we present a generative model that captures the relation between image quality and face recognition performance.
The novelty of this approach is that it directly models the variable of interest (\ie recognition performance measure) instead of modeling intermediate variables like similarity score~\cite{shi2008modeling,scheirer2011meta}.
Furthermore, since the model is based only on image quality features, face recognition performance prediction can be done even before the actual recognition has taken place thereby facilitating many preemptive action.

\item [Instability of performance predictor features derived from similarity scores]
A considerable amount of literature on performance prediction have used features derived solely from similarity scores.
In Section~\ref{dutta2013facial_intro}, we evaluate the influence of image quality variations on the non-match score distribution of several face recognition systems.
The evidence from this study suggests that performance predicting features derived from similarity scores are unstable in the presence of image quality variation and therefore should be used with caution in performance prediction models.

\item [Impact of automatic eye detection error on face recognition performance]
Image quality variations have dual impact on performance of a face recognition system:
\begin{inparaenum}[\itshape a\upshape)]
\item impact on the accuracy of automatic eye detection; and
\item impact on the accuracy of facial feature extraction and comparison.
\end{inparaenum}
The investigation reported in Chapter~\ref{dutta2015impact_intro} has shown that, for a face recognition system sensitive to errors in facial image registration, the performance predictor feature set should include some features that can predict the accuracy of automatic eye detector used in the face recognition system.
This is essential to accurately model and predict the performance variations in a practical face recognition system.
So far, existing work has only focused on using features that predict the performance of face recognition algorithms.
Our work has laid the foundation for future work in this direction.

\item [Forensic face recognition]
The findings reported in this dissertation are also of interest to forensic investigators handling forensic cases involving face recognition.
In Section~\ref{dutta2014automatic_intro}, we present an image quality measure that is particularly useful in the context of forensic face recognition.
Chapter~\ref{dutta2014phdthesis_forensic-intro} discusses a view based strategy that can be applied in forensic cases dealing with surveillance view probe (or, trace) image.

\end{description}

\section{List of Publications}
Each chapter of this dissertation is based on the following published or submitted research papers:
\begin{description}
\item [Chapter~\ref{dutta2014phdthesis_pred-feat-intro}] : 
 \begin{description}
 \item [Section~\ref{dutta2013facial_intro}] : 

\cite{dutta2013facial} A. Dutta, R. N. J. Veldhuis, and L. J. Spreeuwers. Can facial uniqueness be inferred from impostor scores? In Biometric Technologies in Forensic Science, BTFS 2013, Nijmegen, Netherlands.

 \item [Section~\ref{dutta2014automatic_intro}] : 

\cite{dutta2014automatic} A. Dutta, R. N. J. Veldhuis, and L. J. Spreeuwers. Automatic eye detection
error as a predictor of face recognition performance. In 35rd WIC Symposium on Information Theory in the Benelux, Eindhoven, Netherlands, May 2014, pages 89 - 96.

 \end{description}
\item [Chapter~\ref{dutta2015predicting_intro}] : 
\begin{itemize}
\item \cite{dutta2015predicting} A. Dutta, R. N. J. Veldhuis, and L. J. Spreeuwers. Predicting face recognition performance using image quality. IEEE Transactions on Pattern Analysis and Machine Intelligence. \textit{(submitted)}
\item \cite{dutta2014bayesian} A. Dutta, R. N. J. Veldhuis, and L. J. Spreeuwers. A bayesian model for predicting face recognition performance using image quality. In IEEE International Joint Conference on Biometrics (IJCB), pages 1 - 8, 2014.
\end{itemize}

\item [Chapter~\ref{dutta2015impact_intro}] : 

\cite{dutta2015impact} A. Dutta, M. G{\"u}nther, L. E. El Shafey, S. Marcel, R. N. J. Veldhuis, and L. J. Spreeuwers. Impact of Eye Detection Error on Face Recognition Performance, IET Biometrics, 2015.

\newpage
\item [Chapter~\ref{dutta2014phdthesis_forensic-intro}] :
 \begin{description}
 \item [Section~\ref{dutta2012impact_intro}] : 

\cite{dutta2012impact} A. Dutta, R. N. J. Veldhuis, and L. J. Spreeuwers. The Impact of Image Quality
on the Performance of Face Recognition. In 33rd WIC Symposium on Information Theory in the Benelux, Boekelo, Netherlands, May 2012, pages 141 - 148.

 \item [Section~\ref{dutta2012view_intro}] : 

\cite{dutta2012view} A. Dutta, R. N. J. Veldhuis, and L. J. Spreeuwers. View based approach to forensic face recognition. Technical Report TR-CTIT-12-21, CTIT, University of Twente, Enschede, September 2012.

 \end{description}
\end{description}


\chapter{Features for Face Recognition Performance Prediction}
\label{dutta2014phdthesis_pred-feat-intro}
\section{Introduction}
For predicting the performance of a face recognition system, we require features that are correlated to recognition performance.
In this chapter, we investigate different performance predictor features that can be used to predict the performance of a face recognition system.
This study aims to select the features for performance prediction model discussed in~\chaptername~\ref{dutta2015predicting_intro}.

Quality of facial images are quite popular and intuitive features for performance prediction.
In Section~\ref{dutta2014phdthesis_img-qual-features}, we discuss about the merit of using image quality features such as pose, illumination, \etc as a performance predictor feature.
Since, these image quality features have been widely covered by existing literature, we discuss and select these features based on the results from existing work.

Several past work have also used features derived from similarity score as a predictor of recognition performance.
The key observation underpinning these features is that the overlapping region between match and non-match score distribution entail more uncertainty in decision about identity and therefore correspond to poorer recognition performance.
In Section~\ref{dutta2013facial_intro}, we investigate the stability of non-match (or, impostor) scores and a performance predictor feature derived from non-match scores (\ie Impostor-based Uniqueness Measure) when subject to image quality variations.
These investigations are aimed at assessing the stability of performance predictor features derived from similarity scores.
This analysis helps us decide if such features should be used in the performance prediction model of~\chaptername~\ref{dutta2015predicting_intro}.

The accuracy of automatic eye detectors is affected by the quality of facial image on which it operates.
For instance, a facial image captured under uneven illumination condition would entail higher error -- with respect to manually annotated eye location ground truth -- in automatically detected eye location as compared to the facial image captured under studio lighting conditions.
There are many facial image quality variations that affect the performance of both automatic eye detectors an face recognition algorithms.
In Section~\ref{dutta2014automatic_intro}, we investigate if the extent of error in automatic eye detection is correlated to the recognition performance.
If such correlation exists, the Automatic Eye Detection Error (AEDE) can be used as a feature for performance prediction in the model discussed in \chaptername~\ref{dutta2015predicting_intro}.

\section{Image Quality Features as a Predictor of Face Recognition Performance}
\label{dutta2014phdthesis_img-qual-features}
Facial features can be extracted more accurately from images captured under studio conditions -- frontal pose and illumination, sharp, high resolution \etc.
This contributes to more certainty in the decision about identity and therefore results in better recognition performance.
However, when the studio conditions are not met, facial features may be occluded or obscured causing inaccuracies in the extracted facial features which results in more uncertainty in decision about identity.
Therefore, image quality features such as pose, illumination direction, noise, resolution \etc can be used as a predictor of uncertainty in decision about identity.
Recall that, in this dissertation, we use the term \textit{image quality} to refer to all the static or dynamic characteristics of the subject or acquisition process as described in~\cite{iso_iec_29794-5:2010}.

Facial image quality measures like pose, illumination, noise, resolution, focus, \etc have a proven record of being a reliable predictor of face recognition performance.
Previous work such as~\cite{phillips2013existence,beveridge2008focus} have also shown the merit of following image quality features as a performance predicting feature: pose and illumination, image resolution, sharpness (or, focus), noise, \etc.
Of all the available image quality features, we focus our attention on pose and illumination -- two popular and simple image quality features.
This choice of image quality feature is motivated by the existence of publicly available large data sets~\cite{gross2008multipie,gao2008caspeal}~with controlled variations of pose and illumination.
Therefore, we select pose and illumination as two image quality features for performance prediction model of~\chaptername~\ref{dutta2015predicting_intro}.
According to the classification scheme for facial image quality variations proposed in \cite{iso_iec_29794-5:2010}, head pose and illumination correspond to subject characteristics and acquisition process characteristics respectively.
Furthermore, both quality parameters correspond to dynamic characteristics of a facial image.

\section{Can Facial Uniqueness be Inferred from Impostor Scores?}
\label{dutta2013facial_intro}
The appearances of some human faces are more similar to facial appearances of other subjects in a population. Those faces whose appearance is very different from the population are often called a unique face. Facial uniqueness is a measure of distinctness of a face with respect to the appearance of other faces in a population. Non-unique faces are known to be more difficult to recognize by the human visual system \cite{going1974effects}  and automatic face recognition systems \cite[Fig.~6]{klare2012face}. Therefore, in Biometrics, researchers have been actively involved in measuring uniqueness from facial photographs \cite{klare2012face,ross2009exploiting,yager2010biometric,wittman2006empirical}. Such facial uniqueness measurements are useful to build an adaptive face recognition system that can apply stricter decision thresholds for fairly non-unique facial images which are much harder to recognize.

Most facial uniqueness measurement algorithms quantify the uniqueness of a face by analyzing its similarity score (i.e.\ impostor score) with the facial image of other subjects in a population. For example, \cite{klare2012face} argue that a non-unique facial image (i.e.\ lamb\footnote{sheep: easy to distinguish given a good quality sample, goats: have traits difficult to match, lambs: exhibit high levels of similarity to other subjects, wolves: can best mimic other subject's traits} as defined in \cite{doddington1998sheep}) ``will generally exhibit high level of similarity to many other subjects in a large population (by definition)''. Therefore, they claim that facial uniqueness of a subject can be inferred from its impostor similarity score distribution.

In this paper, we show that impostor scores are not only influenced by facial identity (which in turn defines facial uniqueness) but also by quality aspects of facial images like pose, noise and blur. Therefore, we argue that any facial uniqueness measure based solely on impostor scores will give misleading results for facial images degraded by quality variations.

The organization of this paper is as follows: in section \ref{dutta2013facial_relatedwork}, we review some existing methods that use impostor scores to measure facial uniqueness, next in section \ref{dutta2013facial_quality-impostor} we describe the experimental setup that we use to study the influence of facial identity and image quality on impostor scores, in section \ref{dutta2013facial_ium-stability} we investigate the stability of one recently introduced impostor-based uniqueness measure (i.e.\ \cite{klare2012face}). Finally, in section \ref{dutta2013facial_discussion}, we discuss the experimental results and present the conclusions of this study in section \ref{dutta2013facial_conclusion}.

\subsection{Related Work}
\label{dutta2013facial_relatedwork}
Impostor score distribution has been widely used to identify the subjects that exhibit high level of similarity to other subjects in a population (i.e.\ lamb). The authors of \cite{doddington1998sheep} investigated the existence of ``lamb'' in speech data by analyzing the relative difference between maximum impostor score and genuine score of a subject. They expected the ``lambs'' to have very high maximum impostor score. A similar strategy was applied by \cite{wittman2006empirical} to locate non-unique faces in a facial image data set. The authors of \cite{ross2009exploiting} tag a subject as ``lamb'' if its mean impostor score lies above a certain threshold. Based on this knowledge of a subject's location in the ``Doddington zoo'' \cite{doddington1998sheep}, they propose an adaptive fusion scheme for a multi-modal biometric system. Recently, \cite{klare2012face} have proposed an Impostor-based Uniqueness Measure (IUM) which is based on the location of mean impostor score relative to the maximum and minimum of the impostor score distribution. Using both genuine and impostor scores, \cite{yager2010biometric} investigated the existence of biometric menagerie in a broad range of biometric modalities like 2D and 3D faces, fingerprint, iris, speech, etc.

All of these methods that aim to measure facial uniqueness from impostor scores assume that impostor score is only influenced by facial identity. In this paper, we show that impostor scores are also influenced by image quality (like pose, noise, blur, etc). The authors of \cite{paone2011difficult} have also concluded that facial uniqueness (i.e.\ location in the biometric zoo) changes easily when imaging conditions (like illumination) change.


\subsection{Influence of Image Quality on Impostor Score Distribution}
\label{dutta2013facial_quality-impostor}
\begin{figure}
 \centering
 \includegraphics[width=0.2\linewidth]{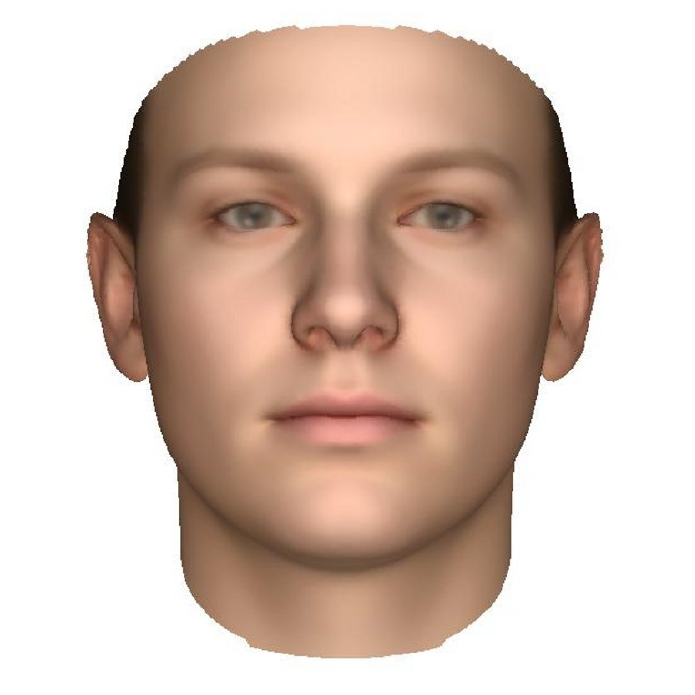}
 \caption{Average face image}
 \label{fig:bfm_average_face}
\end{figure}

In this section, we describe an experimental setup to study the influence of image quality on impostor scores. We fix the identity of query image to an average face image synthesized\footnote{using the code and model provided with \cite{paysan20093dface}} by setting the shape ($\alpha$) and texture ($\beta$) coefficients to zero $(\alpha, \beta = 0)$ as shown in \figurename\ref{fig:bfm_average_face}. We obtain a baseline impostor score distribution by comparing the similarity between the average face and a gallery set (or, impostor population) containing $250$ subjects. Now, we vary the quality (pose, noise and blur) of this gallery set (identity remains fixed) and study the variation of impostor score distribution with respect to the baseline. Such a study will clearly show the influence of image quality on impostor score distribution as only image quality varies while the facial identity remains constant in all the experiments.


We use the MultiPIE neutral expression data set of \cite{gross2008multipie} to create our gallery set. Out of the 337 subjects in MultiPIE, we select 250 subjects that are common in session (01,03) and session (02,04). In other words, our impostor set contains subjects from $(S_1 \cup S_3) \cap (S_2 \cup S_4)$, where $S_i$ denotes the set of subjects in MultiPIE session $i \in \{1,2,3,4\}$ recording $1$. From the group $(S_1 \cup S_3)$, we have $407$ images of $250$ subject and from the group $(S_2 \cup S_4)$, we have $413$ images of the same $250$ subjects. Therefore, for each experiment instance, we have $820$ images of $250$ subjects with at least two image per subject taken from different sessions.

We compute the impostor score distribution using the following four face recognition systems: FaceVACS \cite{facevacs2010}, Verilook \cite{verilook2011}, Local Region PCA and Cohort LDA \cite{bolme2012csu}. The first two are commercial while the latter two are open source face recognition systems. We supply the same manually labeled eye coordinates to all the four face recognition systems in order to avoid the performance variation caused by automatic eye detection error.

In this experiment, we consider impostor population images with frontal view (cam $05\_1$) and frontal illumination (flash $07$) images as the baseline quality. We consider the following three types of image quality variations of the impostor population: pose, blur, and noise as shown in \figurename\ref{fig:qual_var_illus}. For pose, we vary the camera-id (with flash that is frontal with respect to the camera) of the impostor population. For noise and blur, we add artificial noise and blur to frontal view images (cam $05\_1$) of the impostor population. We simulate imaging noise by adding zero mean Gaussian noise with the following variances: $\{0.007, 0.03, 0.07, 0.1, 0.3\}$ (where pixel value is in the range $[0,1.0]$). To simulate $N$ pixel horizontal linear motion of subject, we convolve frontal view images with a $1 \times N$ averaging filter, where $N \in \{3,5,7,13,17,29,31\}$ (using Matlab's \texttt{fspecial('motion', N, 0)} function). For pose variation, camera-id $19\_1$ and $08\_1$ refer to right and left surveillance view images respectively.

In \figurename\ref{fig:fv_csu_qri_rf0}, we report the variation of impostor score distribution of the average face image as box plots. In these box plot, the upper and lower hinges correspond to the first and third quantiles. The upper (and lower) whisker extends from the hinge to the highest (lowest) value that is within $1.5 \times $IQR where IQR is the distance between the first and third quartiles. The outliers are plotted as points.

\subsection{Stability of Impostor-based Uniqueness Measure Under Quality Variation}
\label{dutta2013facial_ium-stability}
In this section, we investigate the stability of a recently proposed impostor-based facial uniqueness measure \cite{klare2012face} under image quality variations. The key idea underpinning this method is that a fairly unique facial appearance will result in low similarity score with a majority of facial images in the population. This definition of facial uniqueness is based on the assumption that similarity score is influenced only by facial identity. 

This facial uniqueness measure is computed as follows: Let $i$ be a probe (or query) image and $J=\{j_1,\cdots,j_n\}$ be a set of facial images of $n$ different subjects such that $J$ does not contain an image of the subject present in image $i$. In other words, $J$ is the set of impostor subjects with respect to the subject in image $i$. If $S = \{s(i,j_1), \cdots, s(i,j_n)\}$ is the set of similarity score between image $i$ and the set of images in $J$, then the Impostor-based Uniqueness Measure (IUM) is defined as:
\begin{equation}
  u(i,J) = \frac{S_{max} - \mu_{S}}{S_{max} - S_{min}}
 \label{eq:ium_score_computation}
\end{equation}
where,  $S_{min}, S_{max}, \mu_{S}$ denote minimum, maximum and average value of impostor scores in $S$ respectively. A facial image $i$ which has high similarity with a large number of subjects in the population will have a small IUM value $u$ while an image containing highly unique facial appearance will take a higher IUM value $u$.

For this experiment, we compute the IUM score of $198$ subjects common in session $3$ and $4$ (i.e.\ $S_3 \cap S_4$) of the MultiPIE data set. The IUM score corresponding to same identity but computed from two different sessions (the frontal view images without any artificial noise or blur) must be highly correlated. We denote this set of IUM scores as the baseline uniqueness scores. To study the influence of image quality on the IUM scores, we only vary the quality (pose, noise, blur as shown in  \figurename\ref{fig:qual_var_illus}) of the session $4$ images and we compute the IUM scores under quality variation. If the IUM scores are stable with image quality variations, the IUM scores computed from session $3$ and $4$ should remain highly correlated despite quality variation in session $4$ images. Recall that the facial identity remains fixed to the same $198$ subjects in all these experiments.

In \cite{klare2012face}, the authors compute IUM scores from an impostor population of $8000$ subjects taken from a private data set. We do not have access to such a large data set. Therefore, we import additional impostors from CAS-PEAL data set (1039 subjects from PM+00 subset) \cite{gao2008cas} and FERET (1006 subjects from Fa subset) \cite{phillips2000feret}. So, for computing the IUM score for subject $i$ in session $3$, we have a impostor population containing the remaining 197 subjects from session $3$, $1039$ subjects from CAS-PEAL and $1006$ subjects from FERET. Therefore, each of the IUM score is computed from an impostor set $S$ containing a single frontal view images of $197+1039+1006=2242$ subjects as shown in \figurename\ref{fig:klare2012face_exp_data_illus}. In a similar way, we compute IUM scores for the same $198$ subjects but with images taken from session $4$. As the Cohort LDA system requires colour images, we replicate the gray scale images of FERET and CAS-PEAL in RGB channels to form a colour image. Note that we only vary the quality of a single query facial image $i$ (from session $4$) while keeping the impostor population quality $J$ fixed to $2242$ frontal view images (without any artificial noise or blur).

In \tablename\ref{tbl:session_3_4_ium_correlation}, we show the variation of Pearson correlation coefficient between IUM scores of $198$ subjects computed from session $3$ and $4$. The bold faced entries correspond to the correlation between IUM scores computed from frontal view (without any artificial noise or blur) images of the two sessions. The remaining entries denote variation in correlation coefficient when the quality of facial image in session $4$ is varied without changing the quality of impostor set. In \figurename\ref{fig:norm_corr_coef_ium}, we show the drop-off of normalized correlation coefficient (derived from \tablename\ref{tbl:session_3_4_ium_correlation}) with quality degradation where normalization is done using baseline correlation coefficient.

\subsection{Discussion}
\label{dutta2013facial_discussion}

\subsubsection{Influence of Image Quality on Impostor Score}
In \figurename\ref{fig:fv_csu_qri_rf0}, we show the variation of impostor score distribution with image quality variations of the impostor population. We consider frontal view (cam $05\_1$) image without any artificial noise or blur (i.e.\ the original image in the data set) as the baseline image quality. The box plot corresponding to cam-id=$05\_1$, blur-length=$0$, noise-variance=$0$ denotes mainly the impostor score variation due to facial identity. From \figurename\ref{fig:fv_csu_qri_rf0}, we observe that, for all the three quality variations, the nature of impostor distribution corresponding to quality variations is significantly different from the baseline impostor distribution. This shows that the impostor score distribution is influenced by both identity (as expected) and image quality.

\subsubsection{Stability of Impostor-based Uniqueness Measure Under Quality Variation}
We observe a common trend in the variation of correlation coefficients with image quality degradation as shown in \tablename\ref{tbl:session_3_4_ium_correlation}. The correlation coefficient is maximum for the baseline image quality (frontal, no artificial noise or blur). As we move away from the baseline image quality, the correlation between IUM scores reduces. This reduction in correlation coefficient indicates the instability of Impostor-based Uniqueness Measure (IUM) in the presence of image quality variations.

The instability of IUM is also depicted by the normalized correlation coefficient plot of \figurename\ref{fig:norm_corr_coef_ium}. For all the four face recognition systems, we observe fall-off of the correlation between IUM scores with variation in pose, noise and blur of facial images. For pose variation, peak correlation is observed for frontal view (camera 05\_1) facial images because all the four face recognition systems are tuned for comparing frontal view facial images.

The instability of IUM measure is also partly due to the use of minimum and maximum impostor scores in equation (\ref{eq:ium_score_computation}) which makes it more susceptible to outliers.

The authors of \cite{klare2012face} report a correlation of $\geq 0.98$ using FaceVACS system on a privately held mug shot database of 8000 subjects. We get a much lower correlation coefficient of $\leq 0.68$ on a combination of three publicly released data set. One reason for this drop in correlation may be due to the use of different data sets in the two experiments. Our impostor population is formed using images taken from three publicly available data set and therefore represents larger variation in image quality as shown in \figurename\ref{fig:klare2012face_exp_data_illus}. To a lesser extent, this difference in correlation could also be due to difference in the FaceVACS SDK version used in the two experiments. We use the FaceVACS SDK version 8.4.0 (2010) and they have not mentioned the SDK version used in their experiments.

\subsection{Conclusion}
\label{dutta2013facial_conclusion}
We have shown that impostor score is influenced by both identity and quality of facial images. We have also shown that any attempt to measure characteristics of facial identity (like facial uniqueness) solely from impostor score distribution will give misleading results in the presence of image quality degradation in the input facial images.

%
%
\clearpage

\begin{figure}
 \centering
 \includegraphics[width=0.7\linewidth]{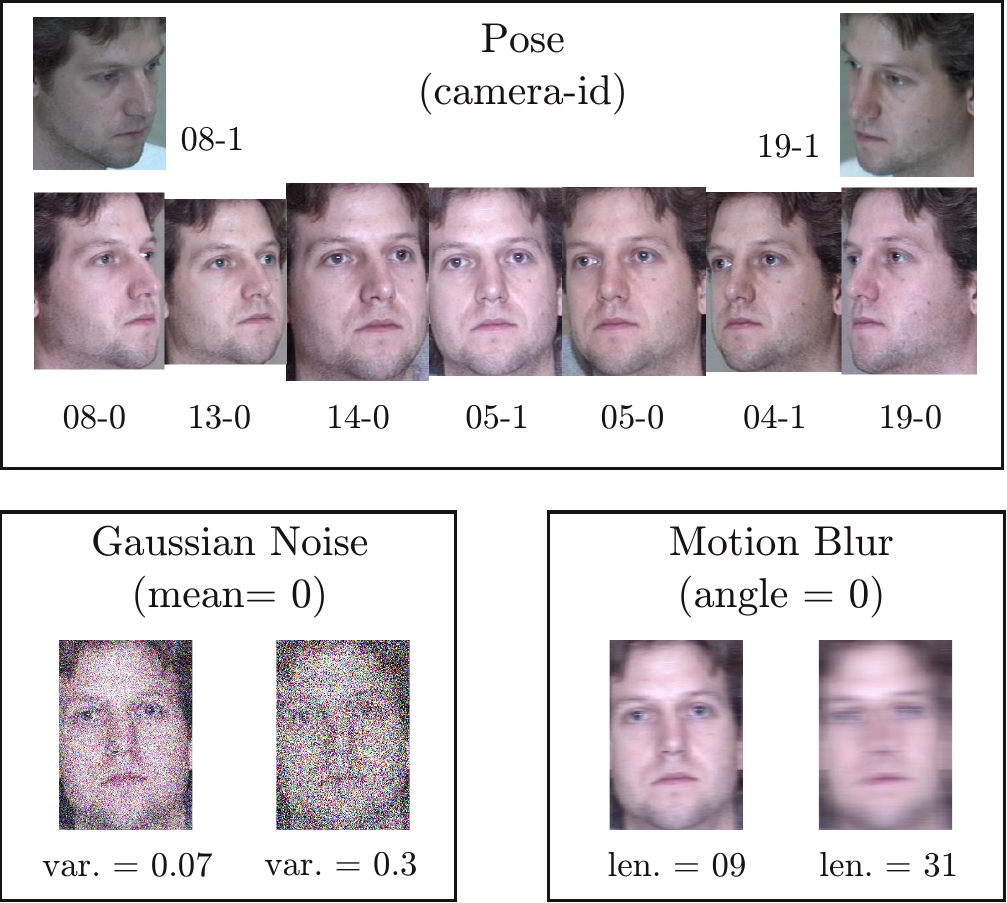}
 \caption{Facial image quality variations included in this study.}
 \label{fig:qual_var_illus}
\end{figure}

\begin{figure*}
 \centering
 \includegraphics[width=0.8\linewidth]{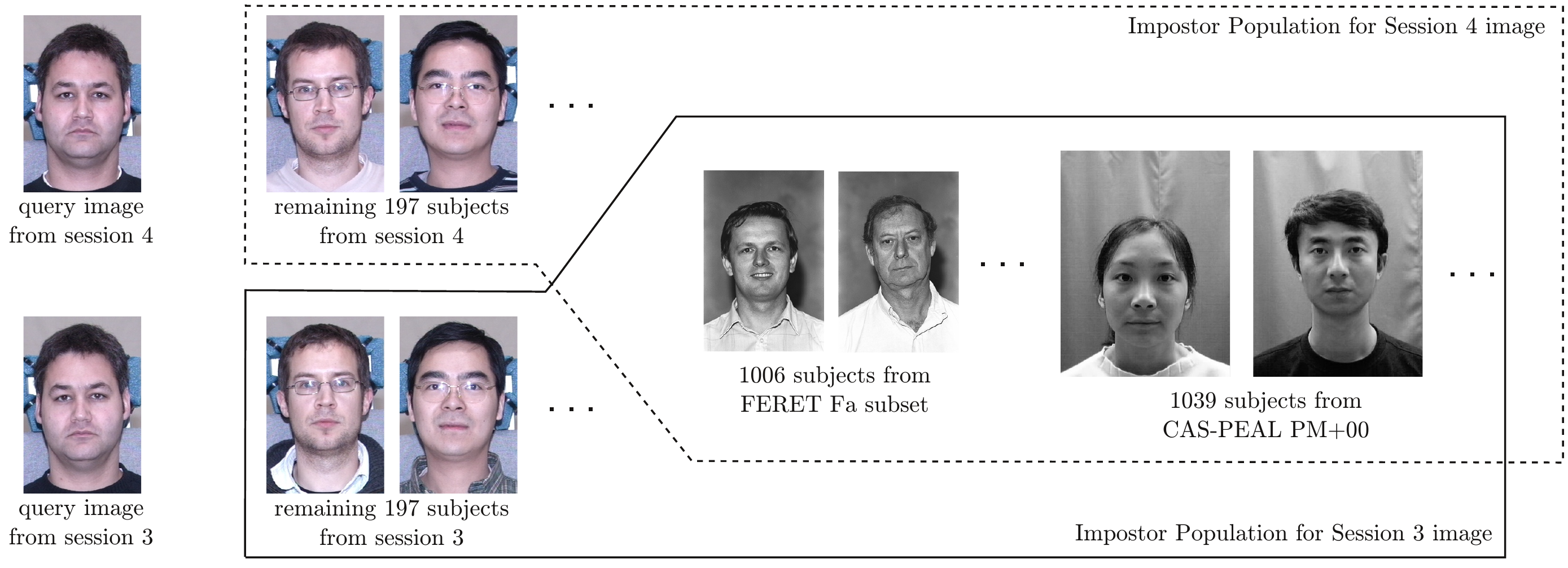}
 \caption{Selection of impostor population for IUM score computation.}
 \label{fig:klare2012face_exp_data_illus}
\end{figure*}

\begin{figure*}[ht]
 \centering
 \includegraphics[width=0.8\linewidth]{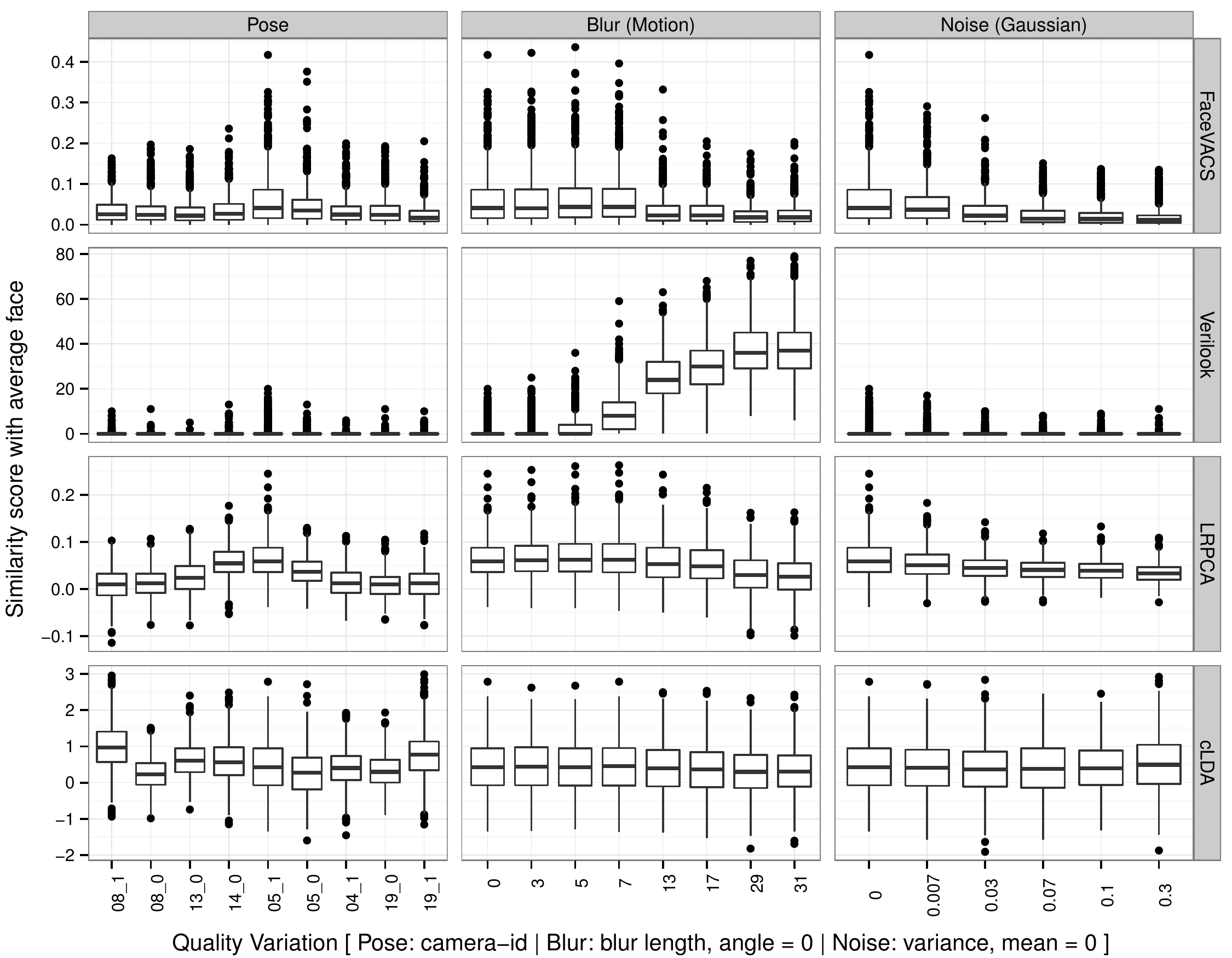}
 \caption{Influence of image quality on impostor score distribution}
 \label{fig:fv_csu_qri_rf0}
\end{figure*}

\begin{table*}[ht]
 \centering
 \subfloat{
 \centering
 \begin{tabular}{l|c|c|c|c|>{\bfseries}c|c|c|c|c}
 \hline
 & 08\_1 & 08\_0 & 13\_0 & 14\_0 & frontal & 05\_0 & 04\_1 & 19\_0 & 19\_1 \\
\hline
FaceVACS & 0.12 & 0.19 & 0.23 & 0.52 & 0.68 & 0.51 & 0.37 & 0.14 & 0.07 \\
Verilook & 0.04 & 0.12 & 0.28 & 0.45 & 0.63 & 0.54 & 0.21 & 0.21 & 0.19 \\
LRPCA & 0.10 & 0.06 & -0.07 & 0.11 & 0.45 & 0.29 & 0.15 & 0.03 & -0.05 \\
cLDA & 0.04 & 0.09 & 0.17 & 0.21 & 0.43 & 0.34 & 0.22 & -0.13 & 0.05 \\
\hline
 \multicolumn{5}{r|}{$\xleftarrow{\textrm{drop in correlation with pose}}$} & baseline & \multicolumn{4}{l}{$\xrightarrow{\textrm{drop in correlation with pose}}$} \\

 \end{tabular}
 }

 \subfloat{
 \begin{tabular}{l|>{\bfseries}c|c|c|c|c}
 \hline
& No blur & length 5 & length 9 & length 17 & length 31 \\
 \hline
FaceVACS & 0.68 & 0.65 & 0.59 & 0.27 & 0.13 \\
Verilook & 0.63 & 0.63 & 0.54 & 0.45 & 0.27 \\
LRPCA & 0.45 & 0.43 & 0.16 & 0.04 & 0.04 \\
cLDA & 0.43 & 0.42 & 0.40 & 0.38 & 0.32 \\
\hline
 & baseline & \multicolumn{4}{l}{$\xrightarrow{\textrm{drop in correlation with blur}}$} \\
 \end{tabular}
 }

 \subfloat{
 \begin{tabular}{l|>{\bfseries}c|c|c|c|c}
 \hline
 & No noise & $\sigma=0.03$ &  $\sigma=0.07$ &  $\sigma=0.1$ &  $\sigma=0.3$ \\
 \hline
FaceVACS & 0.68 & 0.47 & 0.43 & 0.33 & 0.15 \\
Verilook & 0.63 & 0.28 & 0.18 & 0.16 & 0.03 \\
LRPCA & 0.45 & 0.43 & 0.29 & 0.29 & 0.14 \\
cLDA & 0.43 & 0.37 & 0.28 & 0.23 & 0.22 \\
\hline
 & baseline & \multicolumn{4}{l}{$\xrightarrow{\textrm{drop in correlation with noise}}$} \\
 \end{tabular}
 }
 \caption{Variation in correlation of the impostor-based uniqueness measure \cite{klare2012face} for $198$ subjects computed from sessions $3$ and $4$. Note that image quality (pose, noise and blur) of session $4$ images were only varied while session $3$ and impostor population images were fixed to frontal view images without any artificial noise or blur.}
 \label{tbl:session_3_4_ium_correlation}
\end{table*}

\begin{figure*}[ht]
 \centering
 \includegraphics[width=0.8\linewidth]{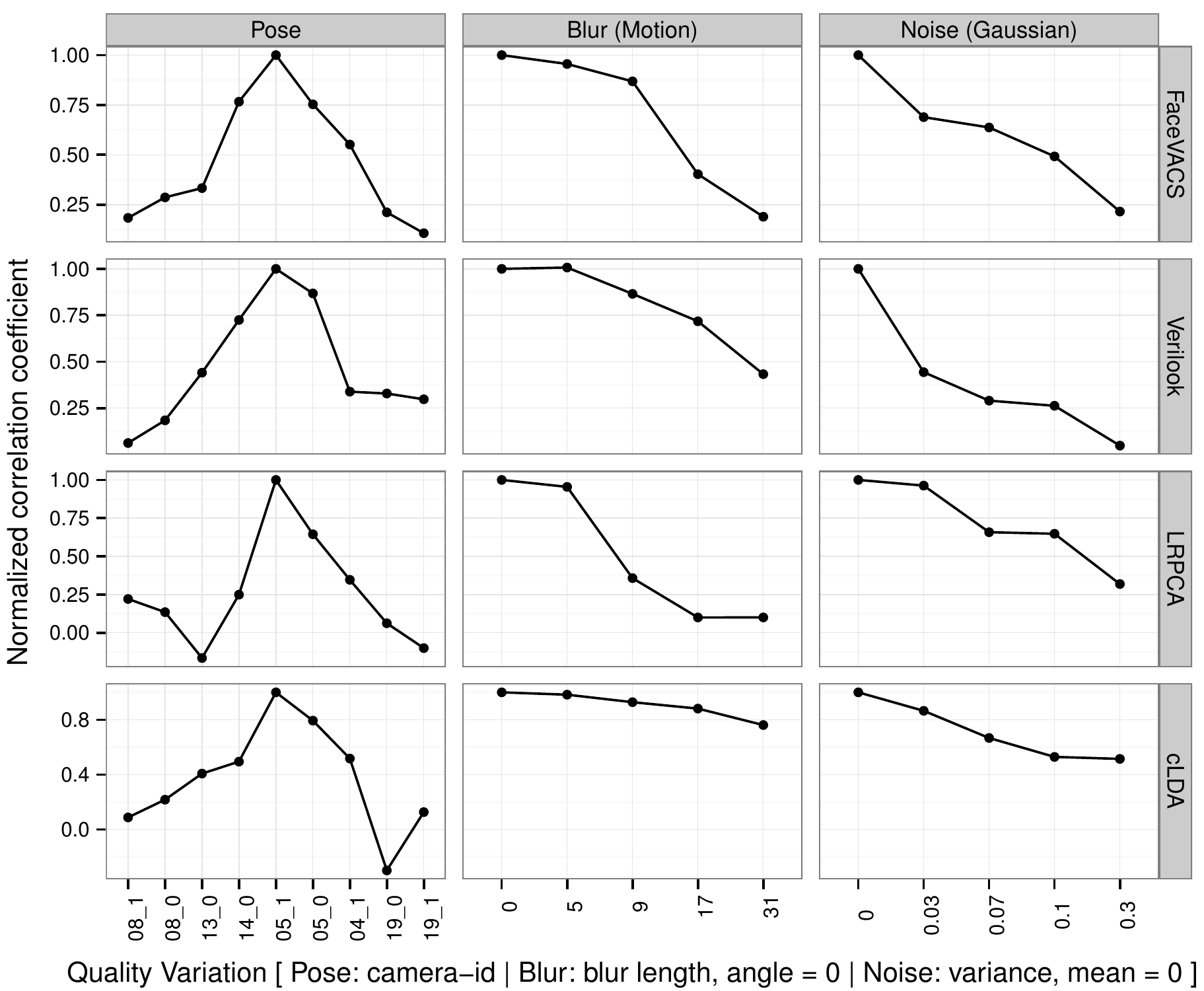}
 \caption{Fall-off of normalized correlation coefficient with quality degradation. Normalization performed using correlation coefficient corresponding to frontal, no blur and no noise case.}
 \label{fig:norm_corr_coef_ium}
\end{figure*}

\begin{figure}
\centering
\includegraphics[width=0.8\linewidth]{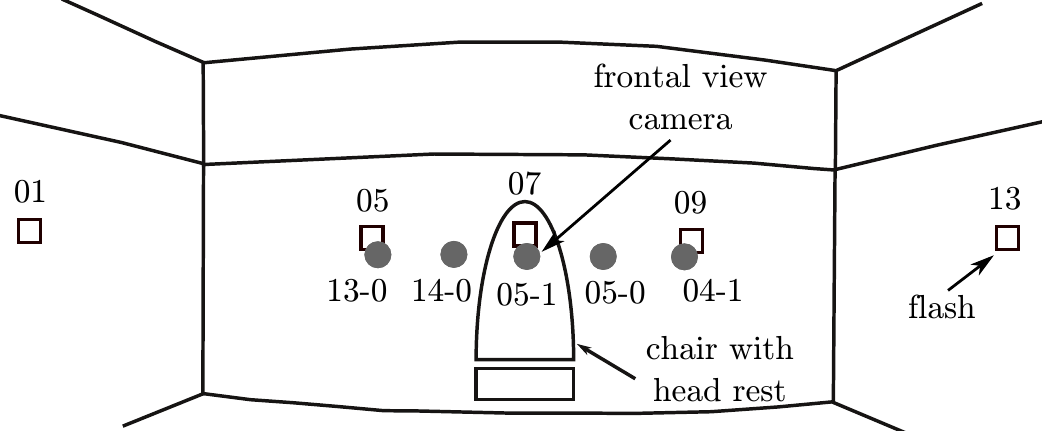}
\caption{MultiPIE camera and flash positions used in this paper.}
\label{fig:images/capture_setup_5cam_5illum.pdf}
\end{figure}


\clearpage

\section{Automatic Eye Detection Error as a Predictor of Face Recognition Performance}
\label{dutta2014automatic_intro}
The quality of facial images is known to affect the performance of a face recognition system. 
A large and growing body of literature has investigated the impact of various image quality parameters on the performance of existing face recognition systems~\cite{beveridge2008focus}. 
The most commonly used image quality parameters are: facial pose, illumination direction, noise, blur, facial expression, image resolution. 
However, some aspects of the recognition performance that cannot be explained by the existing image quality measures remain. 
This shows that still more quality parameters are needed to fully explain the variation in recognition performance.

In this paper, we propose a novel image quality parameter called the \textit{Automatic Eye Detection Error} (AEDE). 
Automatic eye detectors are trained to return the location of two eye coordinates in a facial image. 
To assess the accuracy of automatic eye detectors, we use the manually annotated eye coordinates as the ground truth eye locations. 
The proposed AEDE measures the error in automatically detected eye coordinates. 
The main insight underpinning this novel image quality parameter is as follows: Automatic eye detection becomes more difficult for poor quality facial images and hence the eye detection error should be an indicator of image quality and face recognition performance. 
In other words, we use the knowledge of the accuracy of one classifier (\ie automatic eye detector) as the predictor of the accuracy of another classifier (\ie the face recognition system) when both operate on the same pair of facial images.
The proposed AEDE quality measure can be seen as providing a summary of many, but not all, properties of a facial image.

This paper is organized as follows: 
In Section~\ref{dutta2014automatic_relatedwork}, we review some previous work in this area.
We explain the proposed AEDE quality measure in Section~\ref{dutta2014automatic_methodology}.
We describe experiments to study the relationship between AEDE and face recognition performance in Section~\ref{dutta2014automatic_exp}.

\subsection{Related Work}
\label{dutta2014automatic_relatedwork}
The face recognition research community has been investigating the impact of automatic eye detection error on facial image registration which in turn influences face recognition performance \cite{marques2000effects,wang2005sensitivity,min2005eye,riopka2003eyes,shan2004curse,wang2007modeling,rodriguez2006measuring}.
While some researchers have focused on improving the accuracy of automatic eye detectors \cite{wang2005automatic}, others have explored multiple ways to make face recognition systems inherently robust to facial image registration errors \cite{wagner2012towards,wallace2011intersession}.

To the best of our knowledge, no previous work has proposed the \textit{Automatic Eye Detection Error} (AEDE) as a predictor of face recognition performance.
However,~\cite{wang2005sensitivity} make a concluding remark that points in this direction.
The authors mention that ``a face recognition system suffers a lot when the testing images have the lower face lighting quality, relatively smaller facial size in the image, ...''.
They further note that ``the automatic eye-finder suffers from those kinds of images too''.
This paper is probably the first to observe that some facial image quality parameters (like illumination, resolution, \etc) impact the performance of both face recognition systems and automatic eye detectors.

\subsection{Methodology}
\label{dutta2014automatic_methodology}
Manually annotated eye coordinates are used as the ground truth for the eye locations in a facial image.
Based on this knowledge of true location of the two eyes, we can assess the accuracy of an automatic eye detector.
The error in automatic eye detection gives an indication of how difficult it is to automatically detect eyes in that facial image.
Some of the image quality variations that make the automatic eye detection difficult also contribute towards the uncertainty in decision about identity made by a face recognition system operating on that facial image.
For example: a poorly illuminated facial image not only makes eye detection difficult but it also makes face recognition harder.

Let $p_{\{l,r\}}^{m}$ denote the manually located left and right eye coordinates (\ie the ground truth).
An automatic eye detector is trained to locate the position of the two eye coordinates $p_{\{l,r\}}^{d}$ in a facial image.
The error in automatically detected eye coordinates can be quantified using the Automatic Eye Detection Error (AEDE)~\cite{jesorsky2001robust} as follows:
\begin{equation}
J = \frac{max\{ ||p_{l}^{m}-p_{l}^{d}||, ||p_{r}^{m}-p_{r}^{d}||\}}{||p_{l}^{m}-p_{r}^{m}||}
\label{eq:norm_eye_det_err}
\end{equation}
Let $J_{\{p,g\}}$ denote the AEDE in a probe and gallery image pair respectively.
For this probe and gallery image pair, let $s^{k}$ denote the similarity score computed by face recognition system $k$.
We divide $J$ into $L$ monotonically increasing intervals (based on quantiles, standard deviation of observed $J_{\{p,g\}}$, \etc): $J^{l}$ where $l \in \{1,\cdots,L\}$.
We partition the set of all similarity scores $S$ into $L \times L$ categories of genuine $G$ and impostor $I$ scores defined as follows:
\begin{eqnarray}
G_{(l_1,l_2)} &=& \{ S(i):J_{p}(i) \in J^{l_{1}} \wedge J_{g}(i) \in J^{l_2} \wedge S(i) \; \textnormal{denotes genuine comparison} \}, \\
I_{(l_1,l_2)} &=& \{ S(i):J_{p}(i) \in J^{l_1} \wedge J_{g}(i) \in J^{l_2} \wedge S(i) \; \textnormal{denotes impostor comparison} \},
\end{eqnarray}
where, $l_1,l_2 \in \{1,\cdots,L\}$, $J_{\{p,g\}}(i)$ denotes the normalized eye detection error (or, AEDE) in probe and gallery image respectively corresponding to $i^{\textnormal{th}}$ similarity score $S(i)$.
The performance of a verification experiment is depicted using a Receiver Operating Characteristics (ROC) curve.
The ROC curve corresponding to a particular eye detection error interval $(l_1,l_2)$ is jointly quantified by False Accept Rate (FAR) and False Reject Rate (FRR) defined as follows:
\begin{equation}
\begin{aligned}
FAR_{(l_1,l_2)}(t) &=& \frac{n( \{I_{l_1,l_2}: I_{l_1,l_2} > t\} )}{n(I_{l_1,l_2}}, \\
FRR_{(l_1,l_2)}(t) &=& \frac{n( \{G_{l_1,l_2}: G_{l_1,l_2} < t\} )}{n(G_{l_1,l_2}},
\end{aligned}
\label{eq:roc_far_frr_eq}
\end{equation}
where, $t$ denotes the decision threshold similarity score and $n(A)$ denotes the cardinality of set $A$.

Our hypothesis is that the eye detection error $J$ defined in \eqref{eq:norm_eye_det_err} is correlated with face verification performance defined by \eqref{eq:roc_far_frr_eq}.
Therefore, we expect ROC curves corresponding to different eye detection error intervals to be distinctly different from each other.
Furthermore, we also expect recognition performance to degrade monotonically with increase in eye detection error.

The proposed AEDE quality measure should be used with caution because all the factors that make eye detection difficult are not necessarily always involved in making face recognition harder.
For example, a facial photograph captured under studio conditions but with the subject's eyes closed is a difficult image for automatic eye detector while a face recognition system can still make accurate decisions as most important facial features are still clearly visible.
Therefore, in addition to the automatic eye detection error, we need more quality parameters in order to reliably predict face recognition performance.

\subsection{Experiments}
\label{dutta2014automatic_exp}
In this section, we describe experiments that allow us to study the relationship between Automatic Eye Detection Error (AEDE) and the corresponding face recognition performance.

We use the facial images present in the neutral expression subset of the MultiPIE data set~\cite{gross2008multipie}.
We include all the 337 subjects present in all the four sessions (first recording only).
In our experiments, the image quality (\ie pose and illumination) variations are only present in the probe (or, query) set.
The gallery (or, enrollment) set remains fixed and contains only high quality frontal mugshots of the $337$ subjects.
The probe set contains images of the same $337$ subjects captured by the 5 camera and under 5 flash positions (including no-flash condition) as depicted in~\figurename~\ref{fig:images/capture_setup_5cam_5illum.pdf}.
Since our gallery set remains constant, we only quantify the normalized eye detection error for facial images in the probe set $J_p$.
Of the total $27630$ unique images in the probe set, we discard $69$ images for which the automatic eye detector of FaceVACS fails to locate the two eyes.

We have designed our experiment such that there is minimal impact of session variation and image alignment on the face recognition performance.
We select the high quality gallery image from the same session as the session of the probe image.
Furthermore, we disable the automatically detected eye coordinates based image alignment of FaceVACS by supplying manually annotated eye coordinates for both probe and gallery images.
This ensures that there is consistency in facial image alignment even for non-frontal view images.

We manually annotate the eye locations~$p_{\{l,r\}}^{m}$~in all the facial images present in our data set.
Using the eye detector present in the FaceVACS SDK~\cite{facevacs2010}, we automatically locate position of the two eyes~$p_{\{l,r\}}^{d}$~in all facial images.
Given the manually annotated and automatically detected eye locations, we quantify the eye detection error $J$ using~\eqref{eq:norm_eye_det_err}.
In~\figurename~\ref{fig:images/eye_det_err_stat.pdf}, we show the distribution of normalized eye detection error $J_p$ for images in the probe set categorized according to MultiPIE camera and flash identifier.
The horizontal and vertical axes of~\figurename~\ref{fig:images/eye_det_err_stat.pdf} represent variations in camera and flash respectively.
The inset images show a sample probe image with the given pose and illumination.

Now, using FaceVACS~\cite{facevacs2010} recognition system, We now obtain the verification performance corresponding to each unique pair of probe and gallery images.
For each verification instance, we have $(J_{p}, s_{pg}^{k})$ where $J_p$ denotes the normalized eye detection error in the probe image and $s_{pg}^{k}$ is the similarity score (\ie verification score) computed by $k^{\textnormal{th}}$ face recognition system.
Since we use only one face recognition system in our experiments, we drop the superscript $k$.
Recall that our gallery set remains fixed to high quality images and therefore, we only consider the eye detection error of probe images.
This not only simplifies the analysis and presentation of results but also simulates the conditions of a real world verification experiment.
We partition the set of all similarity scores $S=\{s_{pg}\}$ into four categories based on the corresponding normalized eye detection error of the probe image $J_p$.
If $q_1,q_2,q_3$ denote the $25\%,50\%,75\%$ quantiles of $J_p$, then the four categories correspond to the following interval: $J_1=[0,q_1), J_2=[q_1,q_2), J_3=[q_2,q_3), J_4=[q_3,1)$.
In~\figurename~\ref{fig:images/roc_discrete_eye_err_bin.pdf}, we show the ROC corresponding to the four intervals of $J_p$ as shown in~\tablename~\ref{tbl:eye_det_err_interval}.
The solid lines in~\figurename~\ref{fig:images/roc_discrete_eye_err_bin.pdf} correspond to recognition performance when facial image registration is based on manually annotated eye coordinates.
Section~\ref{dutta2014automatic_discussion} describes , it will be clear that we need this result (\ie the dotted lines o

While discussing our experiment results in Section~\ref{dutta2014automatic_discussion}, we need to rule out one possible explanation for the observed results. Therefore, in~\figurename~\ref{fig:images/roc_discrete_eye_err_bin.pdf}, we also plot the recognition performance when facial images are registered using automatically detected eye coordinates.


\begin{table}[h]
 \small
 \centering
 \caption{Interval of $J_p$}
 \begin{tabular}{c|c|c|c}
 Interval & Range of $J_p$ & \# Genuine & \# Impostor \\
\hline
 $J_{1}$ & $[0.0, 0.0381)$ & $6890$ & $1588511$ \\
 $J_{2}$ & $[0.0381, 0.0495)$ & $6890$ & $1589314$ \\
 $J_{3}$ & $[0.0495, 0.0622)$ & $6890$ & $1589597$ \\
 $J_{4}$ & $[0.0622, 1)$ & $6891$ & $1585740$ \\
\hline
 \end{tabular}
\label{tbl:eye_det_err_interval}
\end{table}

\subsection{Discussion}
\label{dutta2014automatic_discussion}
In this paper, we set out to find if the proposed Automatic Eye Detection Error (AEDE) is a predictor of face recognition performance.
Image quality parameters are very strong indicators of face recognition performance.
Therefore, we first investigate if AEDE responds to controlled pose and illumination variation in facial images.

We first visually inspect the distribution of AEDE to see if it responds to the quality variations present in our data set.
In~\figurename~\ref{fig:images/eye_det_err_stat.pdf}, we show the distribution of AEDE for images in the probe set categorized according to MultiPIE camera and flash identifier.
First, for the frontal camera (05\_1), let us compare the distributions corresponding to frontal flash (07) and no-flash.
For frontal flash, the distribution of $J_p$ is nearly symmetric and centered around $J_p = 0.5$.
For no-flash, the distribution becomes right skewed (\ie right heavy tail) indicating that many samples have high eye detection error.
For other illumination variations also, we observe small increase in right skewness.
This shows that the normalized eye detection error responds to illumination variations.
Furthermore, higher values of AEDE corresponds to degrading illumination condition.
Now let us compare the distributions for different pose variations under no-flash illumination condition.
For frontal pose, the distribution of $J_p$ is already right skewed and it becomes more heavy on the right tail as we move away from the frontal pose.
This indicates that AEDE increases as the pose moves away from frontal view.
Therefore, we conclude that the proposed AEDE measure responds to, at least, pose and illumination quality variations in facial images.

\begin{figure}
\includegraphics[width=\linewidth]{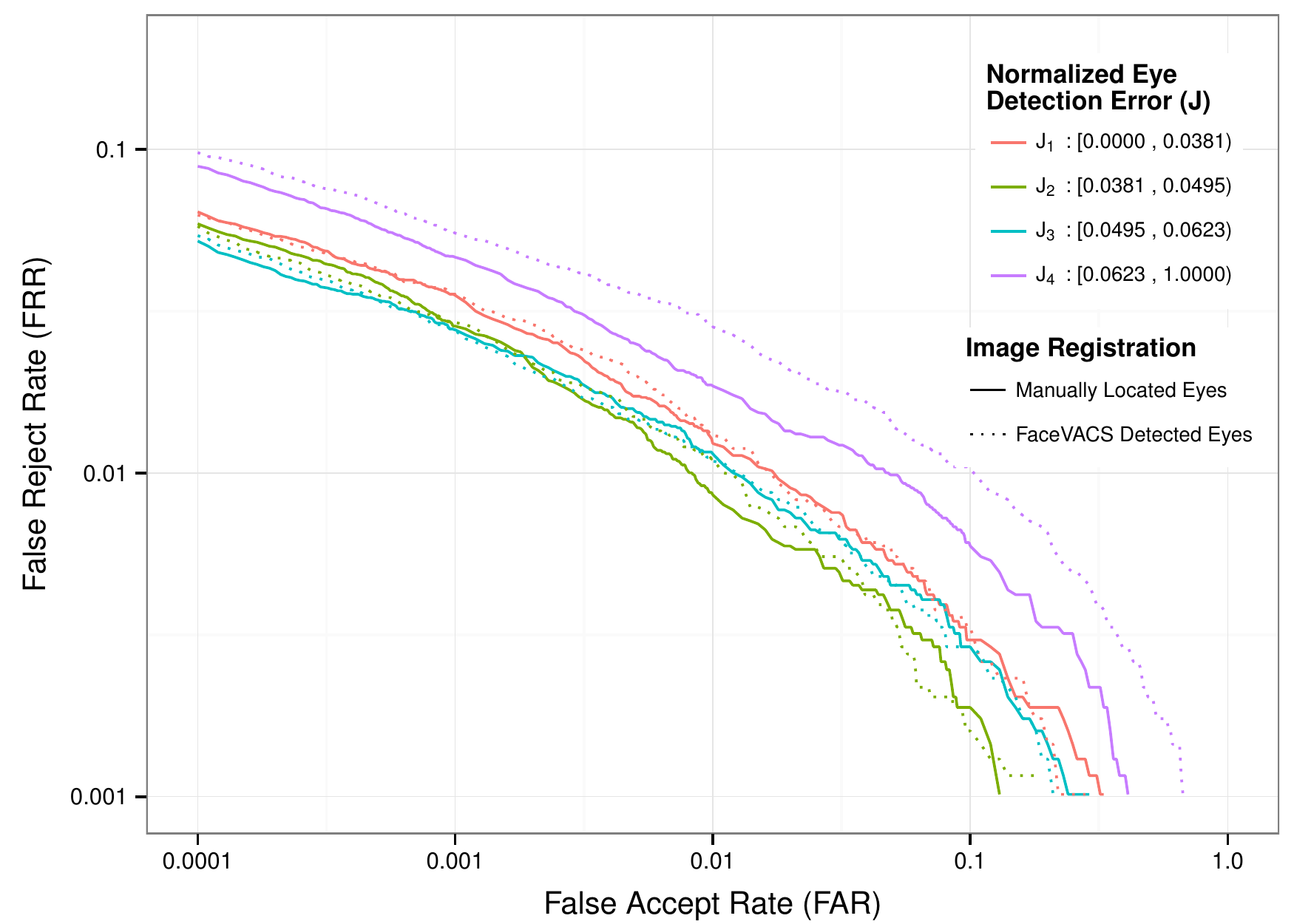}
\caption{Recognition performance variation for each monotonically increasing interval of normalized eye detection error $J$.}
\label{fig:images/roc_discrete_eye_err_bin.pdf}
\end{figure}

In~\figurename~\ref{fig:images/roc_discrete_eye_err_bin.pdf}, we show the ROC corresponding to the four intervals of the normalized eye detection error in probe image $J_p$.
First, we discuss the four ROCs (\ie solid lines) corresponding to facial images registered using manually annotated eye coordinates.
We observe that the four intervals of $J_p$ correspond to four distinct ROC curves.
However, contrary to our expectations, the four monotonically increasing intervals of $J_p$ do not correspond to monotonically degrading ROC curves.
For example, $J_1$ corresponds to the interval with lowest eye detection error but it does not correspond to the best ROC.
In fact, the interval $J_2$ and $J_3$ correspond to best recognition performance.
As expected, the largest eye detection error \ie $J_4$ correspond to the worst recognition performance.
These findings suggests that the normalized eye detection error has a non-linear relationship with face recognition performance.
Our results further support the argument that a single metric is not sufficient to capture all image quality variations that may affect face recognition performance.

One could argue that the observed non-linear relationship is due to bias in the manually annotated eye coordinates and FaceVACS would behave differently if allowed to automatically register facial images.
To check the validity of this argument, in~\figurename~\ref{fig:images/roc_discrete_eye_err_bin.pdf}, we plot the four ROCs (\ie dotted lines) corresponding to facial images automatically registered by FaceVACS using its own detected eye coordinates.
These ROCs also show the same trend and therefore this argument does not explain the non-linear relationship between eye detection error and recognition performance.
Further work is required to determine the causes of this non-linearity.

\subsection{Conclusion}
\label{sec:conclusion}
In this paper, we have proposed Automatic Eye Detection Error (AEDE) as a predictor of face recognition performance.
Our results show that AEDE has a non-linear relationship with face recognition performance and further work is required to fully understand the reasons for this non-linearity.

One of the major limitations of AEDE is that it requires manually annotated eye coordinates in order to quantify the quality of a facial image. 
For real time biometric applications, the manually annotated eye coordinates are usually not available. 
However, for forensic face recognition applications, a forensic investigator can manually annotate a small number of facial images relevant to the casework. 
Availability of such manual eye annotations can greatly help in quantifying the uncertainty in decision about identity using the proposed \textit{Automatic Eye Detection Error} (AEDE) image quality measure.

The proposed eye detection error cannot capture all types of quality variations that may affect face recognition performance.
For example, in a photograph containing facial image with closed eye, the eye detection error will be very high. 
This does not necessarily translate into a difficult verification problem.
Similarly, facial expressions like smile can greatly affect face recognition performance but may not necessarily impact the performance of an automatic eye detector.
Therefore, we need more quality parameters to fully quantify the variability in recognition performance.

\begin{figure}[t]
\setlength{\unitlength}{1pt}
\begin{center}
\includegraphics[width=\linewidth]{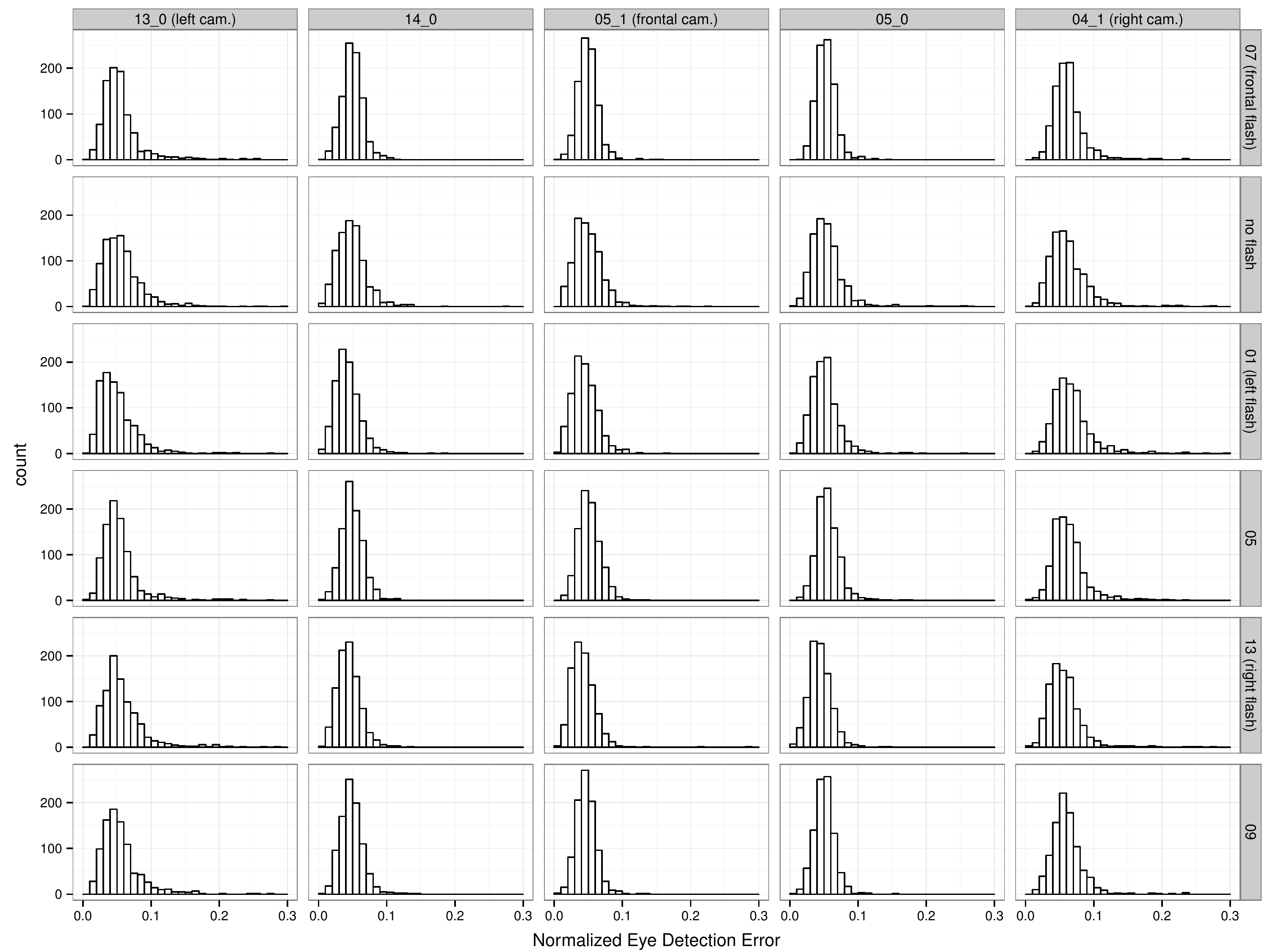}
\end{center}
\begin{picture}(1,1)
  \put(228,298){\includegraphics[width = 0.06\linewidth]{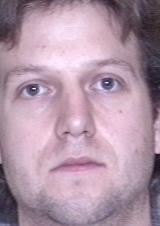}}
  \put(308,298){\includegraphics[width = 0.06\linewidth]{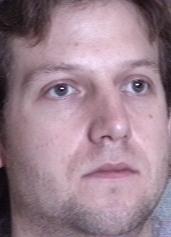}}
  \put(150,299){\includegraphics[width = 0.06\linewidth]{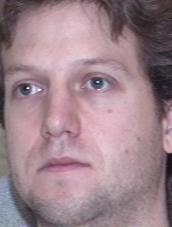}}
  \put(386,299){\includegraphics[width = 0.06\linewidth]{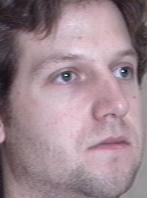}}
  \put(70,299){\includegraphics[width = 0.06\linewidth]{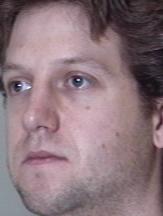}}

  %
  \put(228,246){\includegraphics[width = 0.06\linewidth]{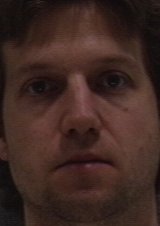}}
  \put(228,197){\includegraphics[width = 0.06\linewidth]{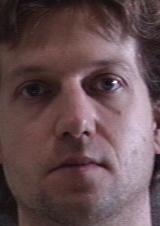}}
  \put(228,148){\includegraphics[width = 0.06\linewidth]{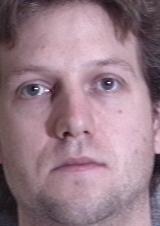}}
  \put(228,98){\includegraphics[width = 0.06\linewidth]{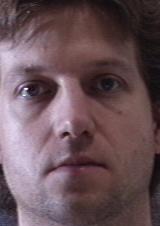}}
  \put(228,50){\includegraphics[width = 0.06\linewidth]{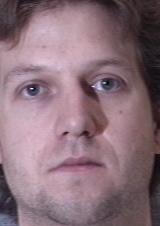}}
\end{picture}
\caption{Distribution of normalized eye detection error $J$ of probe images for different pose and illumination variations from the MultiPIE data set.}
\label{fig:images/eye_det_err_stat.pdf}
\end{figure}


\section{Conclusion}
The investigation of Section~\ref{dutta2013facial_intro} showed that the non-match score distribution is influenced by both identity (which is expected) and image quality.
In presence of image quality variations, it is difficult to discern if a low non-match score is due to a non-match identity or poor image quality.
Therefore, performance prediction features (like the Impostor-based Uniqueness Measure~\cite{klare2012face}) derived from the non-match score distribution may not be reliable in the presence of image quality degradation in the input facial images.
We therefore decide not to include features based on similarity scores for the performance prediction model of~\chaptername~\ref{dutta2015predicting_intro}.

Our analysis of Section~\ref{dutta2014automatic_intro} shows the Automatic Eye Detection Error (AEDE) to be correlated to the face recognition performance.
While this performance prediction feature is quite practical for forensic cases involving face recognition, it quite impractical to expect ground truth manual eye annotations in general biometric applications.
We therefore decide not to include the AEDE performance predictor feature for the performance prediction model of~\chaptername~\ref{dutta2015predicting_intro}.
Recall that we supplied manual eye annotations to face recognition system to ensure minimal error in facial image registration and therefore the recognition performance variations are solely caused by the influence of image quality variations on the face recognition algorithm.

In the performance prediction model of~\chaptername~\ref{dutta2015predicting_intro}, we use the following two image quality features proven to be a predictor of recognition performance: pose and illumination.

\chapter{Predicting Face Recognition Performance Using Image Quality}
\label{dutta2015predicting_intro}
A face verification system compares a pair of facial images and decides whether the image pair is a match (originating from the same individual) or non-match (originating from different individuals) based on their similarity score which is compared with a verification decision threshold.
Given that practical face recognition systems make occasional mistakes in such verification decisions, there is a need to quantify the uncertainty of decision about identity.
In other words, we are not only interested in the verification decision (match or non-match) but also in its uncertainty.

\begin{figure}
 \centering
 \includegraphics[width=0.8\linewidth]{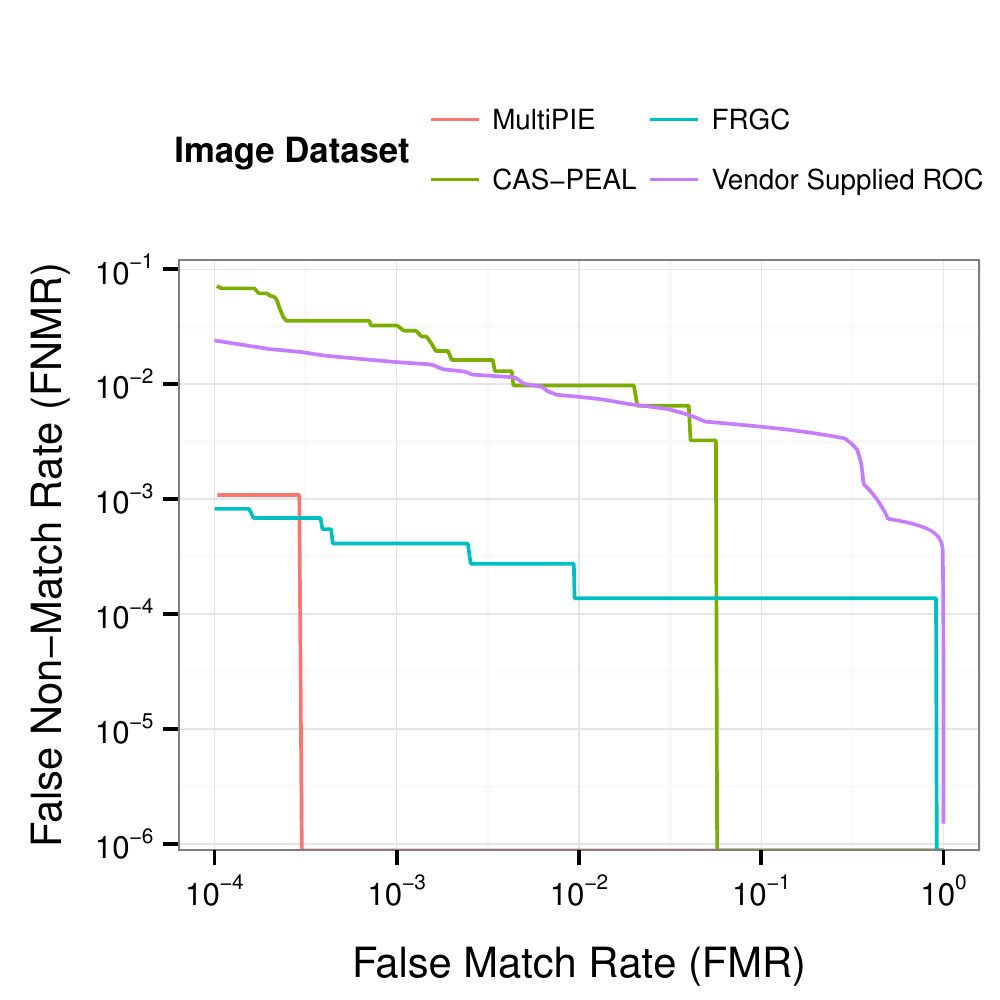}
 \caption{Vendor supplied Receiver Operating Characteristic (ROC) and actual ROC curve of a COTS face recognition system~\cite{facevacs2010} operating on frontal pose, illumination, neutral expression subset of three independent data sets (sample facial images are shown in~\figurename~\ref{fig:prb_ref_img_illus.pdf}).}
 \label{fig:fv_vendor_roc_vs_true_roc.pdf}
\end{figure}

The vendors of commercial off-the-shelf (COTS) face recognition systems provide the Receiver Operating Characteristics (ROC) curve which characterizes the uncertainty of the decision about identity at several operating points in terms of trade-off between false match and false non-match rates.
As shown in~\figurename\ref{fig:fv_vendor_roc_vs_true_roc.pdf}, the vendor supplied ROC for a COTS face recognition system~\cite{facevacs2010} differs significantly from ROCs obtained from frontal image subsets of three facial image data sets~\cite{gross2008multipie,phillips2005overview,gao2008caspeal} that were captured using different devices and under different setup.
Usually, the vendor supplied ROC represents recognition performance that the face recognition system is expected to deliver under ideal conditions.
In practice, the ideal conditions are rarely met and therefore the actual recognition performance varies as illustrated in~\figurename\ref{fig:fv_vendor_roc_vs_true_roc.pdf}.
Therefore, practical applications of verification systems cannot rely on the vendor supplied ROC curve to quantify uncertainty in decision about identity on per verification instance basis.

In this paper, we address this problem by presenting a generative model that predicts the verification performance based on image quality.
In addition to the inherent limitations of a face recognition system, the quality (like pose, illumination direction, noise, etc) of the pair of facial images used in verification process also contribute to the uncertainty in decision about identity.
For example, a verification decision made using a non-frontal image with uneven lighting entails more uncertainty than a verification decision carried out on frontal mugshots captured under studio conditions.
Therefore, in this paper, we use image quality as the feature for predicting performance of a face recognition system.
Throughout this paper, we use the term ``image quality'' to refer to all the static or dynamic characteristics of the subject or acquisition process as described in~\cite{iso_iec_29794-5:2010}, including for instance facial pose, illumination direction, \etc.

A large and growing body of literature has investigated the use of similarity scores (\ie classifier's output) as a feature for performance prediction.
However, there is evidence that non-match scores are influenced by both by identity and by the quality of image pair~\cite{dutta2013facial}.
Therefore, it is not possible to discern if a low non-match score is due to non-match identity or poor quality of image pair.
Hence, we avoid using similarity scores as a performance feature in the proposed solution.
This design decision not only avoids the issues associated with using similarity score as a feature but also allows our model to predict performance even before the actual facial comparison has taken place.

A substantial amount of literature has tried to model the similarity score distribution (match and non-match) conditioned upon image quality in order to predict performance\cite{shi2008modeling,wein2005using,scheirer2011meta}.
Since the parameter of interest (\ie similarity score) is a uni-variate variable, the model is much simpler and any recognition performance measure can be derived from these models of score distributions.
In practice, we rarely need to know about the underlying score distributions and are mostly interested in the recognition performance that can be expected from a particular face recognition system operating on a given image pair.
Therefore, in this paper, we take a more practical approach of directly modeling the recognition performance measure (\eg False Non-Match Rate - FNMR and False Match Rate - FMR at a certain point of operation) of interest rather than modeling intermediate variable (\ie similarity score).
The proposed model is flexible to accommodate any type of recognition performance measure that is of interest to the user like Area Under ROC (AUC), Equal Error Rate (EER), calibrated log-likelihood-ratios, \etc.

There are many applications of models that can predict the performance of a face recognition system.
In forensic cases involving face recognition, it can rank CCTV footage frames based on the image quality of each frame thereby allowing forensic investigators to focus their effort on a smaller set of images with higher evidential value.
When capturing facial images for the reference set (\ie enrollment or gallery set), it can alert the operator whenever a ``poor'' quality image sneaks into the enrollment set.
Such a model can be used to dynamically set the verification decision threshold that adapts according to the sample quality, for instance to maintain a prescribed False Match Rate (FMR).
The tolerance of face recognition algorithms to image quality degradation varies and therefore results from multiple algorithms can be fused based on the predicted performance corresponding to individual face recognition algorithm.

The method we present is based on modeling the relationship between image quality and face recognition performance using a probability density function.
During the training phase, this density function is approximated by evaluating the recognition performance corresponding to the quality variations encountered in practice -- a data driven approach.
A model of this density function learned during the training phase allows us to predict the performance of a face recognition system on previously unseen facial images even before the actual verification has taken place.

This paper is organized as follows:
We review some of the existing literature on performance prediction in Section~\ref{dutta2015predicting_related_work}.
Section~\ref{dutta2015predicting_model_description} describes the proposed generative model which uses image quality features to predict performance of a face recognition system.
In Section~\ref{dutta2015predicting_exp}, we present the result of model training and performance prediction on three independent data sets for six face recognition systems.
The key observations from these experiments are discussed in Section~\ref{dutta2015predicting_discussion} followed by final conclusions in Section~\ref{dutta2015predicting_conclusion}.

\section{Related Work}
\label{dutta2015predicting_related_work}

\begin{figure}
\centering
\includegraphics[width=\linewidth]{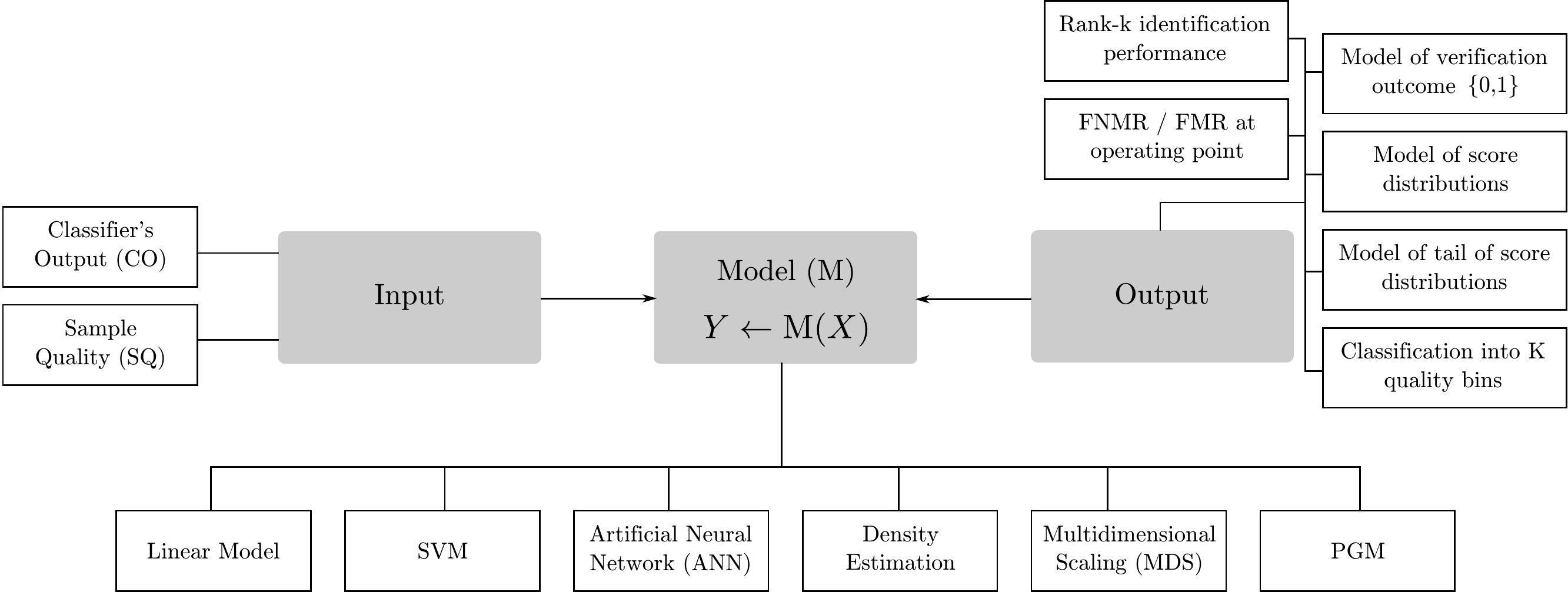}
\caption{Typical components of a system aiming to predict the performance of a biometric system.}
\label{fig:biometric_perf_pred_literature_map.pdf}
\end{figure}

Systems aiming to predict the recognition performance are characterised by three components: \textit{Input} denotes the features with performance predictive capability; \textit{Output} denotes the  recognition performance measure of interest; and \textit{Model} corresponds to a model that represents the functional relationship between Input and Output.
The existing publications on performance prediction differ in the variants of Input, Model and Output as listed in~\tablename~\ref{tbl:literature_classification}. In~\figurename\ref{fig:biometric_perf_pred_literature_map.pdf}, we show all the variants of these components that we found during literature review.

In this paper, we classify the existing literature into two groups based on the type of feature (\ie Input) used for performance prediction.
The first group of performance prediction systems use output of the classifier (CO) itself as a feature for performance prediction while the second group uses biometric sample quality (SQ).

\begin{table}[h]
\renewcommand{\arraystretch}{1.3}
\caption{Classification of existing literature on performance prediction based on the three typical components of such systems as shown~\figurename\ref{fig:biometric_perf_pred_literature_map.pdf}.}
\label{tbl:literature_classification}
\centering
\footnotesize
\begin{tabular}{p{2.5cm} | p{3.0cm} | p{2.8cm} | p{3.8cm}}
\hline
Paper & Input & Output & Model \\
\hline
\cite{li2005predicting} & Classifier's Output & Rank-k & SVM \\
\cite{wang2007modeling} & Classifier's Output & Rank-k & SVM \\
\cite{scheirer2011meta} & Classifier's Output & Model of tail & Density Est. (Wbl) \\
\cite{shi2008modeling} &  Classifier's Output & Model of tail & Density Est. (GPD) \\
\cite{klare2012face} & Classifier's Output & Score dist. model & Density Est. (KDE) \\
\cite{ozay2009improving} & Classifier's Output & Verific. outcome & PGM \\
\hline
\cite{beveridge2008focus},~\cite{beveridge2010quantifying} & Sample Quality & FNMR/FMR & Linear Model (GLMM) \\
\cite{dutta2014bayesian} & Sample Quality & FNMR/FMR & Density Est. (GMM) \\
\textbf{our work} & \textbf{Sample Quality} & \textbf{FNMR/FMR} & \textbf{Density Est. (GMM)} \\
\cite{wein2005using} & Sample Quality & Score dist. model & Density Est. ($\gamma$, $\textnormal{ln} \; \mathcal{N}$) \\
\cite{aggarwal2011predicting} & Sample Quality & Score dist. model & MDS \\
\cite{aggarwal2012predicting} & Sample Quality & Quality bins & Linear Model (PLS) \\
\cite{tabassi2004fingerprint},~\cite{tabassi2005novel} & Sample Quality & Quality bins & ANN \\
\hline
\cite{jammalamadaka2012algorithm}* & CO \& SQ & pred. pose err. & SVM \\
\cite{zuo2010adaptive}* & CO \& SQ & Verific. outcome & ANN \\

\hline
\end{tabular}
\vspace{0.2cm}

* denotes work in domains other than face and fingerprint biometrics \\
\end{table}

The key observation underpinning the first group of existing literature is that the overlapping region between match and non-match score distribution entail more uncertainty in decision about identity.
They begin by creating features from classifier's output (\ie similarity score) that are predictive of recognition performance.
For example,~\cite{li2005predicting} create a set of three features based on similarity score while~\cite{wang2007modeling} uses similarity score based features to quantify the intrinsic factors (properties of algorithm, reference set, \etc) and extrinsic factors (properties of probe set).
Rather than considering the full set of similarity scores, \cite{scheirer2011meta}~and~\cite{shi2008modeling} argue that decision about identity is more uncertain in the overlapping region of the match and non-match distributions and therefore they consider the similarity scores only in the tail region.
The authors of~\cite{ozay2009improving} use the distance of a similarity score from the non-match distribution in units of standard deviation (\ie d-prime value~\cite{jain2007handbook}) while \cite{klare2012face} use the facial uniqueness feature derived from the nature of subject specific non-match distribution as the performance predictor feature.
A major limitation of using features derived from similarity scores is that they become unstable under quality variations~\cite{dutta2013facial} because similarity score is influenced by both identity and quality of the image pair under consideration.
Therefore, with features derived from similarity scores, it is not possible to discern whether a low similarity score is caused by poor sample quality or a non-match pair.

The second group of existing literature's is based on the observation that sample quality influences the uncertainty in decision about identity -- empirical evidence show that poorer sample quality entails more uncertainty in decision about identity.
They begin by externally assessing image quality of probe/reference samples using an Image Quality Assessor (IQA).
For instance,~\cite{tabassi2004fingerprint} and \cite{tabassi2005novel} use fingerprint image quality like clarity of ridges and valleys, number and quality of minutiae, size of image, \etc while~\cite{wein2005using} use fingerprint quality assessments from a propriety IQA as image quality features.
The authors of~\cite{aggarwal2012predicting} use image-specific (like image sharpness, image hue content, image saturation, \etc) and face-specific (like expression) characteristics as image quality features.
A single image quality feature that characterizes the nature of illumination in a facial image was used in~\cite{aggarwal2012predicting}.
Using the term co-variate to denote image quality, \cite{beveridge2008focus}~and~\cite{beveridge2010quantifying} use a wide range of subject co-variates like age, gender, race, wearing glasses and image co-variates like focus, resolution, head tilt as the features for performance prediction.
A major limitation of using image quality as a performance prediction feature is that there are overwhelmingly large number of quality factors that may influence the performance of a face recognition system -- their exact count is still unknown.
Furthermore, accurate measurement of image quality is still an unsolved problem and concerted efforts (like NFIQ2~\cite{nfiq2}) are underway to develop an extensive set of quality feature and to standardize the use and exchange of quality measurements.
The authors of~\cite{phillips2013existence} have proposed the Greedy Pruned Ordering (GPO) scheme to determine the best case upper bound performance prediction capability that can be achieved by any quality measure on a \textit{particular combination of algorithm and data set}.

Some existing works like~\cite{jammalamadaka2012algorithm} and \cite{zuo2010adaptive} belong to both the first and second group because they combine both classifier's output (CO) and image quality features (SQ) to predict performance.

The choice of recognition performance measure (\ie Output) is based on user requirements.
For instance, Rank-k recognition rate and FNMR/FMR at the operating point are the recognition performance measure used for modeling identification and verification performance respectively.
Authors choosing to model the similarity score distribution do not need to define the recognition performance measure because any performance measure can be derived from the model of similarity score distribution.
Some authors model discrete quality bins with distinctly different recognition performance as the output.
In this paper, we model the following recognition performance measure: FNMR and FMR at a particular operating point defined by a decision threshold.
Some existing works like~\cite{ozay2009improving} and \cite{zuo2010adaptive} have tried to directly predict the success/failure of the verification outcome which according to~\cite{phillips2009introduction} is a pursuit equivalent to finding a perfect verification system.

Once the performance predictor feature (\ie Input) and the desired recognition performance measure (\ie Output) is fixed, the final step is to use an appropriate model to learn the relationship between predictor features and recognition performance.
So far, many variants of learning algorithms has been applied to learn the relationship between performance predictor features and the recognition performance measure.
For instance, \cite{wang2007modeling} and \cite{li2005predicting} use Support Vector Machine (SVM) to model this relationship while \cite{tabassi2004fingerprint} and \cite{tabassi2005novel} use the Artificial Neural Network (ANN) to learn the relationship between fingerprint sample quality features and the normalized similarity score -- the distance of match score from non-match score distribution.
The authors aiming to model similarity score distributions conditioned on image quality either use a standard parametric distribution like Weibull~\cite{scheirer2011meta}, General Pareto Distribution (GPD)~\cite{shi2008modeling}, gamma/log-normal distributions~\cite{wein2005using} or use Kernel Density Estimation (KDE)~\cite{klare2012face} when the score distribution cannot be explained by standard parametric distributions.
The authors of~\cite{aggarwal2011predicting} apply Multi-Dimensional Scaling (MDS) to model the relationship between quality features and match score distribution while in~\cite{aggarwal2012predicting}, the authors use regression to model the relationship between quality partition (good, bad and ugly) and image quality features.

In this paper, we extend the work of~\cite{dutta2014bayesian} in many fronts.
We address the issue of limited training data set by using a probabilistic model of quality and recognition performance in small regions of the quality space.
We also report the accuracy of predicted performance on three independent facial image data sets for six face recognition systems.
We use the conditional expectation, instead of maximum a posteriori probability (MAP), to estimate the recognition performance given the image quality.
We also use a simulated IQA to demonstrate the recognition performance achievable by an accurate and unbiased IQA.
Furthermore, our work most closely relates to the work of~\cite{beveridge2008focus} which uses a Generalized Linear Mixed Model (GLMM) to model the relationship between image quality (like focus, head tilt, \etc) and the False Non-Match Rate (FNMR) at a given False Match Rate (FMR).
Their analysis focused on investigating the impact of each quality metric on recognition performance.
Our work focuses on predicting performance for a given sample quality.
We directly model the relationship between image quality and recognition performance (FNMR and FMR at the operating point) using a probability density function.

\section{Model of Image Quality and Recognition Performance}
\label{dutta2015predicting_model_description}
Let $\mathbf{q}^{p} \in \mathbb{R}^{m}$ and $\mathbf{q}^{g} \in \mathbb{R}^{m}$ be the vectors denoting the image quality features (like pose, illumination direction, noise, \etc) of a probe and reference image pair respectively as assessed by an Image Quality Assessment (IQA) system.
We coalesce $\mathbf{q}^{p}$ and $\mathbf{q}^{g}$ to form a single quality feature vector $\mathbf{q}=[\mathbf{q}^{p}; \mathbf{q}^{g}] \in \mathbb{R}^{2m}$ which denotes the image quality features of probe and reference image pair.
We consider the quality features of image pair in the proposed model because in~\cite{beveridge2011when} it has been shown that face recognition performance is a function of image quality pair and not an individual image.
For a particular face recognition system, let $\mathbf{r} \in \mathbb{R}^{n}$ denote the face recognition performance corresponding to a sufficiently large set of different probe (or, query) and reference (or, enrollment) image pairs having same image quality $\mathbf{q}$.
The prosed model is flexible to accommodate any recognition performance parameter of interest to the user in vector $\mathbf{r}$.
For instance, $r_{1}$ and $r_{2}$ would correspond to FMR and FNMR respectively if we want to model and predict the FMR and FNMR at an operating point.
The vector $\mathbf{r}$ can be expanded to accommodate FMR and FNMR at several other operating points if we want to model and predict the full Receiver Operating Characteristics (ROC) curve of a face recognition system.
Other recognition performance measures like Area Under Curve (AUC), Equal Error Rate (EER) \etc can fit equally well in vector $\mathbf{r}$.

Here, we assume that vector $\mathbf{q}$ is sufficient to capture all the relevant quality variations possible in a facial image pair that have an influence on face recognition performance.
Different face recognition systems have varying level of tolerance to image quality degradations and therefore vector $\mathbf{r}$ is a function of a particular face recognition system.

\begin{figure}
 \centering
 \includegraphics[width=0.8\linewidth]{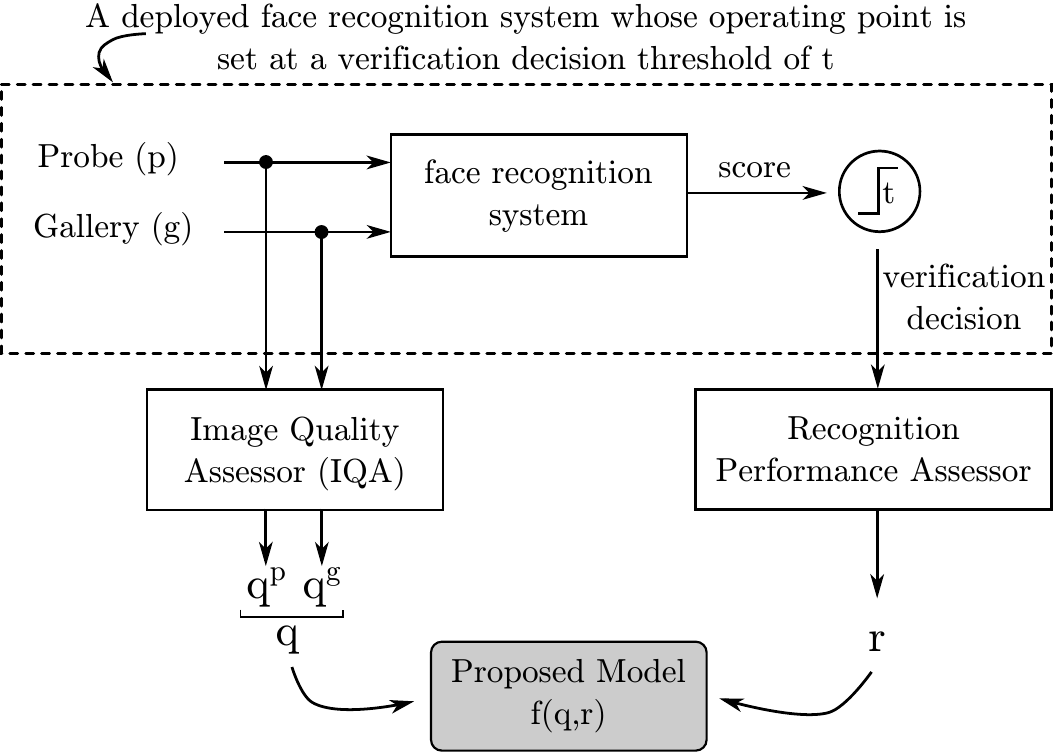}
 \caption{The proposed performance prediction model treats a face recognition system as a ``black box'' and captures the relationship between image quality features $\mathbf{q}$ and recognition performance measures $\mathbf{r}$ using a probability density function $f(\mathbf{q},\mathbf{r})$.}
 \label{fig:black_box_qr_model_illus}
\end{figure}

To model the interaction between image quality features $\mathbf{q}$ and recognition performance $\mathbf{r}$, we coalesce $\mathbf{q}$ and $\mathbf{r}$ and form the Quality-Performance (QR) space.
We model this QR space using a Probability Density Function (PDF) $f(\mathbf{q},\mathbf{r})$ as depicted in~\figurename\ref{fig:black_box_qr_model_illus}.
This PDF defines the probability of observing certain combination of image quality $\mathbf{q}$ and recognition performance $\mathbf{r}$.
Given the quality $\mathbf{q}$ of previously unseen verification instance, we can apply the Bayes' theorem to obtain the posterior distribution of recognition performance $\mathbf{r}$ as follows:
\begin{equation}
f(\mathbf{r}|\mathbf{q}) = \frac{f(\mathbf{q},\mathbf{r})}{f(\mathbf{q})}.
\label{eq:P_r_given_q}
\end{equation}
The conditional expectation of $\mathbf{r}$ with respect to the conditional probability distribution of~\eqref{eq:P_r_given_q} is:
\begin{equation}
\mathbb{E}(\mathbf{r}|\mathbf{q}) = \int \mathbf{r} f(\mathbf{r}|\mathbf{q}) dr,
\label{eq:expectation_r}
\end{equation}
where, $\mathbb{E}(\mathbf{r}|\mathbf{q})$ denotes the expected value of recognition performance for a given image quality pair $\mathbf{q}$.
In this paper, we use $\mathbb{E}(\mathbf{r}|\mathbf{q})$ as an estimate of recognition performance $\mathbf{r}$ given quality features $\mathbf{q}$ of probe and reference image pair.

\subsection{Model Training : Estimating $f(\mathbf{q},\mathbf{r})$ from data}
\label{model_training}
In this paper, we model the Probability Density Function (PDF) $f(\mathbf{q},\mathbf{r})$ using a mixture of $K$ multivariate Gaussian (MOG)
\begin{equation}
f(\mathbf{q},\mathbf{r}) = \sum_{k=1}^{K} \pi_k f_{k}(\mathbf{q},\mathbf{r}),
\label{eq:P_q_r_MOG}
\end{equation}
where, $f_{k}(\mathbf{q},\mathbf{r})=\mathcal{N}([\mathbf{q},\mathbf{r}]; \mu_k, \Sigma_k)$ denotes the $k^{\textnormal{th}}$ Gaussian mixture component with mean $\mu_k$ and covariance $\Sigma_k$ and $\pi_k$ are the mixture coefficients such that $0~\leq~\pi_k~\leq~1$, $\sum_{k}~\pi_k~=~1$.

To learn the parameters of MOG in~\eqref{eq:P_q_r_MOG}, we require a training data set $\mathcal{D}_{\textnormal{train}}=\{ [\mathbf{q}_{i}, \mathbf{r}_{i}] \}$ where each $\mathbf{q}_{i}$ denotes a sample point in quality space and $\mathbf{r}_{i}$ is the corresponding recognition performance.
Such a training data set can be created only if we have sufficiently large number of similarity scores (both match and non-match) at each sampling point $\mathbf{q}$ in the quality space.
In other words, for each point of interest in quality space $\mathbf{q}$, we require a training data set with a large number of unique verification attempts having probe and reference image quality $\mathbf{q}$.
In practice, it is very difficult to obtain such a training data set.
Due to limited nature of practical training data, we cannot reliably evaluate recognition performance at discrete points in the quality space.
Therefore, we build probabilistic models of quality and performance in small regions of the quality space as described in~Section~\ref{prob_model_q_r}.
We randomly sample from these models of $\mathbf{q}$ and $\mathbf{r}$ to build the training data set $\tilde{\mathcal{D}}_{\textnormal{train}} = \{ [\tilde{\mathbf{q}_{i}}, \tilde{\mathbf{r}_{i}}] \}$.
This strategy of building the training data set allows us to capture the uncertainty in quality and performance measurements entailed by IQA and limited training data set respectively.

\begin{figure}
 \centering
 \includegraphics[width=0.5\linewidth]{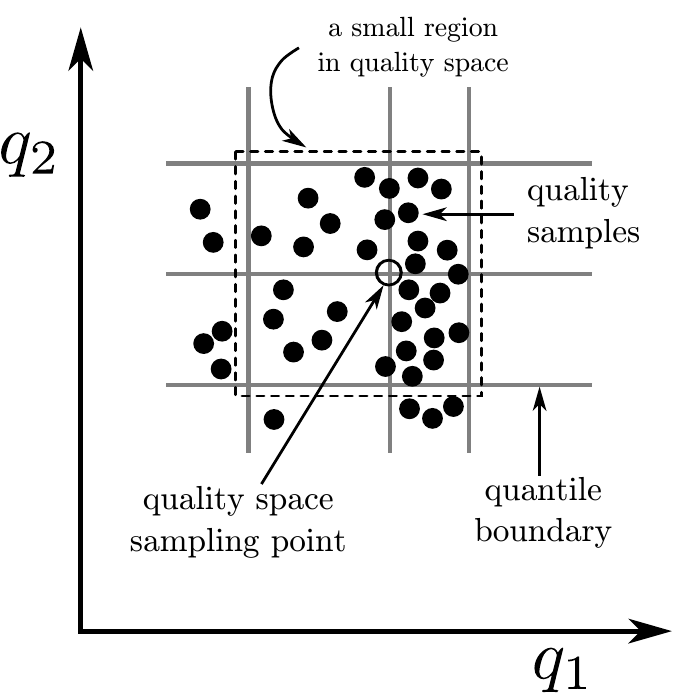}
  \caption{Region formation in the quality space.}
\label{fig:qspace_sampling_region_illus.pdf}
\end{figure}

The size and location of small regions in the quality space is determined by the nature of training data which commonly has densely populated samples in the regions of quality space corresponding to most common types of quality variations and sparse samples in other regions.
We define $N_{\textnormal{qs}}$ quantile points (along each quality axis) based on quantiles of evenly space probabilities.
Unique sampling points are formed in the quality space by taking all the possible combination of these $N_{\textnormal{qs}}$ quantile points along each quality axis.
We form regions around these quality space sampling points such that the adjacent quantile points define the boundary of these overlapping region as shown in~\figurename\ref{fig:qspace_sampling_region_illus.pdf}.

The raw training data is composed of unique verification attempts and therefore each record of the training data set contains: similarity score $s$, quality of probe and reference images $\mathbf{q}$ and the ground truth (match or non-match) of the verification decision.
Let $\mathbf{Q}(\mathbf{q})$ and $\mathbf{S}(\mathbf{q})$ denote the set of quality samples $\mathbf{q}$ and corresponding similarity scores $s$ present in a quality space region formed around a quality space sampling point~\figurename\ref{fig:qspace_sampling_region_illus.pdf}.
As described in Section~\ref{prob_model_q_r}, we build a probabilistic model of $\mathbf{q}$ and $\mathbf{r}$ in each quality space region and randomly sample $N_{\textnormal{rand}}$ samples from these models to create the data set $\tilde{\mathcal{D}}_{\textnormal{train}} = \{ [\tilde{\mathbf{q}}_{i}, \tilde{\mathbf{r}}_{i}] \}$ needed to train the model of (\ref{eq:P_q_r_MOG}).
We pool the QR training data from each quality space region and apply the Expectation Maximization (EM) algorithm~\cite{fraley2012mclust} to learn the parameters $(\pi_{k}, \mu_{k}, \Sigma_{k})$ of Mixture of Gaussian in~\eqref{eq:P_q_r_MOG}.

\subsubsection{Probabilistic Model of quality $\mathbf{q}$ and performance $\mathbf{r}$}
\label{prob_model_q_r}
We now describe the probabilistic model of $\mathbf{q}$ and $\mathbf{r}$ in the quality space region containing quality samples $\mathbf{Q}(\mathbf{q})$ and similarity score samples $\mathbf{S}(\mathbf{q})$.
The random samples from these probabilistic models form the training data set used to learn the model parameters of~\eqref{eq:P_q_r_MOG}.

We assume that the elements of $\mathbf{q}$ are mutually independent and they follow a Gaussian distribution within the quality region.
Based on this assumption, we fit a multivariate Gaussian $\mathcal{N}(\mathbf{q}; \mu_{i}, \Sigma_{i}^{\textnormal{diag}})$ with diagonal covariance matrix parametrization to all quality samples in $\mathbf{Q}$.

Now, we describe a probabilistic model for recognition performance in the quality region.
First, we define the recognition performance measure used throughout this paper.
For face recognition systems deployed in real-world, the operating point is set to achieve certain False Match Rate (FMR) or False Non-Match Rate (FNMR).
In this paper, we assume that the recognition performance $\mathbf{r}$ of interest is the FMR and FNMR at certain decision threshold $t$ (which defines the operating point of a face recognition system): $\mathbf{r} = [\textnormal{FMR}_{t}, \textnormal{FNMR}_{t}]$.

Given an observation of the number of similarity scores below (and above) the decision threshold $t$, we want to know the nature of the distribution of FNMR (and FMR).
First, we consider the set of match scores $\mathbf{M}(\mathbf{q}) \subset \mathbf{S}(\mathbf{q})$ to build a model of FNMR distribution.
Each element in $\mathbf{M}(\mathbf{q})$ is a similarity score corresponding to a pair of probe and reference image containing facial images of same subject (\ie match pair).
For all the elements in $\mathbf{M}(\mathbf{q})$, we can make a verification decision $w \in \{0,1\}$ based on the decision threshold $t$ as follows
\begin{equation}
w_{j} = 
\begin{cases}
1 & \text{if} \; M(\mathbf{q})[j] < t, \\
0 & \text{otherwise},
\end{cases}
\label{eq:bernouli_trial_expression}
\end{equation}
where, $M(\mathbf{q})[j]$ corresponds to the $j^{\textnormal{th}}$ similarity score in set $\mathbf{M}(\mathbf{q})$ and $w_{j}=1$ is used to denote failure in verification of a match pair (\ie False Non-Match) while $w_{j}=0$ denotes success in verification of a math pair.
Therefore, each verification decision $w_{j}$ can be thought of as the outcome of a Bernoulli trial where success and failure corresponds to $w_{j}=1$ and $w_{j}=0$ respectively.
Let $\mathbf{w}=\{w_{j}\}$ denote a Binomial experiment containing a set of $N=|\mathbf{M}(\mathbf{q})|$ statistically independent Bernoulli trials such that $f(w_{j}=1|N, \tau)= \tau$ where $\tau$ is the probability of failure in verification of a match pair which is also called the False Non-Match Rate (FNMR).
Furthermore, let $m$ be a random variable indicating the number of $w_{j}=1$ (\ie success) in the Binomial experiment.
The value of False Non-Match Rate (FNMR) is given by:
\begin{equation}
\textnormal{FNMR} = \frac{m}{N}.
\end{equation}
We are interested in the distribution of FNMR which in turn depends on the distribution of random variable $m$.
The probability of getting $m$ success in $N$ trials follows a Binomial distribution defined as follows
\begin{equation}
\textnormal{Bin}(m|N,\tau) = \binom{N}{m} \tau^{m} (1-\tau)^{N-m},
\label{eq:binomial_likelihood}
\end{equation}
where, $\tau$ denotes the probability of getting success in a Bernoulli trial (\ie FNMR).
Taking a Bayesian perspective on the problem of estimating distribution of $\tau$, we first define the prior distribution over $\tau$.
Since, (\ref{eq:binomial_likelihood}) belongs to the exponential family, we chose the Beta distribution $\textnormal{Beta}(\tau|a,b)$ as the prior for $\tau$ where $a,b$ denote the shape parameters of the Beta distribution.
Based on the property of conjugate priors~\cite[p.70]{bishop2006pattern}, the posterior distribution, which is also a Beta distribution, is
\begin{equation}
f(\tau|m,l,a,b) = \textnormal{Beta}(m+a, l+b) \\
\label{eq:tau_density}
\end{equation}
where, $l=N-m$ denotes the number failures in $N$ Bernoulli trial.
This shows that the underlying uncertainty in FNMR is given by a Beta distribution.
In a similar way, we can show that the uncertainty in FMR is also given by a Beta distribution.

In order to create the training data set $\tilde{\mathcal{D}}_{\textnormal{train}} = \{ [\tilde{\mathbf{q}}_{i}, \tilde{\mathbf{r}}_{i}] \}$, we draw $N_{\textnormal{rand}}$ random samples independently from the multivariate Gaussian distribution model of $\mathbf{q}$ and Beta distributions corresponding to FMR and FNMR.

Since we do not have any prior knowledge about the distribution of FMR and FNMR in a quality region, we assume a uniform prior \ie $\textnormal{Beta}(1,1)$.
Furthermore, since FMR and FNMR values follow a Beta distribution, the recognition performance measure $r_{i}$ has a Bayesian credible interval $(c, d)$ of size $1-\alpha$ such that
\begin{equation}
\int_{c}^{d} \textnormal{Beta}(r; a,b) \; dr = 1 - \alpha
\label{eq:credible_interval}
\end{equation}

\subsection{Performance Prediction}
Given a previously unseen probe and reference image pair with quality $\mathbf{q}$, we now derive an expression for the posterior distribution of recognition performance $f(\mathbf{r}|\mathbf{q})$.

From training, we have a model $f(\mathbf{q},\mathbf{r})$ defined as 
\begin{equation}
\begin{aligned}
f(q,r) &= \sum_{k} \pi_k \mathcal{N}([\mathbf{q},\mathbf{r}]; \mu_k, \Sigma_k) \\
 &= \sum_{k} \pi_k f_{k}(\mathbf{q},\mathbf{r}).
\end{aligned}
\label{eq:f_q_r}
\end{equation}

The marginal distribution $f(\mathbf{q})$ is given by
\begin{align}
f(\mathbf{q}) &= \int_{r} f(\mathbf{q},\mathbf{r}) \; dr \nonumber \\
 &= \int_{r} \sum_{k} \pi_k f_{k}(\mathbf{q},\mathbf{r}) \; dr \qquad \textnormal{from~\eqref{eq:f_q_r}} \nonumber \\
 &= \sum_{k} \pi_k \int_{r} f_{k}(\mathbf{q},\mathbf{r}) \; dr \qquad \textnormal{since $\pi_k f_{k}(\mathbf{q},\mathbf{r}) \geq 0$} \nonumber \\
 &= \sum_{k} \pi_k f_{k}(\mathbf{q})
\label{eq:f_q}
\end{align}

For a given quality $\mathbf{q}$, the conditional distribution of $\mathbf{r}$ is obtained by applying the Bayes' theorem as follows
\begin{align}
f(\mathbf{r}|\mathbf{q}) = \frac{f(\mathbf{r},\mathbf{q})}{f(\mathbf{q})}
\label{eq:f_r_given_q1}
\end{align}
Substituting~\eqref{eq:f_q_r}~and~\eqref{eq:f_q}~in~\eqref{eq:f_r_given_q1},
\begin{align}
f(\mathbf{r}|\mathbf{q}) &= \frac{\sum_{} \pi_k f_{k}(\mathbf{q},\mathbf{r})}{\sum_{} \pi_k f_{k}(\mathbf{q})}
\label{eq:f_r_given_q2}
\end{align}
Applying the Bayes's theorem to $f_{k}(\mathbf{q},\mathbf{r})=f_{k}(\mathbf{r}|\mathbf{q}) f_{k}(\mathbf{q})$ in~\eqref{eq:f_r_given_q2}, the posterior distribution of $\mathbf{r}$ for the given quality $\mathbf{q}$ is
\begin{align}
f(\mathbf{r}|\mathbf{q}) &= \frac{\sum_{k} \pi_k f_{k}(\mathbf{r}|\mathbf{q}) f_{k}(\mathbf{q})}{\sum_{k} \pi_k f_{k}(\mathbf{q})} \nonumber \\
 &= \sum_{k} f_{k}(\mathbf{r}|\mathbf{q}) \left( \frac{\pi_k f_{k}(\mathbf{q})}{\sum_{k} \pi_k f_{k}(\mathbf{q})} \right) \nonumber \\
 &= \sum_{k} \psi_{k} f_{k}(\mathbf{r}|\mathbf{q})
\label{eq:f_r_given_q3}
\end{align}

where, $\psi_{k}$ denotes the new weights for conditional mixture of Gaussian.
The conditional and marginal distribution of each mixture component is given by \cite{petersen2008matrix}:
\begin{equation*}
\begin{aligned}
f_{k}(\mathbf{r}|\mathbf{q}) &= \mathcal{N}([\mathbf{q},\mathbf{r}]; \hat{\mu}_{k,r}, \hat{\Sigma}_{k,r}) \\
f_{k}(\mathbf{q}) &= \mathcal{N}([\mathbf{q}]; \mu_{k,q}, \Sigma_{k,q})
\end{aligned}
\end{equation*}
where, 
\begin{equation*}
\begin{aligned}
\hat{\mu}_{k,r} &= \mu_{k,r} + \Sigma_{k,c}^{T} \Sigma_{k,q}^{-1} (\mathbf{q} - \mu_{k,q}) \\
\hat{\Sigma}_{k,r} &= \Sigma_{k,r} - \Sigma_{k,c}^{T} \Sigma_{k,q}^{-1} \Sigma_{k,c} \\
\mu_{k} &= \begin{bmatrix} \mu_{k,q} \\ \mu_{k,r} \end{bmatrix} \\
\Sigma_{k} &= \begin{bmatrix} \Sigma_{k,q} & \Sigma_{k,c} \\ \Sigma_{k,c}^{T} & \Sigma_{k,r} \end{bmatrix}
\end{aligned}
\end{equation*}

Using~\eqref{eq:expectation_r}, the estimate of recognition performance $\mathbf{r}$ is given by the conditional expectation as follows:
\begin{align}
\mathbb{E}(\mathbf{r}|\mathbf{q}) &= \sum_{k} \psi_{k} \mathbb{E} ( f_{k}(\mathbf{r}|\mathbf{q}) ) \qquad \textnormal{from~(\ref{eq:f_r_given_q3})} \nonumber \\
 &= \sum_k \psi_{k} \; \hat{\mu}_{k,r}
\label{eq:expectation_r_given_q0}
\end{align}

In words, estimate of the predicted recognition performance is equal to the weighed sum of conditional mean of each conditional mixture component.

\subsection{Model Parameter Selection}
\label{model_param_selection}
The proposed Gaussian Mixture model of~(\ref{eq:P_q_r_MOG}), requires selection of the following two parameters:
\begin{inparaenum}[\itshape a\upshape)]
\item Number of mixture components $K$,
\item Parametrization $\Sigma^{p}$ of the covariance matrix $\Sigma$.
\end{inparaenum}
We require parametrization of the covariance matrix because we lack sufficient training data to estimate the full covariance matrix.
Let $\theta = [K, \Sigma^{p}]$ denote the parameter for our optimization, where
\begin{align}
\Sigma^{p} & \in \{ \text{EII}, \text{VII}, \text{EEI}, \text{VEI}, \text{EVI}, \text{VVI}, \text{EEE}, \text{EEV}, \text{VEV}, \text{VVV} \} \nonumber \\
K & \in [ \textnormal{k}_{min}, \textnormal{k}_{max} ] \nonumber
\end{align}
and, the search space for $\Sigma^{p}$ is based on the covariance matrix parametrization scheme presented in~\cite{fraley2012mclust}.
In this parametrization scheme, the three characters denote volume, shape and orientation respectively.
Furthermore, E denotes equal, V means varying across mixture components and I refers to identity matrix in specifying shape or orientation.
For example, EVI corresponds to a model in which all mixture components have equal (E) volume, the shapes of mixture components may vary (V) and the orientation is the identity (I).

We select the optimal model parameter based on the Bayesian Information Criterion (BIC)~\cite{fraley2012mclust} defined as:
\begin{equation}
\textnormal{BIC} \equiv 2 \; \textnormal{ln} \; f(q,r|\theta) - n \; \textnormal{ln} N,
\label{eq:bic}
\end{equation}
where, $\textnormal{ln} f(q,r|\theta_{i})$ denotes the log-likelihood of data under model~\eqref{eq:P_q_r_MOG} with parameter $\theta$, $n$ is the number of independent parameters to be estimated in the model and $N$ is the number of observations in the training data set.
In general, the BIC measure penalizes more complex models and favors the model which is rendered most plausible by the data at hand.
We chose the model parametrization $\theta^{*}$ that has largest BIC value in the $\theta$ search space.

\subsection{Image Quality Assessment (IQA)}
\label{IQA}
In this paper, we consider the following two image quality features of a facial image: pose and illumination.
Pose and illumination have proven record of being a strong performance predictor features for face recognition systems~\cite{beveridge2008focus,phillips2013existence}.
Therefore, these two image quality features are sufficient to demonstrate the merit of the proposed performance prediction model.
Furthermore, the choice of these two quality parameters is also motivated by the availability of a public facial image data set (\ie MultiPIE~\cite{gross2008multipie}) containing systematic pose and illumination variations.
Based on the classification scheme for facial image quality variations proposed by the ISO~\cite{iso_iec_29794-5:2010}, head pose and illumination correspond to subject characteristics and acquisition process characteristics respectively.
Furthermore, both quality parameters correspond to dynamic characteristics of a facial image as described in~\cite{iso_iec_29794-5:2010}.

\begin{figure}
 \centering
 \includegraphics[width=\linewidth]{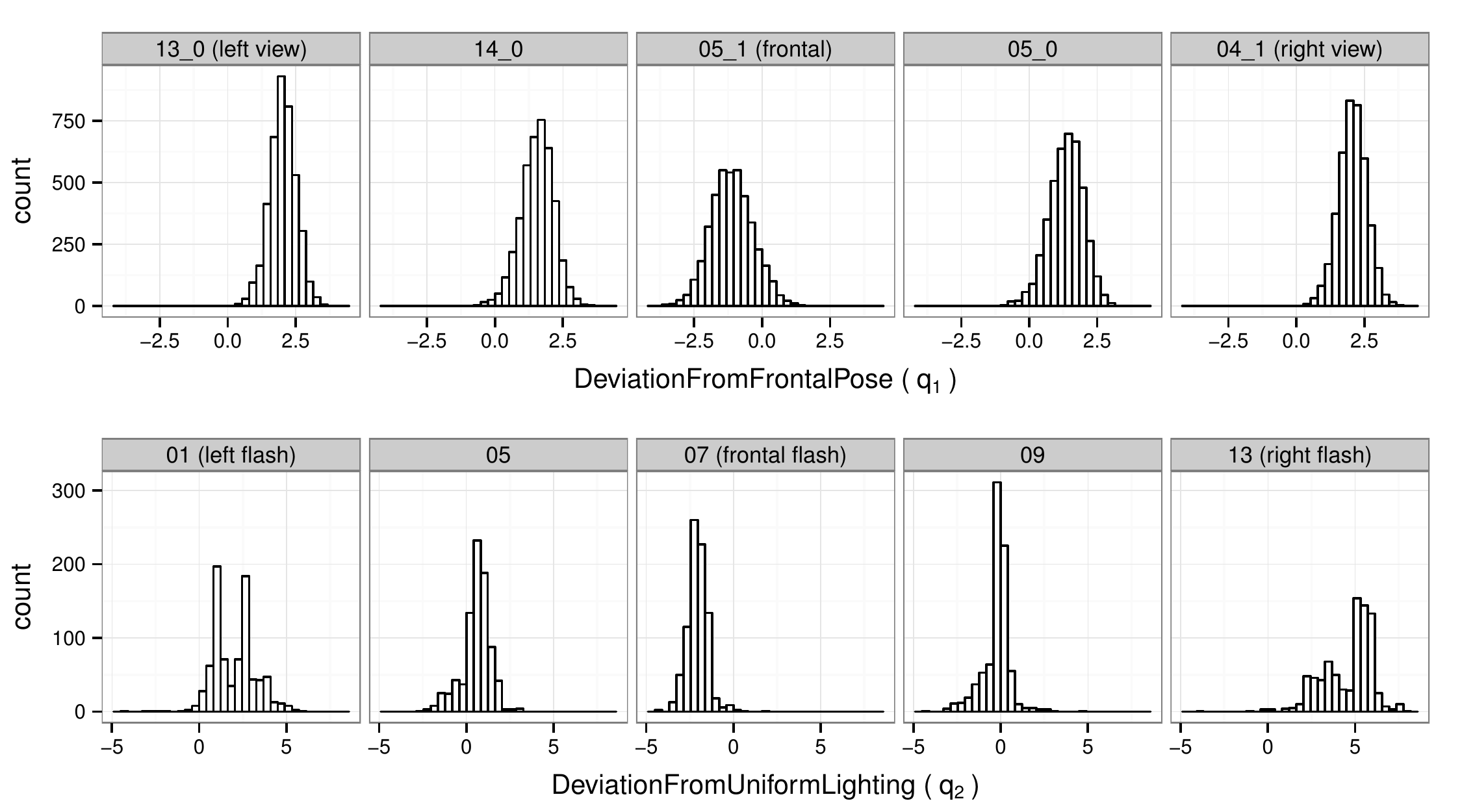}
 \caption{Distribution of image quality features measured by COTS-IQA on the MultiPIE training data set.}
 \label{fig:mpie_train_q1q2_dist_N10_m1.pdf}
\end{figure}

In this paper, we use a Commercial off-the-shelf (COTS) Image Quality Assessment (IQA) tool \texttt{dbassess} by~\cite{facevacs2010}.
Now onwards, we refer to this IQA using the acronym COTS-IQA.
The distribution of \verb+DeviationFromFrontalPose+ ($q_{1}$) and \verb+DeviationFromUniformLighting+ ($q_{2}$) quality features measured by COTS-IQA for the first fold (of 10-fold validation) training data set from MultiPIE data set is shown in~\figurename\ref{fig:mpie_train_q1q2_dist_N10_m1.pdf}.
The distribution of $q_1$ for frontal view images is centered around $-1.0$ while for non-frontal views, it shifts toward $+2.0$.
Similarly, while keeping the pose fixed to frontal view, we vary the illumination and observe that for frontal illumination the distribution of $q_2$ is centered around $-2.0$ while for other illumination conditions it shifts towards values $\geq 0$.
These distributions show that although COTS-IQA is accurate, its quality feature measurements are biased -- both left and right profile views are mapped to similar range of values thereby loosing the distinction between the two types of profile views.

\begin{figure}
 \centering
 \includegraphics[width=0.8\linewidth]{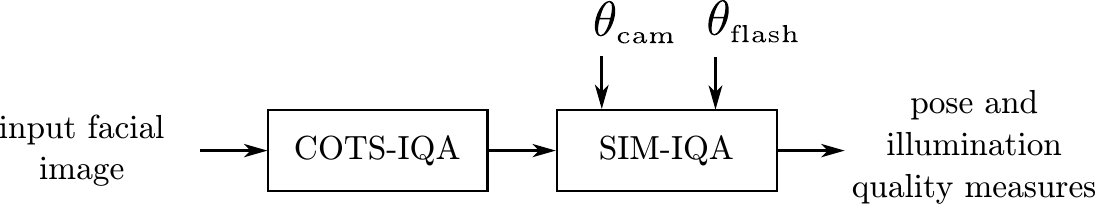}
 \caption{Input/Output features of the unbiased IQA (SIM-IQA) derived from COTS-IQA.}
 \label{fig:sim_iqa_block_dia.pdf}
\end{figure}

\begin{figure}[h]
 \centering
 \includegraphics[width=\linewidth]{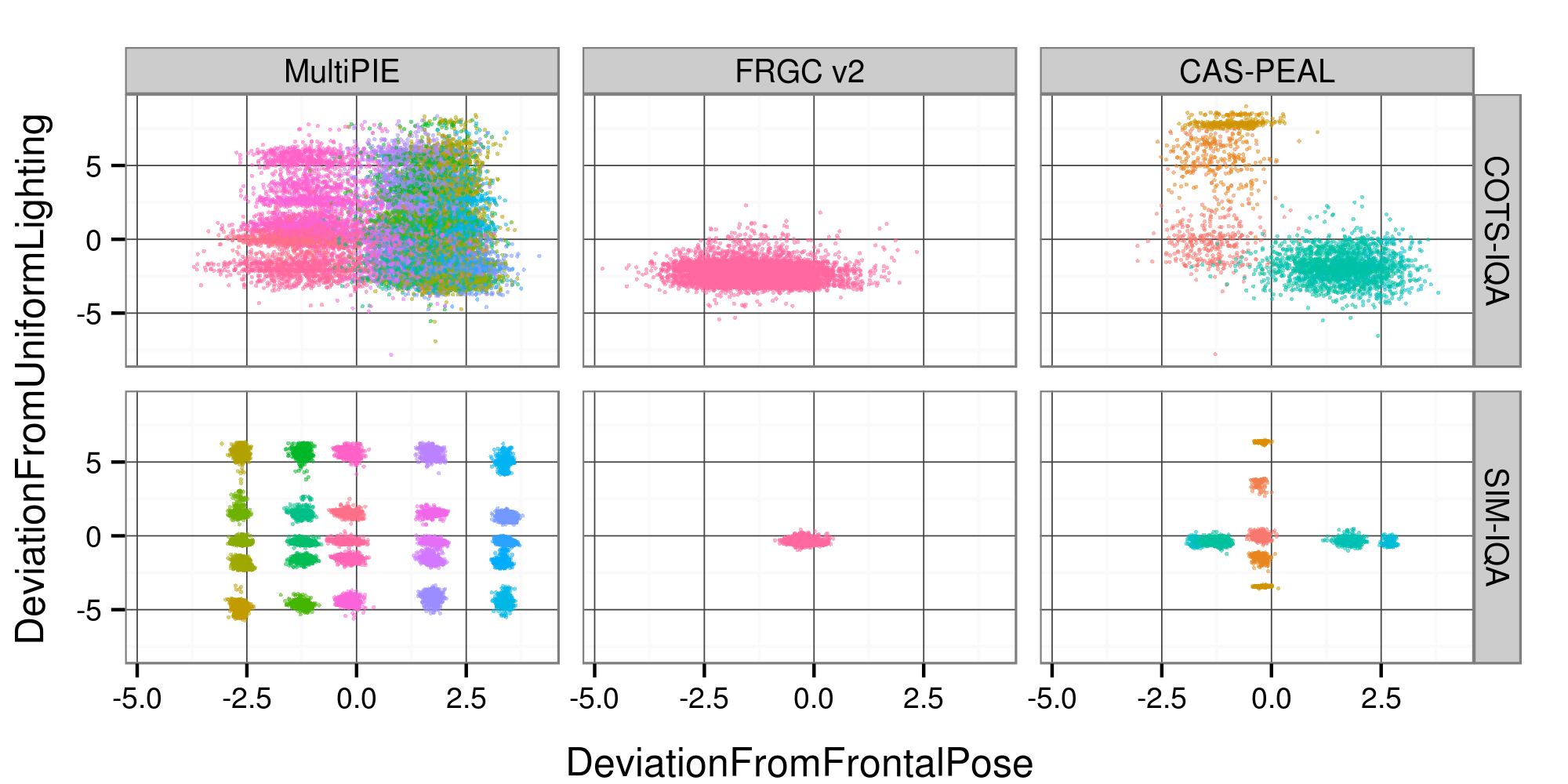}
 \caption{Quality space of COTS-IQA and unbiased IQA (SIM-IQA) which is derived from the COTS-IQA.}
 \label{fig:COTS-IQA_SIM-IQA_illustration.jpg}
\end{figure}

To demonstrate the performance prediction capability achievable by an accurate and unbiased IQA, we derive an unbiased IQA from the COTS-IQA -- henceforward referred using the acronym SIM-IQA.
The SIM-IQA is derived from the COTS-IQA and uses ground truth camera and flash positions, as shown in~\figurename\ref{fig:sim_iqa_block_dia.pdf}, to achieve more accuracy and unbiased quality assessments as shown in~\figurename\ref{fig:COTS-IQA_SIM-IQA_illustration.jpg}.
While the ground truth camera and flash positions are sufficient to simulate an IQA, we use input from COTS-IQA in order capture the characteristics of a realistic IQA.
It is important to understand that we do not use any other specific properties of COTS-IQA tool and therefore any other IQA tool can be easily plugged into this model.

SIM-IQA is obtained by transforming quality feature measurement by COTS-IQA such that images captured by same camera and under same flash map to distinctly separated clusters in the quality space as shown in~\figurename\ref{fig:COTS-IQA_SIM-IQA_illustration.jpg}. Let $\mathbf{A}=\{ \mathbf{a}_{i} \}$, where $\mathbf{a}_{i}=[q_{1}, q_{2}, \gamma_{1}, \gamma_{2}] \in \mathbb{R}^{4}$ such that $q_{1}, q_{2}$ are the pose and illumination quality measurements by COTS-IQA and $\gamma_{1}, \gamma_{2}$ be the corresponding ground truth camera and flash angle with frontal view as the reference (supplied with the image data set) for the $i^{\textnormal{th}}$ facial image sample. The corresponding quality measurements by SIM-IQA is $\mathbf{B}=\{ \mathbf{b}_{i} \}$, where $\mathbf{b}_{i} = [\hat{q}_{1}, \hat{q}_{2}] \in \mathbb{R}^{2}$ such that
\begin{align}
\hat{q}_{1} &= a \gamma_{1} + (q_{1} - \mu_{1}^{\gamma_{1},\gamma_{2}}) \nonumber \\
\hat{q}_{2} &= b \gamma_{2} + (q_{2} - \mu_{2}^{\gamma_{1},\gamma_{2}}) \nonumber
\end{align}
where, $a=1/10, b=1/18$ are scaling factor for the angle measurements $\gamma_{\{1,2\}}$ measured in degree and $\mu_{\{1,2\}}^{\gamma_{1},\gamma_{2}}$ are the mean of $q_{\{1,2\}}$ for each unique combination of ground truth camera and flash azimuth $\gamma_{\{1,2\}}$.
From the MultiPIE training data, we create the matrices $\mathbf{A}$ and $\mathbf{B}$ an compute a transformation matrix $\mathbf{x} \in \mathbb{R}^{4 \times 2}$ such that
\begin{equation}
\argmin_{\mathbf{x}} || \mathbf{A} \mathbf{x} - \mathbf{B} || \nonumber
\end{equation}
whose optimal solution, in the least square sense, is $\mathbf{x} = (\mathbf{A}^{T}\mathbf{A})^{-1} \mathbf{A}^{T} \mathbf{B}$.

\section{Experiments}
\label{dutta2015predicting_exp}
This section deals with the experiments designed to train and test the performance prediction model described in Section~\ref{dutta2015predicting_model_description}.
The description of the facial image data sets and face recognition systems used in these experiments are presented in Section~\ref{exp:datasets} and~\ref{exp:face_recognition_systems} respectively.
Several practical aspects of training a performance prediction model are dealt in Section~\ref{ssec:exp:model_training}.
In Section~\ref{ssec:exp:perf_pred}, we evaluate the accuracy of performance predictions on a test data set that is disjoint from the training data set.

\subsection{Data sets}
\label{exp:datasets}
We use the following three publicly available facial image data sets for all our experiments: MultiPIE~\cite{gross2008multipie}, FRGC v2~\cite{phillips2005overview} and CAS-PEAL~r1~\cite{gao2008caspeal}.
\begin{figure}[h]
\begin{center}
   \includegraphics[width=0.8\linewidth]{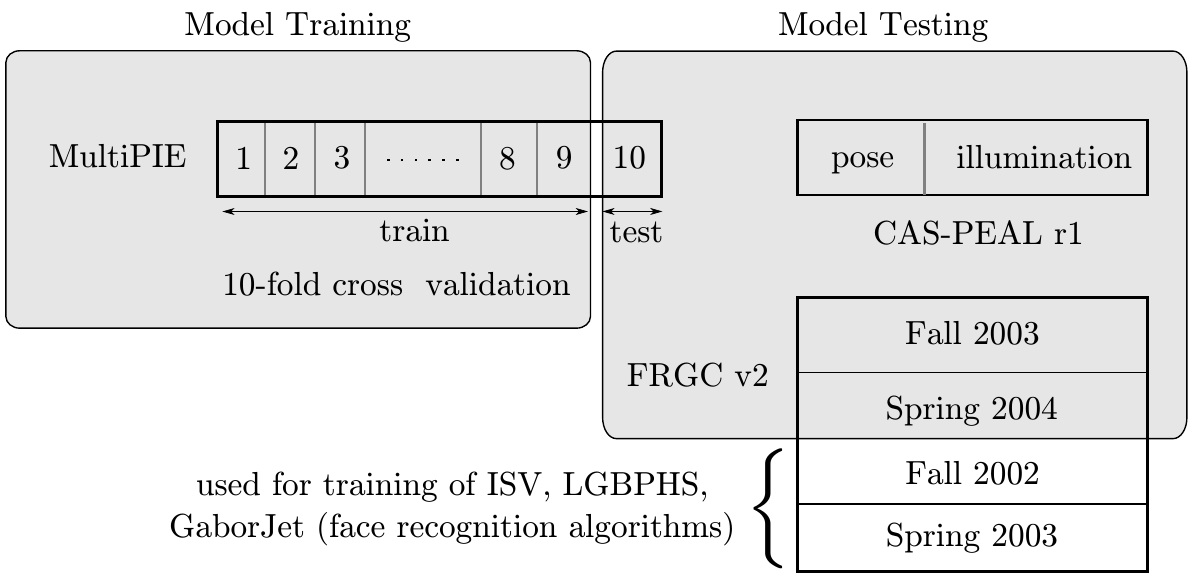}
\end{center}
   \caption{Data sets used for training and testing of the proposed model.}
\label{fig:train_test_dataset_illus.pdf}
\end{figure}

\begin{figure}[h]
 \centering
 \includegraphics[width=0.8\linewidth]{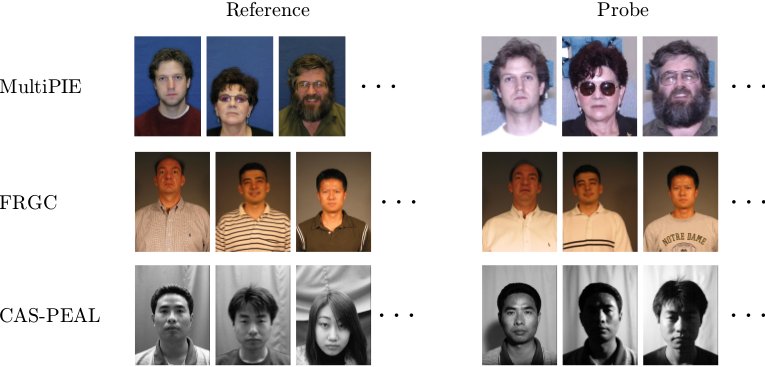}
 \caption{Sample reference and probe images from facial image data sets used in this paper.}
 \label{fig:prb_ref_img_illus.pdf}
\end{figure}

\begin{figure}
\begin{center}
   \includegraphics[width=0.8\linewidth]{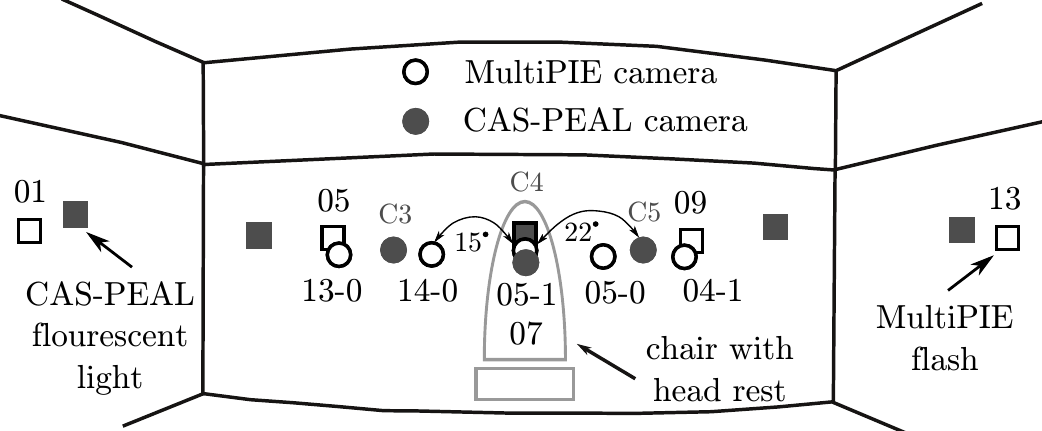}
\end{center}
   \caption{Camera and flash positions of the MultiPIE and CAS-PEAL data set.}
\label{fig:caspeal_mpie_capture_setup.pdf}
\end{figure}

The MultiPIE data set is primarily used for 10-fold cross validation of the model.
This data set contains systematic variations in pose and illumination and therefore has sufficient data to train and test the model based on two image quality parameters -- pose and illumination.
We use images of all the $337$ subjects across four sessions (first recording only) from the neutral expression subset of the MultiPIE.
The impact of session variation is eliminated by choosing both probe (or, query) and reference (or, enrollment) images from the same session.
The reference set contains high quality frontal mugshots and image quality variations exists only in the probe set as shown in~\figurename\ref{fig:prb_ref_img_illus.pdf}.
This simulates a real-world face verification scenario where the gallery is fixed to contain a set of high quality frontal mugshot of known individuals and facial image quality variations exists mainly in the probe set.
Recall that the proposed model can accommodate quality variation in both probe and gallery images.
However, to simulate real-world face verification scenario, we only vary the quality of probe image while keeping the gallery quality fixed.
The probe set contains $22960$ unique images of $337$ subjects captured by five camera (each separated by $15^{\circ}$) and 5 flash positions as depicted in~\figurename\ref{fig:caspeal_mpie_capture_setup.pdf}.
For the N-fold cross validation, we partition the full probe set into $N=10$ blocks such that each block contains $2296$ images randomly sampled from the full probe set.
Of the $10$ blocks, one block is retained as the validation data for testing the model while the remaining $9$ blocks are used for training the model as depicted in~\figurename\ref{fig:train_test_dataset_illus.pdf}.
This cross-validation process is repeated $10$ times such that each block is used as a test set exactly one time.
This ensures that training set has sufficient number of samples distributed in the quality space.
For each fold, the training set contains $20664$ match and $4764188$ non-match scores corresponding to $20664$ unique probe images and the testing set contains $2296$ match and $528381$ non-match scores corresponding to $2296$ unique probe images.

We also test the trained model on two other data sets that are independent from the training data set.
The first data set is the Fall 2003 and Spring 2004 subset (neutral expression, controlled condition only) of the FRGC v2 data set as shown in~\figurename\ref{fig:prb_ref_img_illus.pdf}.
This subset contains frontal view neutral expression images captured under controlled condition and therefore allows us to assess the performance prediction capability of the model on good quality images.
A single image of each subject is used as the reference while the remaining images are used as the probe.
The selected FRGC subset contains $7299$ match and $2596256$ non-match scores corresponding to $7299$ unique probe images.
Again, to minimize the impact of session variation, we chose probe and gallery images from the same session.

The second data set is the CAS-PEAL data set.
Since the model is trained to consider only pose and illumination image quality features, we include facial images of $1040$ subjects from the pose and illumination subset of the CAS-PEAL data set.
For the pose variation, we only include camera $\{C3,C5\}$ (when looking into frontal camera $C4$) corresponding to $PM\{-22^{\circ}, +22^{\circ}\}$ deviation from the frontal pose since the model is trained only on pose variations of $\pm 30^{\circ}$.
Contrary to documentation, the CAS-PEAL data set includes a pose variation of $\pm 15^{\circ}$ for some subjects.
We pool these images in the $\pm 22^{\circ}$ category.
The illumination variation subset contains images illuminated by a fluorescent (F) and incandescent (L) light source with elevation of $0^{\circ}$ and the following variations in azimuth: $IFM\{-90^{\circ}, -45^{\circ}, 0^{\circ}, 45^{\circ}, 90^{\circ} \}$.
Recall that the training data (based on MultiPIE) contained camera flash as the illumination source.
These camera and illumination positions of the CAS-PEAL data set are also depicted in~\figurename\ref{fig:caspeal_mpie_capture_setup.pdf}.

\subsection{Face Recognition Systems}
\label{exp:face_recognition_systems}
The impact of image quality variations on recognition performance also varies according to the capabilities of the face recognition system under consideration.
Therefore, we train and test the proposed model on the following six face recognition systems that have varying levels of tolerance towards facial image quality variations: FaceVACS~\cite{facevacs2010}, Verilook~\cite{verilook2011}, Cohort LDA~\cite{bolme2012csu}, Inter-Session Variability modeling~\cite{wallace2011intersession}, Gabor-Jet~\cite{guenther2012disparity}, Local Gabor Binary Pattern Histogram Sequences~(LGBPHS)~\cite{zhang2005local}.
The first two systems are Commercial off-the-shelf (COTS) and the remaining four systems are open source face recognition systems.
Throughout this paper, we refer to these six face recognition systems by the abbreviations COTS-A, COTS-B, cLDA, ISV, GaborJet, LGBPHS respectively.
We use the implementation of ISV, GaborJet and LGBPHS available in \texttt{FaceRecLib}~\cite{gunther2012open}, which is built on top of the open source signal-processing and machine learning toolbox \texttt{Bob}~\cite{anjos2012bob}.
The COTS-A, COTS-B and cLDA systems are pre-trained and ISV, GaborJet, LGBPHS are trained using the Fall 2002 and Spring 2003 subset of the FRGC v2 data set as defined in the training protocol of \texttt{FaceRecLib}.
We disable the automatic eye detection preprocessing stage of the face recognition systems and supply the same manually annotated eye coordinates to all the six face recognition systems.
This ensures consistency in facial image registration even for non-frontal images.

For face recognition systems deployed in real-world, the vendor (or, the user) sets an operating point by fixing the value of the decision threshold as shown in~\figurename\ref{fig:black_box_qr_model_illus}.
This decision threshold is chosen based on the user requirement of a certain minimum False Match Rate (FMR) or False Non-Match Rate (FNMR).
For the six systems considered in this paper, we simulate such a real world setup by setting the operating point to achieve a FMR of $0.1\%$ for the first three systems and $1\%$ for the remaining three systems.
To generate the Receiver Operating Characteristics (ROC) curve of~\figurename\ref{roc_mpie_Train9_Test1_Nqs12_fv.pdf}~to~\ref{fig:roc_caspeal_Train9mpie_Nqs12.pdf}, we train eight separate models to predict performance for the face recognition systems operating at the following FMR: $\{0.01\%,0.03\%,0.1\%,0.3\%,1\%,3\%,10\%,30\%\}$.
The corresponding decision threshold for COTS-A and COTS-B directly comes from their SDK (\ie vendor).
For the remaining four open source systems, we use frontal view and illumination $(05\_1,07)$ images of the MultiPIE training data set to compute the verification decision threshold corresponding to each FMR.

\subsection{Model Training}
\label{ssec:exp:model_training}
We begin with a coarse sampling of the COTS-IQA quality space based on $N_{qs}=12$ quantiles of evenly spaced probabilities along each dimension of the 2D quality space.
Discarding the first and last sampling points (corresponding to quantiles with probabilities $0.0$ and $1.0$ respectively), we have $10 \times 10$ unique sampling points in the 2D quality space resulting in $100$ overlapping regions around each sampling point.
As described in Section~\ref{model_training}, we draw $N_{\textnormal{rand}}=20$ random samples of $\mathbf{q}$ and $\mathbf{r}$ in each region which results in a training data set $\tilde{\mathcal{D}}_{\textnormal{train}} \in \mathbb{R}^{2000 \times 4}$.
A small value of $N_{\textnormal{rand}}$ ensures that the training process completes quickly ($\sim 5$sec).
Recall that we only consider the quality of probe images as the reference set contains high quality frontal mugshot images.

\begin{figure}
\begin{center} \includegraphics[width=0.7\linewidth]{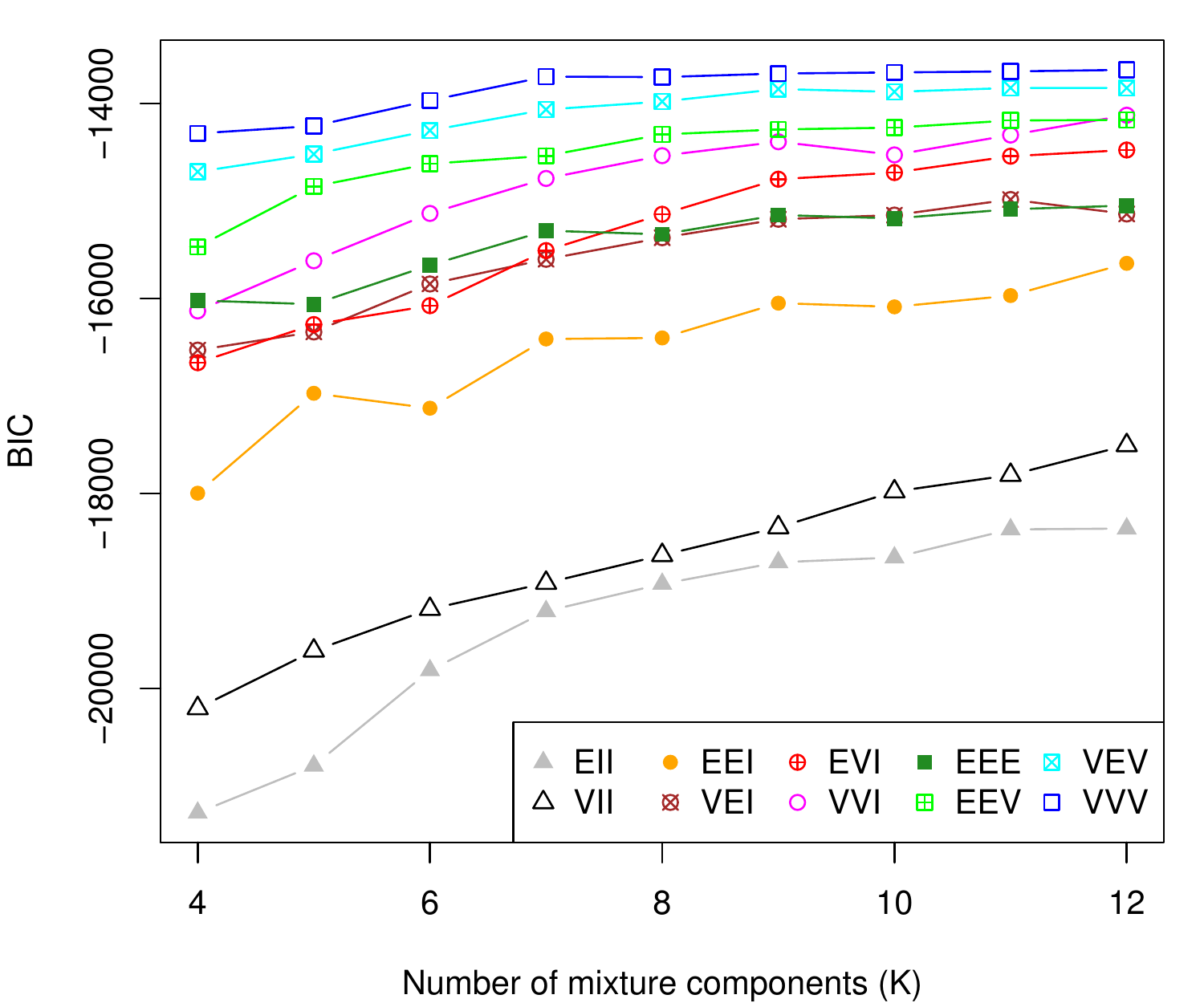}
\end{center}
   \caption{BIC value corresponding to different assignments of model parameter $\theta$.}
\label{fig:fv_BIC_vs_theta_Train9_Test1_Nbeta20_Nqs12_tid3.pdf}
\end{figure}

We select the optimal model parameters based on the BIC criterion as described in Section~\ref{model_param_selection}.
For COTS-A,~\figurename\ref{fig:fv_BIC_vs_theta_Train9_Test1_Nbeta20_Nqs12_tid3.pdf} shows the BIC value for the $\theta$ parameter search space.
We select $\theta^{*}=(9,\textnormal{VVV})$ because the BIC value attains maximum value at this point and saturates beyond it.
Furthermore, the remaining five face recognition systems also have similar trend of BIC values and therefore $\theta^{*}=(9,\textnormal{VVV})$ is selected as the model parameter for all six face recognition systems.
Here, VVV corresponds to a covariance matrix parametrization scheme in which all mixture components have varying (V) volume, the shapes and orientation of mixture components may vary (V).

The quality space sampling for SIM-IQA (\ie simulated IQA) is much simpler.
Quality regions correspond to the $25$ clusters formed by unique combination of $5$ camera and $5$ flash positions as shown in~\figurename\ref{fig:caspeal_mpie_capture_setup.pdf}.
With $N_{\textnormal{rand}}=20$, the training set corresponding to SIM-IQA is $\tilde{\mathcal{D}}_{\textnormal{train}} \in \mathbb{R}^{500 \times 4}$.
To model this nicely clustered quality space, the optimal model parametrization is manually selected to $\theta^{*} = (25,\textnormal{VVI})$.

Using the Expectation Maximization (EM) algorithm implementation available in the \texttt{mclust}~\cite{fraley2012mclust} library, we learn the model parameters of the GMM~(\ref{eq:P_q_r_MOG}) for both COTS-IQA and SIM-IQA.
The original $10 \times 10$ QR space of COTS-IQA and that learned by the model is shown in~\figurename\ref{fig:qr_map_Train9_Test1_Nqs12_K9_SigmaVVV_Nrand20_fv_IQA.pdf}.
This figure shows variation in recognition performance measures (FMR and FNMR at the operating point) for variations in quality of the probe.
The X and Y axis of this plot correspond to \verb+DeviationFromFrontalPose+ ($q_{1}$) and \verb+DeviationFromUniformLighting+ ($q_{2}$)  respectively of the probe image and the color denotes the value of FMR and FNMR at the operating point.
The black regions in QR space corresponding to the training data denotes quality regions where none of the samples were above (or below) the decision threshold for FMR (or FNMR) computation.
Furthermore, the QR space corresponding to the training data is discrete because the recognition performance values correspond to a quality space region.
Recall that, we do not consider the quality of reference as it remains fixed to high quality frontal mugshot images.
To plot the model's QR landscape, we evaluate the conditional expectation of recognition performance $r$ at a dense set of quality space sampling points using~(\ref{eq:expectation_r_given_q0}).
The conditional expectation values are shown in log$_{10}$ scale.
This visualization shows that both FMR and FNMR measures vary with the quality of the probe image.
Similarly, the original $5 \times 5$ QR space of SIM-IQA and that learned by the model are shown in~\figurename\ref{fig:qr_map_Train9_Test1_Nqs5_K25_SigmaVVI_Nrand20_fv_IQA0.pdf}.
This QR space is organized in small patches which correspond to clusters in the quality space as shown in~\figurename\ref{fig:COTS-IQA_SIM-IQA_illustration.jpg}.
Furthermore, \figurename\ref{fig:qr_map_Train9_Test1_Nqs12_K9_SigmaVVV_Nrand20_fv_IQA.pdf} and \ref{fig:qr_map_Train9_Test1_Nqs5_K25_SigmaVVI_Nrand20_fv_IQA0.pdf} also show that a GMM based model with appropriate model parametrization can accurately capture the variations in quality space.

\subsection{Performance Prediction}
\label{ssec:exp:perf_pred}
The test data set in each fold of the 10-fold cross validation set contains the following record for each verification attempt: similarity score $s$, quality of probe image $\mathbf{q},$ ground truth for verification decision (match or non-match).
The trained model can predict recognition performance $\mathbf{r}$ based solely on the quality of the probe images $\mathbf{q}$.
However, the test data set does not contain the value of true recognition performance measure per verification attempt.
Therefore, we resort to assessing the merit of model based performance predictions by pooling results according to the ground truth camera and flash label of each probe quality.
For each ground truth quality pool (\ie camera-id, flash-id), we compute the nature of FMR and FNMR distribution using the Beta distribution model of Section~\ref{prob_model_q_r}.
The mean and $95\%$ credible interval defines the variation in true recognition performance over the ground truth quality pool.
We also accumulate all model predictions of the recognition performance (FMR and FNMR) corresponding to each ground truth quality pool (\ie camera-id, flash-id).
The mean of these predictions and $95\%$ confidence interval (difference between $0.975$ and $0.025$ quantiles) define the variation in predicted performance over each ground truth quality pool.
We train the model at several operating points and plot the full ROC curves corresponding to the true and model predicted (both using COTS-IQA and SIM-IQA) performance.

In Figures \ref{roc_mpie_Train9_Test1_Nqs12_fv.pdf} to \ref{roc_mpie_Train9_Test1_Nqs12_clda.pdf}, we show the ROC curves corresponding to the MultiPIE test set for the six face recognition systems.
The ROC curves are pooled according to the ground truth quality data for probe images.
We only show the $95\%$ confidence and credible interval for FNMR to aid proper visualization of the results.
We also evaluate the merit of predicted performance on a test set derived from the FRGC data set and the CAS-PEAL pose and illumination subset.
For these evaluations on independent data sets, the model was trained on the MultiPIE training set corresponding to the first fold of 10-fold cross validation. 
The FRGC test set contains only frontal view images and therefore~\figurename\ref{fig:roc_frgc_Train9mpie_Nqs12.pdf} shows a single pool of quality.
The CAS-PEAL test set contains more pose and illumination variations and the true and predicted recognition performance are shown in~\figurename\ref{fig:roc_caspeal_Train9mpie_Nqs12.pdf}.

\subsubsection*{Error versus Reject Curve (ERC)} 
The authors of~\cite{grother2007performance} have proposed the Error versus Reject Curve (ERC) as a metric for evaluating the efficacy of a performance prediction system in identifying samples that contribute to verification error (\ie FNMR or FMR).
The ERC plots FNMR (or FMR) against the fraction of verification attempts rejected by a performance prediction system as being ``poor'' quality.
A biometric system with FNMR of $x$ would achieve a FNMR of $0$ by rejecting all the $x$ verification attempts that would lead to a False Non-Match.
This provides a benchmark for evaluating performance prediction systems.

\begin{figure}[t]
 \centering \includegraphics[width=\linewidth]{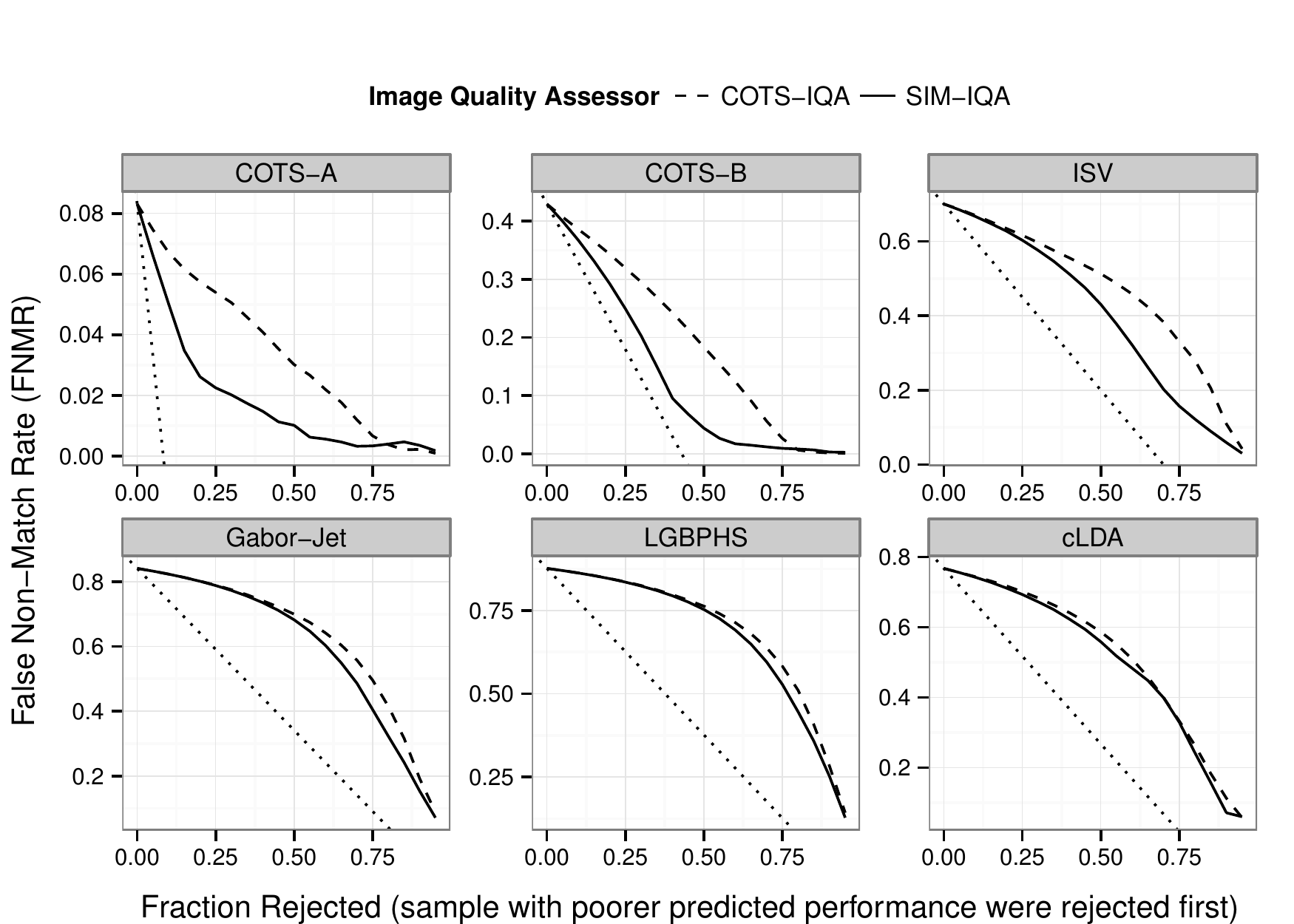}
 \caption{Error versus reject curve for the proposed performance prediction model based on two different Image Quality Assessors (IQA). Note that the fine dotted line denotes a sample rejection scheme based on an ideal performance prediction system (the benchmark).}
 \label{fig:erc_fnmr_mpie_Train9_Test1_K9_SigmaVVV_Nbeta20_Nqs12_tid333555.pdf}
\end{figure}

In this paper, we also use the ERC to evaluate the merit of performance predictions made by our model.
We sort all the verification attempts accumulated from the test set of N-fold cross validation based on the corresponding FNMR predicted by our model.
We sequentially remove verification attempts -- verification attempts with poorest predicted performance are rejected first -- and recompute the FNMR to create the ERC plot shown in~\figurename\ref{fig:erc_fnmr_mpie_Train9_Test1_K9_SigmaVVV_Nbeta20_Nqs12_tid333555.pdf}.

As expected, the ERC plot of~\figurename\ref{fig:erc_fnmr_mpie_Train9_Test1_K9_SigmaVVV_Nbeta20_Nqs12_tid333555.pdf} shows that performance predictions made using SIM-IQA are more accurate.
Furthermore, for COTS-A and COTS-B, we observe an initially sharp decline in FNMR which indicates that the model is good at identifying the poorest quality samples in pose and illumination quality space.
The flattening out of the ERC curves after the initial sharp decline suggests that pose and illumination quality features are not sufficient to identify ``poor'' quality samples containing image quality degradations along other quality dimensions.
We require additional image quality features to capture all the quality variations present in the test set.
For the remaining four face recognition systems, the ERC curves remain flattened until a majority of samples are rejected.
The reason for this is explained by the composition of our test data set and the nature of these systems.
The test set used for generating these ERC curve contains almost $80\%$ (only $4606$ of $22964$ images are frontal) non-frontal images.
The four face recognition systems are known to be highly sensitive to even small pose variations, and therefore a large number of non-frontal samples have to be rejected to bring down the FNMR.
On the contrary, COTS-A and COTS-B have some tolerance towards small deviation from frontal pose (like camera $14\_0$ and $05\_0$) and therefore significant drop in FNMR is achieved after rejecting a small number of extreme non-frontal images (corresponding to camera $13\_0$ and $04\_1$).

\section{Discussion}
\label{dutta2015predicting_discussion}
\begin{figure}[t]
 \centering
 \includegraphics[width=\linewidth]{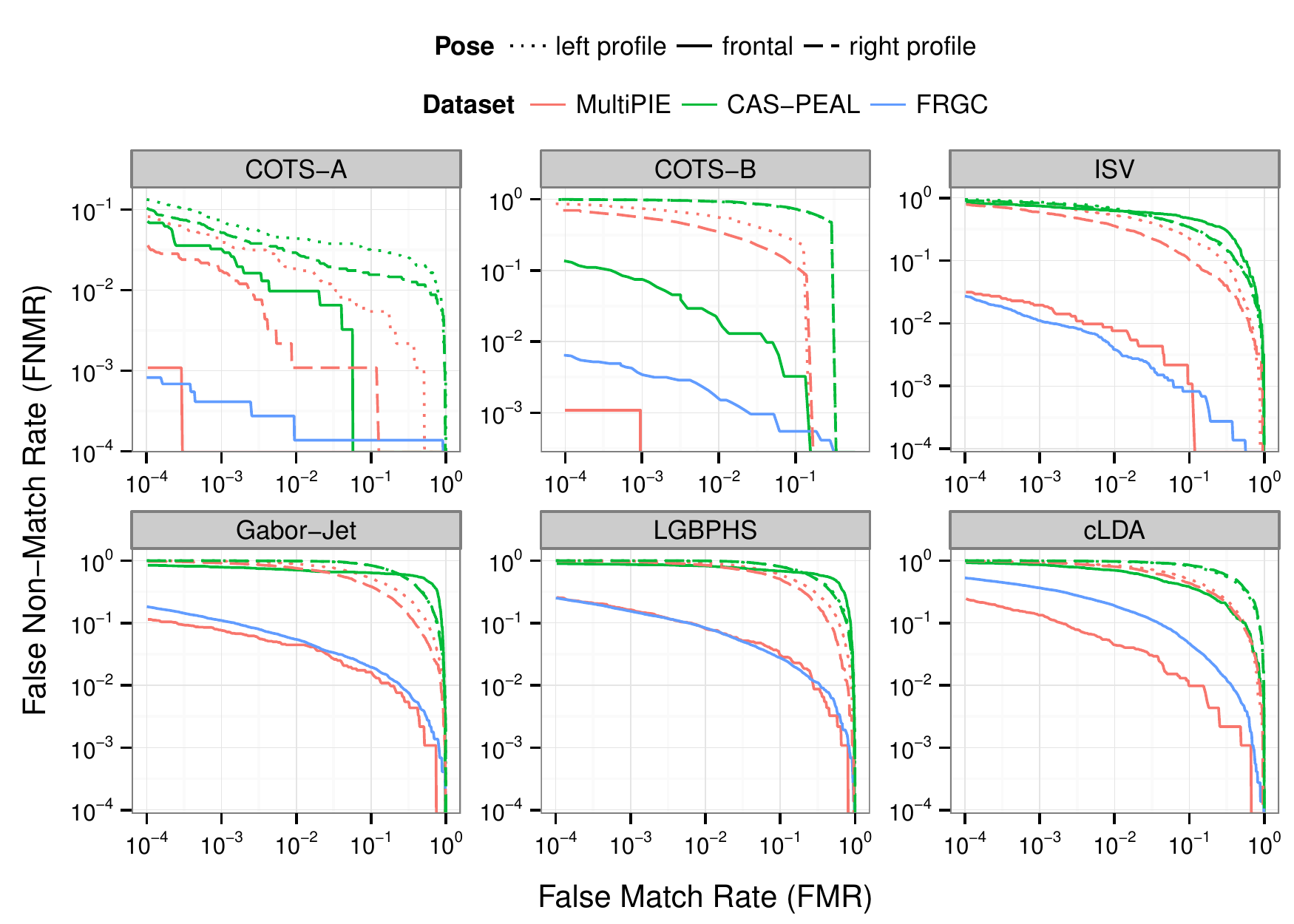}
 \caption{The nature of face recognition systems towards difference in facial pose (left and right profile views) and the differences across independent facial image data sets. Note: left and right view correspond to MultiPIE camera $\{13\_0,04\_1\}$ and CAS-PEAL camera $\{C2,C6\}$.}
 \label{fig:frs_roc_pose_dataset.pdf}
\end{figure}

\begin{figure*}
\centering
\subfloat[False Match Rate (FMR)]{\includegraphics[width=\linewidth]{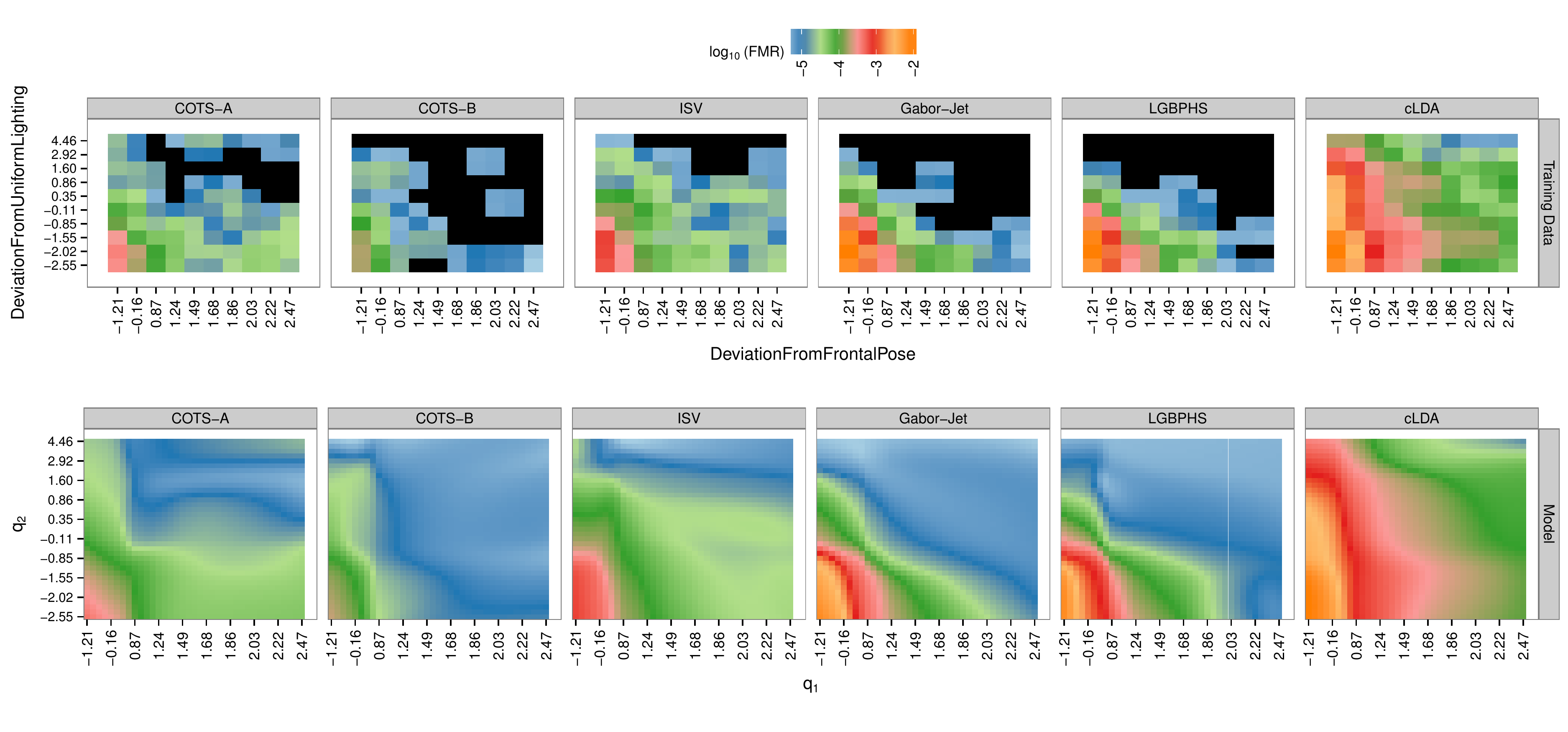}%
\label{fig:fmr_qr_map_Train9_Test1_Nqs12_K9_SigmaVVV_Nrand20_fv_IQA.pdf}}

\subfloat[False Non-Match Rate (FNMR)]{\includegraphics[width=\linewidth]{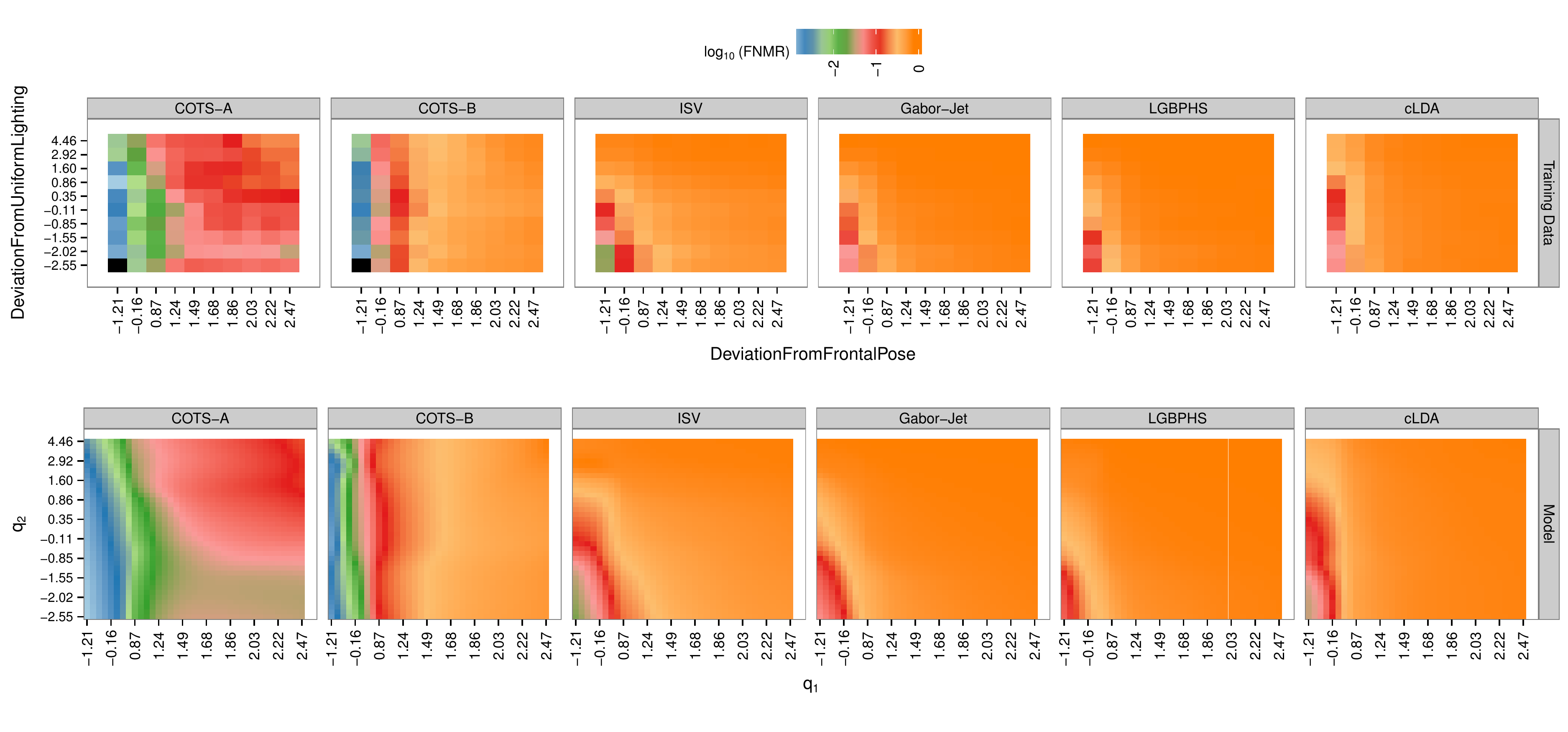}%
\label{fig:fnmr_qr_map_Train9_Test1_Nqs12_K9_SigmaVVV_Nrand20_fv_IQA.pdf}}
\caption{[COTS-IQA] Visualization of recognition performance (FMR and FNMR) in the quality space $q$ of COTS-IQA for the training data (with $N_{qs}=12$) and the GMM based trained Model (with $N_{\textnormal{rand}}=20$).}
\label{fig:qr_map_Train9_Test1_Nqs12_K9_SigmaVVV_Nrand20_fv_IQA.pdf}
\end{figure*}

\begin{figure*}
\centering
\subfloat[False Match Rate (FMR)]{\includegraphics[width=\linewidth]{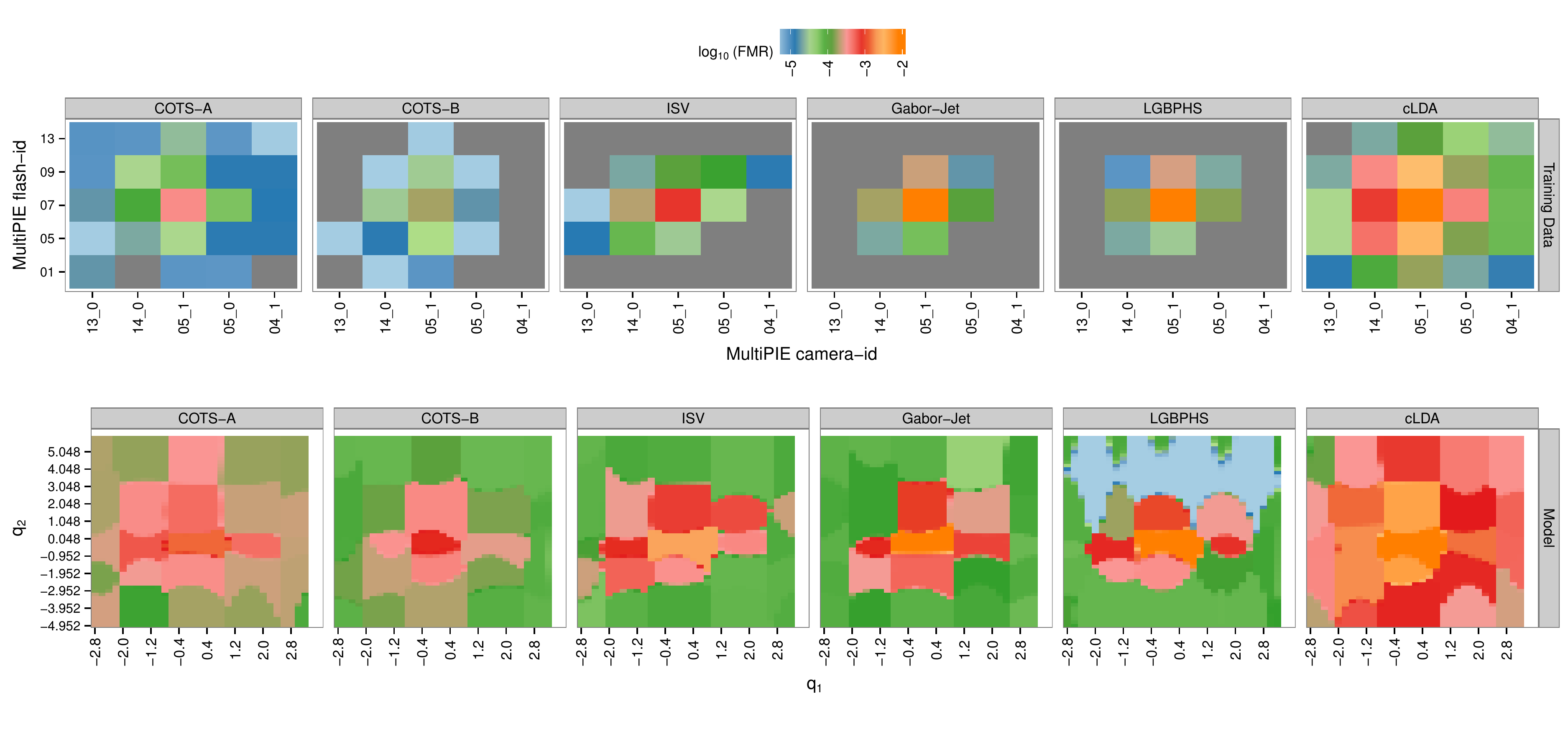}%
\label{fig:fmr_qr_map_Train9_Test1_Nqs5_K25_SigmaVVI_Nrand20_fv_IQA0.pdf}}

\subfloat[False Non-Match Rate (FNMR)]{\includegraphics[width=\linewidth]{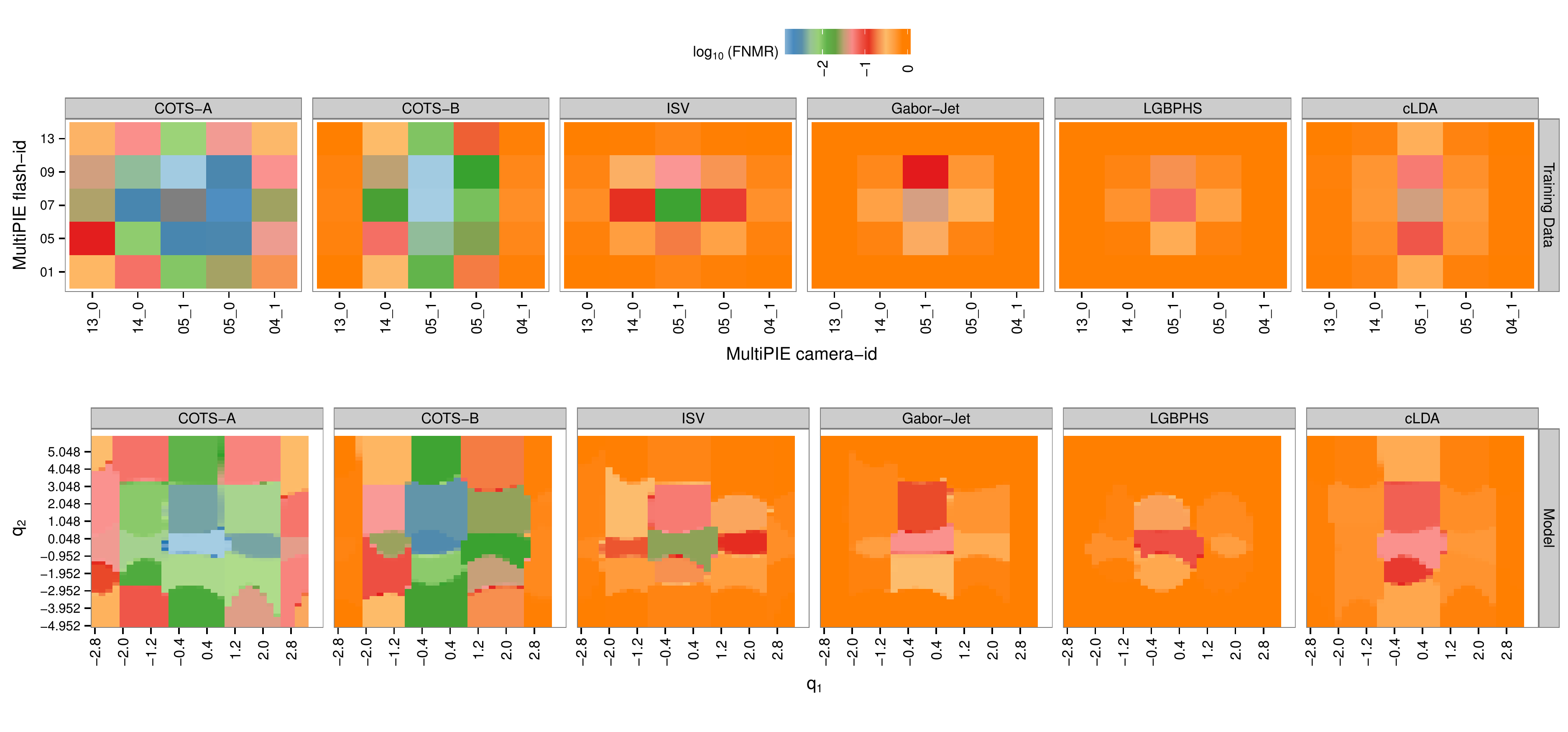}%
\label{fig:fnmr_qr_map_Train9_Test1_Nqs5_K25_SigmaVVI_Nrand20_fv_IQA0.pdf}}
\caption{[SIM-IQA] Visualization of recognition performance (FMR and FNMR) in the quality space $\mathbf{q}$ of the unbiased IQA (SIM-IQA) derived from the COTS-IQA for the training data and the GMM based trained Model (with $N_{\textnormal{rand}}=20$).}
\label{fig:qr_map_Train9_Test1_Nqs5_K25_SigmaVVI_Nrand20_fv_IQA0.pdf}
\end{figure*}

\begin{figure*}
 \centering
 \includegraphics[width=\linewidth]{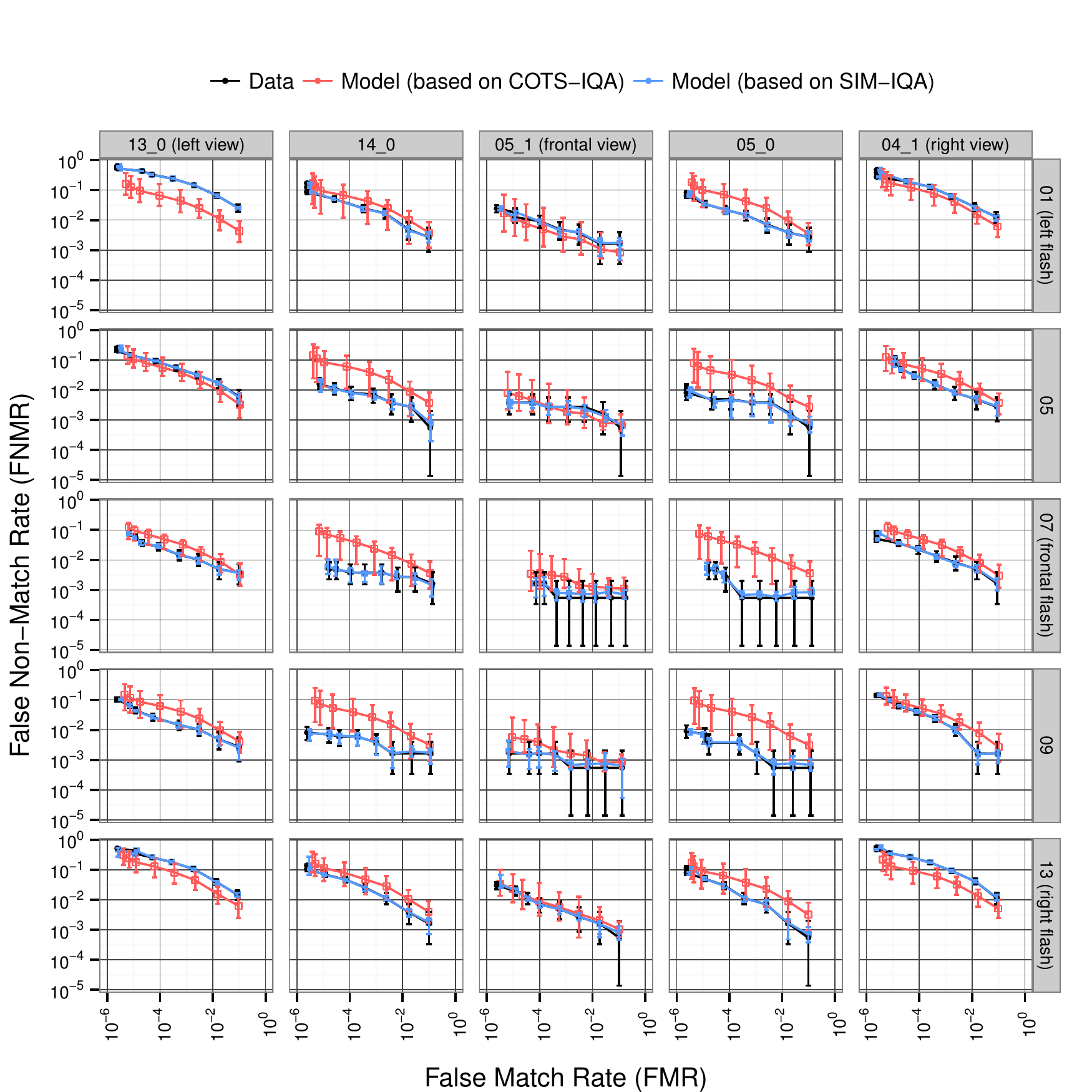}
 \caption{Recognition performance prediction of COTS-A system using COTS-IQA and SIM-IQA for MultiPIE test set pooled from 10-fold cross validation.}
 \label{roc_mpie_Train9_Test1_Nqs12_fv.pdf}
\end{figure*}

\begin{figure*}
 \centering
 \includegraphics[width=\linewidth]{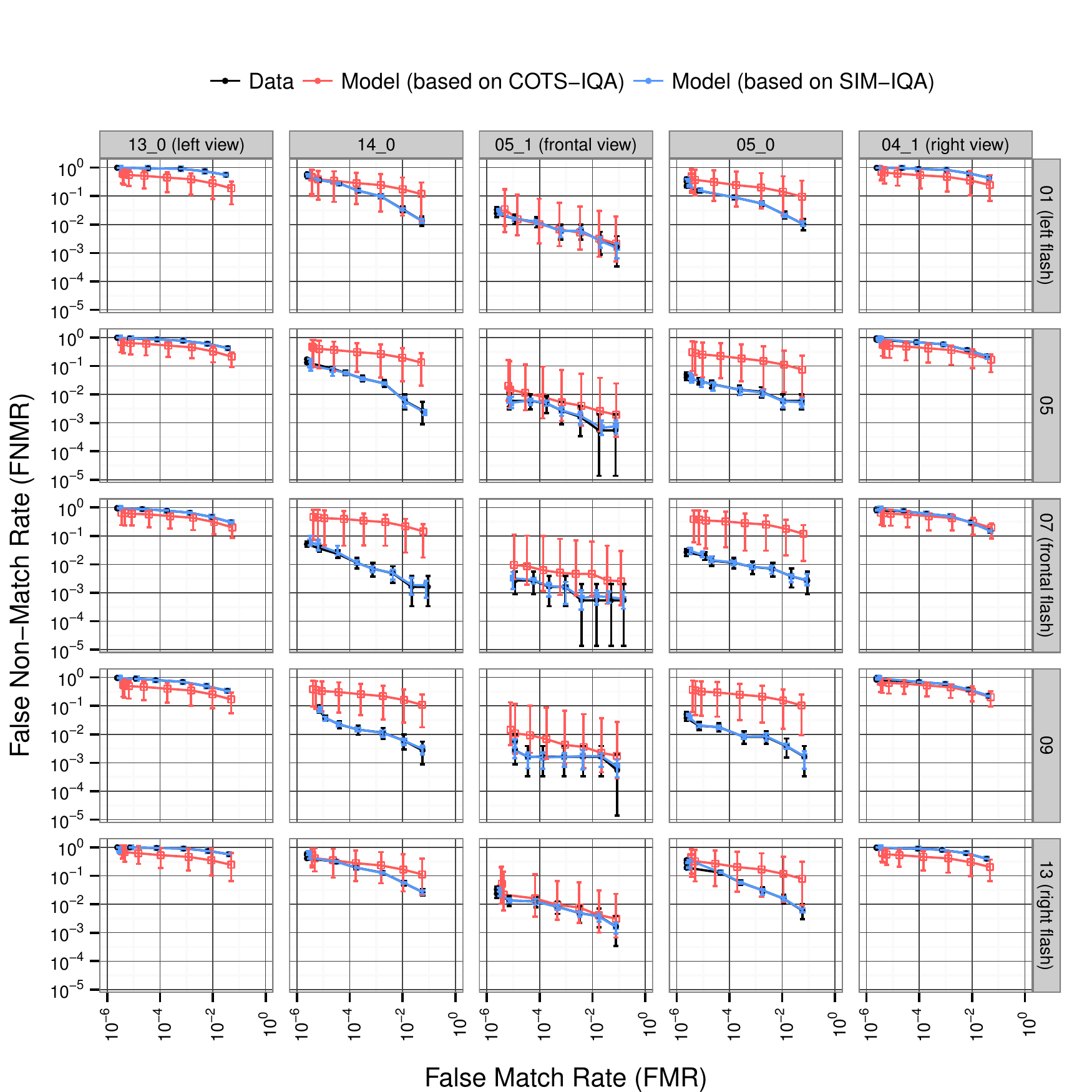}
 \caption{Recognition performance prediction of COTS-B system using COTS-IQA and SIM-IQA for MultiPIE test set pooled from 10-fold cross validation.}
 \label{roc_mpie_Train9_Test1_Nqs12_vl.pdf}
\end{figure*}

\begin{figure*}
 \centering
 \includegraphics[width=\linewidth]{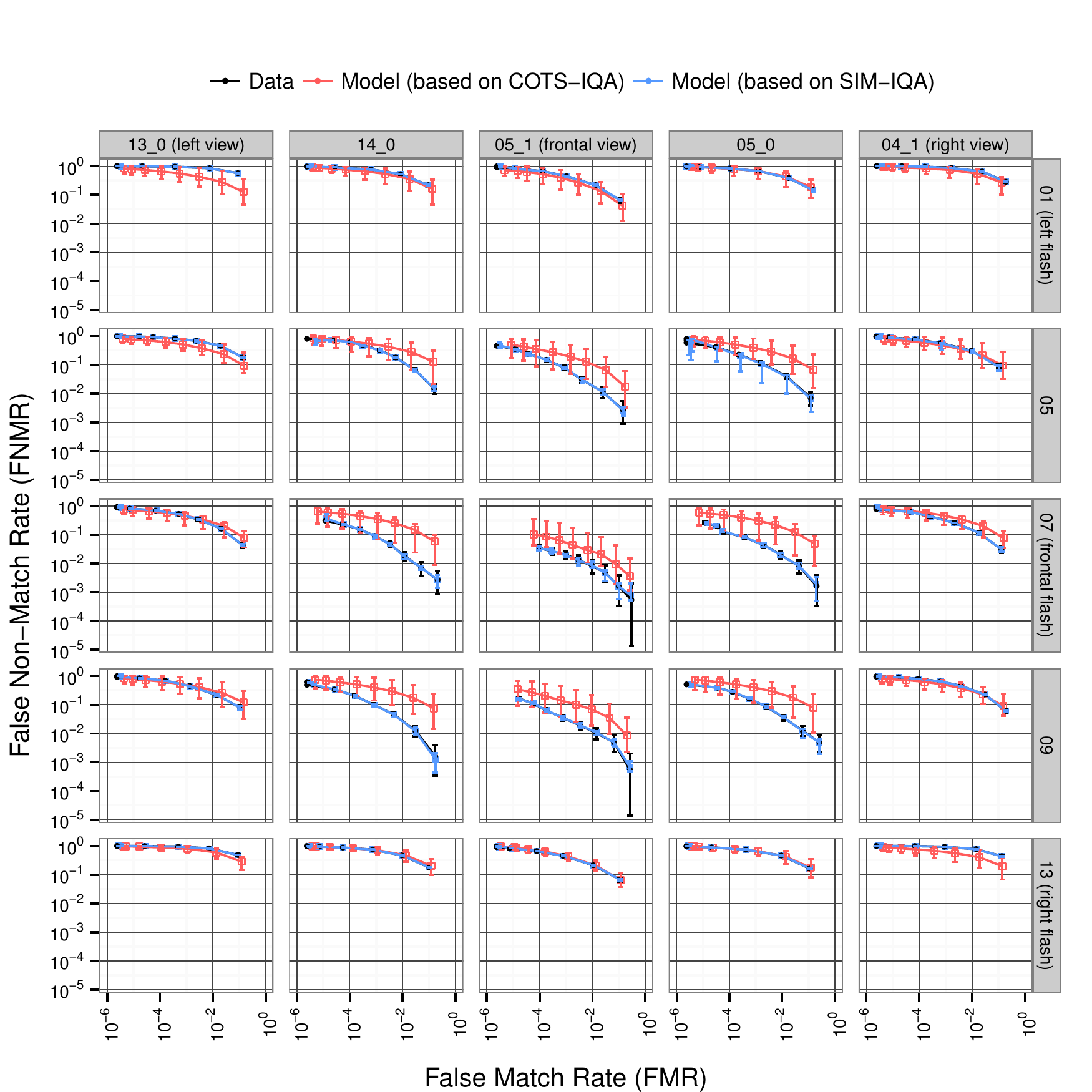}
 \caption{Recognition performance prediction of ISV system using COTS-IQA and SIM-IQA for MultiPIE test set pooled from 10-fold cross validation.}
 \label{roc_mpie_Train9_Test1_Nqs12_isv.pdf}
\end{figure*}

\begin{figure*}
 \centering
 \includegraphics[width=\linewidth]{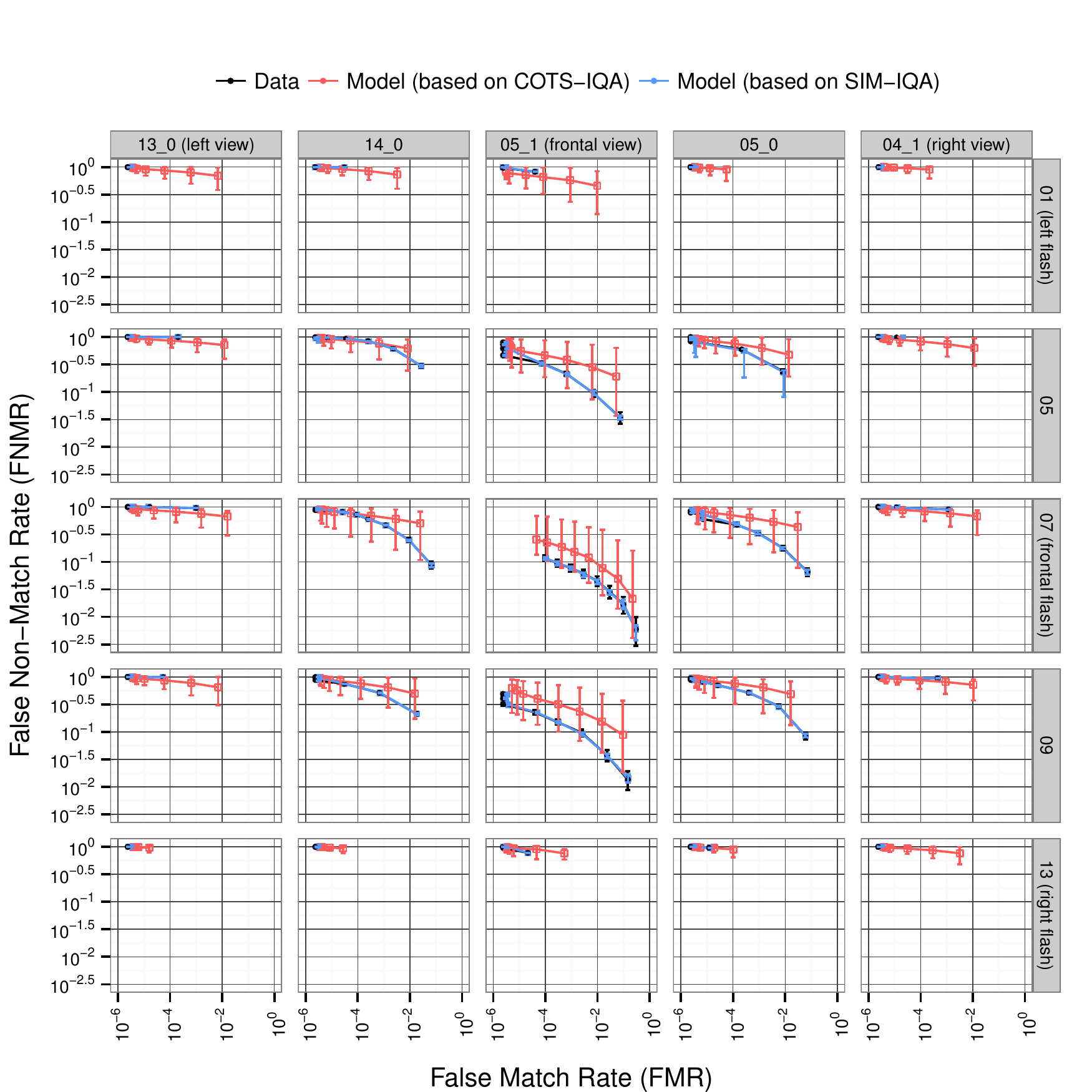}
 \caption{Recognition performance prediction of Gabor-Jet system using COTS-IQA and SIM-IQA for MultiPIE test set pooled from 10-fold cross validation.}
 \label{roc_mpie_Train9_Test1_Nqs12_gabor-jet.pdf}
\end{figure*}

\begin{figure*}
 \centering
 \includegraphics[width=\linewidth]{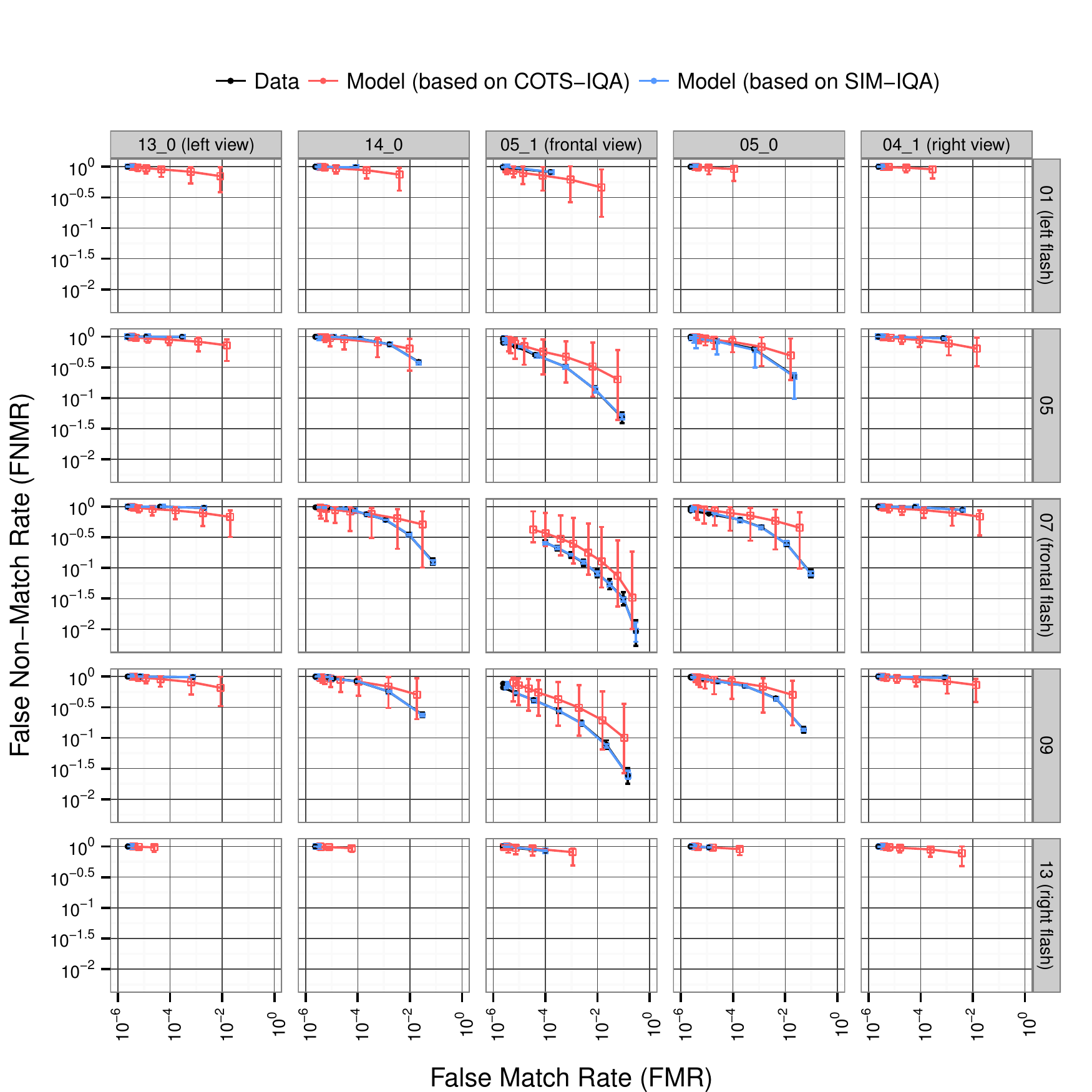}
 \caption{Recognition performance prediction of LGBPHS system using COTS-IQA and SIM-IQA for MultiPIE test set pooled from 10-fold cross validation.}
 \label{roc_mpie_Train9_Test1_Nqs12_lgbphs.pdf}
\end{figure*}

\begin{figure*}
 \centering
 \includegraphics[width=\linewidth]{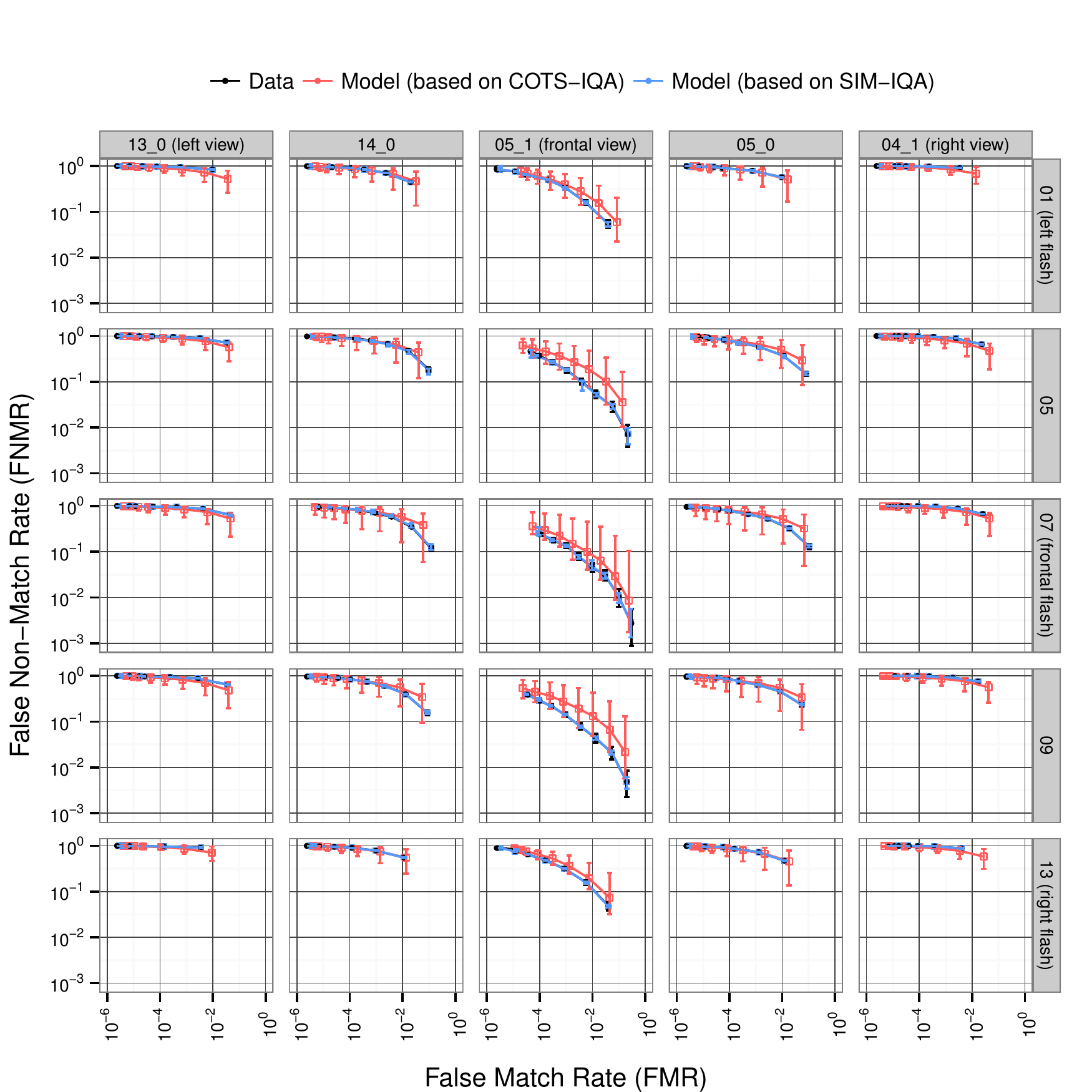}
 \caption{Recognition performance prediction of cLDA system using COTS-IQA and SIM-IQA for MultiPIE test set pooled from 10-fold cross validation.}
 \label{roc_mpie_Train9_Test1_Nqs12_clda.pdf}
\end{figure*}

\figurename\ref{fig:frs_roc_pose_dataset.pdf} is central to the discussions presented in this section.
Therefore, we first describe the contents of this figure.
We compare the performance of face recognition systems when operating on left and right profile images corresponding to probe image subsets taken from the MultiPIE and CAS-PEAL data set.
With the exception of cLDA, all the remaining five face recognition systems have better recognition performance while comparing right profile view images.
cLDA demonstrates consistent recognition performance for both left and right profile view images.
To capture such asymmetric relationship between pose and recognition performance, we require an IQA tool that maps left and right profile views to distinctly different regions of the quality space.
\figurename\ref{fig:mpie_train_q1q2_dist_N10_m1.pdf} shows that COTS-IQA maps both left and right profile to the same region ($q[1] \sim 2$) of the quality space thereby introducing ambiguity between left and right pose in the quality space.
On the other hand, the SIM-IQA maps left and right profile views to distinctly well separated regions of quality space as shown in~\figurename\ref{fig:COTS-IQA_SIM-IQA_illustration.jpg}.
Therefore, we expect the performance predictions based on COTS-IQA to have larger errors to non-frontal views and those based on SIM-IQA to be more accurate.
In Figures~\ref{roc_mpie_Train9_Test1_Nqs12_fv.pdf} to \ref{roc_mpie_Train9_Test1_Nqs12_clda.pdf}, we show the true and model predicted (based on COTS-IQA and SIM-IQA quality assessors) recognition performance for each unique combination of MultiPIE camera-id and flash-id.
As expected, we observe -- except for cLDA -- that model predictions based on COTS-IQA are further away from the true performance for non-frontal views while predictions based on SIM-IQA are more accurate for non-frontal views. 
Since, the recognition performance of cLDA remains consistent for both left/right profile, the performance prediction for non-frontal view is accurate irrespective of the IQA (COTS-IQA and SIM-IQA) used in the predictions as shown in~\figurename\ref{roc_mpie_Train9_Test1_Nqs12_clda.pdf}.
This shows that performance predictions based on an accurate and unbiased IQA are more accurate.

\begin{figure*}
 \centering
 \includegraphics[width=\linewidth]{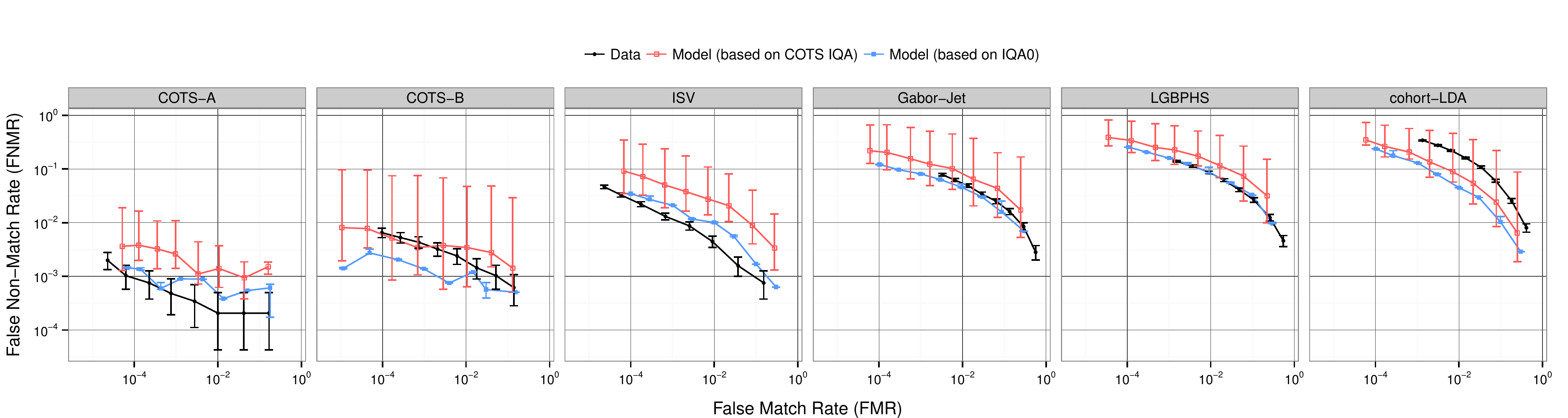}
 \caption{Model predicted and true recognition performance for test set based on the FRGC v2 data set.}
 \label{fig:roc_frgc_Train9mpie_Nqs12.pdf}
\end{figure*}

\begin{figure*}
 \centering
 \subfloat[Pose variation]{\includegraphics[width=\linewidth]{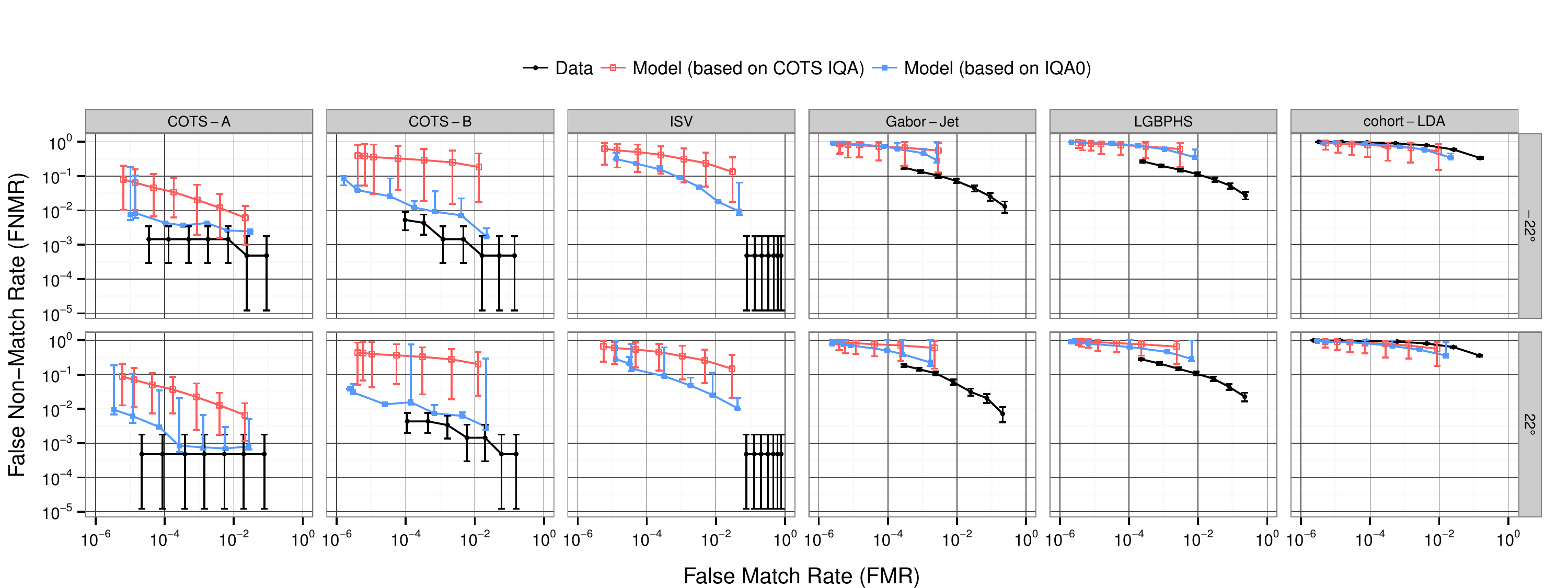}}

 \subfloat[Illumination variation]{\includegraphics[width=\linewidth]{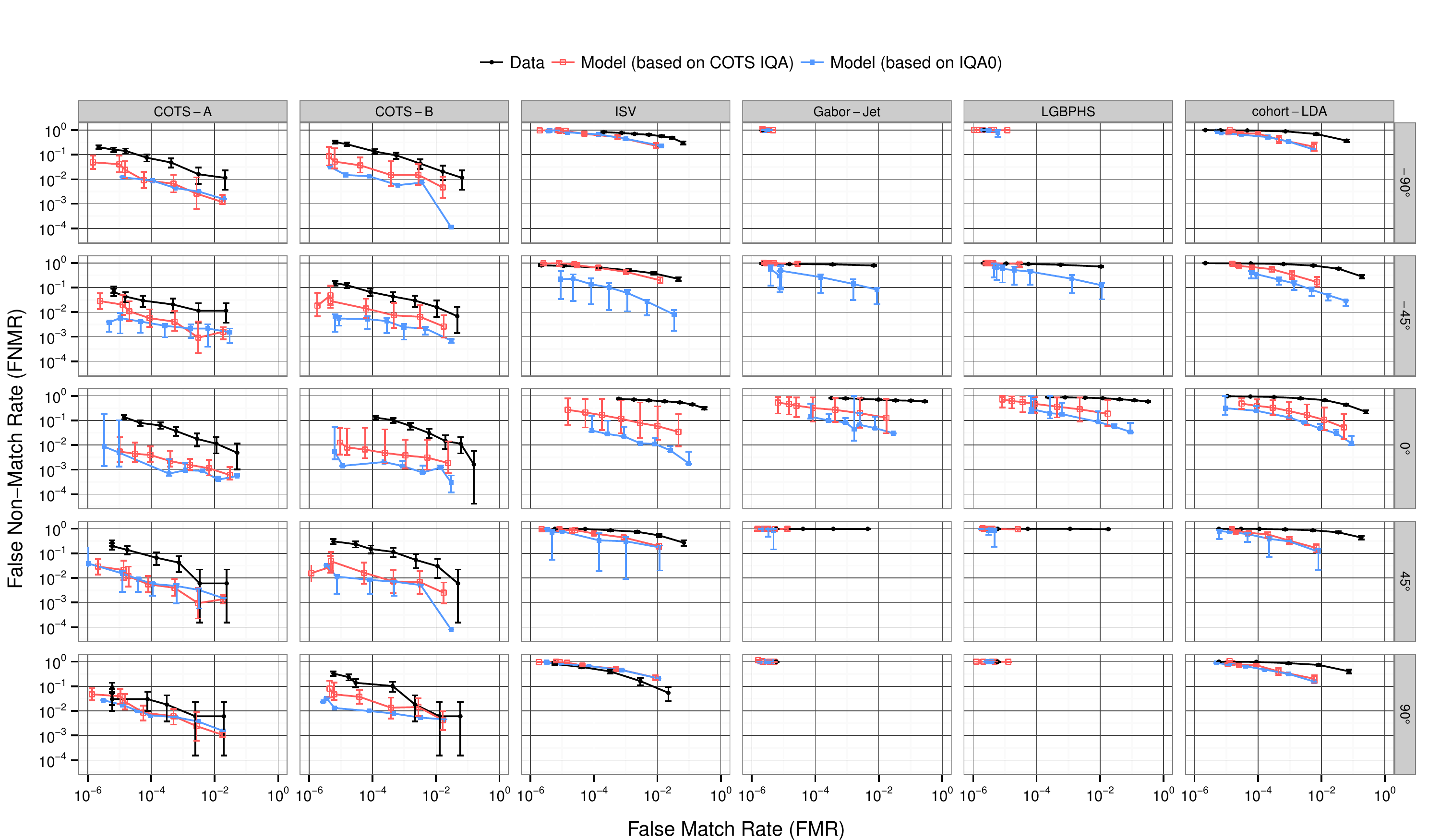}}
 \caption{Model predicted and true recognition performance for test set based on the CAS-PEAL data set.}
 \label{fig:roc_caspeal_Train9mpie_Nqs12.pdf}
\end{figure*}

In our performance prediction model, we have only considered two image quality features: pose and illumination.
However, there may exist variability in the \textit{unaccounted quality space} which is formed by image quality parameters other than pose and illumination.
For example, in this context, the unaccounted quality space may be composed of image quality features like resolution, capture device characteristics, facial uniqueness, \etc.
Furthermore, in a controlled data set like MultiPIE, FRGC or CAS-PEAL, it is reasonable to assume that variability in unaccounted quality space within a data set remains constant while such variability differs among the data sets.
Often variability among data sets is caused by difference in the capture device and capture environment.
Such variability is the reason why some data sets are more challenging than others in the context of face recognition.
To investigate the extent of variation present in the unaccounted quality space of the MultiPIE, FRGC and CAS-PEAL data sets, we compare the recognition performance of six face recognition systems on frontal view and frontal illumination images of these data sets in~\figurename\ref{fig:frs_roc_pose_dataset.pdf}.
Since we have selected the frontal pose and illumination subset of these controlled data sets, any performance difference for a particular face recognition system can be attributed to the variability present in the unaccounted quality space of these data sets.
Furthermore, since the performance prediction model is trained solely on the MultiPIE subset, in the following analysis, we assume the unaccounted quality space of MultiPIE frontal subset to be the reference.
In~\figurename\ref{fig:frs_roc_pose_dataset.pdf}, we observe that the performance of all six face recognition systems are consistently much poorer on the CAS-PEAL data set as compared to the corresponding recognition performance on MultiPIE.
This shows that the unaccounted quality space of CAS-PEAL data set is significantly different from that of the MultiPIE or FRGC data set.
Therefore, we expect that a performance prediction model trained solely on the MultiPIE data set (using the SIM-IQA) will perform poorer on the CAS-PEAL data set.
As expected, \figurename~\ref{fig:roc_caspeal_Train9mpie_Nqs12.pdf} confirms that performance predictions (using models trained on MultiPIE subset with COTS-IQA or SIM-IQA) on the CAS-PEAL data set are erroneous because of the large difference in the unaccounted quality space variability of CAS-PEAL as compared to that of the MultiPIE data set.
Surprisingly,~\figurename\ref{fig:frs_roc_pose_dataset.pdf} reveals that there is very small difference between the performance corresponding to MultiPIE and FRGC for Gabor-Jet and LGBPHS systems while the performance is significantly different for the remaining four face recognition systems.
This suggests that while there is difference in the variability of the unaccounted quality space between FRGC and MultiPIE data sets, the Gabor-Jet and LGBPHS systems are tolerant to this difference.
Therefore, for the Gabor-Jet and LGBPHS systems, we expect a performance prediction model trained solely on the MultiPIE data set (using the SIM-IQA) will make more accurate predictions on the FRGC data set.
Furthermore, the prediction error on the FRGC data set will be high for COTS-B and cLDA because these systems are highly sensitive to the difference in unaccounted quality space of FRGC data set as shown in~\figurename\ref{fig:frs_roc_pose_dataset.pdf}.
The performance predictions on the FRGC data set are shown in Figure~\ref{fig:roc_frgc_Train9mpie_Nqs12.pdf} and the accuracy of the predictions are exactly as we expected -- more accurate predictions for Gabor-Jet and LGBPHS while high prediction error on COTS-B and cLDA.
These findings suggest that reliability of performance predictions is highly dependent on the variability that exists in the unaccounted quality space.
Therefore, to make accurate predictions for a face recognition system, we must consider all the image quality features that have an influence on the performance of that system.

The CAS-PEAL data set mainly consists of subjects from East Asia.
There is evidence that face recognition algorithms (like the six systems used in this paper) trained mainly on Caucasian faces are less accurate when applied to East Asian faces~\cite{phillips2011otherrace}.
This suggests that race of subjects contained in a verification attempt is potentially an important image quality feature (static subject characteristics according to ~\cite{iso_iec_29794-5:2010}) that is essential to address such a performance bias present in existing face recognition systems.

\section{Conclusion}
\label{dutta2015predicting_conclusion}
In this paper, we present a generative model to capture the relation between image quality features $\mathbf{q}$ (\eg pose, illumination, \etc) and face recognition performance $\mathbf{r}$ (\eg FMR and FNMR at operating point).
Such a model allows performance prediction even before the actual recognition has taken place because the model is based solely on image quality features.
A practical limitation of such a data driven generative model is the limited nature of training data.
To address this limitation, we have developed a Bayesian approach to model the nature of FNMR and FMR distribution based on the number of match and non-match scores in small regions of the quality space.
Random samples drawn from the models provide the initial data essential for training the generative model $P(\mathbf{q},\mathbf{r})$.

We evaluated the accuracy of performance predictions based on the proposed model using six face recognition systems operating on three independent data sets.
The evidence from this study suggests that the proposed performance prediction model can accurately predict face recognition performance using an accurate and unbiased Image Quality Assessor (IQA).
An unbiased IQA is essential to capture all the complex behaviours of face recognition systems.
For instance, our results show that the performance of some face recognition systems on right view is better than the recognition performance on left view.
Such a complex and unexpected behaviour can only be captured by an IQA that maps left and right views to different regions of the quality space.

We also investigated the reason behind high performance prediction error when the performance prediction model is applied to other independent data.
We found variability in the \textit{unaccounted quality space} -- the image quality features not considered by the IQA -- as the major factor causing inaccuracies in predicted performance.
Even controlled data sets have large amount of variability in the unaccounted quality space.
Furthermore, face recognition systems differ in their tolerance towards such variability.
Therefore, in general, to make accurate predictions on a large range of test data set, we should either consider all the relevant image quality features in order to minimize the variability in unaccounted quality space or use a classifier that is agnostic to variability in the unaccounted quality space.

This work has pointed out future work in many directions.
Clearly, the most significant effort needs to be concentrated in the direction of discovering novel features that can summarize a large number of image quality variations.
This is essential for limiting the variations present in the unaccounted quality space.
Furthermore, there is a clear need to develop accurate and unbiased Image Quality Assessment systems.
Although our model can accept image quality parameter measurements from off-the-shelf and uncalibrated quality assessment systems, more transparent and standardized quality metrics are needed to facilitate standardized exchange of image quality information as proposed in~\cite{iso_iec_29794-1:2009}.
Future work could also investigate methods to directly incorporate probabilistic models of quality and recognition performance into the EM based training procedure.
It would also be interesting to apply the proposed model to predict the performance of other biometric systems and other classifiers in general.


\chapter{Impact of Eye Detection Error on Face Recognition Performance}
\label{dutta2015impact_intro}
Facial image registration is a common preprocessing applied by many face recognition systems.
It corrects for scale and orientation of facial images to ensure consistency in the spatial location of facial features.
In other words, this preprocessing ensures that facial features like eyes, nose, \etc occupy same location in the facial image used for facial feature extraction.
This critical preprocessing can greatly influence the performance of a face recognition system that is sensitive to variations in scale and orientation of facial features.
Most face recognition systems~\cite{facevacs2010,verilook2011,bolme2012csu} use the two eyes coordinates as the landmarks for facial image registration where the two eyes coordinates are obtained using an automatic eye detector.
Therefore, the performance of such face recognition systems depend not only on capabilities of the facial feature extraction/comparison stages -- the core components of a face recognition algorithm -- but also on the accuracy of automatic eye detectors.
The accuracy of automatic eye detectors is known to be influenced by image quality variations~\cite{dutta2014automatic}.
Furthermore, image quality variations also influence the accuracy of face recognition algorithms by either occluding or obscuring facial features present in a image~\cite{beveridge2008focus}.
Image quality variations therefore have this dual impact on face recognition performance:
\begin{inparaenum}[\itshape a\upshape)]
\item impact on the accuracy of automatic eye detection; and
\item impact on the accuracy of facial feature extraction/comparison.
\end{inparaenum}

In this chapter, we investigate the influence of automatic eye detection error on the performance of face recognition systems.
We simulate the error in automatic eye detection by performing facial image registration based on perturbed manually annotated eye coordinates.
Since the image quality of probe images are fixed to frontal pose and ambient illumination, the performance variations is solely due to the impact of facial image registration error on face recognition performance. 
Furthermore, we also assess the accuracy of automatic eye detectors included in two commercial face recognition systems~\cite{facevacs2010,verilook2011}.

This study helps us understand how image quality variations can amplify its influence on recognition performance by having dual impact on both facial image registration and facial feature extraction/comparison stages of a face recognition system.
Such an understanding is important to model and predict the recognition performance variations of an automatic face recognition system.

In addition to the two eye coordinates, some modern face recognition systems employ other facial landmarks for the registration of facial images.
These systems are naturally more capable of accurate facial image alignment~\cite{kazemi2014one}.
However, many face recognition systems (both commercial and open source) are still based on the two eye coordinates based facial image registration.
Therefore, in this chapter, we focus our study on face recognition systems that rely on the detection of the two eye coordinates for facial image registration.

\section{Introduction}
\begin{figure}[ht!]
 \centering
 \includegraphics[width=\linewidth]{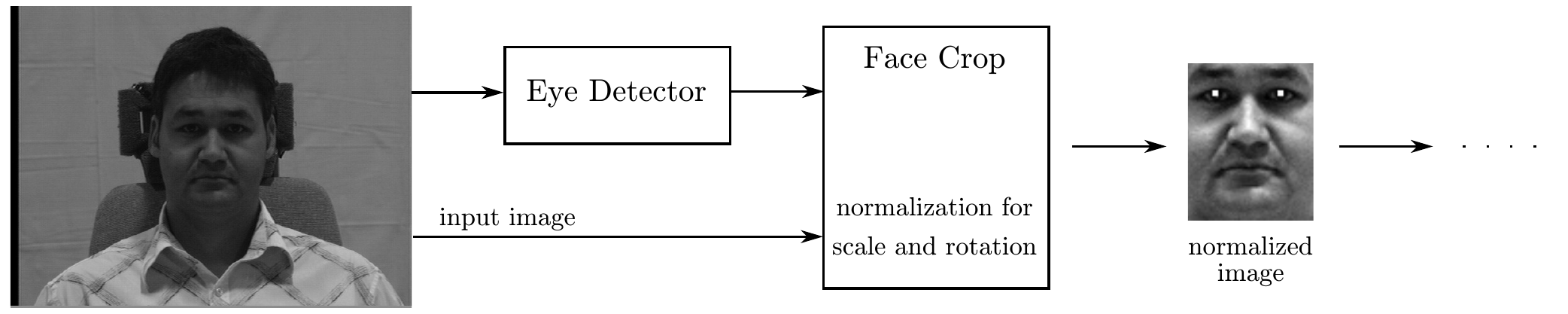}
 \caption{Basic facial image preprocessing pipeline used by most face recognition systems.}
 \label{fig:facecrop_illustration}
\end{figure}

The normalization of a facial image for scale and rotation is a common preprocessing stage of many face recognition systems.
This preprocessing stage, often called geometric normalization of the face, ensures that facial features like nose or eyes occupy similar spatial positions in all images.
The locations of at least two facial landmarks are required to normalize a facial image for translation, scale and in-plane rotation.
Most commercial and open source face recognition systems use the centers of the eyes as landmarks for face normalization because the inter-ocular distance can be used to correct scale, while the orientation of the line between the eyes allows correction of in-plane rotation~\cite{riopka2003eyes}, as indicated in Figure~\ref{fig:facecrop_illustration}.

A face normalization scheme based on the centers of the eyes is known to contribute to decrease face recognition performance if supplied with inaccurate eye locations~\cite{min2005eye,rodriguez2006measuring,riopka2003eyes,wang2005sensitivity}.
So far, this observation was based on results of a small number of face recognition system operating on a limited facial image data set.
However, face recognition systems have different tolerances to misalignment of facial features.
Therefore, in this paper, we study the impact of misalignment that is caused by eye localization errors on the performance of the following five open source face recognition systems, which are spanning a wide range of popular and state-of-the-art systems:
Eigenfaces~\cite{turk1991eigenfaces}, Fisherfaces~\cite{belhumer1997eigenfaces}, Gabor-Jet~\cite{guenther2012disparity}, Local Gabor Binary Pattern Histogram Sequence (LGBPHS)~\cite{zhang2005local} and Inter-Session Variability modeling (ISV)~\cite{wallace2011intersession}.
All methods are evaluated using two evaluation metrics, the (unbiased) Half Total Error Rate (HTER) and the (biased) Area Under Curve (AUC).

It is common practice to consider manually located eye coordinates as ground truth when assessing the accuracy of automatically detected eye coordinates.
We investigate the merit of this practice by analyzing the difference in manual eye annotation performed by two independent institutions.
We also analyze the accuracy of automatic eye detectors present in two commercial face recognition systems.
Based on a novel experiment methodology, we confirm that ``eye locations are essentially ambiguous''~\cite{shan2004curse}.
Furthermore, we investigate the impact of using different types of annotations (manual, automatic) for the training/enrollment and the query phases on face recognition.

Our experiments provide an insight into the limits of geometric face normalization schemes based on eye coordinates.
Our aim is to motivate face recognition system designers to build algorithms that are robust to a minimum amount of misalignment that is unavoidable due to the ambiguity in eye locations.
The major contributions of this paper are:
\begin{enumerate}
 \item We reveal the sensitivity of five open source face recognition systems towards eye localization errors causing various types of misalignment as translation, rotation and scaling of the face.
We also show that unbiased and biased evaluation techniques -- HTER and AUC -- have different characteristics in evaluating query images with inaccurate eye locations.

 \item We explore the inherent limitations of facial image alignment schemes based on eye coordinates.
To the best of our knowledge, this is the first study to analyze the difference between two independent manual eye annotations carried on same set of images.
Our study shows an ambiguity of four pixels (with an average inter-ocular distance of $70$ pixels) in both horizontal and vertical location of manual eyes center annotations in frontal facial images.

 \item We show that the automatic eye detection system included in a commercial face recognition system achieves the eye detection accuracy of manual eye annotators.
Furthermore, we show that such a fairly accurate automatic eye detector can help to achieve recognition performance comparable to manually annotated eyes, given that the same automatic detector is used for annotating all training, enrollment and query images.

 \item Our work lays the foundation for reproducible research in this avenue of research by allowing the extension of our experiments with additional face recognition systems and other facial image data set.
The results of all experiments presented in this paper can be reproduced by using open source software\footnote{The scripts to reproduce the experiments that are described in this paper can be downloaded from \url{http://pypi.python.org/pypi/xfacereclib.paper.IET2015}} including a detailed description on how to regenerate the plot and tables.
Such an open platform encourages researchers to pursue similar studies with other face recognition systems or image databases.
We also aim to achieve reproducible research to allow the research community to reproduce our work on public databases and open source software hence excluding commercial systems that are not available to the large majority of researchers.
With our framework other researchers can test or proof their claimed stability against eye localization errors by simply re-running the experiments presented in this study using their algorithms.

\end{enumerate}

This paper is organized as follows: We review related studies about face recognition with eye localization errors in Section~\ref{sec:related_work}.
In Section~\ref{sec:methodology}, we describe our methodology to study the impact of misalignment on face recognition performance and to assess the accuracy of manual and automatic eye detectors.
Sections~\ref{sec:exp_impact_translation_rotation} and~\ref{sec:exp_impact_translation_rotation_scaling} describe the experiments designed to study the influence of misalignment (translation, rotation and scale) on face recognition performance.
In Section~\ref{sec:exp_ambiguity_eye_loc}, we analyze the ambiguity in manual and automatic eye annotations. Section~\ref{sec:exp_choice_eye_det} studies the impact of different sources of eye annotations for training/enrollment and query phases of a face recognition system.
In Section~\ref{sec:discussion}, we discuss the results from our experiments and, finally, Section~\ref{sec:conclusion} concludes the presented work.

\section{Related Work}
\label{sec:related_work}
Face normalization is a critical preprocessing stage of most 2D face recognition systems because subsequent processing stages like feature extraction and comparison rely on an accurate spatial alignment of facial features.
Therefore, other researchers have investigated the influence of this preprocessing stage on face recognition performance.

The impact of translation, rotation and scaling on the similarity value of an Eigenfaces~\cite{turk1991eigenfaces} algorithm was studied in~\cite{marques2000effects}.
There, the analysis was limited to a small data set (seven human subjects and one synthetic face) and a single face recognition algorithm.
Moreover, only the variation in the raw similarity score was studied, and not the actual face recognition performance.

The authors of~\cite{wang2005sensitivity} compare the performance of one Eigenfaces based and one commercial system for automatically detected and manually labeled eye annotations.
They also compare the accuracy of automatic eye detection under controlled and uncontrolled lighting conditions by considering the manual annotations as ground truth.
Their experimental results show that image quality (like illumination) affects face recognition performance in two ways: by contributing to misalignment caused by error in eye detection and by directly influencing a classifiers ability to extract facial features from misaligned images.
We do not include image quality variations in our study, but we perform experiments on a larger number of face recognition systems and use two automatic eye detectors.

In~\cite{min2005eye}, the authors perturb initial eye locations to generate multiple face extractions from a single query image and select a match using nearest neighbor scheme.
At the expense of an increased computational cost, they show that this approach consistently outperforms face recognition systems based on both manually located and automatically detected eye coordinates.
Hence, manual eye annotations do not always guarantee optimal face recognition performance.

The authors of~\cite{shan2004curse} show that the performance of a Fisherfaces~\cite{belhumer1997eigenfaces} based face recognition system drops drastically already under small misalignment.
They argue that ``eye locations are essentially ambiguous'' and even the manual eye annotations have large variance.
Therefore, they put forward a case to explicitly model the effects of misalignment during the training of the classifier.
To make a Fisherfaces based system robust to small misalignment, they deliberately introduce misaligned samples corresponding to each facial image during training.
They show that such a system has a higher tolerance towards misaligned query images.

The sensitivity of Eigenfaces, Elastic Bunch Graph Matching~(EBGM) and a commercial system's face recognition performance to misalignment was investigated by~\cite{riopka2003eyes}.
They systematically perturb manual eye annotations and report performance variation for three face recognition systems.
They find that misalignment caused by scaling have a higher influence on face recognition performance than rotation or translation.

The authors of \cite{wang2005automatic} propose an automatic eye detection system and report that their automatically detected eye coordinates achieve face recognition performance comparable to manually annotated eye coordinates on the FRGC 1.0 data set.
However, it is not clear which eye coordinates, manual or automatically detected, were used for training the face recognition systems.
Furthermore, their analysis is only limited to Eigenfaces and Fisherfaces baseline face recognition systems.
In this paper, we report performance variations for five face recognition systems, which span a wide range from classic systems to state-of-the-art systems of different kinds.

Based on the analysis of similarity scores, the authors of \cite{wang2007modeling} build a model to predict recognition performance for a set of probe images.
They use this model to predict the performance corresponding to probe sets built from different perturbed eye coordinates.
This allows them to adjust the image alignment in order to attain best possible recognition performance.
They conclude that manually located eye coordinates do not necessarily provide the best alignment and that performance prediction systems can be used to select the best alignment that can outperform systems based on manually located eye coordinates.
However, their analysis is limited to a single Eigenfaces based face recognition system.
Furthermore, it is not clear which eye coordinates (manual or automatic) they use for training.

Some face recognition systems are more tolerant to misalignment.
Therefore, the authors of \cite{rodriguez2006measuring} argue that performance of a face localization system should be evaluated with respect to the final application, \eg, face verification.
They build a model that directly relates face localization errors to verification errors.
It is possible to build such a model using the data reported in this paper.
However, in this paper, we only aim to compare the tolerance of different face recognition systems towards misalignment.
Therefore, we do not build such a model and only report verification errors parametrized for the following two types of localization errors: (a) errors involving only translation and rotation (without scaling) and (b) errors belonging to a normal distribution of landmark locations with variable variances.

More recently, there have been efforts to build face recognition algorithms that are more robust to misalignment.
In \cite{wagner2012towards}, the authors highlight the importance of image alignment for correct recognition and propose a modification to the sparse representation-based classification algorithm such that it is more robust to misalignment.
Similarly, by extracting features from parts of the facial image independently, the authors of \cite{wallace2011intersession} achieve natural robustness to misalignment -- a claim that is also validated by the experiment results presented in this paper.
In the pursuit for robustness against misalignment, several works like \cite{ekenel2009face}, \cite{shan2004curse} and \cite{min2005eye} train the classifier on multiple instances of same image cropped using perturbed eye coordinates.
At the expense of increased computational cost, they achieve some performance improvement in handling misaligned images.

In this paper, we study the impact of misalignment caused by errors in eye localization on the verification performance of face recognition systems.
Similar studies carried out in the past were either limited by the number of face recognition systems or the size of facial the image database.
Our study is based on five open source face recognition algorithms using an unbiased verification protocol based on a much larger data set containing images of $272$ subjects ($208$ in training and $64$ in development set).
Furthermore, our experiments are solely based on open source software and our results are reproducible and extensible to include more algorithms and facial image databases in the future.

\section{Methodology}
\label{sec:methodology}
We want to study the influence of eye annotation error on face recognition performance.
Therefore, we keep all other parameters (like training and enrollment images, algorithm configuration, etc.) fixed and only perturb the manual eye annotations of the query images.
This allows us to analyze the contribution of misalignment caused by eye annotation error in degrading the performance of a face recognition system.

We study the difference between manual eye annotations performed independently at two institutions.
Furthermore, we compare the accuracy of automatic eye detectors by considering manual eye annotations as ground truth.
This analysis reveals the ambiguity present in the location of the eyes for both manual annotators and automatic eye detectors.

\subsection{Face recognition systems}
\label{sec:methodology_frs}
We evaluate the performance of five open source face recognition algorithms implemented in the \texttt{FaceRecLib}~\cite{gunther2012open}, which itself is built on top of the open source signal-processing and machine learning toolbox \texttt{Bob}~\cite{anjos2012bob}.
In our experiments, we extend the \texttt{FaceRecLib} to allow systematic perturbation of the, otherwise fixed, eye coordinates in the normalized images.
The five systems considered are listed and succinctly described below.
These algorithms were chosen for the following two reasons:
(a) the recognition performance of these algorithms span from baseline performance (Eigenfaces) to state-of-the-art face recognition performance (ISV), and
(b) their open source implementation is available in a single stable package called \texttt{FaceRecLib}, which allows to reproduce and extend the results presented in this paper.
In our experiments we use the stock implementations of the algorithms with their default configuration, we do not adapt any parameter to the database or the experimental setup.
For a more detailed description of the algorithms, please refer to the cited papers, or have a look into their implementations.\footnote{The latest stable version of the \texttt{FaceRecLib} can be downloaded from \url{http://pypi.python.org/pypi/facereclib}}

The first algorithm we investigate is the well-known Eigenfaces~\cite{turk1991eigenfaces} algorithm.
It uses raw pixel gray values as features by concatenating all pixels of the normalized image to one vector.
It extracts a basis of eigenvectors from a training set of samples using a Principal Component Analysis~(PCA).
For feature extraction, the dimensionality of samples is reduced by projecting them into this basis.
Finally, classification in this lower dimensional space is performed using the Euclidean distance.

Fisherfaces~\cite{belhumer1997eigenfaces} use a similar approach to face recognition.
This system first extracts a basis of eigenvectors from a training set of samples using a combination of PCA and Linear Discriminant Analysis~(LDA) techniques.
As for Eigenfaces, the dimensionality of the samples is reduced by projecting them into this basis, and classification is performed in the exact same way.

A more complex strategy for recognizing faces is given in the Gabor-Jet~\cite{guenther2012disparity} algorithm.
Here, responses of several Gabor wavelets are computed on a set of points located on a grid on the original image.
At each point, a descriptor is obtained by concatenating the response values, which is referred to as a Gabor-Jet.
The comparison of these Gabor-Jets is then performed using a similarity function that is based on both absolute values and phase values of Gabor wavelet responses.
Please note that this algorithm requires no training.

The Local Gabor Binary Pattern Histogram Sequences~(LGBPHS)~\cite{zhang2005local} algorithm combines two different feature extraction techniques.
It first extracts Gabor filtered images from the original sample, before applying the Local Binary Pattern~(LBP) operator on them.
Each of this filtered images is decomposed into overlapping blocks, from which local LGBP histograms are gathered.
Finally, all local histograms are concatenated to form the LGBPHS, and classification is performed using the histogram intersection measure.
As for Gabor-Jet, this algorithm requires no training either.

A completely different approach to face recognition is given by the Inter-Session Variability modeling (ISV)~\cite{wallace2011intersession}.
This generative parts-based algorithm models the distribution of local DCT block features using Gaussian Mixture Models~(GMMs), where many DCT features are extracted independently from overlapping image blocks.
At training time, a Uniform Background Model~(UBM) is learnt from a set of samples from several identities, as well as a subspace that describes the variability caused by different recording conditions (session variability).
At enrollment time, for each client a specific GMM is computed by adapting the UBM to the enrollment samples of the client.
In particular, ISV enrolls these models by suppressing session-dependent components to yield true session-independent client models.
Finally, classification relies on the computation of log-likelihood ratios.

\subsection{Image database and evaluation protocol}
\label{sec:methodology_img_db_protocol}
For all the experiments discussed in this paper, we chose the training and development set according to the protocol\footnote{An open source implementation of the protocols for the Multi-PIE database is available at \url{http://pypi.python.org/pypi/bob.db.multipie}} \texttt{M} of the Multi-PIE data set~\cite{gross2008multipie}.
The full Multi-PIE data set contains images of $337$ subjects captured in four sessions with various variations in pose, illumination and expression.

This unbiased face verification protocol \texttt{M} is defined as follows: The training set contains those $208$ subjects that do not appear in all four sessions.
The development and evaluation sets contain $64$ and $65$ disjoint subjects, respectively, all of which are not included in the training set.
The enrollment images are taken from session $01$, while query images stem from sessions $\{02,03,04\}$, the pose is fixed to the frontal camera \texttt{05\_1} with no flash \texttt{00}.
Hence, the training set consists of $515$ images, while the development set contains $1$ enrollment image per subject and $256$ query images of the same $64$ subjects, where all enrolled models are compared with all query samples.
To keep our investigations unbiased, we report results only for the development set and do not perform any experiment on the evaluation set.

\subsection{Performance measures}
\label{sec:methodology_perf_measures}
To evaluate the face verification performance, we use the False Acceptance Rate~(FAR) and the False Rejection Rate~(FRR).
For a given similarity score threshold $s_t$, these metrics are defined as follows:

\begin{equation}
 \textrm{FAR}(s_t) = \frac{| \{ s_{imp} | s_{imp} \geq s_t\} |}{| \{s_{imp}\} |},\quad \textrm{FRR}(s_t) = \frac{|\{s_{gen} | s_{gen} < s_t\}|}{| \{s_{gen}\} |}, \\
\end{equation}
where $s_{gen}$ and $s_{imp}$ denote genuine (same source comparison) and impostor (different source comparison) scores, respectively.
In this paper, we use two evaluation metrics, both of which are based on FAR and FRR.
The first metric is the Half Total Error Rate~(HTER).
Let $s_{t}^{*}$ denote the threshold for development set, without any eye perturbations, such that:
\begin{equation}
 s_{t}^{*} = \argmin_{s_t} \frac{\textrm{FAR}(s_t) + \textrm{FRR}(s_t)}{2}\,,
\end{equation}
then HTER with perturbed eye locations is defined as:
\begin{equation}
 \textrm{HTER}_{(\theta,t_X,t_Y)} = \frac{\textrm{FAR}_{(\theta,t_X,t_Y)}(s_{t}^{*}) + \textrm{FRR}_{(\theta,t_X,t_Y)}(s_{t}^{*})}{2}\,,
\end{equation}
where $(\theta,t_X,t_Y)$ are the rotation and translation parameters as defined in Section~\ref{sec:exp_impact_translation_rotation}.
Note that a perfect system has an HTER of $0.0$, while the HTER of a random system is $0.5$.

The second evaluation metric is the Receiver Operating Characteristics~(ROC), where the Correct Acceptance Rate~(CAR) with $\textrm{CAR}=1.0-\textrm{FRR}$ is plotted against the FAR.
Usually, we are interested in the CAR values at low FAR values and, therefore, we plot the FAR axis in logarithmic scale.
Additionally, the ROC can be characterized by a single number called the Area Under Curve~(AUC), which -- as the name implies -- measures the region covered by the ROC curve.
The AUC can be approximated as:
\begin{equation}
 \textrm{AUC} = \sum_{i=1}^{n-1} (F[i+1] - F[i]) \left(\frac{C[i] + C[i+1]}{2} \right),
 \label{eq:auc_eqn}
\end{equation}
where $C[i]$ denotes the CAR value corresponding to FAR of $F[i]$ and the $n$ values in $F$ are sorted in ascending order.
A perfect verification system has an AUC of $1.0$.

Though both measures HTER and AUC are based on the same FAR and FRR, they have different characteristics.
The AUC measures performance directly using the perturbed scores, which makes this measure biased.
A more realistic and unbiased measure is given by the HTER, which defines a threshold $s_t$ using clean conditions, but measures performance in perturbed conditions.

\subsection{Measures of misalignment}
\label{sec:methodology_misalignment}
\begin{figure}[t!]
 \centering
 \includegraphics[width=0.98\linewidth]{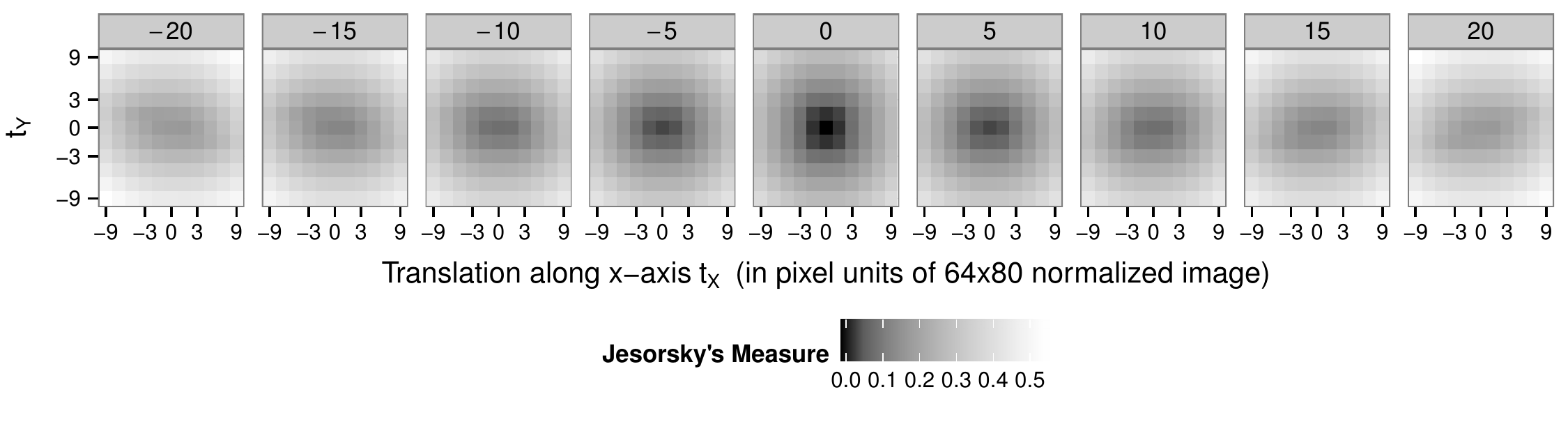}
 \caption{Relationship between the Jesorsky measure and annotation transformation $(\theta,t_X,t_Y)$ when same transformation is applied to both eye coordinates for the rotation $\theta$ varying between $-20^\circ$ and $20^\circ$ and translation $t_X,t_Y$ between $-9$ and $9$ pixels.}
 \label{fig:jerosky_theta_tx_ty_map}
\end{figure}

The authors of \cite{jesorsky2001robust} have proposed the Jesorsky measure of eye annotation misalignment, which is defined as follows:

\begin{equation}
  J = \frac{\max\{\| p_{l}^{m} -p_{l}^{d} \|, \| p_{r}^{m} - p_{r}^{d} \|\}}{\|p_{l}^{m} - p_{r}^{m}\|},
 \label{eq:jesorsky_measure}
\end{equation}
where $p_{\{l,r\}}^{m}$ denote manually annotated left and right eye coordinates, while $p_{\{l,r\}}^{d}$ denote automatically detected eye coordinates (as defined in Table~\ref{tbl:list_of_symbols}).
In \figurename~\ref{fig:jerosky_theta_tx_ty_map}, we show the correspondence between the Jesorsky measure $J$ and the transformation parameters $(\theta,t_X,t_Y)$ (see next section), when same transformation is applied to both eye coordinates. This measure of misalignment cannot differentiate between errors caused by translation, rotation or scale.
This is also evident from \figurename~\ref{fig:jerosky_theta_tx_ty_map}, which shows that multiple transformation parameters map to same $J$ value.
Therefore, in this paper we quantify the amount of misalignment in the normalized image space using translation $(t_X,t_Y)$ and rotation $\theta$ parameters.

Misalignment in the original image space is defined as the difference between the manual eye annotations and automatically detected eye locations.
However, we report misalignment in units of the normalized image because misalignment in the original image depends on the inter-ocular distance, which in turn varies with image resolution.
For the normalized image, the inter-ocular distance remains fixed and, therefore, our results are not affected by resolution of the original image.
Here, we establish the relationship between misalignment in the original image space and misalignment in the normalized image space.
Such a relationship allows us to express eye detection errors in units of the normalized image space and, therefore, a comparison with the results presented in Section~\ref{sec:exp_impact_translation_rotation} and \ref{sec:exp_impact_translation_rotation_scaling} is possible.

\begin{table}
 \centering
 \small
 \caption{List of symbols}
 \begin{tabular}{l l}
 $(x,y)$ & Cartesian coordinate in original image space (also in subscripts) \\
 $(X,Y)$ & Cartesian coordinate in normalized image space (also in subscripts) \\
 $p$ & position vector $[x,\; y]$ in original image space \\
 $P$ & position vector $[X,\; Y]$ in normalized image space \\
 $c$ & center between the two eyes in original image space \\
 $C$ & center between the two eyes in normalized image space \\
 $*^{\{m,d\}}$ & superscripts to denote \emph manually and automatically \emph detected eyes\\
 $*_{\{l,r\}}$ & subscripts to denote \emph left and \emph right eye
 \end{tabular}
 \label{tbl:list_of_symbols}
\end{table}

Geometric normalization of facial images involves scaling and rotating the original image space $p$ such that the manually annotated left and right\footnote{with respect to the viewer} eyes in the original image $p_{\{l,r\}}^{m}$ get transformed to a predefined location $P_{\{l,r\}}^{m}$ in the normalized image:

\begin{equation}
 P = \frac{1}{s} R_{-\alpha} (p - c) + C
 \label{eqn:geom_norm_eq}
\end{equation}
where scale $s$, rotation angle $\alpha$ and rotation matrix $R_{\alpha}$ are defined as:
\begin{align}
 s  &= \frac{\| P_{r}^{m} - P_{l}^{m} \|}{\| p_{r}^{m} - p_{l}^{m} \|},&
 \alpha &= \tan^{-1} \left( \frac{y_r - y_l}{x_r - x_l} \right),&
 R_{\alpha} &=
  \left[\!\begin{array}{rr} \cos \alpha & -\sin \alpha \\ \sin \alpha & \cos \alpha \end{array}\!\right].
  \label{eqn:geom_norm_eq_val}
\end{align}
Using the coordinate space transformation of \eqref{eqn:geom_norm_eq}, we transform the original image and extract a predefined region around $P_{\{l,r\}}^{m}$ to obtain the final geometrically normalized facial image.
For our experiments, we use $P_{l}^{m}=[15,16]$ and $P_{r}^{m}=[48,16]$ in a normalized image of dimension $64 \times 80$.
An exemplary normalized image including the locations of the normalized eye positions $P_{l,r}$ can be found in Figure~\ref{fig:facecrop_illustration}.

In this paper, we also study the accuracy of automatic eye detectors with respect to manual eye annotations.
We report face recognition performance results parametrized by eye position error in the normalized image space.
The method to convert eye detection errors from pixel units in the original image to errors in the normalized image is as follows:
We first compute scale $s$ and rotation $\alpha$ using the manually annotated eye coordinates $p^{m}$ in the original image and predefined eye locations $P^{m}$ in the normalized image as given in \eqref{eqn:geom_norm_eq_val}.
Using \eqref{eqn:geom_norm_eq}, we transform the automatically detected eye coordinate $p^{d}$ in original image space to obtain its position $P^{d}$ in the normalized image space.
We do not differentiate between errors in left and right eye coordinates and, therefore, define the eye detection error in normalized image space as:
\begin{align}
 \Delta X &= X^d - X^m & \Delta Y &= Y^d - Y^m,
 \label{eqn:misalignment_delXY}
\end{align}
where $(\Delta X,\Delta Y)$ denote the difference between manually annotated and automatically detected eye coordinates in the normalized image space.

\section{Experiments}
\label{sec:experiments}
In our experiments, we quantify misalignment in units of normalized image space $(\Delta{X},\Delta{Y})$ as defined in \eqref{eqn:misalignment_delXY} and evaluate their influence on face recognition performance in Sections~\ref{sec:exp_impact_translation_rotation} and \ref{sec:exp_impact_translation_rotation_scaling}.
Note that we only perturb the eye coordinates of query images, while for training and enrollment images, we use the manually annotated eye coordinates provided by~\cite{shafey2013scalable}.\footnote{The manual eye coordinates can be downloaded from \url{http://www.idiap.ch/resource/biometric}, they are also included in our source code package \url{http://pypi.python.org/pypi/xfacereclib.paper.IET2015}.}
In Section~\ref{sec:exp_ambiguity_eye_loc}, we study the variability between two independently performed manual eye annotations.
Furthermore, in Section~\ref{sec:exp_choice_eye_det}, we  study the accuracy of automatic eye detectors by considering manual eye annotations as ground truth for eye locations.
In Section~\ref{sec:discussion}, we present a list of key observations from all these experiments.

\subsection{Impact of Translation and Rotation}
\begin{figure}[t!]
 \centering
 \subfloat{\includegraphics[width=0.2\linewidth]{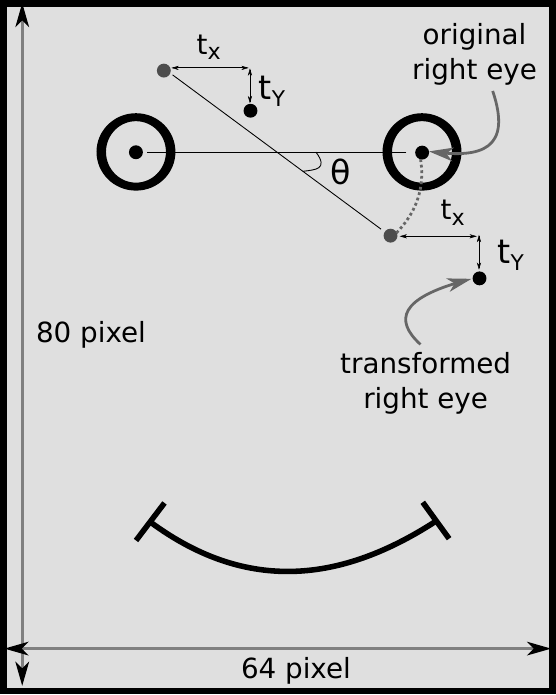}}
 $\qquad$
 \subfloat{
  \includegraphics[width=0.6\linewidth]{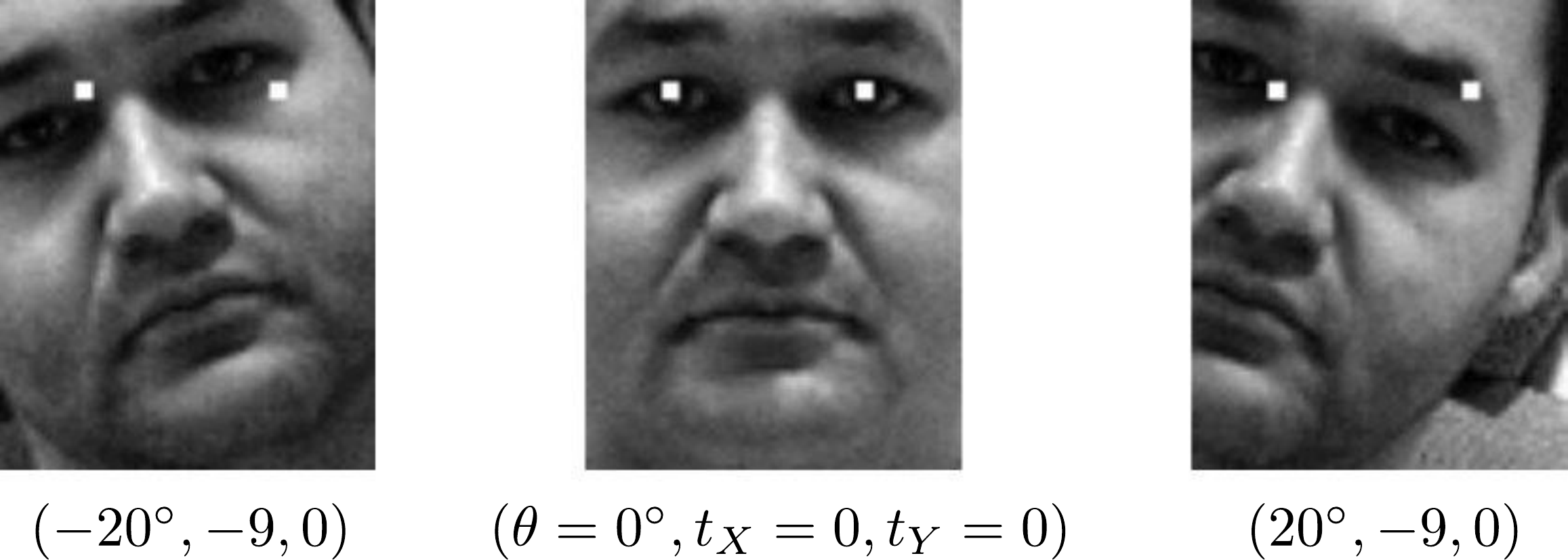}
 }
 \caption{Rotation (about center between the eyes) followed by translation of query images in the normalized image space. The two white dots denote the untransformed location of left and right eyes in the normalized image.}
 \label{fig:same_eye0_eye1_illustration}
\end{figure}

\label{sec:exp_impact_translation_rotation}
In our first experiment, we systematically rotate $(\theta)$ and translate $(t_X,t_Y)$ the hand-labeled eye coordinates $P_{\{l,r\}}^{m}$ in normalized image (depicted as two white dots in \figurename~\ref{fig:same_eye0_eye1_illustration}) to simulate misalignment.
We apply the same transformation $(\theta,t_X,t_Y)$ to both eye coordinates $P_{\{l,r\}}^{m}$ and, hence, the size of the faces in the misaligned facial images is not varied.
The perturbed eye coordinates $\mathcal{P}_{\{l,r\}}^{m}$ in the normalized image are computed as follows:
\begin{equation}
 \mathcal{P}_{\{l,r\}}^{m} = T(t_X,t_Y)T(C)R(\theta)T(-C)P_{\{l,r\}}^{m}
\end{equation}
where $C$ denotes the coordinate of the center of two eyes $P_{\{l,r\}}^{m}$, while $T$ and $R$ denote the translation and rotation operator, respectively.


\begin{figure}[p!]
 \centering
 \subfloat[Half Total Error Rate]{\label{fig:fixed_hter_theta_tx_ty} \includegraphics[width=0.98\linewidth]{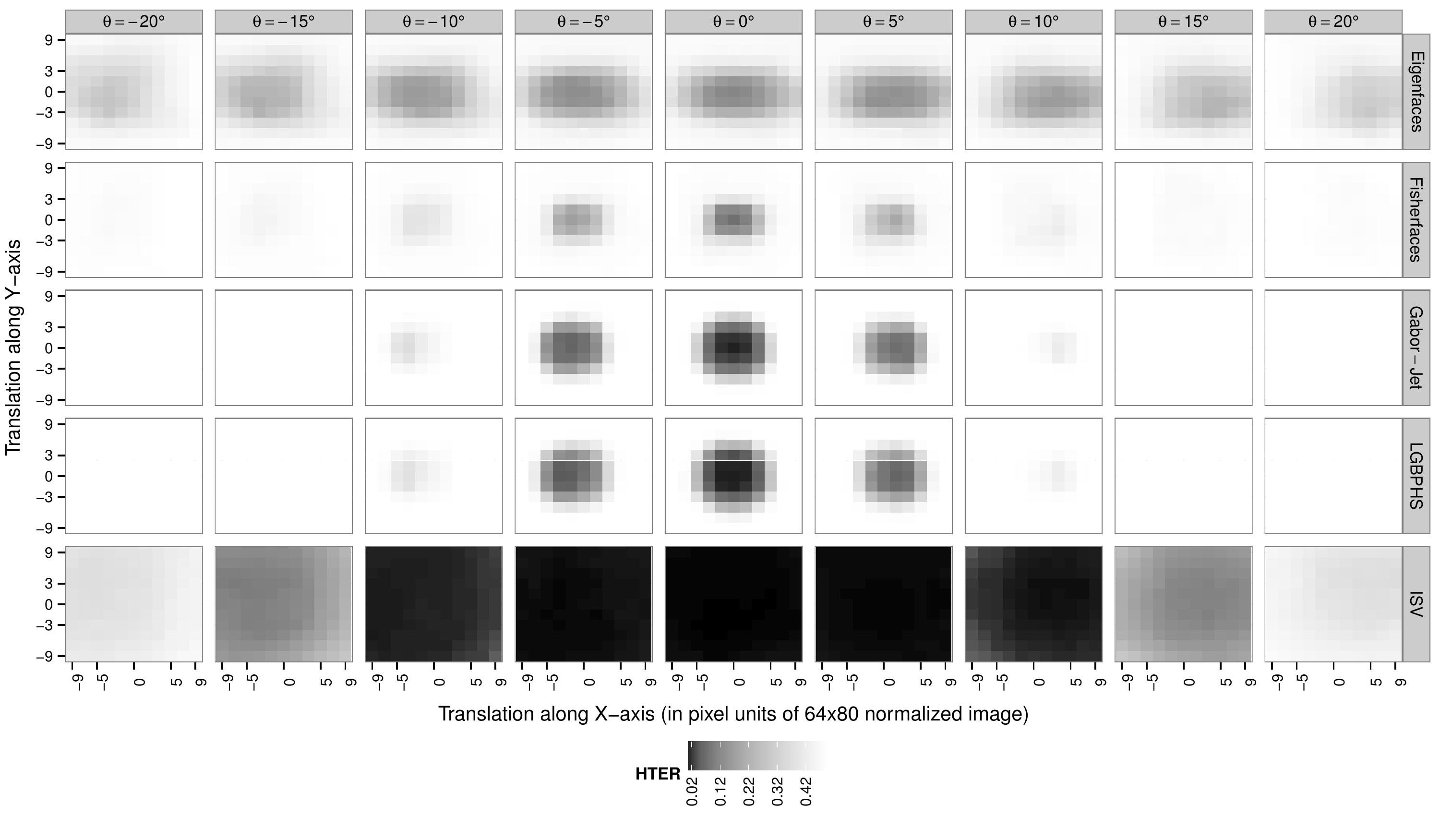}}

 \subfloat[Area Under Curve]{\label{fig:fixed_auc_theta_tx_ty} \includegraphics[width=0.98\linewidth]{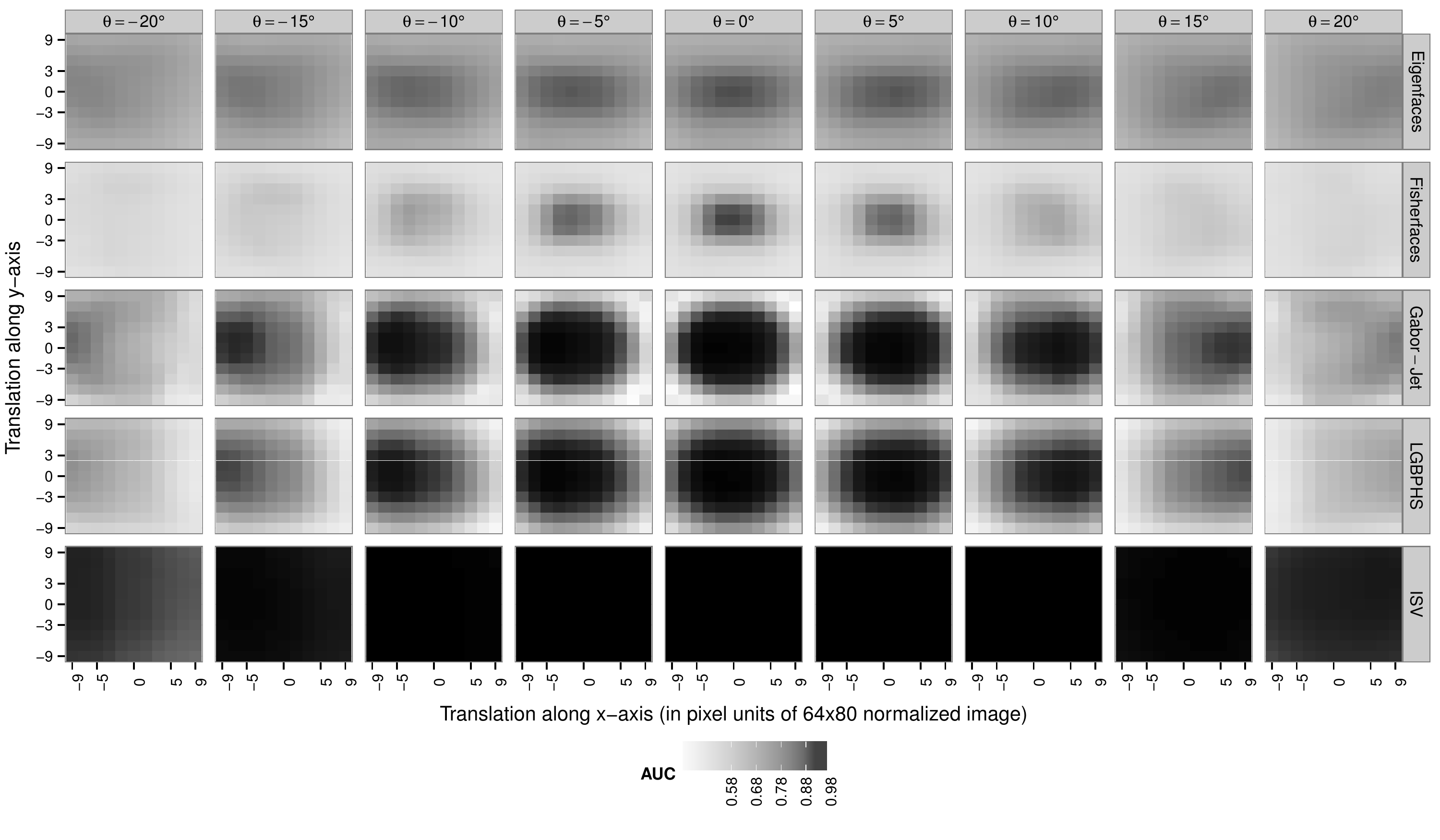}}
 \caption{Impact of same rotation and translation applied to both left and right eye coordinates on performance of five face recognition systems (along rows). Each cell denotes recognition performance for the query images misaligned by the applying the transformation of $(\theta,t_X,t_Y)$ to manually annotated eye coordinates.}
 \label{fig:fixed_performance_theta_tx_ty}
\end{figure}

In \figurename~\ref{fig:fixed_performance_theta_tx_ty}, we report the recognition performance for all possible variations of $(\theta,t_X,t_Y)$:
\begin{equation}
  \begin{split}
    \theta &\in \{-20^{\circ}, -15^{\circ}, -10^{\circ}, -5^{\circ}, 0^{\circ}, 5^{\circ}, 10^{\circ}, 15^{\circ}, 20^{\circ}\}\\
    t_X,t_Y &\in \{-9, -7, -5, -3, -1, 0, 1, 3, 5, 7, 9\}
  \end{split}
\end{equation}
in terms of HTER and AUC.
For HTER, the threshold $s_{t}^{*}$ is computed on the basis of untransformed images, \ie, for $\theta=0,t_X=0,t_Y=0$.

For the Eigenfaces system, which is the first row in \figurename~\ref{fig:fixed_performance_theta_tx_ty}, translation along vertical direction ($t_Y$) has more impact on performance as compared to horizontal translation ($t_X$).
For small angles $-5^{\circ} \leq \theta \leq +5^{\circ}$ and small translations $-5 \leq t_X \leq +5$ and $-3 \leq t_Y \leq +3$, the Eigenfaces algorithm has stable (though comparably low) performance.

Fisherfaces has better recognition performance as compared to Eigenfaces for aligned facial images.
However, its performance drops more rapidly for small misalignment.
For instance, the HTER of Fisherfaces increases from $0.08$ (AUC $=0.95$) to $0.39$ (AUC $=0.80$) for a horizontal translation of $t_X=3$ (with $t_Y=0,\theta=0$).
For the same misalignment, the HTER of Eigenfaces increases from $0.12$ (AUC $=0.94$) to $0.30$ (AUC $=0.87$).
The authors of~\cite{martinex2001pca} have shown that ``when the training data set is small, PCA [Eigenfaces] can outperform LDA [Fisherfaces]''.
Our results highlight another property of Eigenfaces, that it is more robust to misalignment in the input image.

Both Gabor-Jet and LGBPHS achieve an HTER of $0.01$ (AUC $>0.99$) for properly aligned images.
For a horizontal misalignment of $3$ pixels ($t_X=3,t_Y=0,\theta=0$), the HTER of Gabor-Jet increases to $0.08$ and of LGBPHS to $0.04$, while the AUC is stable (AUC $>0.99$).
From \figurename~\ref{fig:fixed_performance_theta_tx_ty}, it is clear that both Gabor-Jet and LGBPHS have similar tolerance towards misalignment and both of them have higher recognition performance and a better robustness towards misalignment than Eigenfaces and Fisherfaces.

From the five recognition systems included in this study, ISV clearly has the best tolerance towards misalignment for all possible combinations of rotation and translation.
For properly aligned images, ISV achieves HTER of $0.00018$ (AUC $>0.99$).
For a 3 pixel horizontal misalignment ($t_X=3$, $t_Y=0$, $\theta=0$), the HTER increases only to $0.00217$ (AUC $>0.99$).
In ISV, features are independently extracted from each part of the facial image and, therefore, this approach is naturally robust to face misalignment, occlusion and local transformation.
It also explicitly models and removes the effect of session variability, \eg, changes in environment, expression, pose and image acquisition.
Only for an extreme misalignment of ($t_X=9,t_Y=9,\theta=20^{\circ}$), the HTER grows to $0.39$ (which is close to chance level $0.5$), but the AUC $=0.98$ still shows very good discrimination abilities.
From this effect one can infer that the similarity values change with the transformation, but for both genuine and impostor accesses in the same way.
One way to improve the unbiased HTER in this case is given by categorical calibration \cite{mantasari2014calibration}.

\begin{figure}[t!]
  \centering
  \includegraphics[width=0.8\linewidth]{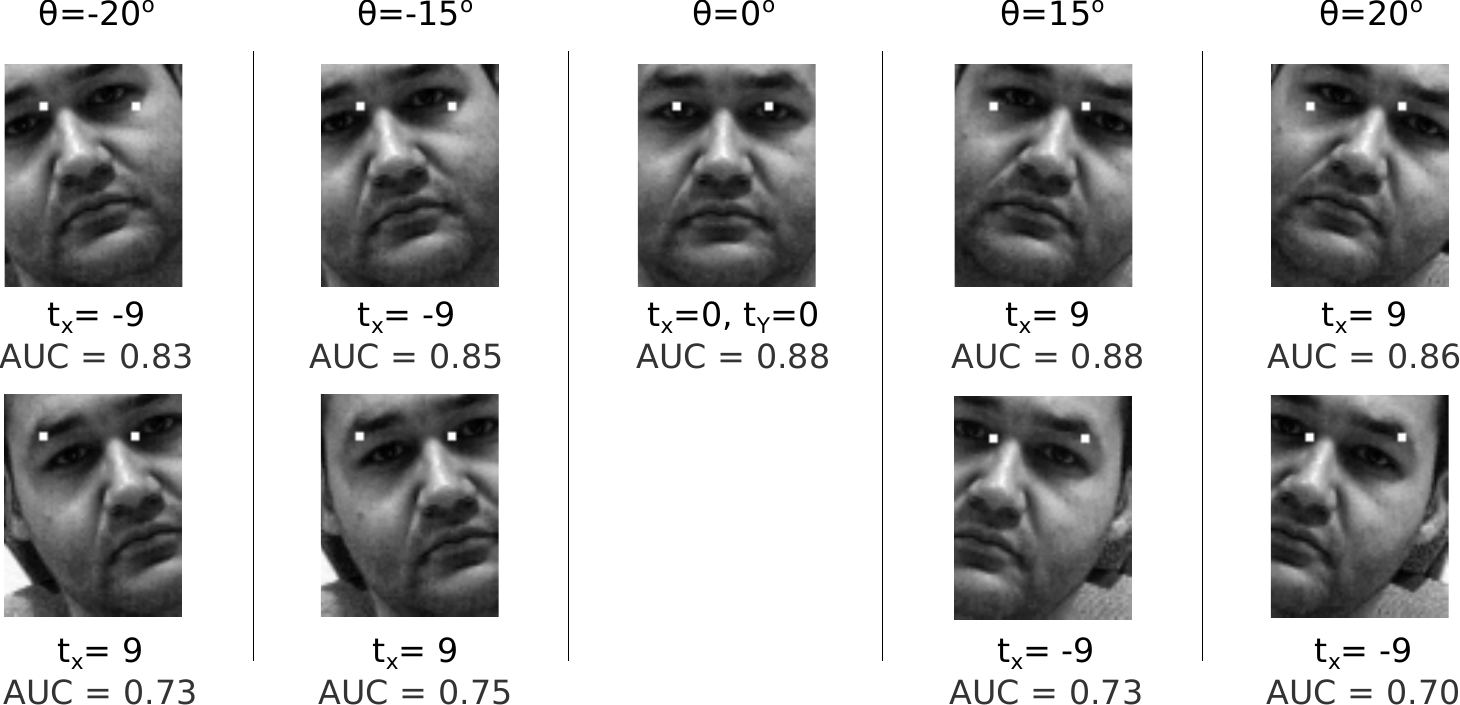}
  \caption{Examples of transformations with large $(t_X,t_Y,\theta)$ parameters that cause low (first row) and high (second row) misalignment of facial features. The AUC values of the Eigenfaces algorithm for those transformations are added to the plots.}
  \label{fig:facecrop_theta_tx_ty_viz}
\end{figure}

For larger rotations ($\theta \geq \pm 10^\circ$), we can observe that the highest recognition performance deviates from the translation center ($t_X=0,t_Y=0$).
This effect can best be seen in the AUC plots for Gabor-Jet and LGBPHS in \figurename~\ref{fig:fixed_performance_theta_tx_ty}.
To investigate this behavior, we display exemplary images of a subject cropped under these eye perturbations in \figurename~\ref{fig:facecrop_theta_tx_ty_viz}.
We observe that some transformations, which are shown in the first row of figure, cause less misalignment of facial features like nose tip, mouth center, etc. and, hence, the performance drop is small.
On the contrary, some transformations introduce large amount of misalignment, which even might lead to facial features being outside of the cropped image, as shown in the second row of \figurename~\ref{fig:facecrop_theta_tx_ty_viz}.
For those transformations, the recognition performance severely degrades.

\subsection{Impact of Translation, Rotation and Scaling}
\label{sec:exp_impact_translation_rotation_scaling}

\begin{figure}[t!]
 \centering
 \subfloat[Sampled eye perturbations]{\includegraphics[width=0.45\linewidth]{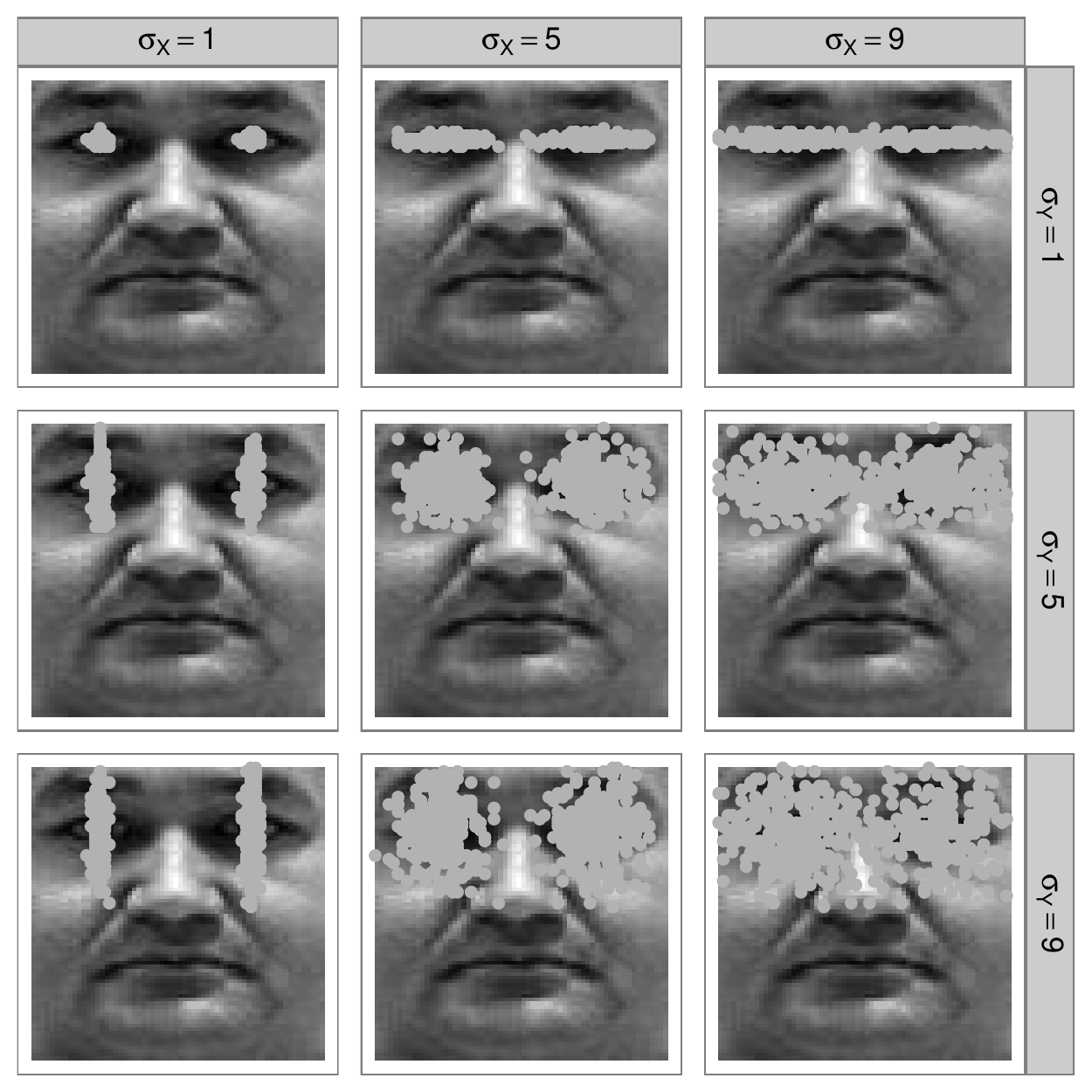}} \hspace*{.03\linewidth}
 \subfloat[Images after Transformation]{\includegraphics[width=0.45\linewidth]{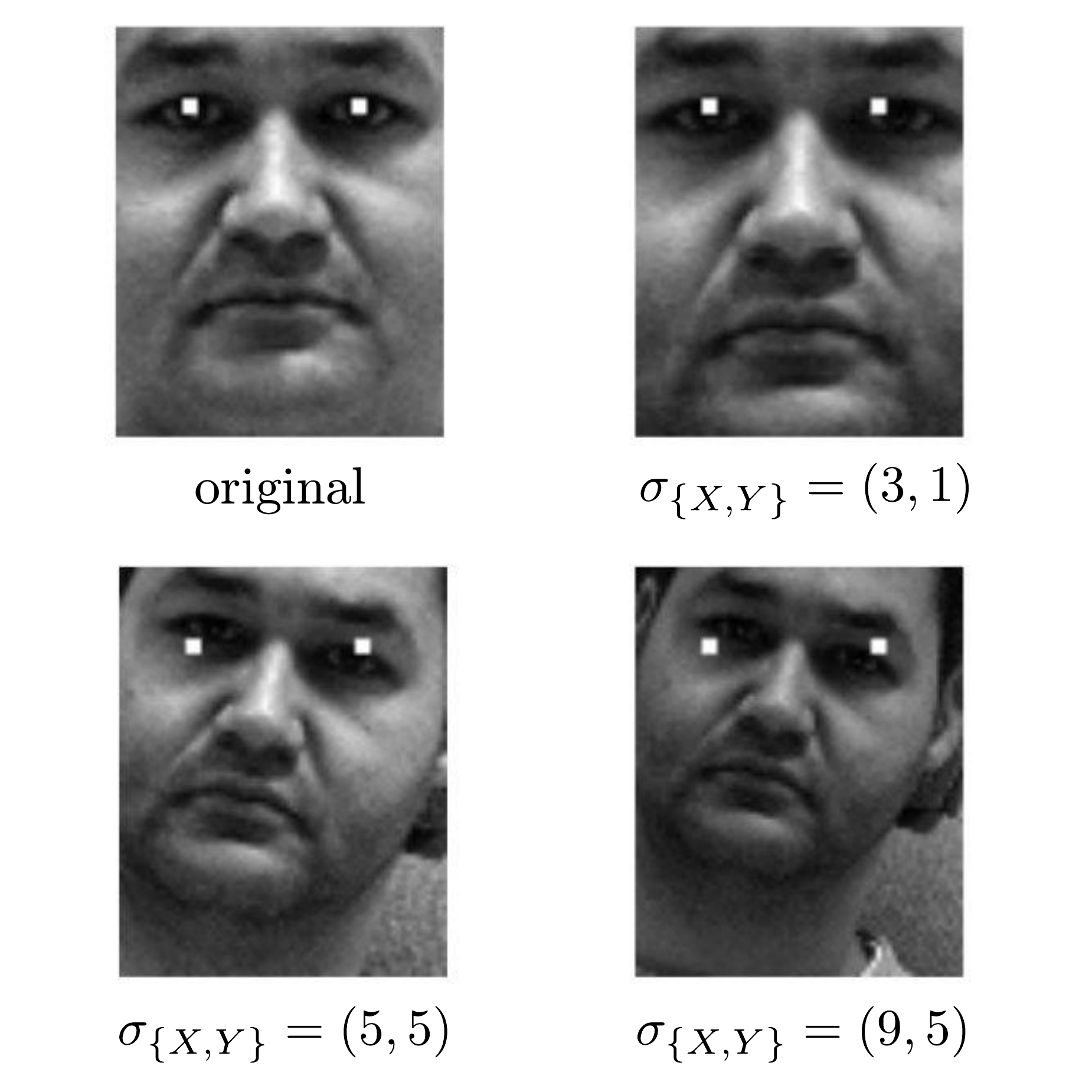}}
 \caption{Random transformation applied to left and right eye coordinates independently, where random samples are drawn from a normal distribution with $\mu_{\{X,Y\}}=0$.}
 \label{fig:rand_eye0_eye1_illustration}
\end{figure}

In the experiments of Section~\ref{sec:exp_impact_translation_rotation}, we apply the same transformation to both eyes in the normalized image.
This eye coordinate transformation strategy can only simulate rotation and translation error, but not the error caused by scaling.
Typically, in a practical automatic eye detector, all three types of transformations are present.
Therefore, in order to simulate the misalignment caused by an automatic eye detector, we independently apply random translation to the left and right eye coordinates as follows:
\begin{align}
   \mathcal{X}^{m} &= X^{m} + \epsilon_{X} \,,&
   \mathcal{Y}^{m} &= Y^{m} + \epsilon_{Y} \,,
\end{align}
where $\epsilon_{\{X,Y\}}$ follows the normal distribution $\mathcal{N}(\mu=0,\sigma_{\{X,Y\}})$, and $\mathcal{P}^m = (\mathcal{X}^{m}, \mathcal{Y}^{m})$ denotes the perturbed eye coordinates of the normalized image.
During random sampling, we discard all samples that move the eye coordinate location beyond the boundary of the normalized image ($64 \times 80$).
In \figurename~\ref{fig:rand_eye0_eye1_illustration}, we show the randomly perturbed eye locations superimposed on a sample facial image.
Additionally, we display exemplary normalized images obtained from such a random eye transformation scheme.
It clearly shows the misalignment caused by all three types of transformations: translation, rotation and scaling.

\begin{figure}[p!]
 \centering
 \subfloat[Half Total Error Rate]{\label{fig:rand_hter_theta_tx_ty} \includegraphics[width=0.76\linewidth]{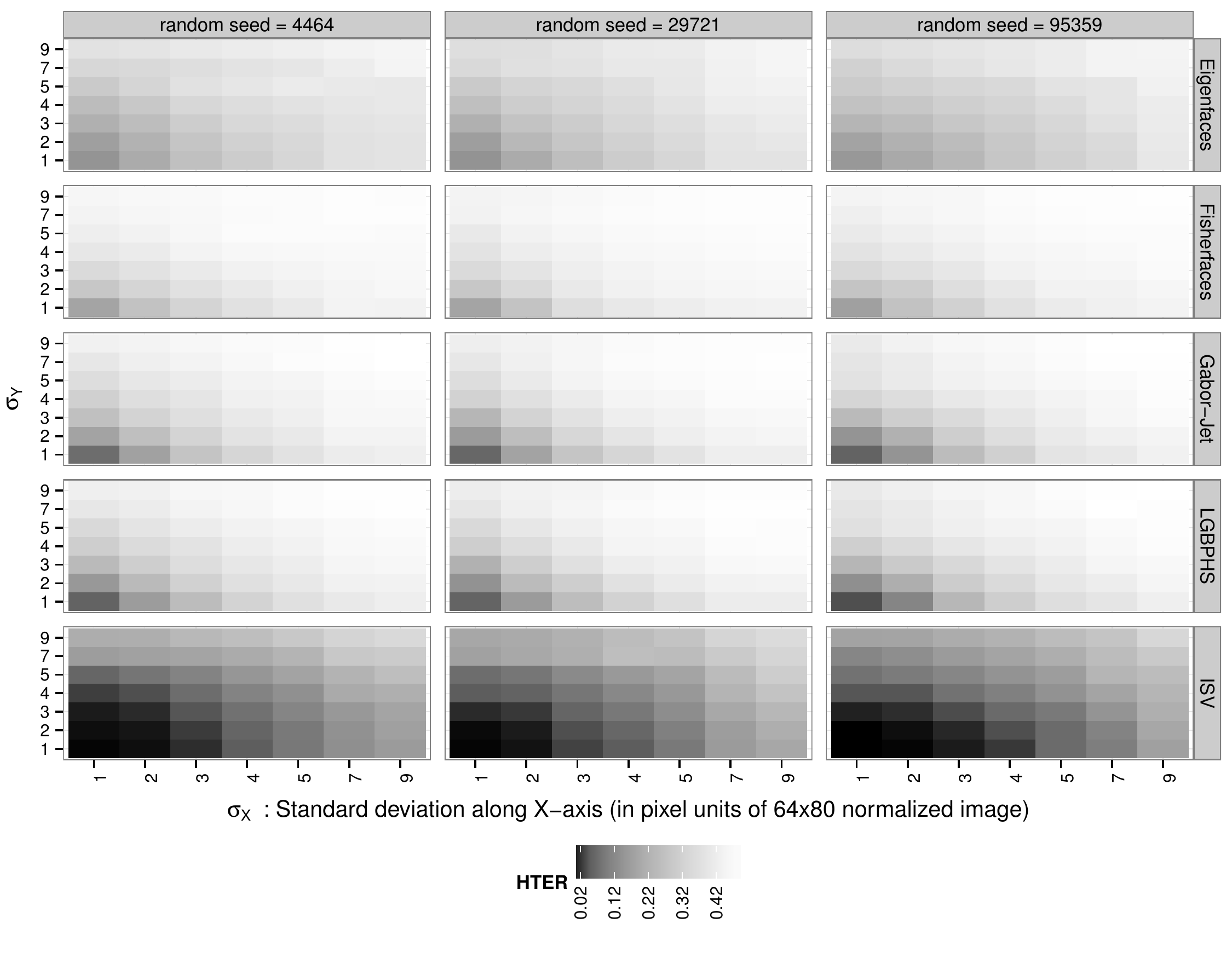}}

 \subfloat[Area Under Curve]{\label{fig:rand_auc_theta_tx_ty} \includegraphics[width=0.76\linewidth]{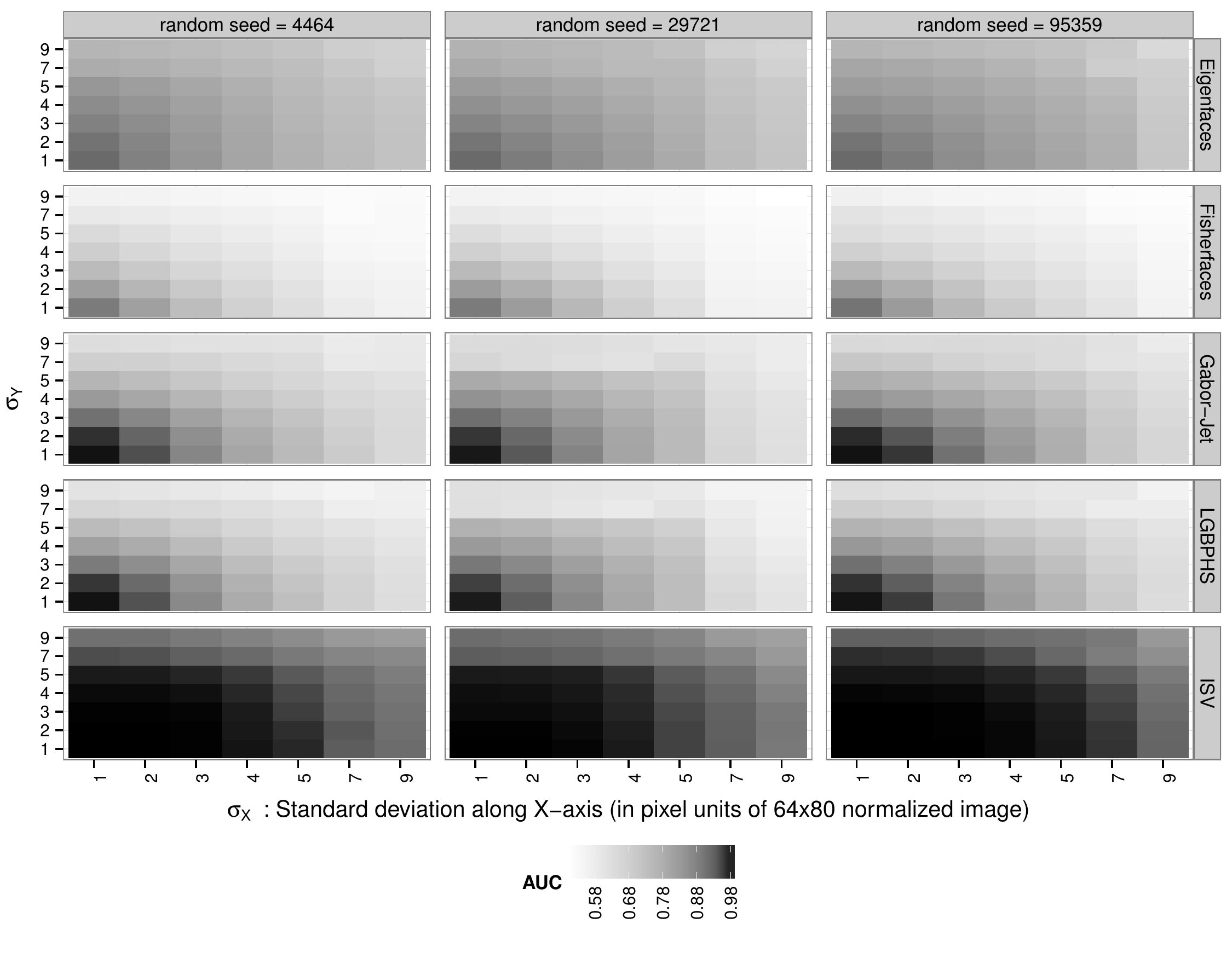}}

 \caption{Impact of random eye perturbations applied to left and right eye coordinates independently on performance of five face recognition systems (along rows). Random perturbations are sampled from a normal distributions with $(\mu=0, \sigma_{\{X,Y\}})$.}
 \label{fig:rand_performance_theta_tx_ty}
\end{figure}

In \figurename~\ref{fig:rand_performance_theta_tx_ty}, we report the face recognition performance corresponding to each possible combination of $(\sigma_X,\sigma_Y) \in \{1,2,3,4,5,7,9\}$ (in pixel units of the normalized image) for three different sets of random positions.
We obtain consistency in performance variations of systems across three random seeds, which shows that our random samples and, hence, our random eye perturbations are not biased.

When changing random eye perturbations from $\sigma_{\{X,Y\}}=1$ to $\sigma_{\{X,Y\}}=3$, the HTER of Eigenfaces increases from $0.14$ (AUC $=0.93$) to $0.26$ (AUC $=0.86$), whereas the HTER of Fisherfaces experiences a more drastic increase from $0.12$ (AUC $=0.93$) to $0.37$ (AUC $=0.71$).
In the previous section, we observed that Eigenfaces, as compared to Fisherfaces, is more robust to misalignment caused by translation and rotation.
The results from the random eye perturbation experiment show that Eigenfaces is more robust towards all types of misalignment.

The performance drop for both Gabor-Jet and LGBPHS are similar for change in random eye perturbations from $\sigma_{\{X,Y\}}=1$ to $\sigma_{\{X,Y\}}=3$.
The HTER of Gabor-Jet increases from $0.05$ (AUC $>0.99$) to $0.33$ (AUC $=0.87$), while that of LGBPHS increases from $0.04$ (AUC $>0.99$) to $0.32$ (AUC $=0.86$).
Comparing this large drop in performance with results from the previous experiment (involving only translation and rotation), we can conclude that the performance of both Gabor-Jet and LGBPHS is more susceptible to scaling variations.
Additionally, we can observe that both Gabor-Jet and LGBPHS drop performance similarly for misalignment of the same kind.

ISV has the best tolerance to misalignment involving translation, rotation and scaling.
From \figurename~\ref{fig:rand_performance_theta_tx_ty}, this property of ISV is evident from the larger dark region (corresponding to good performance) as compared to the remaining four face recognition systems.
Its HTER increases from $0.002$ (AUC $>0.99$) to $0.045$ (AUC $>0.99$) when the random eye perturbations change from $\sigma_{\{X,Y\}}=1$ to $\sigma_{\{X,Y\}}=3$.
For larger perturbations of $\sigma_{\{X,Y\}}=9$, the HTER still is $0.33$ (AUC $=0.84$).
This shows that ISV experiences a significant drop in performance only for extreme misalignment.

\subsection{Ambiguity in the Location of Eyes}
\label{sec:exp_ambiguity_eye_loc}
\begin{figure}[t!]
 \centering
 \subfloat{\includegraphics[width=0.8\linewidth]{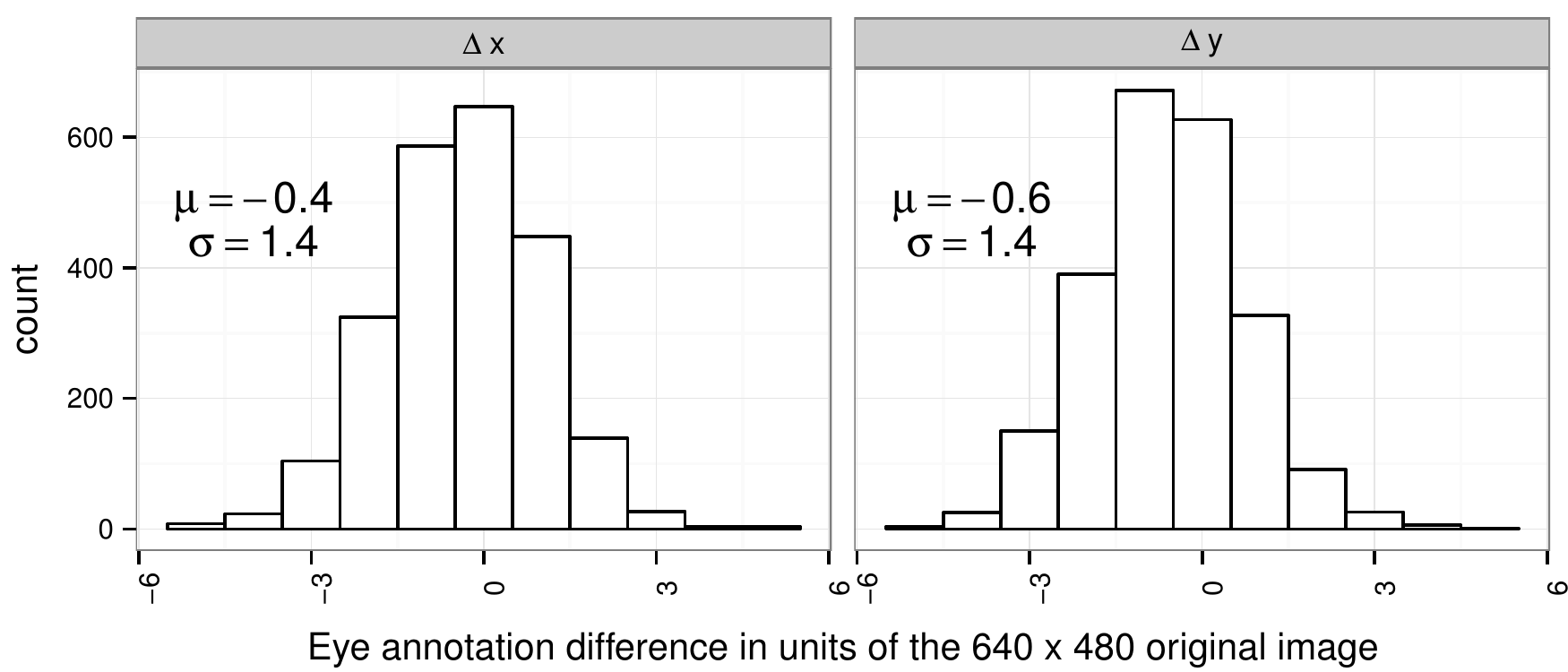}}

 \subfloat{\includegraphics[width=0.8\linewidth]{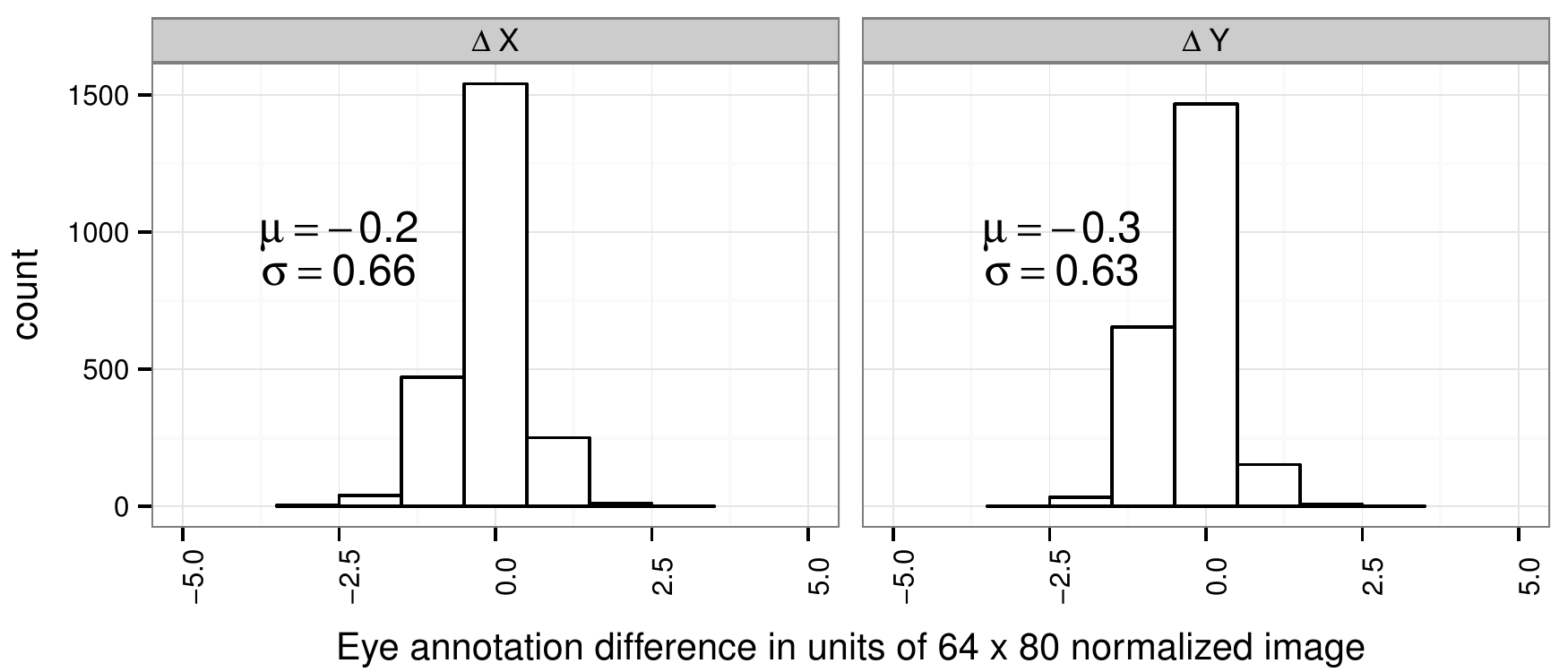}}
 \caption{Statistical difference between manual eye annotations carried out independently at Idiap Research Institute (Switzerland) and University of Twente (Netherlands).}
 \label{fig:mpie_051_ut_idiap_eye_err}
\end{figure}

With this experiment, we investigate the ambiguity in location of the eyes by comparing the manual eye annotations performed by two independent institutions.
For the $1160$ frontal images in the Multi-PIE M protocol, we possess the manual eye annotations from two independent sources: from the Idiap Research Institute (Switzerland)~\cite{shafey2013scalable} and from the University of Twente - UT (Netherlands).
In \figurename~\ref{fig:mpie_051_ut_idiap_eye_err}, we show the distribution of the difference in $x$ and $y$ coordinate of the two eye annotations.
In order to allow comparisons with results from Sections~\ref{sec:exp_impact_translation_rotation} and~\ref{sec:exp_impact_translation_rotation_scaling}, we obtain a mapping from the original $640 \times 480$ pixel image space to the $64 \times 80$ pixel normalized image space by first computing $s$ and $\alpha$ with \eqref{eqn:geom_norm_eq_val} using the Idiap eye annotations as the base, and then transforming the UT eye annotations to the normalized space using \eqref{eqn:geom_norm_eq} and computing the difference with the manual eye locations in the normalized image.

\begin{figure}[t!]
 \centering
 \includegraphics[width=0.8\linewidth]{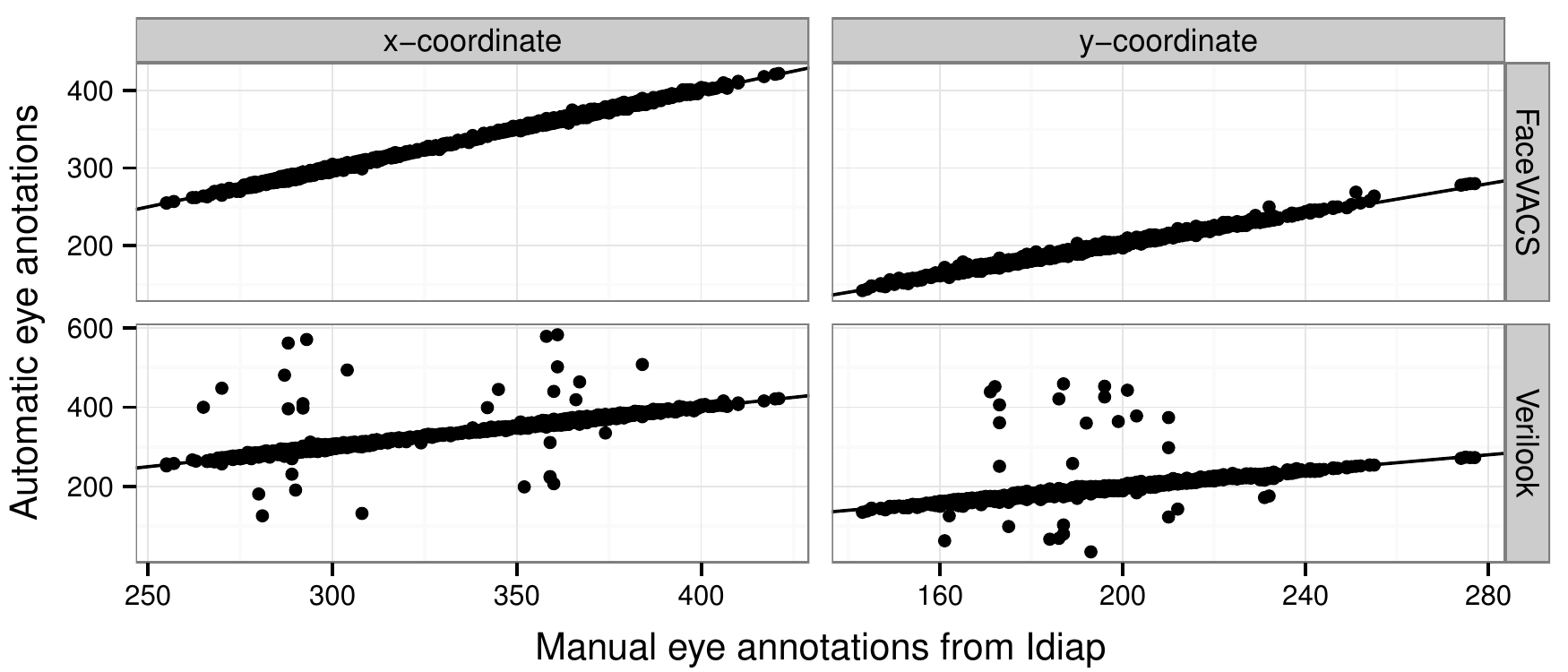}
 \caption{Correlation between manually annotated \cite{shafey2013scalable} and automatically detected eye coordinates in the pixels units of the original image space. The black solid line indicates a correlation of 1.}
 \label{fig:mpie_051_fv_vl_correlation_wrt_idiap_eye}
\end{figure}

Most face recognition systems employ a carefully tuned automatic eye detector to obtain the location of the eyes.
In the second part of this experiment, we investigate the accuracy of automatic eye detectors present in the commercial face recognition systems FaceVACS~\cite{facevacs2010} and Verilook~\cite{verilook2011}.
The correlation between automatically detected eye coordinates and manually annotated eye locations is shown in \figurename~\ref{fig:mpie_051_fv_vl_correlation_wrt_idiap_eye}, where the distribution of errors of the two automatic eye detectors are shown considering the manual eye annotation of Idiap~\cite{shafey2013scalable} as ground truth.
We transform this error distribution to normalized image space using the same procedure as for the UT annotations.

Note that the Idiap manual annotations and automatic eye annotations from FaceVACS and Verilook were carried out on no-flash images (\ie, under ambient illumination), while the UT manual annotations were performed on images captured using frontal flash (\ie, flash=$07$).

\begin{figure}[t!]
 \centering
 \includegraphics[width=0.8\linewidth]{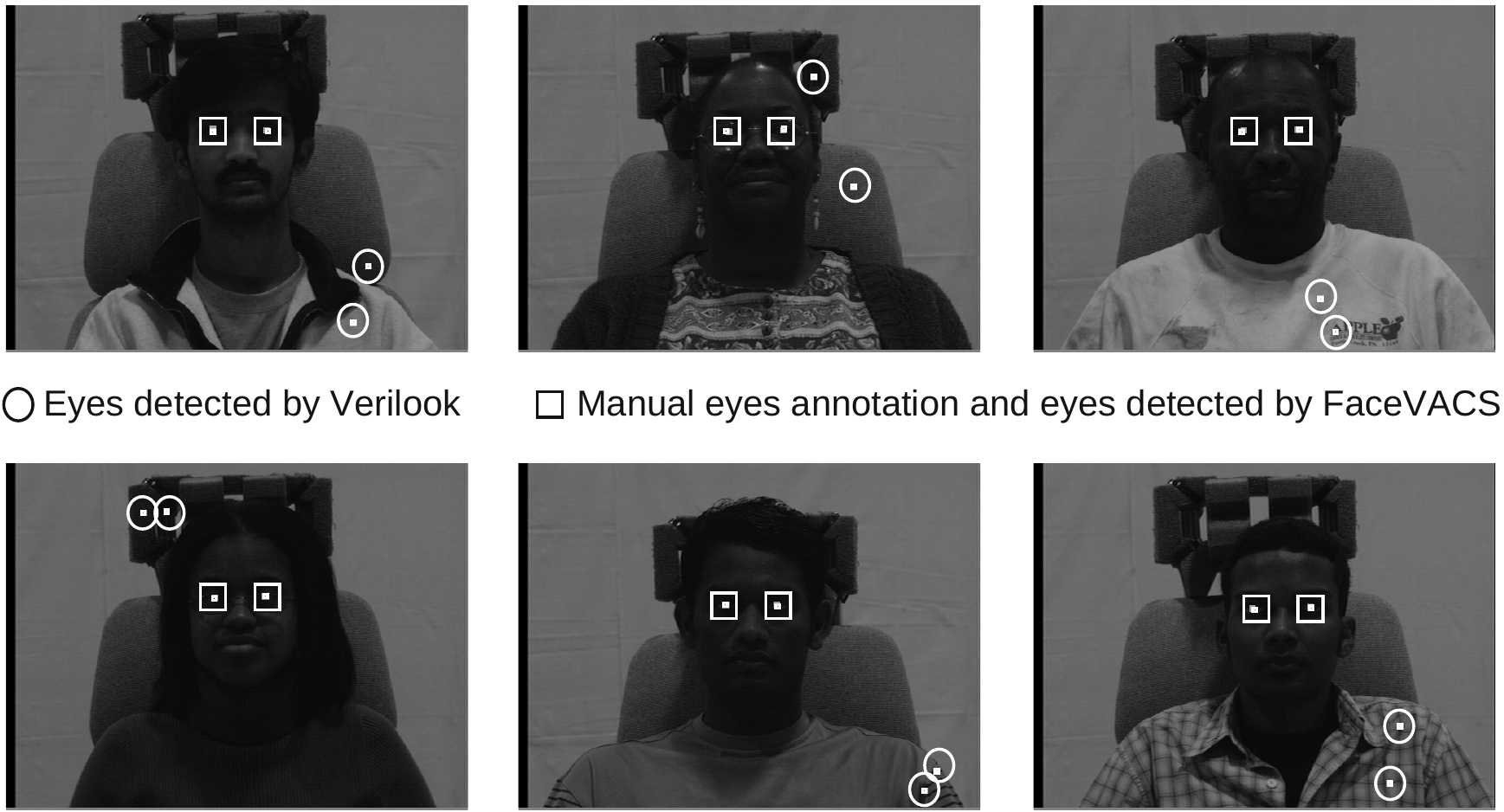}
 \caption{Some sample images, for which the Verilook eye detector has an eye detection error $> 50$ pixels.}
 \label{fig:vl_err_gr_50px}
\end{figure}

\begin{figure}[p!]
 \centering
 \subfloat{\includegraphics[width=0.8\linewidth]{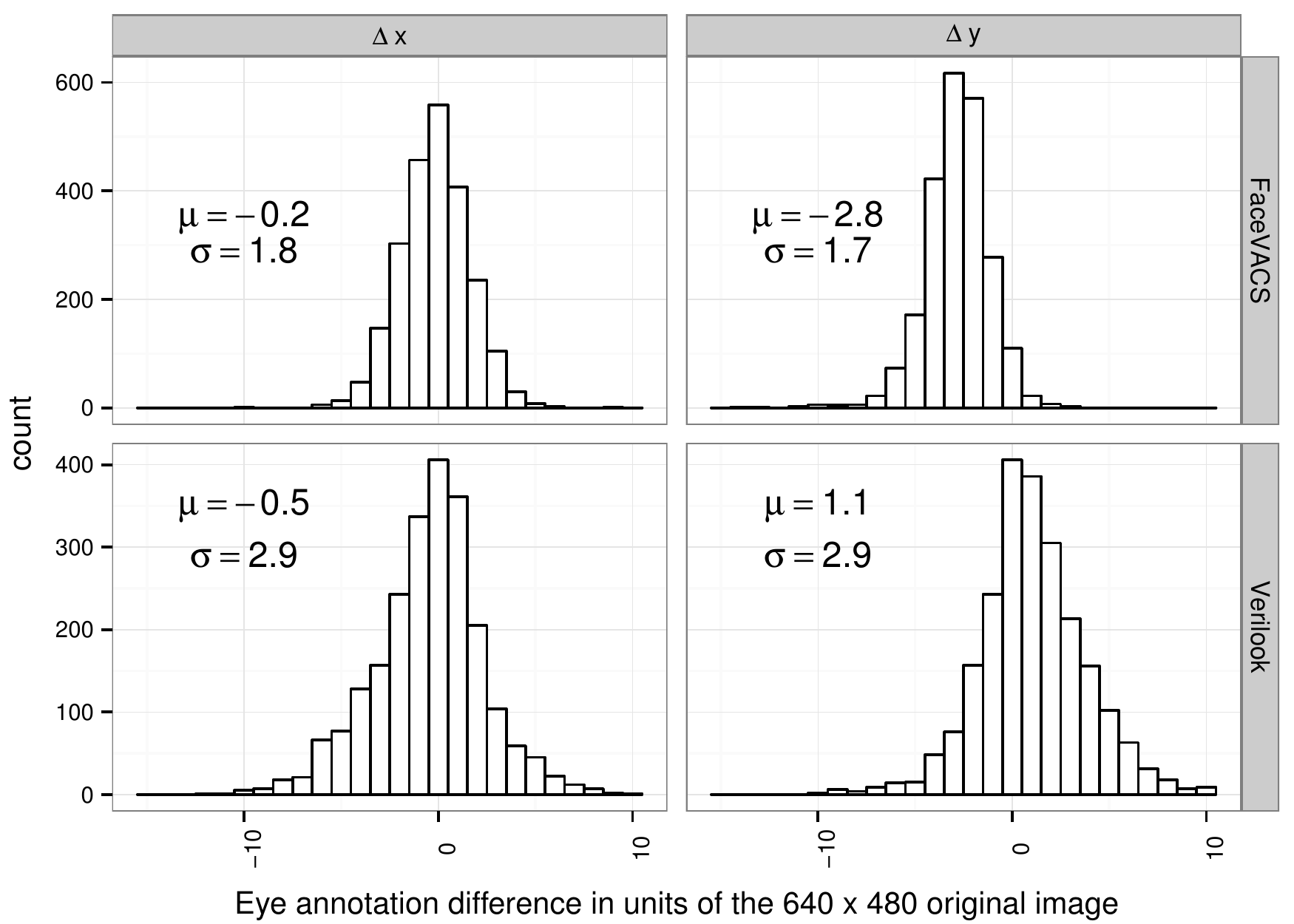} }

 \subfloat{\includegraphics[width=0.8\linewidth]{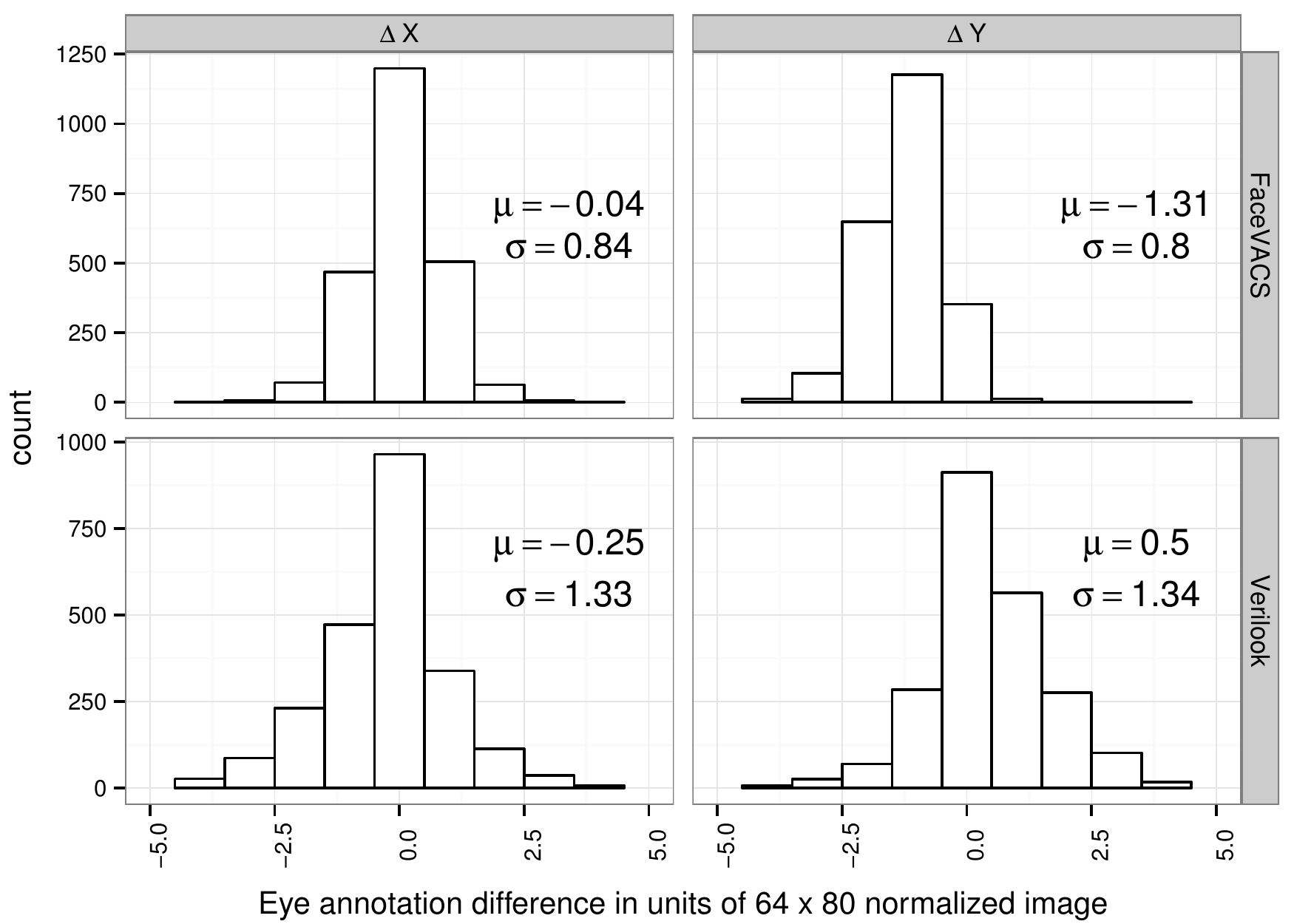} }
 \caption{Difference in eye locations detected by the FaceVACS and Verilook eye detector with respect to manual eye annotations~\cite{shafey2013scalable}. For Verilook, 15 samples with eye detection error $> 50$ pixels are excluded.}
 \label{fig:mpie_051_00_fv_vl_err_wrt_idiap}
\end{figure}

In \figurename~\ref{fig:mpie_051_fv_vl_correlation_wrt_idiap_eye} it can be seen that out of 1160 image samples, Verilook generated eye detection errors $> 50$ pixel for $15$ images.
Some of these are shown in \figurename~\ref{fig:vl_err_gr_50px}, which reveals that dark skin color combined with no-flash photographs contribute to large errors in automatic eye detection by Verilook.
For visual clarity of Verilook's detection error histogram in \figurename~\ref{fig:mpie_051_00_fv_vl_err_wrt_idiap}, we exclude those samples.

\begin{figure}[t!]
 \centering
 \includegraphics[width=0.7\linewidth]{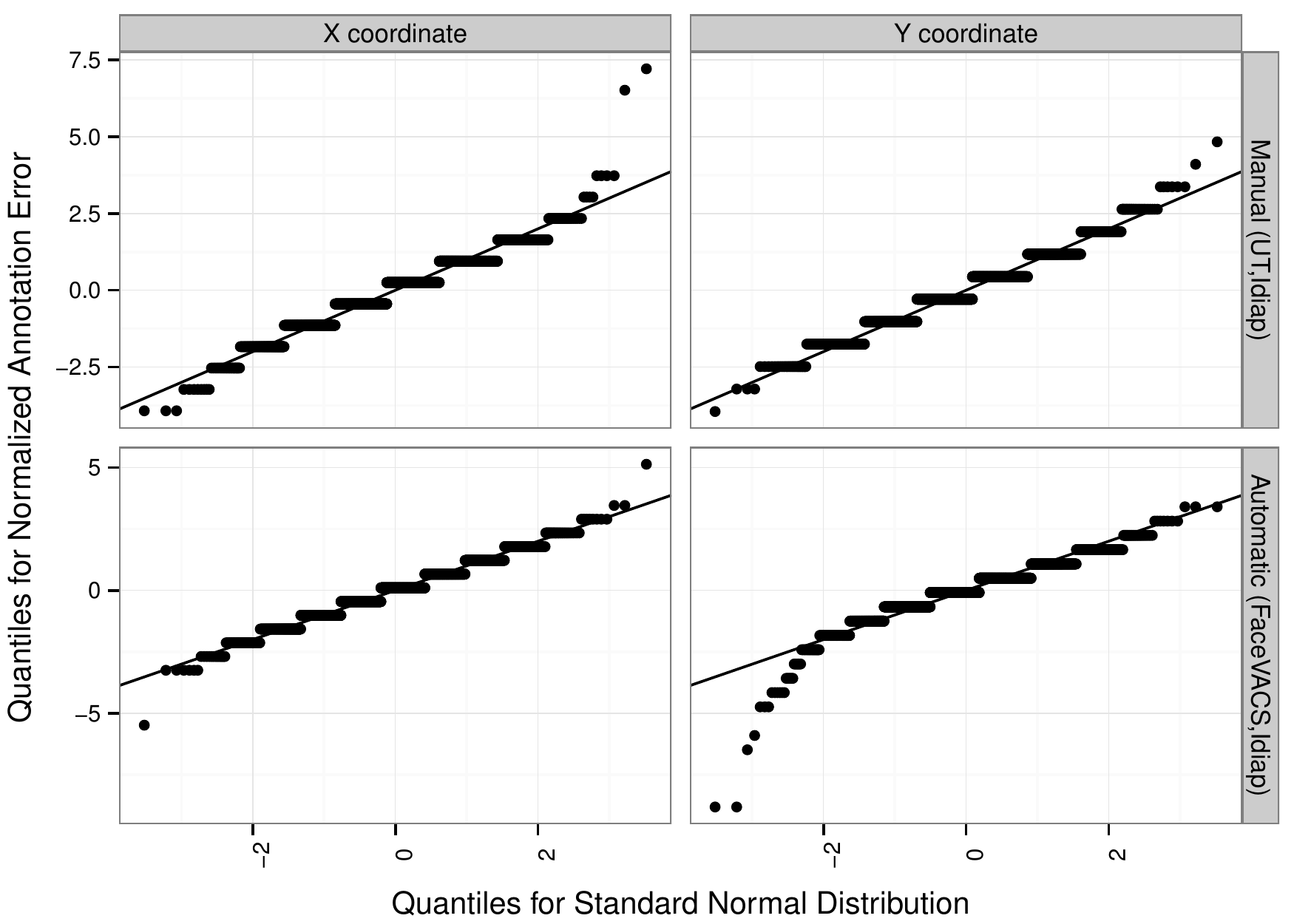}
 \caption{Quantile-Quantile plot of standard normal distribution $(\mu=0,\sigma=1)$ and normalized manual annotation difference distribution $(\frac{x-\mu}{\sigma})$ shown in Figures \ref{fig:mpie_051_ut_idiap_eye_err} (UT) and \ref{fig:mpie_051_00_fv_vl_err_wrt_idiap} (FaceVACS only). Note that the staircase pattern is caused by discrete pixel location values.}
 \label{fig:mpie_051_00_manual_fv_QQ}
\end{figure}

To see, whether the difference between the two independent manual eye annotations shown in \figurename~\ref{fig:mpie_051_ut_idiap_eye_err} follows a normal distribution, we plot the quantiles of this manual annotation difference (normalized such that $\mu=0$ and $\sigma=1$) against the quantiles of a standard normal distribution.
If the manual annotation difference follows the normal distribution, the points on this plot, which is also called a Quantile-Quantile (QQ) plot, should lie on the line $y=x$.
The QQ plot of \figurename~\ref{fig:mpie_051_00_manual_fv_QQ} (top row), confirms that the manual annotation difference follows a normal distribution. We subtract the mean value from this distribution to remove systematic offset, if any, in the eye annotations and claim that:

\begin{equation}
  \begin{array}{@{Pr(}c@{\ \leq \delta_{x} \leq\ }c@{)\ = \ }c}
    -1 & 1 & 0.51 \\
    -2 & 2 & 0.83 \\
    -4 & 4 & 0.99
  \end{array}\qquad \qquad
  \begin{array}{@{Pr(}c@{\ \leq \delta_{y} \leq\ }c@{)\ = \ }c}
    -1 & 1 & 0.53 \\
    -2 & 2 & 0.85 \\
    -4 & 4 & 0.99
  \end{array}
 \label{eq:man_err_prob}
\end{equation}
where $[\delta_{x}, \delta_{y}]=p^{m,idiap}-p^{m,ut}$ denote the difference between Idiap and UT manual eye annotation in the original image space, and $Pr(-4 \leq \delta_{x} \leq 4)$ denotes the probability of four pixel difference in manual annotations along horizontal direction for frontal facial images, which in Multi-PIE have an average inter-ocular distance of $70$ pixels.
From this experiment, we have empirical evidence for an ambiguity of four pixels in the location of the eyes.
Currently, we have access to only two independent sources of manual annotations for the Multi-PIE data set.
However, we would need more independent sources of manual annotation to check if this conclusion generalizes to a larger population of manual annotators.

Now we investigate the error characteristics of automatic eye detectors as shown in \figurename~\ref{fig:mpie_051_00_fv_vl_err_wrt_idiap}.
The correlation plot of \figurename~\ref{fig:mpie_051_fv_vl_correlation_wrt_idiap_eye} shows that the Verilook eye detector has large errors for many image samples, while FaceVACS is fairly accurate.
Hence, we only include the error distribution of FaceVACS (as shown in \figurename~\ref{fig:mpie_051_00_fv_vl_err_wrt_idiap}) in our further analysis.
The QQ plot of \figurename~\ref{fig:mpie_051_00_manual_fv_QQ} (bottom row) shows that the error distribution of FaceVACS follows a normal distribution.
For some large errors in vertical location (\ie{} $(x^{m,idiap}-x^{d,fv}) < -3$), the distribution deviates from this normal distribution.
Assuming normality and subtracting the mean from this distribution to remove systematic offset in eye detections, we can claim that:

\begin{equation}
  \begin{array}{@{Pr(}c@{\ \leq \delta_{x} \leq\ }c@{)\ = \ }c}
    -1 & 1 & 0.42 \\
    -2 & 2 & 0.73 \\
    -4 & 4 & 0.97 \\
    -6 & 6 & 0.99 \\
  \end{array}\qquad \qquad
  \begin{array}{@{Pr(}c@{\ \leq \delta_{y} \leq\ }c@{)\ = \ }c}
    -1 & 1 & 0.43 \\
    -2 & 2 & 0.75 \\
    -4 & 4 & 0.97 \\
    -6 & 6 & 0.99 \\
  \end{array}
 \label{eq:fv_err_prob}
\end{equation}
where $[\delta_{x}, \delta_{y}]=p^{m,idiap}-p^{d,fv}$ denote the difference between Idiap and FaceVACS eye annotations.
Comparing~\eqref{eq:man_err_prob} and~\eqref{eq:fv_err_prob}, we observe that -- after compensation for a systematic offset -- the accuracy of FaceVACS eye detector comes very close to the accuracy of manual annotators.
This observation is also evident in \figurename~\ref{fig:mpie_051_00_fv_vl_err_wrt_idiap}, which shows that FaceVACS achieves a standard deviation of approximately 1.8 pixels, while manual eye annotators achieve a standard deviation of 1.4 pixels as shown in \figurename~\ref{fig:mpie_051_ut_idiap_eye_err} in both horizontal and vertical directions.

For the FaceVACS eye detector, we observe a systematic offset of around 3 pixels along the vertical direction as shown in \figurename~\ref{fig:mpie_051_00_fv_vl_err_wrt_idiap}.
This shows that the FaceVACS detector is trained with a different notion of eye center, revealing the lack of consistency in the definition of eye center in a facial image: Does the eye center refer to center of the pupil, or to center between the two eye corners or eyelids, or to something else?

\subsection{Choice of Eye Detector for Training, Enrollment and Query}
\label{sec:exp_choice_eye_det}
\begin{figure}[p!]
 \centering
 \includegraphics[width=.95\textwidth]{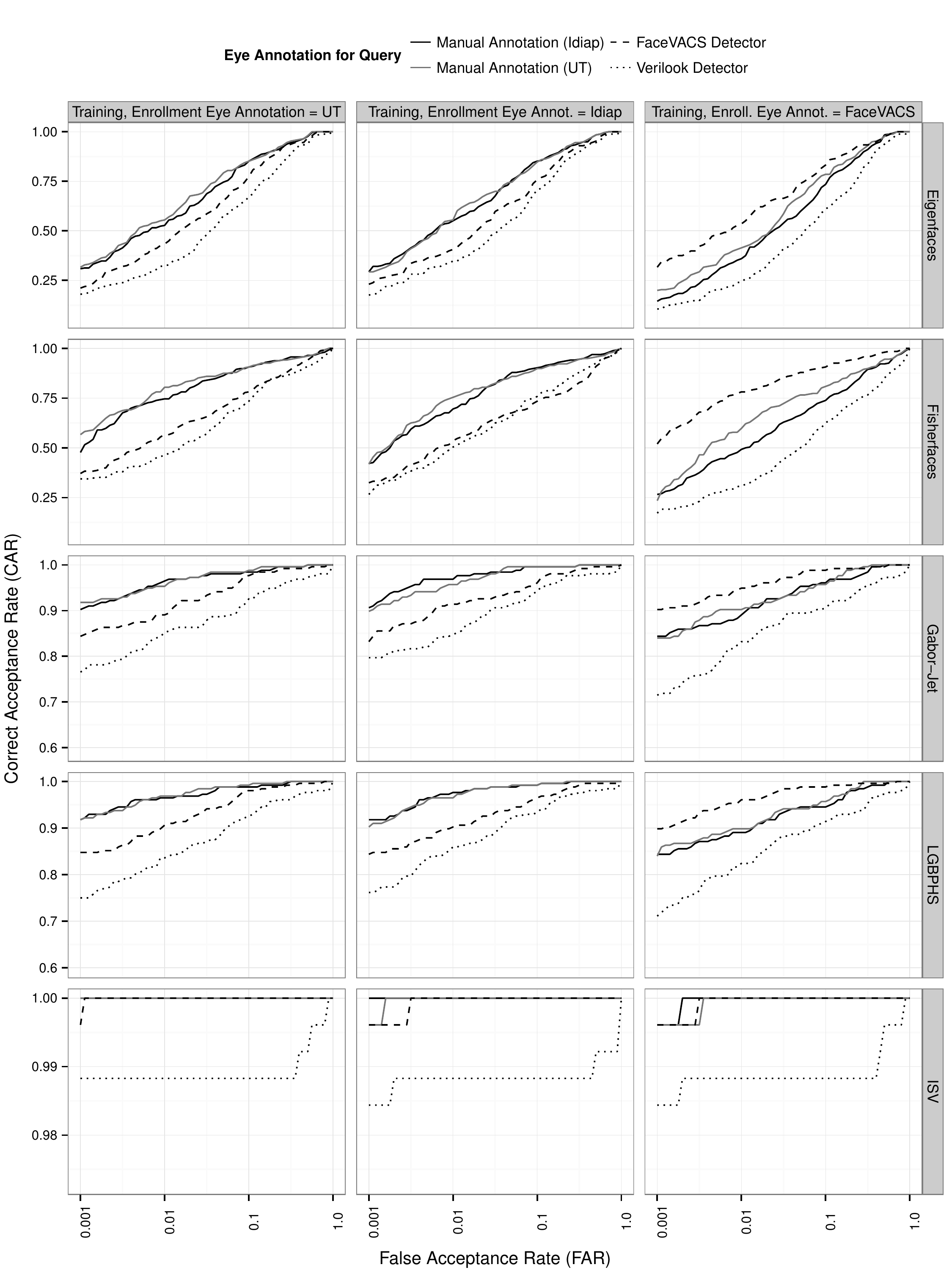}
 \caption{Face recognition performance with eye annotation provided by UT (manual), Idiap (manual), FaceVACS (automatic) and Verilook (automatic). Note that the range of CAR values are different in each row of the above plot.}
 \label{fig:perf_eval_with_fv_vl_idiap_ut_eye_all_alg_roc}
\end{figure}

Most face recognition systems go through the following three phases of operations:
\begin{inparaenum}[\itshape a\upshape)]
 \item a training phase to learn the representation of facial features,
 \item an enrollment phase to enroll models from facial images of known identities and
 \item a query phase to verify the identity in a given query image.
\end{inparaenum}
All three phases exploit the location of the eyes for alignment of images.
We consider the training and enrollment phase as one group because they are defined during the off-line development of a face recognition system.
In this section, we investigate the impact of using different eye detectors for on-line query phase and the off-line training/enrollment phase.

For the Multi-PIE M protocol images, we investigate the following combinations of eye annotations: \begin{inparaenum}[\itshape a\upshape)]
 \item training and enrollment images aligned using manual eye annotations from Idiap, UT or automatic eye locations from FaceVACS eye detector
 \item query images aligned using two types of manual annotations (Idiap, UT) or automatic eye annotations (FaceVACS and Verilook).
\end{inparaenum}

With this experiment, we investigate the impact of using different sources of eye annotations for the training/enrollment and the query phase of face recognition systems.
In the first column of \figurename~\ref{fig:perf_eval_with_fv_vl_idiap_ut_eye_all_alg_roc}, we show the recognition performance for the five open source systems when training and enrollment images are annotated using manual annotations from Idiap and the query images are labeled using manual (Idiap,UT) and automatic (FaceVACS, Verilook) eye annotations.
Note that the range of CAR values are different in each row of this plot.

In \figurename~\ref{fig:perf_eval_with_fv_vl_idiap_ut_eye_all_alg_roc} (first and second column), we observe that both Idiap and UT manual annotations in the query set have similar recognition performance, independent on which of both were used in the training/enrollment stage.
This shows that all of the investigated face recognition systems, when trained using manual eye annotations, are tolerant to small differences up to four pixels (with average inter-ocular distance of $70$ pixels) in manual annotation caused by inter-observer variability.
Moreover, when training/enrollment is done using manual annotations, the recognition performance for query images annotated using manual annotation is always higher than that obtained using automatic eye detectors.
For the accurate eye detector of FaceVACS, we observe a large drop in performance, but this is primarily due to inconsistency in the definition of eye center.
The Verilook eye detector is less accurate (as evident from \figurename~\ref{fig:mpie_051_00_fv_vl_err_wrt_idiap}) and, therefore, all algorithms operating on facial images aligned using Verilook eye coordinates have poor performance.

The third column of \figurename~\ref{fig:perf_eval_with_fv_vl_idiap_ut_eye_all_alg_roc} shows the performance variation when training and enrollment images are labeled using FaceVACS automatic eye annotations.
We observe that FaceVACS eye annotations for query achieves best recognition performance as compared to other sources of annotation for query images.
The large drop in performance for manually annotated query images is, in turn, due to inconsistency in the definition of the eye centers.
Moreover, for Eigenfaces and Fisherfaces, these performances are comparable to the performance of manually annotated query as shown in the first and second columns of \figurename~\ref{fig:perf_eval_with_fv_vl_idiap_ut_eye_all_alg_roc}.
For Gabor-Jet and LGBPHS, there is a slight drop in performance as compared to the manual annotations cases, but still the performance with the FaceVACS query images is best among the other three annotation sources.
This shows that an accurate automatic eye detector (like that of FaceVACS) can help achieve recognition performance comparable to that obtained using manually annotated eyes, given that the same automatic detector is used for annotating the training, enrollment and query images.

The ISV algorithm seems to be unaffected by the source of eye annotations.
Except for the complete misdetections of Verilook discussed in Section~\ref{sec:exp_ambiguity_eye_loc}, the ROC curves are stable at a very high level in the last row of \figurename~\ref{fig:perf_eval_with_fv_vl_idiap_ut_eye_all_alg_roc}.
Not even the different definition of eye centers disturbs the recognition capabilities of ISV.
Most probably this stability comes through the fact that facial features are extracted locally from the image, and the distribution of these features is modeled independently.

\section{Discussion}
\label{sec:discussion}
For different types of misalignment in a $64 \times 80$ normalized image space, we evaluated the performance of following five open source face recognition systems: Eigenfaces, Fisherfaces, Gabor-Jet, LGBPHS and ISV.
We simulated different types of facial image misalignment by scaling, rotating and translating manually annotated eye locations.
We found that Eigenfaces is more robust to misalignment (caused by scaling, rotation or translation of eye locations) as compared to Fisherfaces.
However, Fisherfaces has higher recognition performance (HTER=$0.08$, AUC=$0.95$) for properly aligned images (\ie, $t_X=0,t_Y=0,\theta=0$) as compared to Eigenfaces (HTER=$0.12$, AUC=$0.94$).
Furthermore, we found that Eigenfaces is more sensitive to vertical misalignment as compared to misalignment along horizontal direction.
However, in any case the Eigenfaces and Fisherfaces algorithms showed performance inferior to the other investigated methods.
Both Gabor wavelet based methods, Gabor-Jet and LGBPHS, have similar tolerance towards misalignment with both of them being more susceptible to misalignment caused by scaling.
We found that ISV has the best tolerance for misalignment since it is able to maintain a consistent level of performance for a large range of misalignment $(\mu=0,\sigma \leq 3$).
ISV demonstrates such a natural robustness to misalignment because features from all parts of the facial image are modeled independently.

We investigated two different evaluation measures, AUC and HTER.
While the former measure is biased since it evaluates performance after all scores of a certain database have been collected,
the HTER is unbiased and more application oriented since it requires a threshold to be selected prior to evaluation.
We have found that the biased AUC was often stable, while the HTER dropped drastically.
This shows that the threshold for HTER cannot simply be selected based on hand-labeled eye locations when the query set is detected automatically.
Hence, other strategies as score normalization \cite{wallace2012ztnorm} or calibration \cite{mantasari2014calibration} need to be applied to the scores from the face recognition systems in order to make use of the face recognition algorithms in case of automatically detected eyes.

In practical face recognition systems, an automatic eye detector is used to localize the two eyes and then perform automatic registration of facial images.
Therefore, we investigated the accuracy of automatic eye detectors present in two commercial face recognition systems: FaceVACS and Verilook.
Additionally, we analyzed the difference between two independent sources of manual eye annotations for the frontal images of the Multi-PIE M protocol.
This allows us to understand the inherent limitations of using the two eye coordinates as the landmarks for facial image registration.
We found an ambiguity of around $4$ pixels in manual eye annotation of frontal view images with an average inter-ocular distance of 70 pixels.
Therefore, face recognition systems should be built to tolerate at least this amount of error in eye coordinates.
Those $4$ pixels translate to approximately 2 pixels in the normalized image, which means that the algorithms Gabor-Jet, LGBPHS and ISV still perform reasonably well, cf. \figurename~\ref{fig:rand_performance_theta_tx_ty}.
The automatic eye detector of FaceVACS achieves a detection accuracy $(\sigma=1.8)$ that is close to the accuracy of manual annotators $(\sigma=1.4)$.
However, the eye locations were detected on well-illuminated frontal images and, thus, this result needs to be verified in presence of illumination or non-frontal pose.
We observed higher eye detection error in the automatic eye detector included in the Verilook system.
We found that the FaceVACS eye detector showed a systematic offset of $3$ pixels in vertical location of the eyes, which reveals lack of consistency in the definition of the eye center in frontal facial images.

We also explored the impact of using different sources (automatically detected or manually located) of eye annotations for training, enrollment and query phases of a face recognition system.
We found that using manual eye annotations for training and enrollment while utilizing automatic eye annotations for query results in a large drop in performance, but we discussed that this is caused by the inconsistent definition of the eye centers.
We also found that using eye annotations from a accurate automatic eye detector (like that of FaceVACS) for all training, enrollment and query images results in face recognition performance that is comparable to the performance achieved using manual eye annotations.
Therefore, our results underline the importance of consistent definition of eye center in a facial image and also highlights the performance gain achieved by using same automatic eye detector for training, enrollment and query images.
Furthermore, the performance of ISV remains consistently high for all combination of eye annotation sources.
This shows that a combination of moderately accurate eye detector and a face recognition system that is naturally robust to moderate misalignment can potentially be a solution for practical applications.

One important fact of a face recognition algorithm is its complexity.
The execution time of the five face recognition systems on a Intel i7 3.5 GHz (4 cores) machine for training, enrollment and scoring operation has been recorded as: ISV: 159.8 min., LGBPHS: 9.7 min., Gabor-Jet: 2.8 min., Fisherfaces: 1.8 min. and Eigenfaces: 1.7 min.
Hence, the robustness of ISV towards misalignment comes at the expense of very high computational costs.
The best trade-off between accuracy and complexity in our tests was achieved by the Gabor-Jet algorithm.

Another important point of this evaluation is that all experiments solely rely on open source software -- except for the automatically detected eye locations, which were generated by third party software.
Effectively, we provide the scripts and documentation to install the required software, to rerun all face recognition experiments presented in this paper, and to regenerate Figures \ref{fig:fixed_performance_theta_tx_ty}, \ref{fig:rand_performance_theta_tx_ty} and \ref{fig:perf_eval_with_fv_vl_idiap_ut_eye_all_alg_roc}.
Additionally, the source code can be easily adapted to run the same experiments using a different image database (for which at least the hand-labeled eye positions must be available) or to investigate the stability of other face recognition algorithms towards eye localization errors.

\section{Conclusion}
\label{sec:conclusion}
In this investigation, the aim was to determine the impact of misalignment caused by errors in eye localization on the performance of face recognition systems.
Similar studies carried out in the past were either limited by the number of face recognition systems or the size of facial image database.
Our study is based on five open source face recognition algorithms operating on a larger facial image database.
One of the more significant findings to emerge from this study is the ambiguity in the definition of eye centers in a facial image.
The two eye centers are widely used as the landmarks for registration of facial images.
However, a commonly agreed definition of the ``eye center'' is still missing.
This causes inconsistency in the eye locations detected by different automatic eye detectors thereby reducing performance when eye detectors and face recognition systems of different origin are mixed.
Perhaps, this is the most serious limitation of using the two eye centers as the landmarks for facial image registration.
If the same automatic eye detector is used for annotating the training, enrollment and query images, our experiment results show that it is possible to achieve recognition performance comparable to that obtained using manually annotated eyes, given that the facial images are well-illuminated and show a frontal pose.

We compared the manual eye annotations obtained from two independent sources to study the ambiguity in manual eye annotations.
To the best of our knowledge, such a study has not been reported before.
We found that there exists an ambiguity of four pixels in manual annotations of the two eyes when the frontal facial images have an average inter-ocular distance of $70$ pixels.
Therefore, assuming that humans are the best possible eye detectors, face recognition systems that use the location of the eyes for alignment should be built to handle at least this level of ambiguity, which was the case for the five investigated face recognition systems.
Our results also show that the accuracy of FaceVACS automatic eye detector is very close to that of manual eye annotators.

It has been demonstrated that the Jesorsky measure is insufficient to distinguish between landmark localization errors that cause the normalized image to be shifted or rotated.
On the other hand, we have shown that most algorithms have a higher tolerance towards translation and rotation than towards scaling.
Hence, the Jesorsky error of an automatic eye detector has a limited correlation with the actual face recognition performance of a face recognition system.

The results reported in this paper reveal the nature of five open source face recognition algorithms towards misalignment caused by errors in eye localization.
ISV demonstrates excellent tolerance towards large amount of misalignment caused by errors in eye localization.
Its performance drops only for extreme misalignment of facial images, but this performance comes at a cost of long execution time and, hence, might not be usable under real-time requirements.
Gabor-Jet shows good tolerance towards misalignment and has very low execution time as compared to ISV.
Both Gabor-Jet and LGBPHS have similar tolerance towards misalignment and they have higher recognition performance and are more robust to misalignment as compared to Eigenfaces and Fisherfaces.

Due to the availability of independent hand-labeled sources of eye landmarks, the present study was performed on the Multi-PIE image database.
A further study could include more face recognition systems or more challenging image conditions like different illumination, facial expressions and head pose.
Since the tools used in this study are open source and released with this paper, it is possible to perform such a study with minimal effort.
Also the impact of score normalization or calibration on the performance of the unbiased evaluation needs to be addressed.

\chapter{Notes on Forensic Face Recognition}
\label{dutta2014phdthesis_forensic-intro}
A forensic case involving face recognition commonly contains a surveillance view trace (usually a frame from CCTV footage) and a frontal suspect reference set containing facial images of suspects narrowed down by police and forensic investigation.
When a forensic investigator is tasked to compare the surveillance view trace (or, probe) to the suspect reference set, it is quite common to manually compare these images.
However, if the forensic investigator chooses to use an automatic face recognition system for this task, there are two choices available: a model based approach or a view based approach.
In a model based approach, a frontal view probe image is synthesized based on a 3D model reconstructed from the surveillance view trace.
Most face recognition systems are fine tuned for optimal recognition performance for comparing frontal view images and therefore the model based approach, with synthesized frontal probe and frontal suspect reference images, ensures high recognition performance.
In a view based approach, the reference set is adapted such that it matches the pose of the surveillance view trace.
This approach ensures that a face recognition system always gets to compare facial images under similar pose -- not necessarily the frontal view.
In this chapter, we explore several aspects of the view based approach in a forensic context.

Section~\ref{dutta2012impact_intro} presents a preliminary investigation on the impact of image quality variations on the performance of a face recognition system.
In particular, we investigate recognition performance variation for all possible combination of pose and illumination variation in the probe (or test) and reference set.
Furthermore, to simulate a forensic scenario, we fix the probe to contain surveillance view image and reference set to contain frontal or near-frontal surveillance view image and analyze the impact of noise, blur and resolution variation of both probe and reference set on the recognition performance.
These experiments provide an insight on whether a view based approach can potentially be useful for forensic cases.

In Section~\ref{dutta2012view_intro}, we compare the performance of both model and view based approaches.
For the model based approach, we reconstruct a 3D face model using a surveillance view probe image followed by the synthesis of a frontal view probe image while the reference set contain frontal view images.
For the view based approach, we keep the probe set fixed to the original surveillance view image while the gallery is changed to match the pose of probe set (\ie surveillance view).
We evaluate the performance of both model and view based approaches on five face recognition system.
In the experiment results reported in Section~\ref{dutta2012view_intro}, the two commercial off-the-shelf (COTS) face recognition systems A and B are~\cite{facevacs2010}~and~\cite{verilook2011} respectively.
Furthermore, LDA-IR~\cite{bolme2012csu} has been recently renamed to cohort-LDA (cLDA) by the authors.

s\section{The Impact of Image Quality on the Performance of Face Recognition}
\label{dutta2012impact_intro}
Although there are CCTV cameras everywhere, they rarely contribute to strong evidence in the court of law because even the best trained forensic investigators find it difficult to compare and interpret these low quality face images. Automatic face recognition systems are rarely used in evaluation of forensic cases because they are tuned to deliver good accuracy for well illuminated and sharp frontal face images.

It is known that the performance of a face recognition system depends on the quality of both test and reference face images participating in the face comparison process. In a forensic evaluation case involving face recognition, the test image is usually captured by a CCTV camera and the forensic investigators have no control over its quality. But, they have some control over the quality of the reference image (i.e. face images of the suspects). We investigate how this capability of controlling the reference image quality can be exploited to improve face recognition accuracy under the constraint that quality of the test image cannot be modified.

For a given quality of test image, there exists a reference image quality that would deliver optimal recognition performance over all the other possible reference image qualities using a particular face recognition system. In this paper, we evaluate the performance of a commercial face recognition system \cite{facevacs2010} for variations in the following five image quality parameters of the test and reference images: pose, illumination, noise (Gaussian), blur (motion), and, resolution. Such an evaluation provides answer to the following two questions commonly encountered by forensic investigators: (a) for a given test image, what reference image quality would deliver best recognition performance using a particular face recognition system? (b) for such image quality pair, what is the expected recognition performance from that face recognition system?

\subsection{Related Work}
\label{dutta2012impact_relatedwork}
Face recognition systems are fine tuned to achieve optimal recognition performance for frontal view test (probe) and reference (gallery) images. Therefore, a common 
approach to handle pose and illumination variation in test or reference image is to reconstruct 3D models of faces from non-frontal views and synthesize frontal view images for use with view based face recognition systems. \cite{zhao2000sfs}, \cite{blanz2005face}, and, \cite{park20073dmodel} have shown that this approach delivers superior performance as compared to the case of comparing non-frontal view images.

This approach of reconstruction of a 3D face model from non-frontal view image followed by synthesis of frontal view image is very difficult to apply in a real 
forensic face recognition cases. In a typical forensic evaluation case involving face recognition, the trace (image captured by CCTV camera at the crime scene) is often of very low quality and therefore it is very difficult, and often impossible, to locate adequate numbers of feature points (like nose tip, eye corners, etc): a prerequisite for reconstruction of a 3D model using \cite{blanz2005face}, \cite{park20073dmodel}. Even if we succeed in locating at least 6 feature points, there is a possibility that the costly model fitting algorithm would not converge to a stable solution. \cite{blanz2005face} and \cite{park20073dmodel} were able to apply this approach because both test and reference images were of good quality and therefore it was easy to locate feature points.

In this paper, we investigate the possibility of transforming frontal view mug shots in the suspect reference set (gallery) to match the pose of the trace in order to achieve near optimal recognition performance. In a typical forensic case, the suspect reference set consists of good quality frontal view mug shot of individuals suspected to be present in the trace which is usually of low quality (surveillance view, motion blur, low resolution, etc). As it is difficult to synthesize frontal view images from such a low quality trace, we investigate if transforming the frontal view suspect reference images to a pose similar to that present in the trace can improve recognition performance. If true, this will allow us to apply the approach of \cite{blanz2005face} to the frontal view suspect reference images to synthesize surveillance view images in order to improve recognition performance when comparing to low quality trace using a view based face recognition system.

Recently, \cite{beveridge2011when} have shown that if we consider quality as being predictive of face recognition performance, then quality is the property of an image 
pair and not of an individual image. Therefore, in this study, we evaluate the performance variation of a commercial face recognition system \cite{facevacs2010} for image quality variation in both test and reference images.

\subsection{Performance Evaluation Setup}
\label{dutta2012impact_perf-eval}
In this paper, we evaluate the performance of a commercial face recognition system \cite{facevacs2010} for test and reference images varying in the following 5 image quality parameters: Pose, Illumination, Resolution, Motion Blur, and, Gaussian Noise. In a typical forensic evaluation case involving face recognition system, these 5 quality parameters are dominant in the trace.

\begin{figure}[t]
  \centering
  
  \subfloat[Properties of all the test and reference sets]{	
  \footnotesize
	\centering
	\begin{tabular}{l | c c}
	 & Test Set (Probe) & Reference Set (Gallery)\\
	 \hline
	 size (image count) & 479 & 442 \\
	 person count & 319 & 268 \\
	 session & \texttt{01,03} & \texttt{02,04} \\
	 expression & neutral & neutral \\
	 eye annotation & manual & manual \\
	 \hline
	\end{tabular}
	\label{tbl:test_ref_set_spec}
	}

  \subfloat[Sample of facial image quality variations included in this study]{\label{fig:qual_var_illus} \includegraphics[width=0.45\linewidth, trim=0 0 0 0, clip = true]{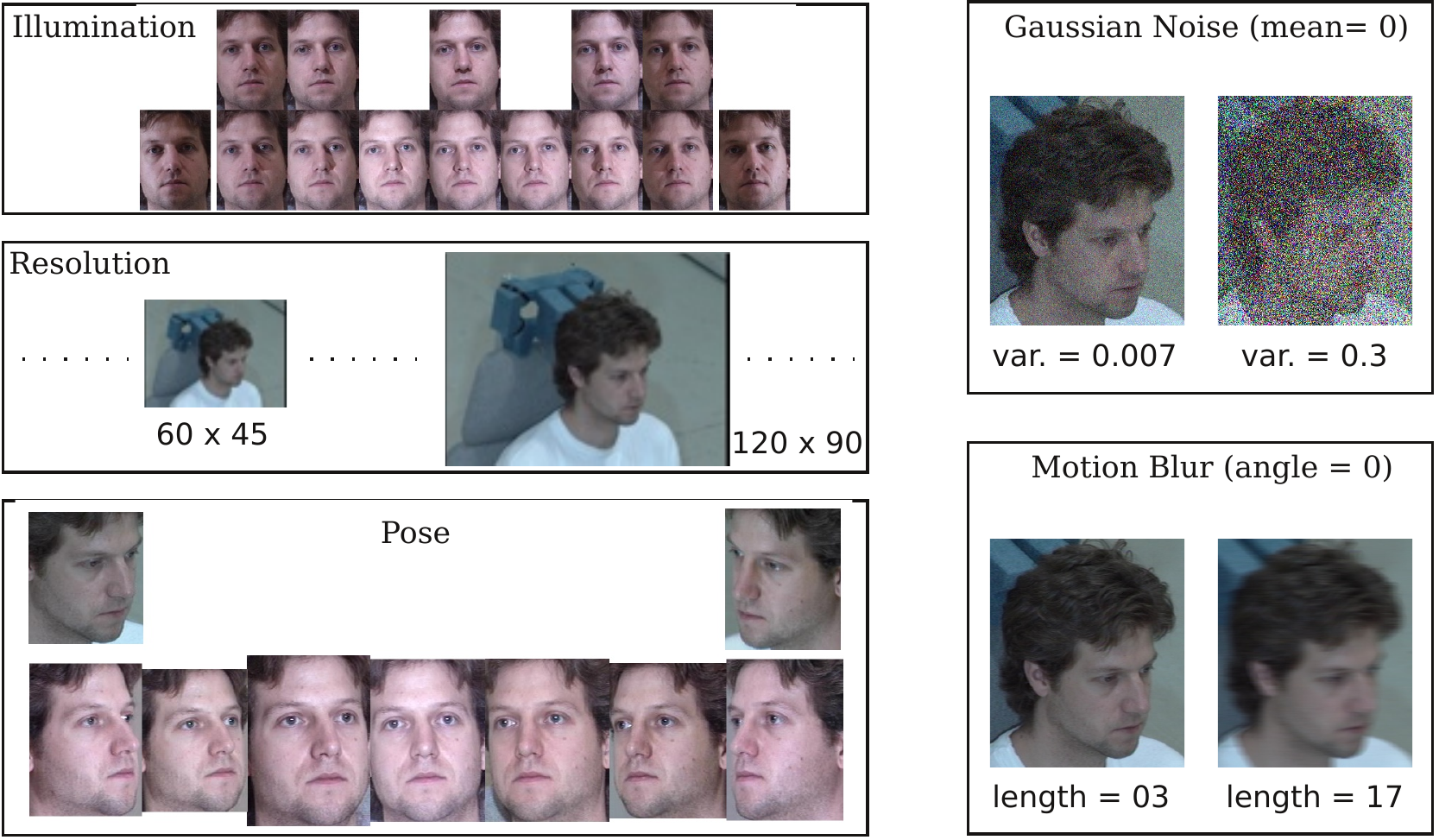}}
  $\;$
  \subfloat[Camera and flash location for all the images used in this experiment (source : MultiPIE \cite{gross2008multipie})]{\label{fig:mpie_cap_env_spec} \includegraphics[width=0.5\linewidth, trim=10 0 10 0, clip = true]{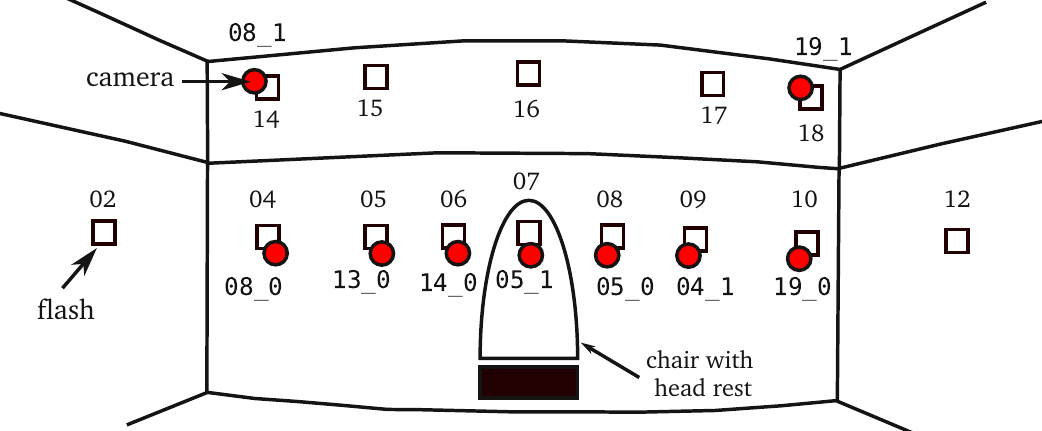}}
  
  \subfloat[Image quality variations included in this study]{	
  \footnotesize
  \centering
  \label{tbl:test_ref_set_spec}
  \begin{tabular}{p{2.4cm} | c c | c c | c c | c c | c c | c }
  \hline
  \multirow{2}{*}{Quality} & \multicolumn{2}{c|}{Camera} & \multicolumn{2}{c|}{Flash} & \multicolumn{2}{c|}{Resolution} & \multicolumn{2}{c|}{Motion Blur} & \multicolumn{2}{c|}{Gaus. Noise} & \multirow{2}{*}{Result} \\
  \cline{2-11}
   & \multicolumn{2}{c|}{Test$_{i}$, Ref$_{j}$} & \multicolumn{2}{c|}{Test$_{i}$, Ref$_{j}$} & \multicolumn{2}{c|}{Test$_{i}$, Ref$_{j}$} & \multicolumn{2}{c|}{Test$_{i}$, Ref$_{j}$} &  \multicolumn{2}{c|}{Test$_{i}$, Ref$_{j}$} & \\
  \hline
  Pose and $\qquad$ Illumination & \multicolumn{2}{c|}{$c_{i}, c_{j} \in \textbf{C}$} & \multicolumn{2}{c|}{$f_{i}, f_{j} \in \textbf{F}$} & \multicolumn{2}{c|}{$r_{i}, r_{j} = D_0$} & \multicolumn{2}{c|}{0, 0} & \multicolumn{2}{c|}{0, 0}  & \figurename\ref{fig:pi_man_eye_fv} \\
  \hline
  Resolution & \multicolumn{2}{c|}{$19\_1, \{*\}$} & \multicolumn{2}{c|}{$18, \{**\}$} & \multicolumn{2}{c|}{$r_{i}, r_{j} \in \textbf{R}$} & \multicolumn{2}{c|}{0, 0} & \multicolumn{2}{c|}{0, 0} & \figurename\ref{fig:re_auc_var} \\
  \hline
  Gaussian Noise & \multicolumn{2}{c|}{$19\_1, \{*\}$} & \multicolumn{2}{c|}{$18, \{**\}$} & \multicolumn{2}{c|}{$r_{i}, r_{j} = D_0$} & \multicolumn{2}{c|}{0, 0} & \multicolumn{2}{c|}{$\bar{\sigma}_{i}, \bar{\sigma}_{j} \in \textbf{N}_{\bar{\sigma}}$} & \figurename\ref{fig:ng_auc_var} \\
  \hline
  Motion Blur & \multicolumn{2}{c|}{$19\_1, \{*\}$} & \multicolumn{2}{c|}{$18, \{**\}$} & \multicolumn{2}{c|}{$r_{i}, r_{j} = D_0$} & \multicolumn{2}{c|}{$l_i, l_j \in \textbf{B}_{l}$} & \multicolumn{2}{c|}{0, 0} & \figurename\ref{fig:bm_auc_var} \\
  \hline
  
  \multicolumn{12}{l}{
  where, $\textbf{C} = [19\_1,\; 19\_0,\; 04\_1,\; 05\_0,\; 05\_1,\; 14\_0,\; 13\_0,\; 08\_0,\; 08\_1], \hfill$
  } \\
  \multicolumn{12}{l}{$\textbf{F} = [02,\; 04,\; 14,\; 05,\; 15,\; 06,\; 07,\; 16,\; 08,\; 09,\; 17,\; 10,\; 18,\; 12]$, $\textbf{R} = [640\times480,\; \cdots ,\; 60\times45], \hfill$} \\
  \multicolumn{12}{l}{$\textbf{B}_{l} \textnormal { (length in pixels)} = [1,\; 3,\; 5,\; 7,\; 13,\; 17,\; 21],\quad$ Note: angle= 0 $\hfill$} \\
  \multicolumn{12}{l}{$\textbf{N}_{\bar{\sigma}} \textnormal { (variance)} = [0.001,\; 0.007,\; 0.03,\; 0.07,\; 0.1,\; 0.2]. \quad$ Note: mean = 0 $\hfill$} \\
  \multicolumn{12}{l}{$\{*\}$ = \{19\_1, 05\_1\}, $\quad\{**\}$ = \{10, 07\}, $\quad D_0 = 640 \times 480$  $\hfill$} \\
  \end{tabular}
  }  
  \caption{Specification of all facial images used in this study}
  \label{fig:all_img_specs}
\end{figure}

All the test and reference images used in this experiment were taken from the MultiPIE data set \cite{gross2008multipie}. Selection of test and reference set images was based on the criteria shown in \figurename\ref{tbl:test_ref_set_spec}. MultiPIE data set provides good sampling of pose and illumination for 337 subjects using an image capture setup shown in \figurename\ref{fig:mpie_cap_env_spec}. We simulated the open set recognition scenario, commonly encountered in forensic cases, by creating test and reference set such that not all the individuals in the test set are present in the reference set.

For all the experiment scenarios, we supplied manually annotated eye coordinates to \cite{facevacs2010}. Eye detection is a critical pre-processing stage of \cite{facevacs2010} and it failed to detect eyes in a majority of surveillance view, low resolution, noisy and blurred images present in our experiment. Therefore, to perform an experiment of this nature, we disabled automatic eye detection and provided manually annotated eye locations to \cite{facevacs2010} for all the test and reference images used in this study. Also, it is important to mention that \cite{facevacs2010} is robust against pose deviation of $\pm 15^{\circ}$ from the frontal view and it has not been optimized to handle the pose variations included in this study.

We evaluate the performance of \cite{facevacs2010} for test and reference image quality variation as shown in \tablename\ref{tbl:test_ref_set_spec}. By varying one quality parameter (for example: resolution) at a time and keeping all the remaining four quality parameters constant, we report the recognition performance in terms of Area Under the ROC - AUC (for example: \figurename\ref{fig:re_auc_var}). For pose and illumination, we report AUC variation in \figurename\ref{fig:pi_man_eye_fv} for all possible combinations of pose and illumination in the test and reference set.  

For evaluation of resolution, motion blur and Gaussian noise, we select surveillance view (i.e. camera 19\_1) test images and the following two views for reference images: (a) frontal view (i.e. camera 05\_1 or mug shot view); (b) near surveillance view (i.e. camera 19\_0). These two pose variations in the reference set were included in our study in order to simulate different choices available to a forensic investigator in selecting the pose of the reference image. We report the corresponding recognition performance results in \figurename\ref{fig:re_ng_bm_results}. 

Some sample images used in this study are shown in \figurename\ref{fig:qual_var_illus}. Note that the cropped images in all the figures in this paper are only for illustration purpose and in the actual experiment, we used full view image (as shown for resolution variation in \figurename\ref{fig:qual_var_illus}). Also, the reported value of variance in zero mean Gaussian noise is for image intensity value $\in [0,1]$.

\subsection{Results}
\label{dutta2012impact_results}
A summary of overall difference in area under ROC (AUC) for individual image quality parameters is given in \tablename\ref{tbl:auc_variation_summary}. In the following sections, we analyze the recognition performance data corresponding to each quality parameter:

\begin{figure}[h]
\setlength{\unitlength}{1pt}
\centering
\begin{picture}(400,600)
  \put(-15,130){\includegraphics[width = 1\linewidth]{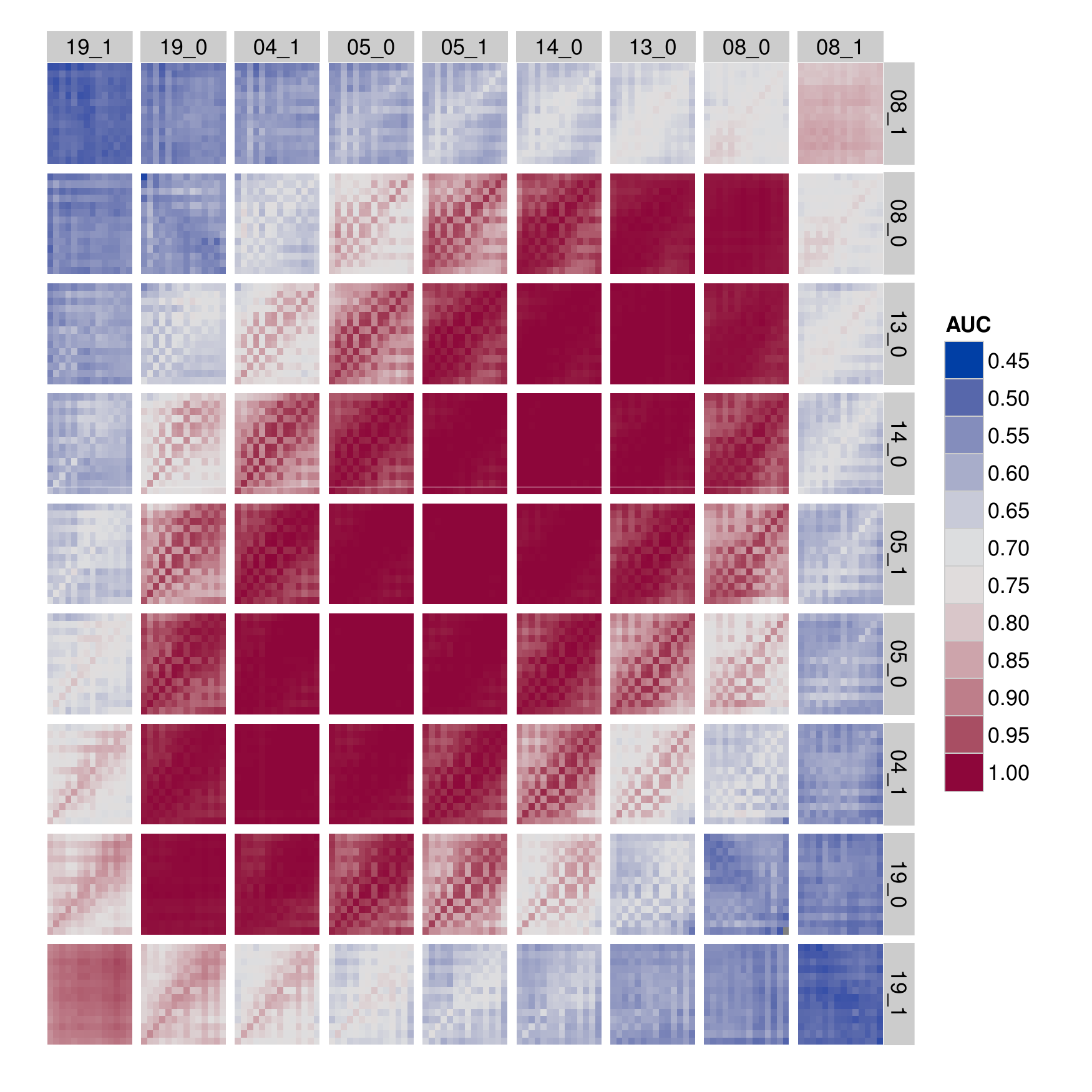}}
  \put(82,604){{reference camera id (pose variation)}}
  \put(-10,290){{\begin{sideways}
      {test camera id (pose variation)}
    \end{sideways}}}
  \put(10,-10){\includegraphics[width = 0.8\linewidth]{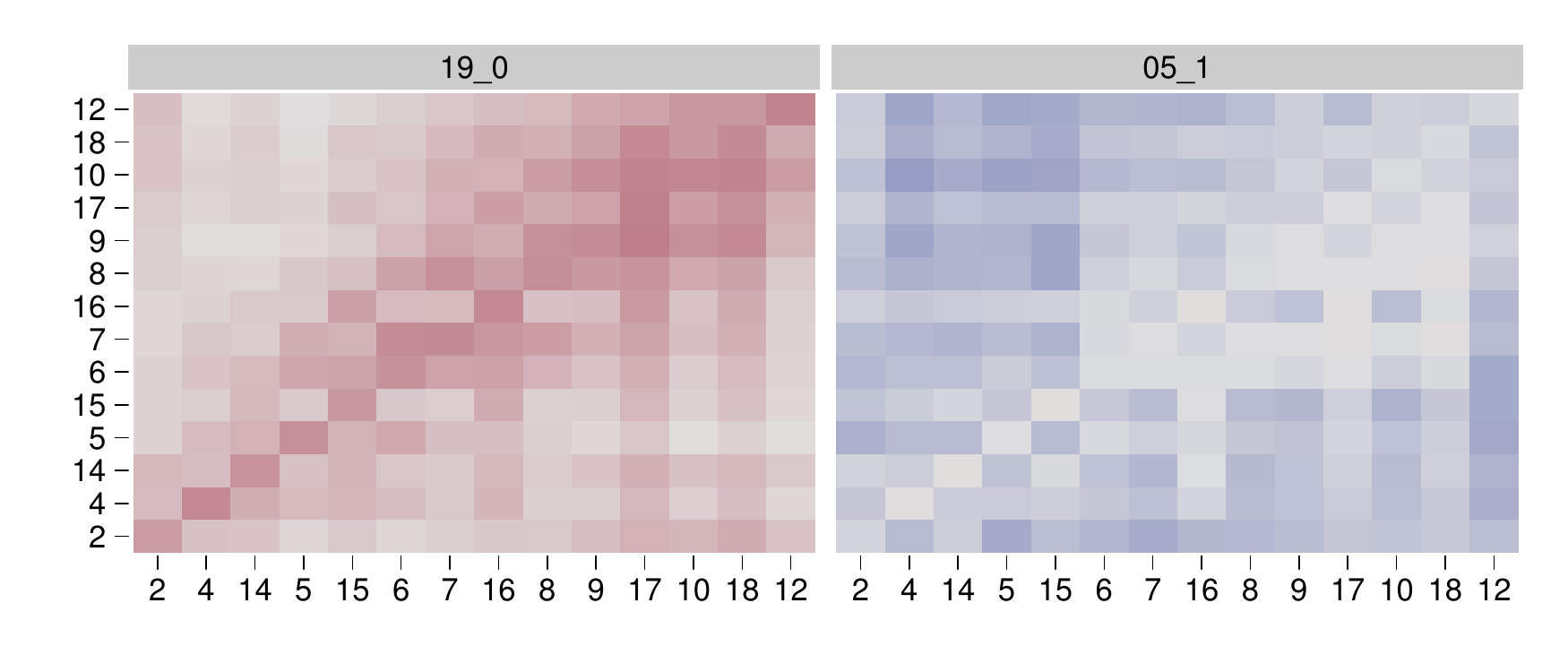}}
  
  \put(41,147){\framebox(37,44){}}
  \put(154,147){\framebox(37,44){}}
  \put(60, 144){\vector(1,-1){20}}
  \put(190, 144){\vector(1,-1){20}}
  
  \put(160,-6){reference flash id}
  \put(6,42){\begin{sideways}
      {test flash id}
    \end{sideways}}
    
	\put(162,560){\includegraphics[width=0.06\linewidth]{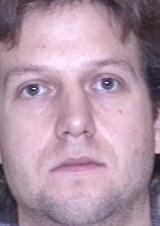}}
	\put(197,561){\includegraphics[width=0.06\linewidth]{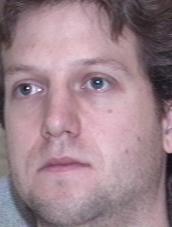}}
	\put(235,561){\includegraphics[width=0.06\linewidth]{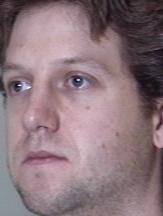}}
	\put(274,561){\includegraphics[width=0.06\linewidth]{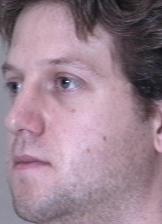}}
	\put(312,581){\includegraphics[width=0.06\linewidth]{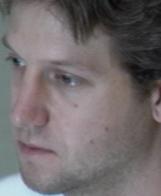}}
		
	\put(125, 560){\includegraphics[width=0.06\linewidth]{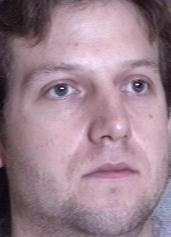}}		
	\put(88,561){\includegraphics[width=0.06\linewidth]{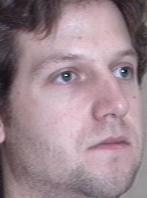}}
	\put(47,561){\includegraphics[width=0.06\linewidth]{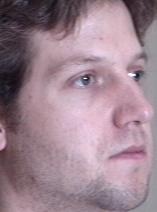}}
	\put(8,581){\includegraphics[width=0.06\linewidth]{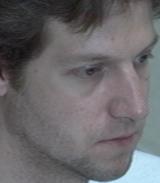}}
\end{picture}
\caption{Face recognition performance variation of \cite{facevacs2010} in terms of Area Under ROC(AUC) for all possible combination of pose and illumination variation.}
\label{fig:pi_man_eye_fv}
\end{figure}

\subsubsection{Pose and Illumination}
To compare recognition performance for pose and illumination variation, we show the AUC value in \figurename\ref{fig:pi_man_eye_fv} for all possible combination of pose and illumination in the test and reference set. Here, each cell block represents performance variation under all possible illumination variation for a fixed test and reference pose.
\begin{itemize}
 \item As expected, the frontal pose (i.e. camera 05\_1) test set has good recognition performance ($\sim 90\%$) for a large range of pose variation ($\pm 45^{\circ}$) in reference set. The recognition performance drops significantly for the surveillance view (i.e. camera 19\_1) reference set. Note that even near frontal pose trace images (captured by a CCTV camera at a crime scene) are rare in real forensic cases.
 
 \item We observe gradual reduction in recognition performance if the reference set pose moves away from the pose in the test set. This implies that near-optimal recognition performance can still be achieved with a reference set having a pose very close to the pose in the test set. In practice, it is very difficult to exactly match pose between test and reference images and therefore this result is very encouraging for practical forensics face recognition.

 \item For surveillance view test images (i.e. camera 19\_1), optimal recognition performance $\sim 95\%$ is achieved if the reference images are also captured by the same camera (i.e. 19\_1) -- irrespective of the illumination condition in the test and reference set. In real forensic cases, it is often not possible to acquire the CCTV camera that captured the trace. In such a case, sub-optimal recognition performance can be still be achieved with a suspect reference set having near surveillance view pose (camera 19\_0 : reference pose close to the original pose in test images). Performance can be further improved by matching the illumination direction in the test and reference images (AUC along the diagonal in bottom left plot of \figurename\ref{fig:pi_man_eye_fv}).
 
 \item It is common practice in the forensic community to chose frontal pose reference image (i.e. mug shots from police database based on intuition) irrespective of the pose in the test image. \figurename\ref{fig:pi_man_eye_fv} shows that comparison between a surveillance view (i.e. camera 19\_1) test set and the frontal view (i.e. camera 05\_1) reference set can only achieve maximum performance (i.e. AUC) of $\sim 75\%$. While the same surveillance view (i.e. camera 19\_1) test set when compared with near surveillance view (i.e. camera 19\_0) reference set can achieve performance $\sim 95\%$ by also matching the illumination condition.
 
 \item Worst possible recognition performance occurs if images captured by symmetrically opposite view are compared (for example: when images from camera 19\_1 and 08\_1 are compared, performance drops to $\sim 50\%$.).
 
 \item If there is an exact match between test and reference pose, the role of illumination is insignificant (Note, in the MultiPIE data set, if we exactly match the pose, we are also matching all the imaging characteristics). However, if there is a slight mismatch in pose, matching illumination between test and reference set can significantly improve the performance (see along diagonal for test and ref. pose 19\_1 and 19\_0 respectively).
\end{itemize}

\subsubsection{Resolution}
\label{ss:res}
\begin{figure}[htbp]
  \centering
  \subfloat[Image Resolution]{\label{fig:re_auc_var}\includegraphics[width=0.8\linewidth, trim=10 20 20 10, clip = true]{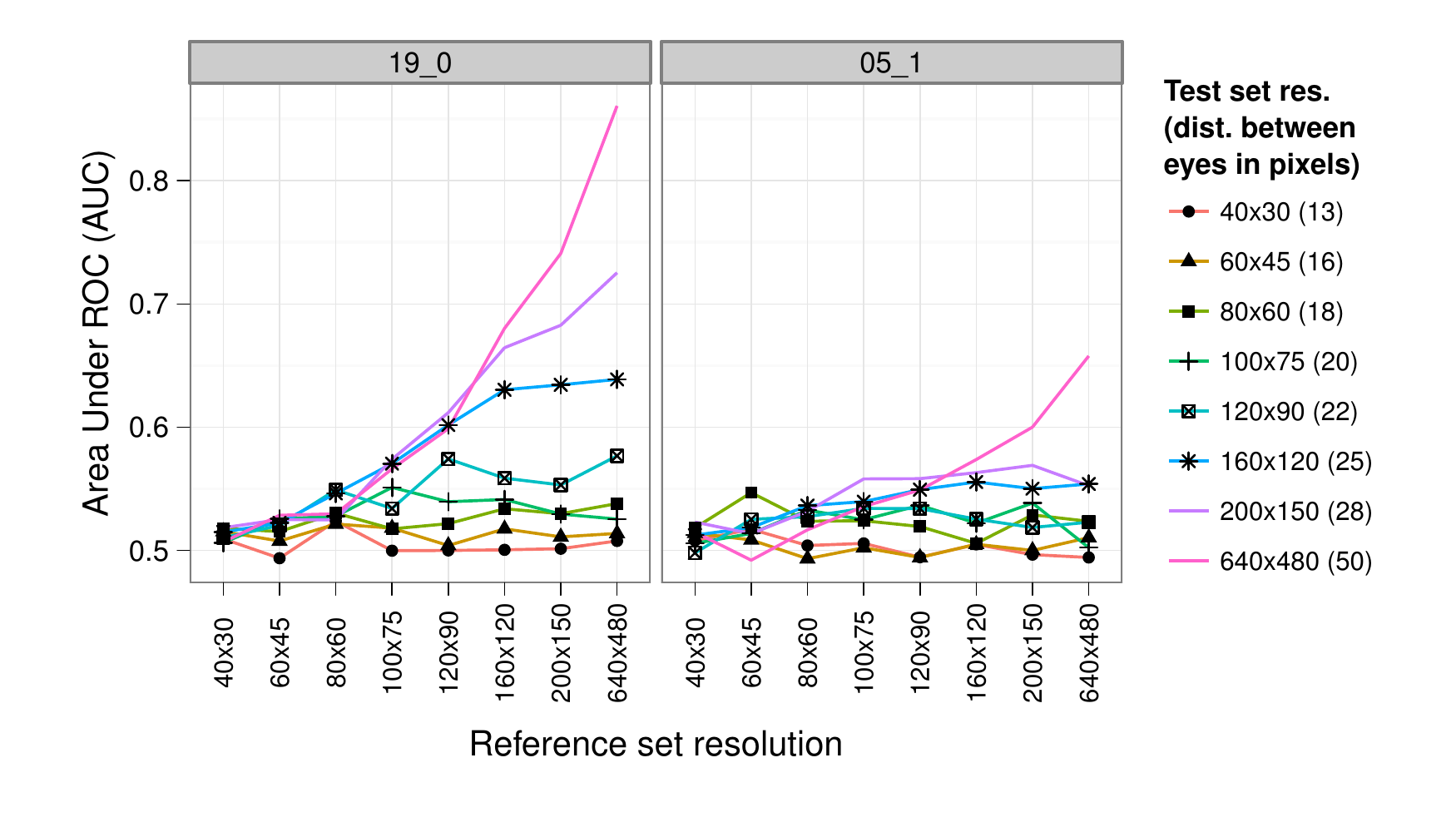}}

  \subfloat[Noise (Gaussian)]{\label{fig:ng_auc_var}\includegraphics[width=0.8\linewidth, trim=10 20 20 10, clip = true]{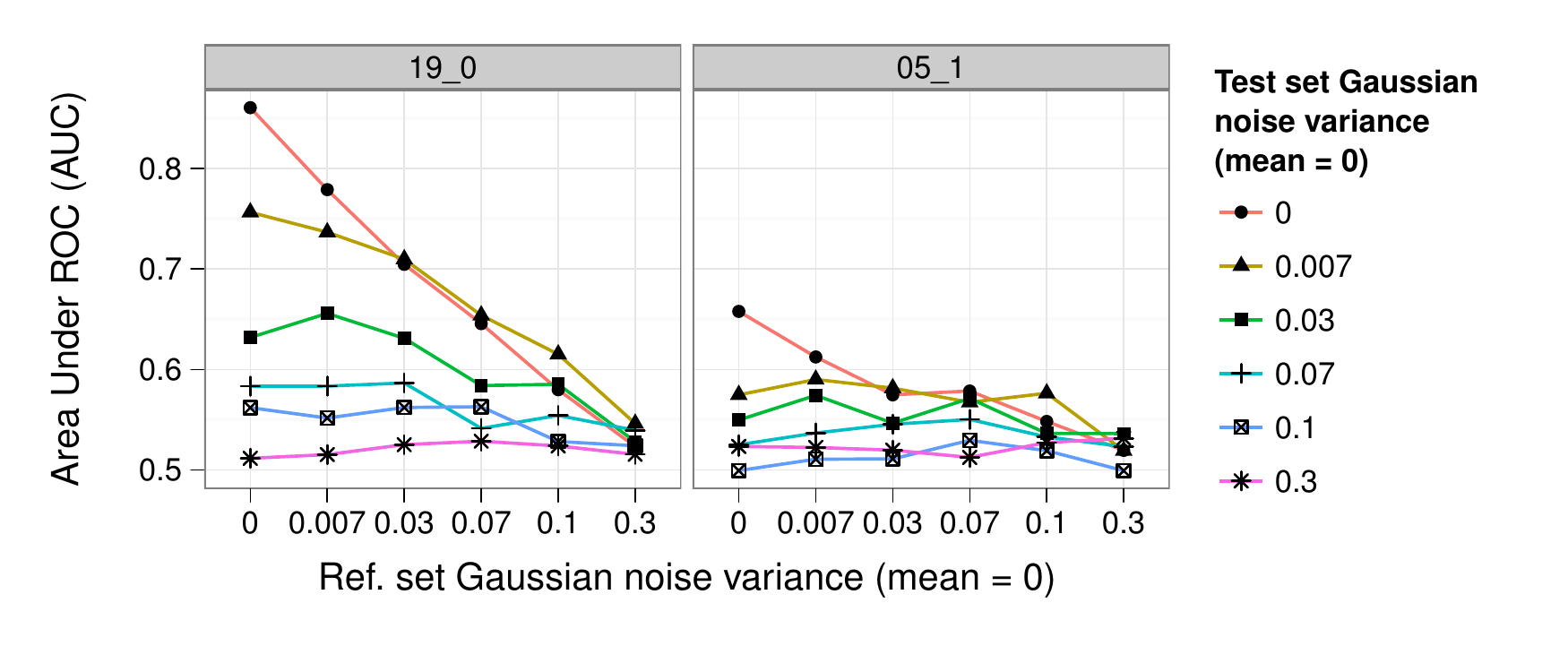}}

  \subfloat[Blur (Motion)]{\label{fig:bm_auc_var}\includegraphics[width=0.8\linewidth, trim=10 20 20 10, clip = true]{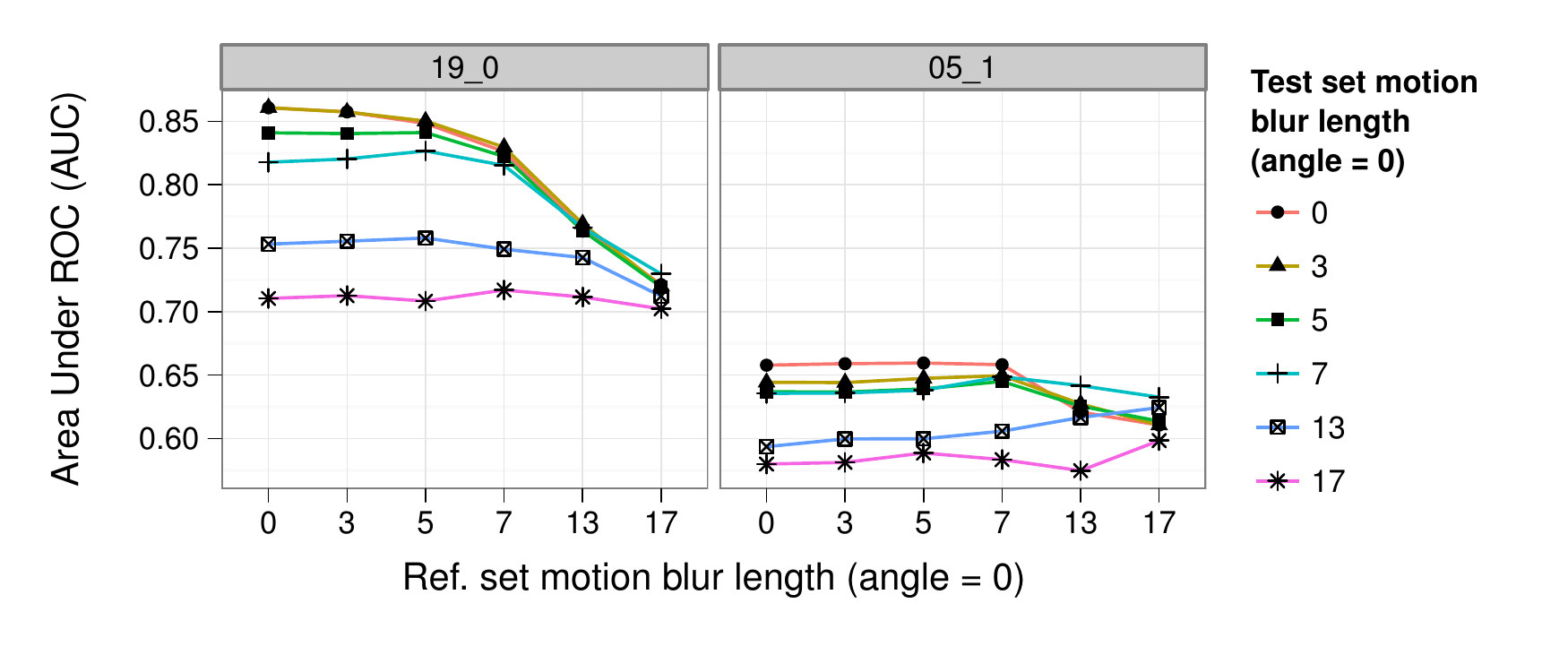}}
  
  \begin{picture}(1,1)
    \put(-130,540){\includegraphics[width = 0.08\linewidth]{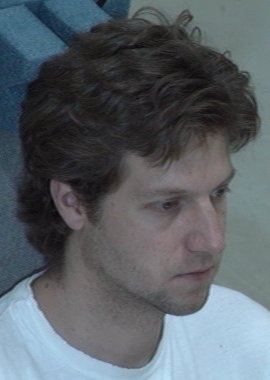}}
    \put(-80,540){\includegraphics[width = 0.08\linewidth]{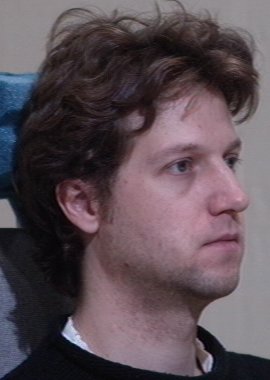}}

    \put(6,540){\includegraphics[width = 0.08\linewidth]{./images/dutta2012impact/mpie_img_sample/cropped_001_01_01_191_18.jpg}}
    \put(56,540){\includegraphics[width = 0.08\linewidth]{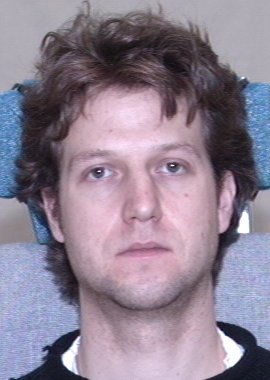}}

    \put(-160,540){\begin{sideways}
      {\small{Test \& ref.}}
      \end{sideways}}
    \put(-150,546){\begin{sideways}
      {\small{samples}}
      \end{sideways}}
  \end{picture}
  \caption{Face recognition performance variation of \cite{facevacs2010} in terms of Area Under ROC(AUC) for all possible combination of image resolution, noise, and, blur.}
  \label{fig:re_ng_bm_results}
\end{figure}

In \figurename\ref{fig:re_auc_var}, we report AUC value for different combinations of test and reference image resolution.


As expected, recognition performance improves with the resolution of the test and reference set. The resolution of test (or, reference) set constraints the maximum recognition performance achievable by varying the reference (or, test) set image resolution.

If test and reference set have similar pose (for example: test camera = 19\_1 and ref. camera = 19\_0), resolution variation has a more dramatic effect on recognition performance as compared to the case if they have very large difference in pose (for example: test camera = 19\_1 and ref. camera = 05\_0). In other words, the effect of resolution variation on recognition performance is very large if test and reference pose are similar.


\subsubsection{Noise (Gaussian)}
To study the effect of noise on recognition performance, in \figurename\ref{fig:ng_auc_var} we report AUC value for different combinations of noise in the test and reference set. We report this result for two combination of test and reference pose as described in \ref{ss:res}.

After pose, noise has the most significant effect on recognition performance. This implies that \cite{facevacs2010} is highly sensitive to noise in test or reference set images. As was the case with resolution, the effect of zero mean Gaussian noise on recognition performance is significant if the test and reference set have similar pose.


\subsubsection{Blur (Motion)}
Similarly, the effect blur on the recognition performance shown in \figurename\ref{fig:bm_auc_var} for all the possible combinations of motion blur in the test and reference image.

As expected, recognition performance degrades gradually as we increase motion blur in the test or reference set. Again, similar to the behaviour of resolution and noise, the effect of motion blur on recognition performance is significant if the test and reference set have similar pose.


\begin{table}[h]\footnotesize
  \vspace{-4mm}
  \centering
  \caption{Summary of difference in AUC}
  \begin{tabular}{l | c}
  \hline
  Quality & Difference in Area Under ROC \\
  \hline
  Pose & $\sim 50\%$ \\
  Resolution & $\sim 35\%$ \\
  Noise (Gaussian) & $\sim 35\%$ \\
  Blur (Motion) & $\sim 20\%$ \\
  Illumination & $\sim 20\%$ \\
  \hline
  \end{tabular}
  \label{tbl:auc_variation_summary}
\end{table}

\subsection{Conclusion}
\label{dutta2012impact_conclusion}
In this study, we have shown that if the pose between the test (trace) and reference (suspect set) images match exactly, we get the best recognition performance achievable using a particular face recognition system. We also observed gradual decrease in recognition performance as the difference in pose between test and reference set increased. This implies that even with a small mismatch in pose, we can still attain near optimal recognition performance. Therefore, in a real forensic evaluation cases involving face recognition, it is sufficient to approximately match the pose between the test and reference set.

If synthesis of frontal view images from a low quality trace (e.g. using \cite{blanz2005face}) is difficult, we recommend applying the method of \cite{blanz2005face} to the frontal view mug shots in the suspect reference set in order to synthesize non-frontal view images having pose similar to the trace image for use with view based face recognition system. We expect that this approach would helps attain near-optimal recognition performance.

Our study has also shown that the relative pose difference between test and reference images plays a critical role in determining the extent of performance degradation that is caused by variations in other quality parameters like illumination, noise, motion blur, and resolution.

Our findings in this paper are subject to at least three limitations. First, we have assumed that the image quality parameters are independent. In reality, all the quality parameters co-exist and presence or absence of one quality parameter (like pose, blur, etc) might affect the behavior of other quality parameters (like resolution, noise, etc). Second, all the images used in this study were taken from a single image data set. Although test and reference images differed by session, ideally both test and reference images should have been taken from the different data set in order to simulate the conditions present in a real forensic case. And, finally, these findings are limited by the inclusion of a specific commercial face recognition system in this study.

\clearpage

\section{View Based Approach to Forensic Face Recognition}
\label{dutta2012view_intro}
Forensic investigators now have access to video recordings of many crime scenes -- thanks to the omnipresent CCTV cameras. Such video recordings are often of very low quality and therefore rarely contribute to a strong evidence in the court of law, because forensic investigators find it difficult to compare and interpret the low quality facial images contained in these recordings. Automatic face recognition systems also have poor performance because they are fine tuned for optimal recognition performance when comparing good quality frontal view images.

One solution to deal with low quality face images is to reconstruct 3D face model from the CCTV images and synthesize the corresponding frontal view image. This strategy ensures that a face recognition system always gets to compare frontal view images, thereby ensuring optimal recognition performance. This approach is known as the model based approach. If the 3D model reconstruction is accurate and the synthesized frontal view image is of good quality, such a model based approach is known to deliver good recognition accuracy \cite{blanz2003face,blanz2005face,park20073dmodel,asthana2011fully}. In most forensic cases, the images extracted from the CCTV footage have surveillance view (range of azimuth $\pm45^{\circ}$, elevation $\sim30^{\circ}$) as shown in \figurename\ref{fig:test_set_illus}. Therefore it is very difficult to synthesize the corresponding good quality frontal view images that can be compared to the reference image that is usually a frontal mug shot.

In the forensic context, little attention has been paid to the view based approach first examined by \cite{pentland1994view}. This approach involves adapting the test and reference images so that a face recognition system always gets to compare images under similar view -- not necessarily the frontal view. The basis for the view based approach is that, given appropriate training and suitable classifiers, comparing non-frontal view facial images is no more difficult than comparing frontal view images and some face recognition algorithms (for example LBP \cite{ahonen2006face}) perform equally well in both tasks. This approach has not been studied well because it is often not practical to capture reference images from all possible pose and illumination variations. 

In this paper, we study the use of the view based approach for forensic cases where there is a possibility of capturing suspect reference images from desired pose and illumination using a desired camera. Our results on the MutltiPIE data set \cite{gross2008multipie} shows that exactly matching pose, illumination and camera between test and reference images delivers improved recognition performance across five different types of face recognition systems.

\subsection{Related Work}
A forensic evaluation case involving face recognition often involves surveillance view images. There are generally two approaches available to deal with non-frontal view (or, pose) facial images in a face comparison process using a pre-trained view based face recognition system: \begin{inparaenum}[\itshape a\upshape)] \item Model based approach \item View based approach \end{inparaenum}. 

The model based approach \cite{blanz2003face,blanz2005face,park20073dmodel,asthana2011fully} exploits the fact that most face recognition systems are fine tuned for optimal recognition performance when comparing frontal view images. This approach begins with reconstruction of a 3D face model from non-frontal view test image followed by synthesis of a frontal view test image (also called virtual test image) for comparison with the frontal view reference images. This approach is applied to all the non-frontal view images present in either the test or the reference set so that a view based face recognition system only compares frontal view face images -- thereby ensuring optimal recognition performance.

To the best of our knowledge, results based on the model based approach has only been reported for non-surveillance view images. In \cite{blanz2005face}, the authors synthesized frontal view images corresponding to the non-frontal view images using a 3D Morphable Model (3DMM) and reported large improvement in recognition performance due to this view transformation. The results were based on good quality images captured at the eye level (i.e. non-surveillance view). More recently, \cite{asthana2011fully} proposed a 3D pose normalization method based on a view based Active Appearance Model (AAM) in order to synthesize a frontal image from a non-frontal view and reported improved recognition performance on five different image data sets. Although performance improvement was reported for $\pm45$ pose variation, surveillance-view images were not included in the study. In \cite{park20073dmodel}, the authors used Structure from Motion (SFM) to infer 3D face shape information by tracking a large number of feature points in a video sequence. Again, the reported improvement in recognition performance were based on a non-surveillance view video sequence.

The authors of \cite{pentland1994view} used the view-based approach to address the problem of recognition under general viewing orientation. They partitioned the face space into view-specific regions and compared a given non-frontal test image using eigenfaces of a particular region of the face space (corresponding to the view in the test image). The basic idea was to compare face images under similar view. 

In this paper, we investigate whether it is useful to apply a similar view-based approach in forensic cases where the test image is usually of very low quality. In section \ref{dutta2012view_recog-exp}, we describe the experimental setup that we used to study the performance of the model and view based approach for the surveillance view test set taken from the MultiPIE data set \cite{gross2008multipie}. In section \ref{dutta2012view_discussion}, we discuss the performance of five pre-trained face recognition systems for this setup. Finally, based on these results, we present our recommendations for the forensic community.

\subsection{Recognition Experiment and Results}
\label{dutta2012view_recog-exp}
With the experiment described in this section, we want to test the performance of the model and view based approaches in a scenario commonly encountered in forensic cases. For both approaches, we evaluate the performance of the following five face recognition systems: two commercial face recognition systems FaceVACS (denoted by A) and Verilook (denoted by B), Local Region PCA (LR-PCA) and LDA - I/Red (LDA-IR) \cite{phillips2011introduction}, and Local Binary Pattern (LBP) \cite{ahonen2006face} where, PCA and LDA are holistic methods while LBP is a local method.
We use the value of True Positive Rate (TPR) at False Positive Rate (FPR) of 0.001 as the metric for recognition performance comparison.
In addition to the FaceVACS system, the authors of~\cite{klontz2013case} have also used NEC NeoFace 3.1 for their case study on unconstrained facial recognition.
We limited our study to these four face recognition systems because of their availability in our research group.

\begin{figure}
  \centering
  \includegraphics[width = 0.8\linewidth, trim=0 0 0 0, clip = true]{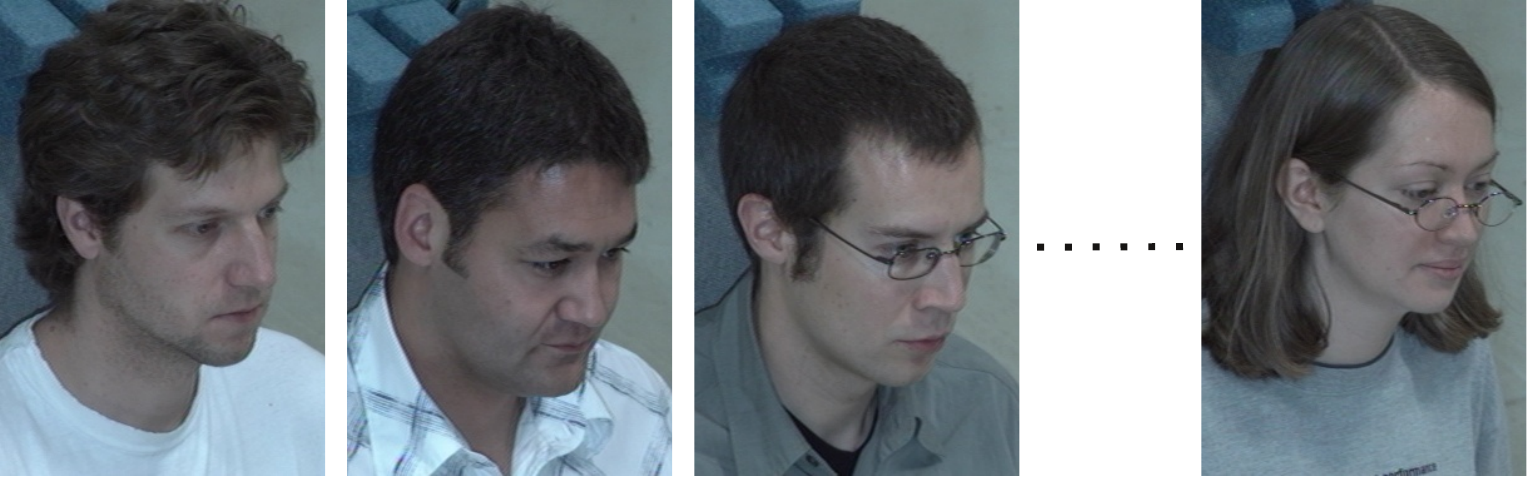}
  \caption{Sample of surveillance view images commonly encountered in forensic cases (taken from MultiPIE \cite{gross2008multipie})}
  \label{fig:test_set_illus}
\end{figure}

\begin{figure}
  \centering
  \includegraphics[width = 0.8\linewidth, trim=0 0 0 0, clip = true]{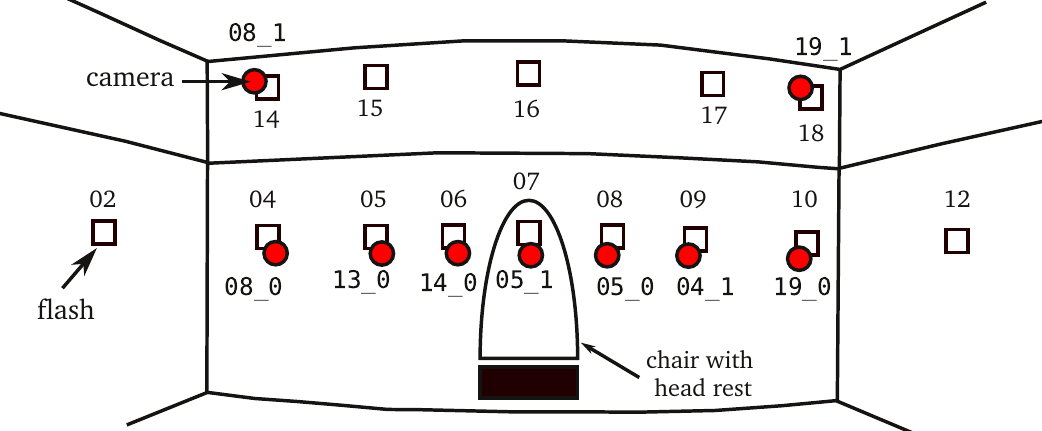}
  \caption{Position of camera (red circles, e.g. 19\_1) and flash (black squares, e.g. 18) in the MultiPIE collection room \cite{gross2008multipie}.}
  \label{fig:mpie_cap_setup}
\end{figure}

Our test set (or probe) consists of surveillance view images of 249 subjects in session 01 with illumination that is frontal with respect to the face: \texttt{(01,19\_1,18)}\footnote{We use the notation \texttt{(session-id,cam-id,flash-id)} to denote a subset of MultiPIE data set with neutral expression} as shown in \figurename\ref{fig:test_set_illus}. The reference set consists of frontal images of 239 subjects in session 04 with frontal illumination: \texttt{(04,05\_1,07)}. The camera and flash positions of the MultiPIE capture environment are shown in \figurename\ref{fig:mpie_cap_setup}. Note that session 01 (test set) and session 04 (ref. set) were captured six months apart\footnote{based on communication with an author of \cite{gross2008multipie}} and therefore this experiment simulates the session variation present in real forensic cases.

The model and view based approaches differ in the way they transform the reference images. In the following sections, we discuss the details of this process by which the reference set is transformed in these two approaches:
\\

\noindent\textbf{Model Based Approach :} There are several methods to implement the model based approach \cite{blanz2005face,park20073dmodel,asthana2011fully}. In this paper, we use the 3D Morphable Model (3DMM) based method of \cite{blanz2003face,blanz2005face} to synthesize frontal view image corresponding to a given surveillance view image shown in  \figurename\ref{fig:test_set_illus}. We manually annotate 10 landmarks in the test image and then fit the Basel Face Model \cite{paysan20093dface}. We then synthesize a frontal view image using the estimated pose, shape, texture and illumination. To study the effect of texture on recognition performance, we synthesize two images as shown in \figurename\ref{fig:effect_of_tex_illus}. The first image contains texture from the morphable model and since \cite{paysan20093dface} is based on 200 faces, it is unable to reproduce local characteristics such as moles or scars. The second image contains partial texture from the original image supplemented with morphable model texture in the occluded regions. Therefore, we observe some artifacts in the synthesized image which has also been reported in \cite{paysan20093dface}. One possible reason for this artifact is the mapping of non-face (e.g. background) pixels to the model because the shape fitting process was not $100\%$ accurate. 

The result of face comparison between synthesized frontal view image and frontal view reference photograph using the five face recognition systems is shown in \figurename\ref{fig:roc_model_approach} (with only morphable model texture) and \figurename\ref{fig:roc_model_approach_with_tex} (with partial original texture supplemented by morphable model texture). The corresponding true positive rate values (at FPR = 0.001) are shown in \tablename\ref{tbl:auc_values}.\\

\noindent\textbf{View Based Approach :} Recall that in a view based approach, the reference image is chosen such that its pose closely matches the pose in the test set (i.e the surveillance view). In this paper, we investigate two scenarios relevant to real forensic cases. The first is a more ideal case where we have access to the original CCTV camera (that captured the test image) and it is possible to capture the suspect's photograph from exactly the same pose and illumination condition present in the test image. Second, is a more constrained case where we neither have access to the original CCTV camera nor are able to photograph suspects under exactly the same pose and illumination condition present in the test image. The second case is often encountered in real forensic cases.

The first case can be simulated with a reference set consisting of surveillance view images taken from session 04: \texttt{(04,19\_1,18)}. This reference set not only exactly matches the pose and illumination in the test set but also matches the camera as shown in \figurename\ref{fig:roc_view_approach_191} (top). Recognition performance for such a test and reference set is shown in \figurename\ref{fig:roc_view_approach_191} and the corresponding true positive rate values (at FPR = 0.001) are shown in \tablename\ref{tbl:auc_values}.

To simulate the second case, we create a reference set consisting of near-surveillance view images taken from session 04 with illumination that is frontal with respect to the face: \texttt{(04,19\_0,10)}. The 19\_1 and 19\_0 camera positions in the MultiPIE data set differ by an elevation and azimuth angle of $25.9^{\circ}$ and  $0.3^{\circ}$ respectively as shown in \figurename\ref{fig:mpie_cap_setup}. In reality, we can more closely match the pose between test and reference images. Recognition performance for such a test and reference set is shown in \figurename\ref{fig:roc_view_approach_190} (bottom) and  the corresponding true positive rate values (at FPR = 0.001) are shown in \tablename\ref{tbl:auc_values}.

\subsection{Discussion}
\label{dutta2012view_discussion}
For the model based approach, performance across all five systems degrades dramatically when the synthesized frontal view image contains texture from the morphable model as shown in \figurename\ref{fig:roc_model_approach}. With partial texture from the original test image mapped to the synthesized frontal view image, the performance improves for commercial systems A (0.36) and B (0.13) as shown in \figurename\ref{fig:roc_model_approach_with_tex} which shows that, to some extent, these systems are robust to the artifacts present near the boundary of the actual and synthesized texture. On the other hand, LR-PCA, LDA-IR (the holistic methods) and LBP (the local method) show virtually no improvement (at FPR = 0.001) in performance because they are only trained for comparing near frontal views and are also unable to handle the artifacts. These results also show that texture in the synthesized frontal view image is critical to face recognition performance in the model based approach. It is important to realize that our true positive rate values for the model based approach (for instance: TPR $=0.36$ at FPR of $0.1\%$) are significantly lower than that reported in \cite{blanz2003face,blanz2005face} because our test set contains surveillance view images while \cite{blanz2003face,blanz2005face} used non-frontal images captured at the eye-level.

View based approach delivers improved performance across all the five face recognition systems when pose, illumination and camera match exactly between the test and reference images as shown in \figurename\ref{fig:roc_view_approach_191}. For the test and reference set captured by different camera and having large mismatch in pose and illumination, only the commercial system A (and to some extent LBP) shows slight improvement in performance at FPR = 0.001. This reflects the capability of system A (and to some extent of LBP) to handle pose mismatch when comparing non-frontal view images.

\begin{figure}
  \centering
  \includegraphics[width = 0.6\linewidth, trim=0 0 0 0, clip = true]{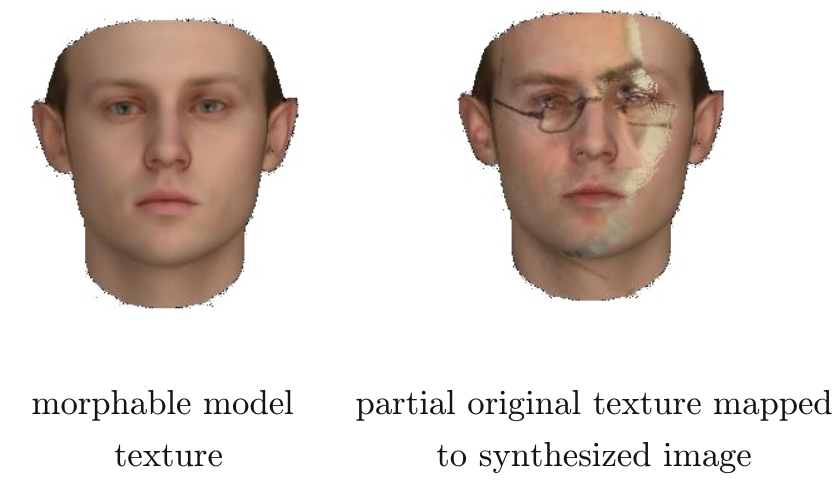}
  \caption{Synthesized frontal view image with two different types of texture}
  \label{fig:effect_of_tex_illus}
\end{figure}

\begin{figure*}
  \centering
  \subfloat[Model Based Approach (with only morphable model texture)]{\label{fig:roc_model_approach} 
    \includegraphics[width = 0.48\linewidth, trim=20 20 0 10, clip = true]{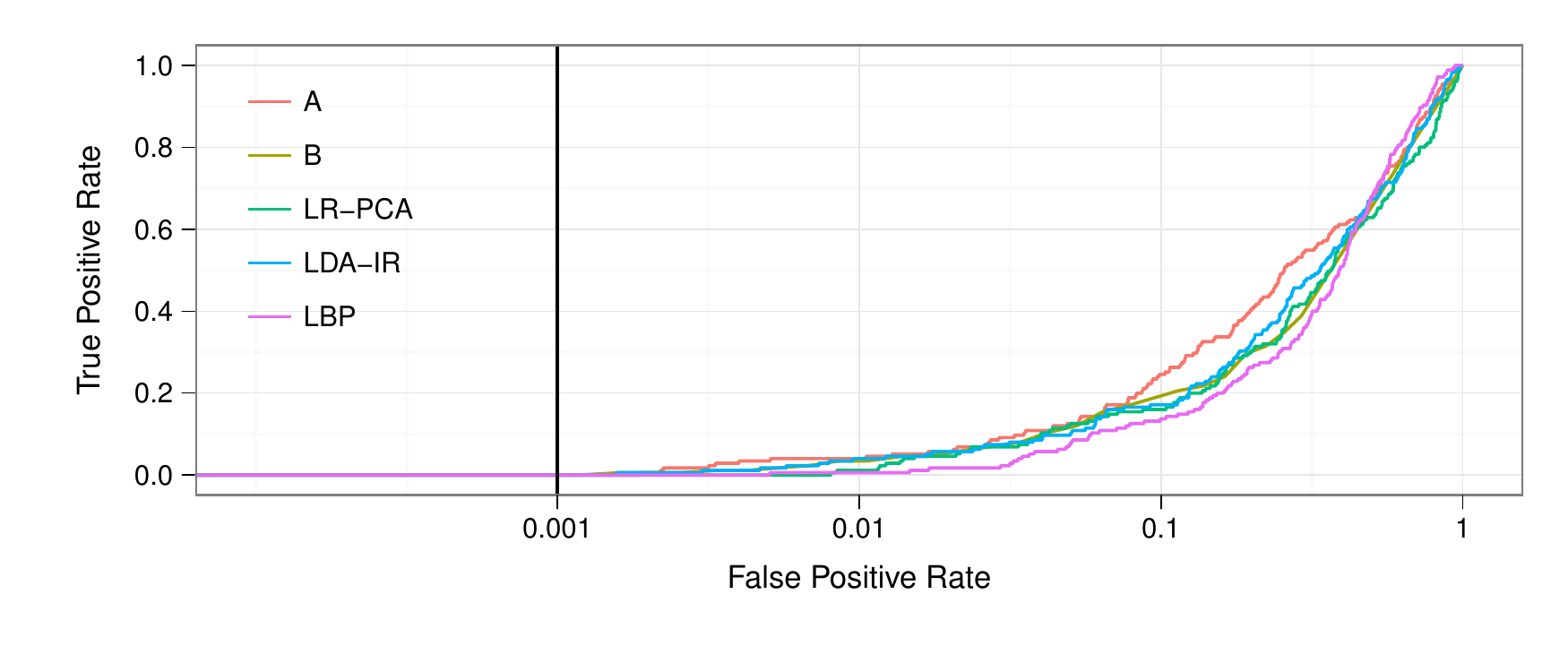}
    \includegraphics[width = 0.3\linewidth, trim=0 0 0 0, clip = true]{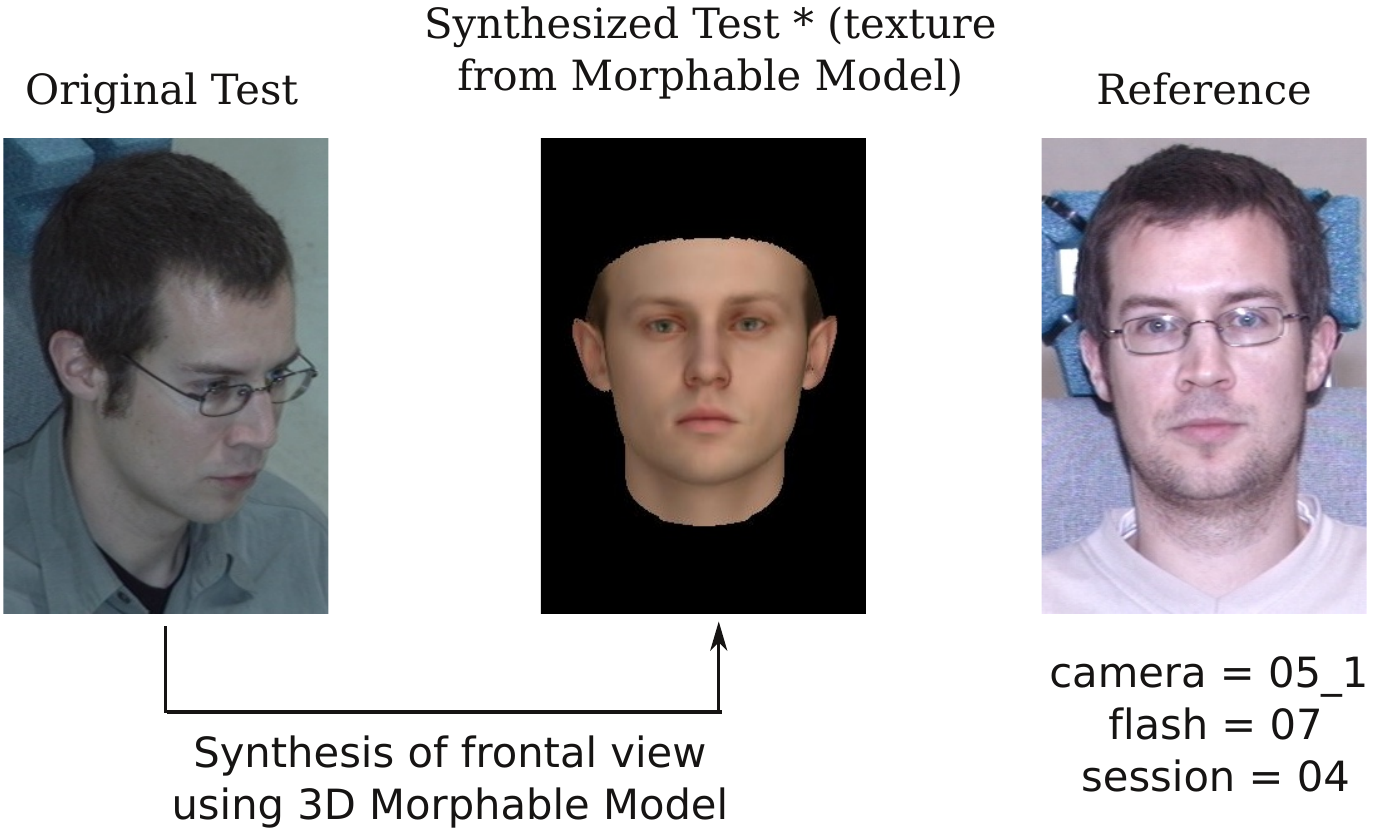}
  }

  \subfloat[Model Based Approach (with partial original texture supplemented by morphable model texture)]{\label{fig:roc_model_approach_with_tex} 
    \includegraphics[width = 0.48\linewidth, trim=20 20 0 10, clip = true]{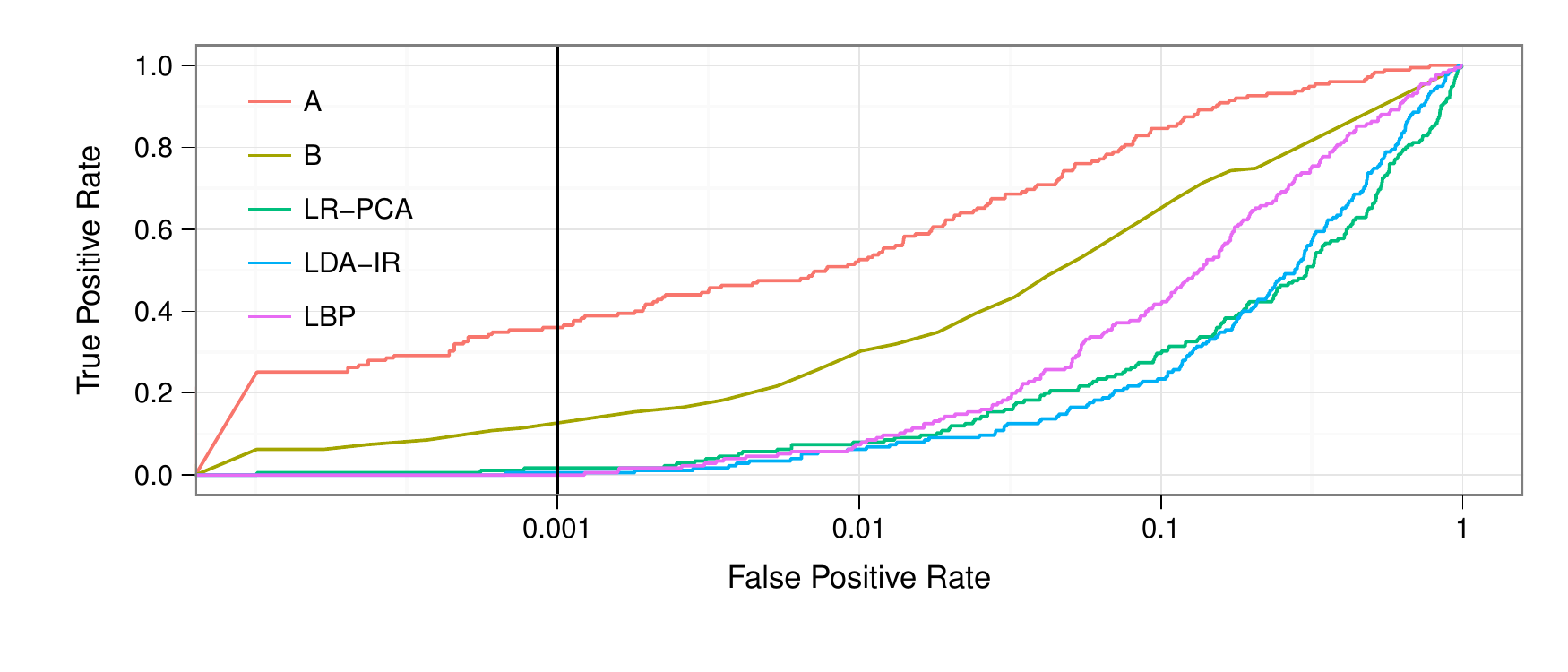}
    \includegraphics[width = 0.3\linewidth, trim=0 0 0 0, clip = true]{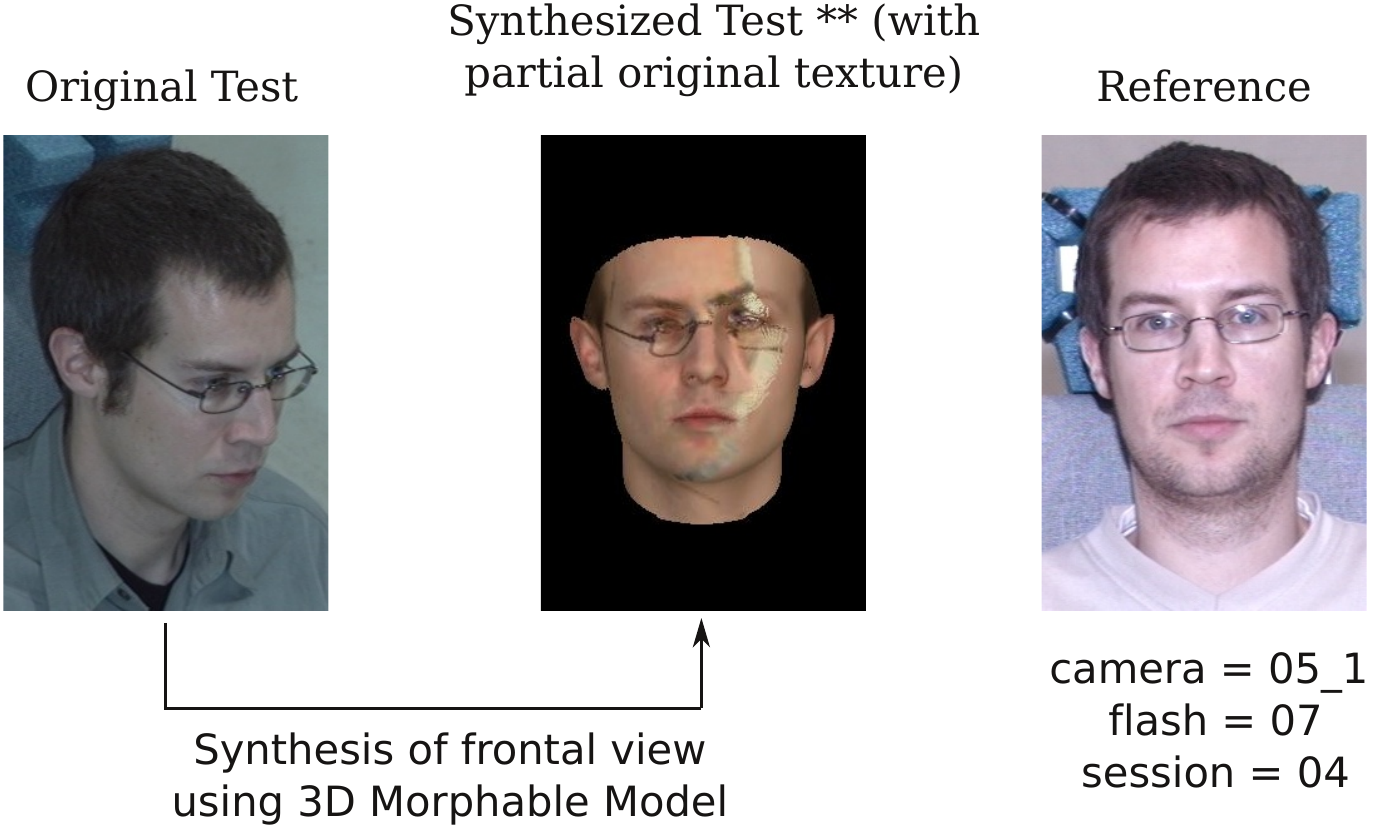}
  }

  \subfloat[View Based Approach (exact match of pose, illumination and camera)]{\label{fig:roc_view_approach_191} 
    \includegraphics[width = 0.48\linewidth, trim=20 20 0 10, clip = true]{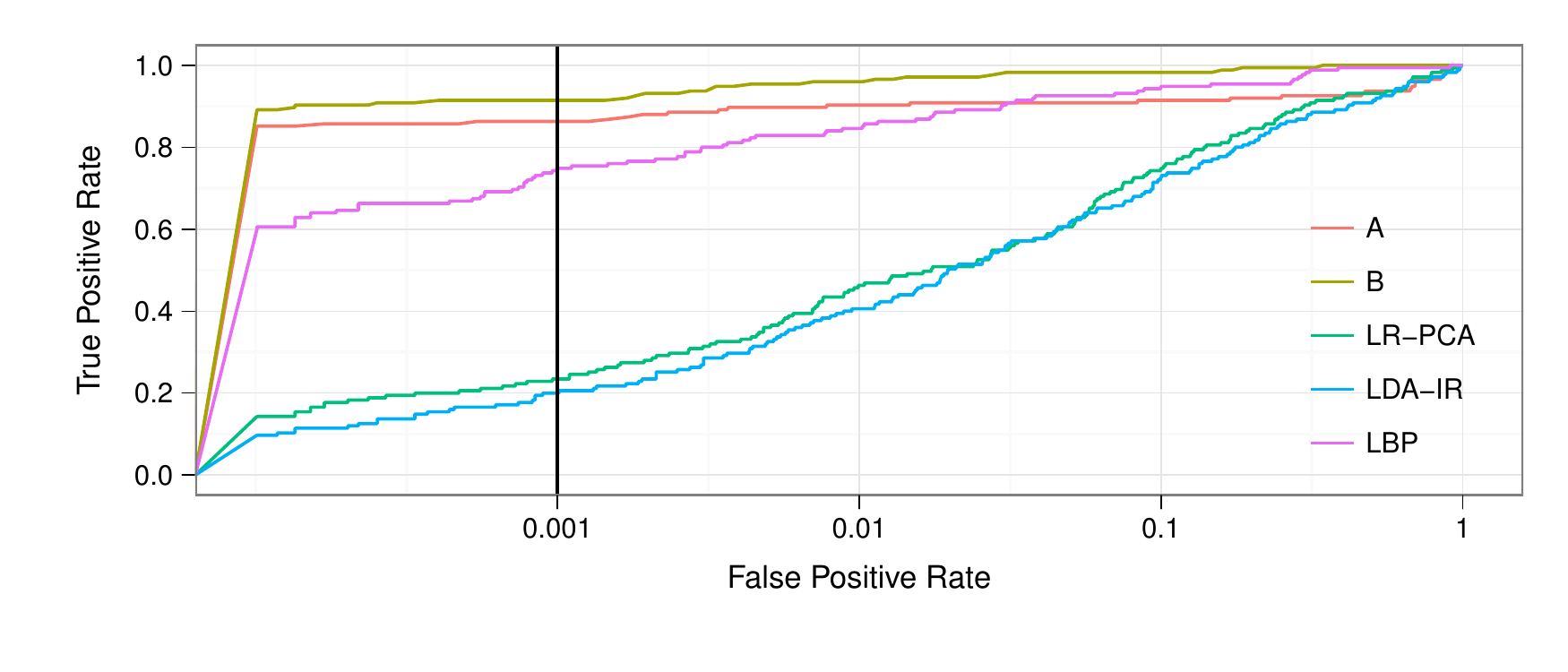}
    \includegraphics[width = 0.3\linewidth, trim=0 0 0 0, clip = true]{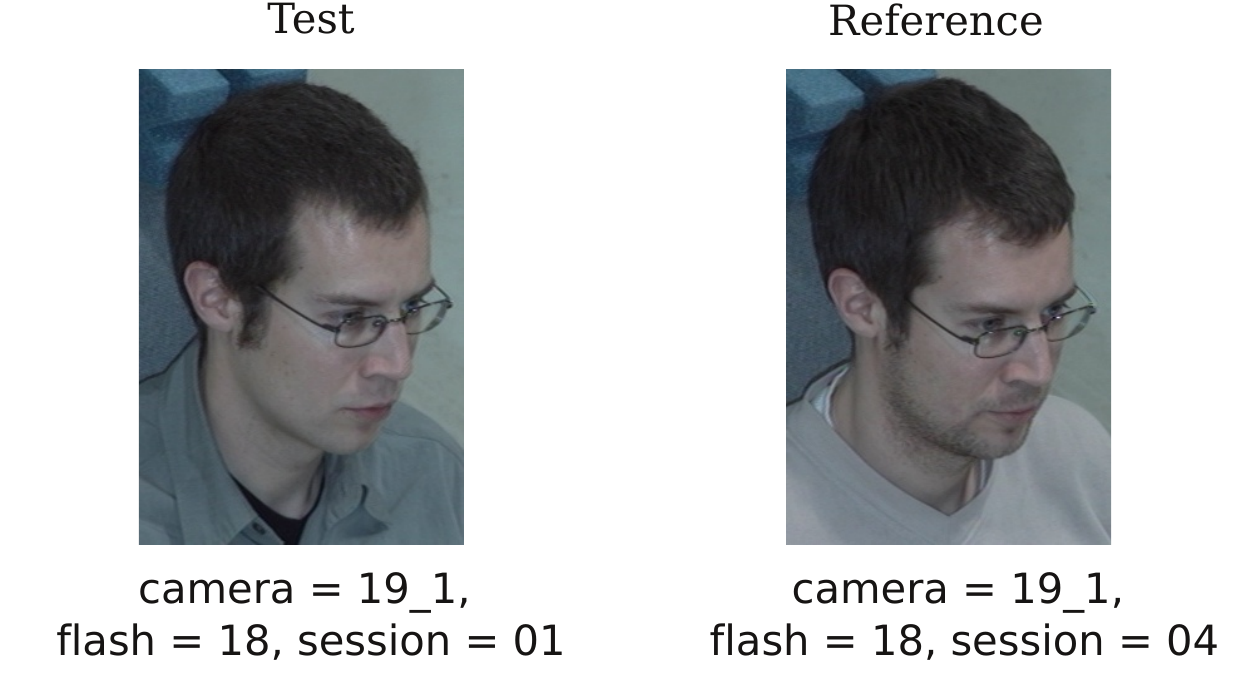}
  }

  \subfloat[View Based Approach (large mismatch in pose and illumination)]{\label{fig:roc_view_approach_190} 
    \includegraphics[width = 0.48\linewidth, trim=20 20 0 10, clip = true]{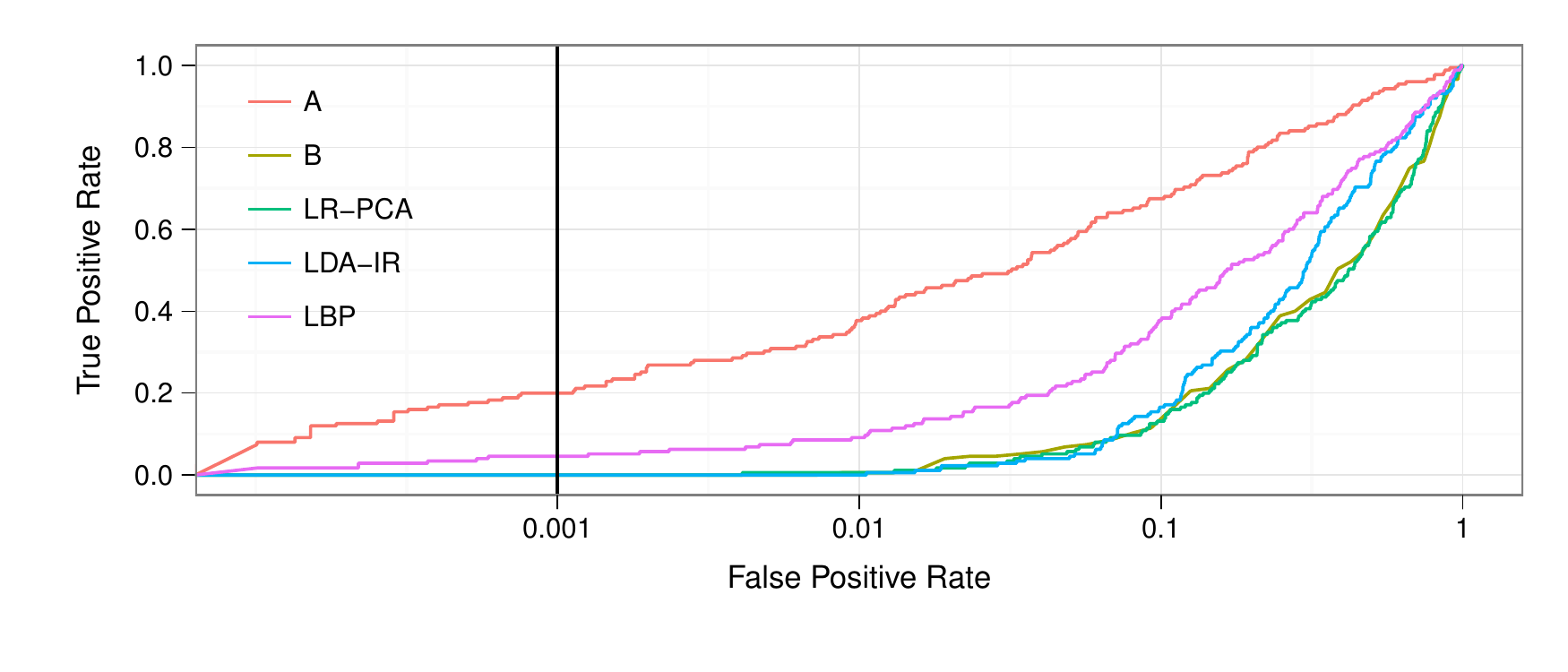}
    \includegraphics[width = 0.3\linewidth, trim=0 0 0 0, clip = true]{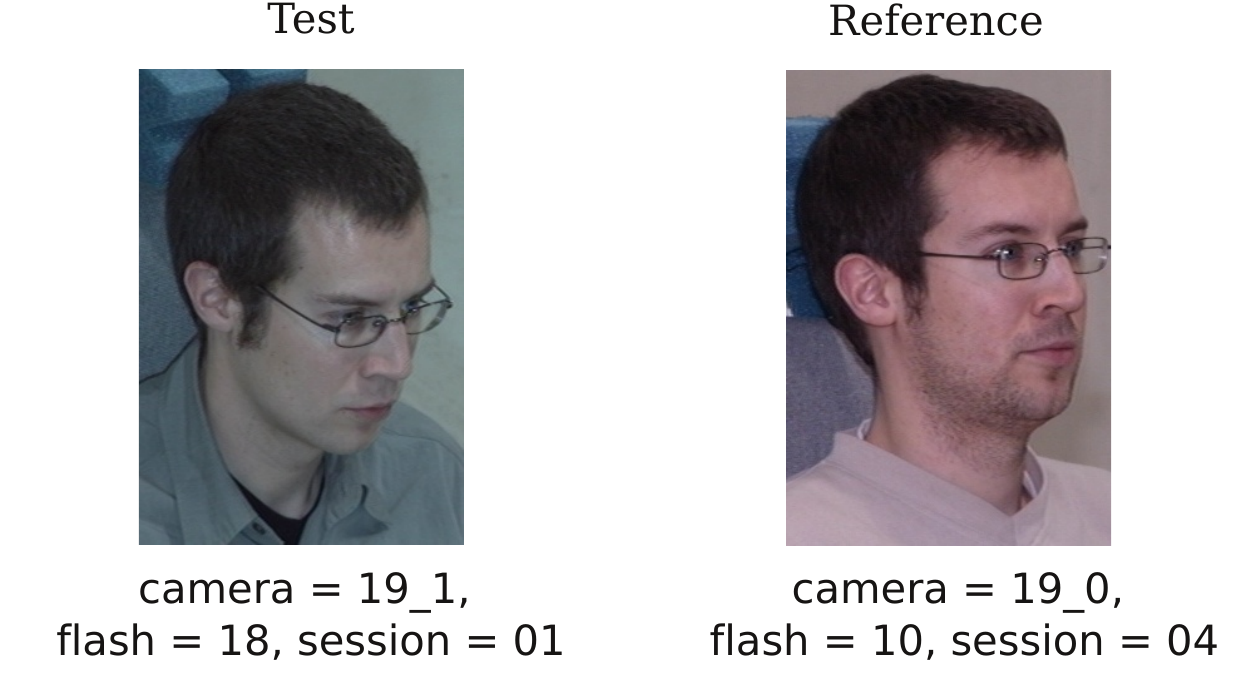}
  }
  
  \caption{Face recognition performance using the model and view based approaches applied to a surveillance view test set. Note: A and B are commercial face recognition systems and the False Accept Rate axis is in log scale.}
  \label{fig:result}
\end{figure*}

In forensic cases, if we can synthesize good quality frontal view image with original texture, then a face recognition system robust to image synthesis artifacts (as shown in \figurename\ref{fig:effect_of_tex_illus} - right) can provide good recognition performance. However, in most real forensic cases, the test image is of very low quality and it is difficult (and often not possible) to synthesize good quality frontal view image with the original texture.

Under such a constraint, our study shows that a forensic investigator has the following two options. First, is to acquire the camera that captured the original test image (i.e. the trace) and photograph the suspects from exactly the same pose and illumination. Our results show that this approach results in improved performance across all five face recognition systems included in this study. Second, is to approximately match the pose and illumination in the test and reference images captured by different camera. Our results shows that performance of some face recognition systems (for instance system A and LBP) show a sign of improvement even if the test and reference images are captured by different camera have large mismatch in pose and illumination.

\subsection{Conclusion}
\label{dutta2012view_conclusion}
For a forensic evaluation case involving face recognition, our results show that the proposed view based approach delivers improved recognition performance if: \begin{inparaenum}[\itshape a\upshape)] \item it is possible to exactly match pose, illumination and camera between the test and reference set images, and \item you have access to a face recognition system that can compare non-frontal view images\end{inparaenum}. It is still possible to attain good performance by approximately matching pose and illumination in the test and reference images captured by different camera. Our results also show that the model based approach should only be applied if: \begin{inparaenum}[\itshape a\upshape)] \item it is possible to synthesize good quality frontal view image with the original texture, and \item you have access to a face recognition system that can handle artifacts caused by image synthesis techniques \end{inparaenum}.

Future research could investigate how the proposed view based approach performs with non-frontal reference images synthesized by applying image synthesis techniques used in the model based approach. In real forensic cases, quite often, it is not possible to photograph the suspects and the forensic investigator has access only to frontal view (mug shot) images of the suspects. Under such a constraint, we expect the synthesized non-frontal view images to be of good quality because of the relatively better quality of frontal test images (mug shot) in the reference set. The case when we exactly match pose, illumination and camera depicts the performance achievable using photo-realistic synthesis of non-frontal view reference image.

\begin{table}
  \centering
  \footnotesize
  \begin{tabular}{ c| c | c| c| c| c| c| c|}
      \multirow{2}{*}{Fig.} & \multirow{2}{*}{Test Pose} & \multirow{2}{*}{Ref. Pose} & \multicolumn{5}{c}{True Positive Rate (at FPR = 0.001)} \\
		  \cline{4-8}
		  &  &  & A & B & LR-PCA & LDA-IR & LBP \\
      \hline\noalign{\smallskip}
			 Fig.~\ref{fig:roc_model_approach} & *synth. frontal & frontal & 0 & 0 & 0 & 0 & 0 \\

			 
			 Fig.~\ref{fig:roc_model_approach_with_tex} & **synth. frontal & frontal & 0.36 & 0.13 & 0.01 & 0.01 & 0 \\
			 
			 Fig.~\ref{fig:roc_view_approach_190} & surveillance & surveillance & \textbf{0.86} & \textbf{0.91} & 0.23 & 0.20 & \textbf{0.75}  \\

			 Fig.~\ref{fig:roc_view_approach_191} & surveillance & near-surveillance & 0.2  & 0 & 0 & 0 & 0.05  \\
			\hline
  \end{tabular}
  \caption{Numerical value of True Positive Rate corresponding to False Positive Rate of 0.001 for the model and view based approaches}
  \label{tbl:auc_values}
\end{table}

\clearpage

\section{Conclusions}
The preliminary experiments presented in Section~\ref{dutta2012impact_intro} were designed to reveal the influence of pose and illumination variations in probe and reference set.
The results from these experiments show that facial pose mismatch between probe and reference image played a key role in determining overall recognition performance and the influence of other quality parameters like resolution, noise, blur on performance.

Our findings in Section~\ref{dutta2012view_intro} suggests that a view based approach is potentially more useful than a model based approach in forensic cases involving face recognition.
Recall that a view based approach~\cite{pentland1994view} involves adapting the reference set such it matches the pose of the probe set.
In forensic face recognition cases involving surveillance view probe image, a view based approach requires a reference set containing images of suspects captured from the same pose as present in the surveillance view probe image.
On the other hand, a model based approach involves reconstruction of a 3D face model from the surveillance view probe image followed by synthesis of a corresponding frontal view probe image while the suspect reference set remains fixed to frontal view.

We recommend using the view based approach if
\begin{inparaenum}[\itshape a\upshape)]
\item it is possible to exactly match the pose, illumination condition and camera of the suspect reference set to that of the probe image (or, forensic trace acquired from CCTV footage); and
\item one uses a face recognition system that is capable of comparing non-frontal view facial images with high accuracy.
\end{inparaenum}
A view based approach in forensic cases may not always be practical because matching pose and camera requires cooperative suspects and access to the same camera that captured the probe image.

The model based approach should only be considered if
\begin{inparaenum}[\itshape a\upshape)]
\item it is possible to synthesize good quality frontal view image with the original texture; and
\item one uses a face recognition system that can handle the texture artifacts in the synthesized frontal view image.
\end{inparaenum}
In forensic cases, the probe image is often of very low quality and therefore it is difficult to synthesize corresponding good quality frontal view images.

\chapter{Conclusions}
\label{dutta2014phdthesis_conclusions}
In this chapter, we revisit our main research question and the resulting subordinate research questions presented in Section~\ref{dutta2014phdthesis_intro-resques} and present our conclusions as follows:
\begin{enumerate}
\item \textit{\ResearchQuestionTwo{}}

In Chapter~\ref{dutta2015predicting_intro}, we present a generative model to capture the relation between image quality features $\mathbf{q}$ (\eg pose, illumination, \etc) and face recognition performance $\mathbf{r}$ (\eg FMR and FNMR at operating point).
This model is based solely on image quality features because results presented in Chapter~\ref{dutta2014phdthesis_pred-feat-intro} showed that features derived solely from similarity scores are unstable under image quality variations.
This design decision not only avoids the issues associated with features derived from similarity scores but also allows our model to predict performance even before the recognition has taken place.
A practical limitation of such a data driven generative model is the limited nature of training data set.
To address this limitation, we have developed a Bayesian approach to model the nature of FNMR and FMR distribution based on the number of match and non-match scores in small regions of the quality space.
Random samples drawn from the models provide the initial data essential for training the generative model $P(\mathbf{q},\mathbf{r})$.

We evaluated the accuracy of performance predictions based on the proposed model using six face recognition systems operating on three independent data sets.
The evidence from this study suggests that the proposed performance prediction model can accurately predict face recognition performance using an accurate and unbiased Image Quality Assessor (IQA).
An unbiased IQA is essential to capture all the complex behaviours of face recognition systems.
For instance, our results show that the performance of some face recognition systems on right view is better than the recognition performance on left view.
Such a complex and unexpected behaviour can only be captured by an IQA that maps left and right profile views to different regions of the quality space.

We also investigated the reason behind high performance prediction error when the performance prediction model is applied to other independent data.
We found variability in the \textit{unaccounted quality space} -- the image quality features not considered by the IQA -- as the major factor causing inaccuracies in predicted performance.
Even controlled data sets have large amount of variability in the unaccounted quality space.
Furthermore, face recognition systems differ in their tolerance towards such variability.
Therefore, in general, to make accurate predictions on a large range of test data set, we should either consider all the relevant image quality features in order to minimize the variability in unaccounted quality space or use a classifier that is agnostic to variability in the unaccounted quality space.

 \begin{enumerate}
 \item \textit{\ResearchQuestionOne{}}

Performance prediction models are commonly based on the two types of performance predictor features: 
\begin{inparaenum}[\itshape a\upshape)]
\item image quality features, and
\item features derived solely from similarity scores.
\end{inparaenum}
In Chapter~\ref{dutta2014phdthesis_pred-feat-intro}, we investigated the merit of these two types of performance predictor feature.

A considerable amount of literature on performance prediction are based on features derived solely from similarity scores.
The evidence from experiments presented in Section~\ref{dutta2013facial_intro} show that the non-match score distribution is influenced by both identity and image quality.
Therefore, it is difficult to discern if a low similarity score is due to non-match identity or poor image quality.
This ambiguity causes instability in performance predictor features derived solely from similarity scores.
The evidence of this instability is seen in our experiments with the Impostor-based Uniqueness Measure (IUM) -- a performance predictor feature derived from non-match scores -- subject to image quality variations.
We therefore do not use features derived from similarity score in the performance prediction model proposed in Chapter~\ref{dutta2015predicting_intro}.

Facial image quality measures like pose, illumination, noise, resolution, focus, \etc have a proven record of being a reliable predictor of face recognition performance~\cite{phillips2013existence,beveridge2008focus}.
Of all the available image quality features, we focus our attention on pose and illumination -- two popular and simple image quality features.
This choice of image quality feature is motivated by the availability of publicly available large data sets~\cite{gross2008multipie,gao2008caspeal}~with controlled variations of pose and illumination.
Such a data set is essential for training a generative model.
Therefore, we select pose and illumination as two image quality features for performance prediction model of~\chaptername~\ref{dutta2015predicting_intro}.
According to the classification scheme for facial image quality variations proposed in \cite{iso_iec_29794-5:2010}, head pose and illumination correspond to subject characteristics and acquisition process characteristics respectively.
Furthermore, both quality parameters correspond to dynamic characteristics of a facial image.

In Section~\ref{dutta2014automatic_intro}, we present a novel image quality measure based on the accuracy of automatic eye detectors.
This quality measure is not practical for general biometric applications because it requires manually annotated eye location \ie ground truth for eye location.
Therefore, we do not use it in our performance prediction model proposed in Chapter~\ref{dutta2015predicting_intro}.
However, this quality measure may be useful for forensic face recognition because a forensic investigator can easily manually annotate the location of eyes in a small suspect reference set.

 \item \textit{\ResearchQuestionThree{}}

The evidence from the study presented in Chapter~\ref{dutta2015impact_intro} suggests that some face recognition systems (like ISV~\cite{wallace2011intersession}) are largely tolerant while other systems like Gabor-Jet~\cite{guenther2012disparity} and LGBPHS~\cite{zhang2005local} are sensitive towards facial image registration errors caused by error in automatic eye detection.
We also studied the accuracy of automatic eye detectors included in two commercial face recognition systems.
The accuracy of one eye detector was close to manually annotated eye while the other eye detection occassionally had large error in detected eye location.
A systematic bias in the detected eye location revealed the ambiguity present in the definition of eye center: Does the eye center refer to center of the pupil, or to center between the two eye corners or eyelids, or to something else?
Such ambiguity existed even in manual eye annotations of same sets of images carried out by two independent manual annotators.
This shows that certain amount of error is unavoidable in facial image registration based on the two eye coordinates.

These findings have important implications for modelling and predicting performance of a automatic face recognition system that :
\begin{inparaenum}[\itshape a\upshape)]
\item uses automatically detected eye coordinates for facial image registration; and
\item is sensitive to facial image registration errors.
\end{inparaenum}
For such face recognition systems, the performance prediction model presented in Chapter~\ref{dutta2015predicting_intro} must also include features that predict the accuracy of automatic eye detector being used by that system.
This is essential to fully explain the performance variability of the face recognition system.
For instance, the eye detection error is high for frontal view facial images with uniform illumination but closed eyes.
However, the facial feature extraction/comparison may not necessarily be difficult for such images because most facial features are clearly visible.
Therefore, although common image quality metrics like pose, illumination, noise, focus, \etc will attribute ``good'' quality to such images, the recognition performance will still remain poor due to facial image registration errors caused by automatic eye detection error.
Hence, performance prediction models should include features that not only predict the accuracy of feature extraction/comparison but also of the automatic eye detector. 

 \item \textit{\ResearchQuestionFour{}}

The preliminary experiments presented in Section~\ref{dutta2012impact_intro} of Chapter~\ref{dutta2014phdthesis_forensic-intro} showed that facial pose mismatch between probe and reference image played a key role in determining overall recognition performance and the influence of other quality parameters like resolution, noise, blur on performance.

Our findings in Section~\ref{dutta2012view_intro} of Chapter~\ref{dutta2014phdthesis_forensic-intro} suggests that a view based approach is potentially more useful than a model based approach in forensic cases involving face recognition.
Recall that a view based approach~\cite{pentland1994view} involves adapting the reference set such it matches the pose of the probe set.
In forensic face recognition cases involving surveillance view probe image, a view based approach requires a reference set containing images of suspects captured from the same pose as present in the surveillance view probe image.
On the other hand, a model based approach involves reconstruction of a 3D face model from the surveillance view probe image followed by synthesis of a corresponding frontal view probe image while the suspect reference set remains fixed to frontal view.

We recommend using the view based appproach if
\begin{inparaenum}[\itshape a\upshape)]
\item it is possible to exactly match the pose, illumination condition and camera of the suspect reference set to that of the probe image (or, forensic trace acquired from CCTV footage); and
\item one uses a face recognition system that is capable of comparing non-frontal view facial images with high accuracy.
\end{inparaenum}
A view based approach in forensic cases may not always be practical because matching pose and camera requires cooperative suspects and access to the same camera that captured the probe image.

The model based approach should only be considered if
\begin{inparaenum}[\itshape a\upshape)]
\item it is possible to synthesize good quality frontal view image with the original texture; and
\item one uses a face recognition system that can handle the texture artifacts in the synthesized frontal view image.
\end{inparaenum}
In forensic cases, the probe image is often of very low quality and therefore it is difficult to synthesize corresponding good quality frontal view images.

The results from Section~\ref{dutta2014automatic_intro} and Chapter~\ref{dutta2015impact_intro} show that accuracy of automatic eye detectors play a crucial role in determining the recognition performance of a face recognition system that is sensitive to facial image registration errors.
Therefore, as far as practicable, forensic investigators should manually annotate eye locations in probe and suspect reference set in order to ensure optimal recognition performance.
Furthermore, the automatic eye detection error based image quality measure, introduced in Section~\ref{dutta2014automatic_intro}, can be used along with other relevant quality mesures to quantify the uncertainty in decision about identity in such forensic cases where manual eye annotations are available.

 \end{enumerate}
\end{enumerate}

\newpage
\section*{Future Work}
This research presented in this dissertation has pointed out future work in many directions.
\begin{itemize}
\item Clearly, the most significant effort needs to be concentrated in the direction of discovering novel features that can summarize a large number of image quality variations.
This is essential for limiting the amount of variations present in the unaccounted quality space.
Results from Chapter~\ref{dutta2015impact_intro} suggests that we also need to include features that predict the performance of automatic eye detectors in order to accurately predict the performance of face recognition systems using facial image registration based on automatically detected eye locations.

\item The experiment results presented in Chapter~\ref{dutta2015predicting_intro} suggests a clear need to develop accurate and unbiased Image Quality Assessment systems.
Although our model can accept image quality parameter measurements from off-the-shelf and uncalibrated quality assessment systems, more transparent and standardized quality metrics are needed to facilitate standardized exchange of image quality information as proposed in~\cite{iso_iec_29794-1:2009}.

\item Future work could also investigate methods to directly incorporate the probabilistic models of quality and recognition performance into the EM based training procedure presented in Chapter~\ref{dutta2015predicting_intro}.
It would also be interesting to apply the proposed model to predict the performance of other biometric systems and other classifiers in general.

\item Our work in Section~\ref{dutta2013facial_intro} only examined the characteristics of one feature (\ie Impostor-based Uniqueness Measure) derived solely from similarity scores.
The results from this study showed that this feature is unstable in presence of image quality variations.
More work is needed to investigate the characteristics of other performance predictor features derived solely from similarity scores.

\item We explored the possibility of using a view based approach in forensic cases involving face recognition in Chapter~\ref{dutta2014phdthesis_forensic-intro}.
This investigation was based on surveillance view images taken from a controlled facial image data set~\cite{gross2008multipie}.
A further study could assess the feasibility of a view based approach on real forensic cases.

\end{itemize}


\appendix


\addcontentsline{toc}{chapter}{Bibliography}

\bibliographystyle{plain}  
\bibliography{dutta_phd_references}

\begin{thebibliography}{10}

\bibitem{iso_iec_29794-5:2010}
ISO/IEC JTC~1/SC 37.
\newblock Information technology — biometric sample quality — part 5: Face
  image data.
\newblock Technical Report ISO/IEC TR 29794-5:2010, International Organization
  for Standardization (ISO), 2010-04-01.

\bibitem{iso_iec_29794-1:2009}
ISO/IEC JTC~1/SC 37.
\newblock Information technology -- biometric sample quality -- part 1:
  Framework.
\newblock Technical Report ISO/IEC 29794-1:2009, International Organization for
  Standardization (ISO), 2012-09-11.

\bibitem{aggarwal2011predicting}
G.~Aggarwal, S.~Biswas, P.~J. Flynn, and K.~W. Bowyer.
\newblock {Predicting performance of face recognition systems: An image
  characterization approach}.
\newblock In {\em {Computer Vision and Pattern Recognition Workshops (CVPRW),
  2011 IEEE Computer Society Conference on}}, pages 52--59, {2011}.

\bibitem{aggarwal2012predicting}
G.~Aggarwal, S.~Biswas, P.~J. Flynn, and K.~W. Bowyer.
\newblock {Predicting good, bad and ugly match Pairs}.
\newblock In {\em {Applications of Computer Vision (WACV), 2012 IEEE Workshop
  on}}, pages 153--160, {2012}.

\bibitem{ahonen2006face}
T.~Ahonen, A.~Hadid, and M.~Pietikainen.
\newblock {Face Description with Local Binary Patterns: Application to Face
  Recognition}.
\newblock {\em {Pattern Analysis and Machine Intelligence, IEEE Transactions
  on}}, {28}({12}):2037--2041, {2006}.

\bibitem{anjos2012bob}
A.~Anjos, L.~El Shafey, R.~Wallace, M.~G\"unther, C.~McCool, and S.~Marcel.
\newblock Bob: a free signal processing and machine learning toolbox for
  researchers.
\newblock In {\em 20th ACM Conference on Multimedia Systems (ACMMM), Nara,
  Japan}, pages 1449--1452. ACM Press, October 2012.

\bibitem{asthana2011fully}
A.~Asthana, T.~K. Marks, M.~J. Jones, K.~H. Tieu, and M.~Rohith.
\newblock {Fully automatic pose-invariant face recognition via 3D pose
  normalization}.
\newblock In {\em {Computer Vision (ICCV), 2011 IEEE International Conference
  on}}, pages 937--944, {2011}.

\bibitem{belhumer1997eigenfaces}
P.N. Belhumeur, J.P. Hespanha, and D.~Kriegman.
\newblock Eigenfaces vs. {F}isherfaces: recognition using class specific linear
  projection.
\newblock {\em IEEE Transactions on Pattern Analysis and Machine Intelligence
  (TPAMI)}, 19(7):711--720, 1997.

\bibitem{beveridge2008focus}
J.~R. Beveridge, G.~H. Givens, P.~J. Phillips, B.~A. Draper, and Yui~Man Lui.
\newblock {Focus on quality, predicting FRVT 2006 performance}.
\newblock In {\em {Automatic Face Gesture Recognition, 2008. FG ’08. 8th IEEE
  International Conference on}}, pages 1--8, {2008}.

\bibitem{beveridge2011when}
J.~R. Beveridge, P.~J. Phillips, G.~H. Givens, B.~A. Draper, M.~N. Teli, and
  D.~S. Bolme.
\newblock {When high-quality face images match poorly}.
\newblock In {\em {Automatic Face Gesture Recognition and Workshops (FG 2011),
  2011 IEEE International Conference on}}, pages 572--578, {2011}.

\bibitem{beveridge2010quantifying}
J.R. Beveridge, D.S. Bolme, B.A. Draper, G.H. Givens, Yui~Man Lui, and P.J.
  Phillips.
\newblock Quantifying how lighting and focus affect face recognition
  performance.
\newblock In {\em Computer Vision and Pattern Recognition Workshops (CVPRW),
  2010 IEEE Computer Society Conference on}, pages 74--81, June 2010.

\bibitem{bolme2012csu}
Ross Beveridge.
\newblock {CSU Baseline Algorithms - Jan. 2012 Releases}.
\newblock
  \url{http://www.cs.colostate.edu/facerec/algorithms/baselines2011.php}.

\bibitem{bishop2006pattern}
Christopher~M Bishop.
\newblock {\em Pattern recognition and machine learning}, volume~1.
\newblock Springer New York, 2006.

\bibitem{blanz2005face}
V.~Blanz, P.~Grother, P.~J. Phillips, and T.~Vetter.
\newblock {Face recognition based on frontal views generated from non-frontal
  images}.
\newblock In {\em {Computer Vision and Pattern Recognition, 2005. CVPR 2005.
  IEEE Computer Society Conference on}}, pages 454--461, {2005}.

\bibitem{blanz2003face}
V.~Blanz and T.~Vetter.
\newblock {Face recognition based on fitting a 3D morphable model}.
\newblock {\em {Pattern Analysis and Machine Intelligence, IEEE Transactions
  on}}, {25}({9}):1063--1074, {2003}.

\bibitem{burton1999face}
A.~Mike Burton, Stephen Wilson, Michelle Cowan, and Vicki Bruce.
\newblock Face recognition in poor-quality video: Evidence from security
  surveillance.
\newblock {\em Psychological Science}, 10(3):243--248, 1999.

\bibitem{facevacs2010}
{Cognitec Systems}.
\newblock {FaceVACS C++ SDK Version 8.7.0}, {2012}.

\bibitem{doddington1998sheep}
George Doddington, Walter Liggett, Alvin Martin, Mark Przybocki, and Douglas
  Reynolds.
\newblock Sheep, goats, lambs and wolves: A statistical analysis of speaker
  performance in the nist 1998 speaker recognition evaluation.
\newblock In {\em Proceedings of International Conference on Spoken Language
  Processing}, 1998.

\bibitem{dutta2012impact}
A.~Dutta, R.~N.~J. Veldhuis, and L.~J. Spreeuwers.
\newblock {The Impact of Image Quality on the Performance of Face Recognition}.
\newblock In {\em {33rd WIC Symposium on Information Theory in the Benelux,
  Boekelo, The Netherlands}}, pages 141--148, {Enschede, the Netherlands},
  {May} {2012}. {Centre for Telematics and Information Technology, University
  of Twente}.

\bibitem{dutta2012view}
A.~Dutta, R.~N.~J. Veldhuis, and L.~J. Spreeuwers.
\newblock View based approach to forensic face recognition.
\newblock Technical Report TR-CTIT-12-21, Centre for Telematics and Information
  Technology, University of Twente, Enschede, September 2012.

\bibitem{dutta2013facial}
A.~Dutta, R.~N.~J. Veldhuis, and L.~J. Spreeuwers.
\newblock Can facial uniqueness be inferred from impostor scores?
\newblock In {\em Biometric Technologies in Forensic Science, BTFS 2013,
  Nijmegen, Netherlands}, Nijmegen, October 2013. Centre for Language and
  Speech Technology, Radboud University.

\bibitem{dutta2014automatic}
A.~Dutta, R.~N.~J. Veldhuis, and L.~J. Spreeuwers.
\newblock Automatic eye detection error as a predictor of face recognition
  performance.
\newblock In {\em 35rd WIC Symposium on Information Theory in the Benelux,
  Eindhoven, Netherlands}, pages 89--96. Centre for Telematics and Information
  Technology, University of Twente, May 2014.

\bibitem{dutta2014bayesian}
A.~Dutta, R.~N.~J. Veldhuis, and L.~J. Spreeuwers.
\newblock A bayesian model for predicting face recognition performance using
  image quality.
\newblock In {\em IEEE International Joint Conference on Biometrics (IJCB)},
  pages 1--8, Sept 2014.

\bibitem{dutta2015predicting}
A.~Dutta, R.~N.~J. Veldhuis, and L.~J. Spreeuwers.
\newblock {Predicting Face Recognition Performance Using Image Quality}.
\newblock {\em {IEEE Transactions on Pattern Analysis and Machine
  Intelligence}}, {submitted}.

\bibitem{dutta2015impact}
Abhishek Dutta, Manuel G{\"u}nther, Laurent El~Shafey, S{\'e}bastien Marcel,
  Raymond Veldhuis, and Luuk Spreeuwers.
\newblock Impact of eye detection error on face recognition performance.
\newblock {\em IET Biometrics}, January 2015.

\bibitem{ekenel2009face}
H.K. Ekenel and R.~Stiefelhagen.
\newblock Face alignment by minimizing the closest classification distance.
\newblock In {\em Biometrics: Theory, Applications, and Systems, 2009. BTAS
  '09. IEEE 3rd International Conference on}, pages 1--6, Sept 2009.

\bibitem{shafey2013scalable}
L.~El~Shafey, C.~McCool, R.~Wallace, and S.~Marcel.
\newblock A scalable formulation of probabilistic linear discriminant analysis:
  Applied to face recognition.
\newblock {\em IEEE Transactions on Pattern Analysis and Machine Intelligence
  (TPAMI)}, 35(7):1788--1794, July 2013.

\bibitem{fraley2012mclust}
Chris Fraley, Adrian~E. Raftery, T.~Brendan Murphy, and Luca Scrucca.
\newblock mclust version 4 for r: Normal mixture modeling for model-based
  clustering, classification, and density estimation.
\newblock Technical Report 597, Department of Statistics, University of
  Washington, 2012.

\bibitem{gao2008caspeal}
Wen Gao, Bo~Cao, Shiguang Shan, Xilin Chen, Delong Zhou, Xiaohua Zhang, and
  Debin Zhao.
\newblock The cas-peal large-scale chinese face database and baseline
  evaluations.
\newblock {\em Systems, Man and Cybernetics, Part A: Systems and Humans, IEEE
  Transactions on}, 38(1):149--161, Jan 2008.

\bibitem{gao2008cas}
Wen Gao, Bo~Cao, Shiguang Shan, Xilin Chen, Delong Zhou, Xiaohua Zhang, and
  Debin Zhao.
\newblock The cas-peal large-scale chinese face database and baseline
  evaluations.
\newblock {\em Systems, Man and Cybernetics, Part A: Systems and Humans, IEEE
  Transactions on}, 38(1):149--161, 2008.

\bibitem{going1974effects}
Merideth Going and JD~Read.
\newblock Effects of uniqueness, sex of subject, and sex of photograph on
  facial recognition.
\newblock {\em Perceptual and Motor Skills}, 39(1):109--110, 1974.

\bibitem{grgic2011scface}
Mislav Grgic, Kresimir Delac, and Sonja Grgic.
\newblock {SCface – surveillance cameras face database}.
\newblock {\em {Multimedia Tools and Applications}}, {51}({3}):863--879,
  {2011}.

\bibitem{gross2008multipie}
Ralph Gross, Iain Matthews, Jeffrey Cohn, Takeo Kanade, and Simon Baker.
\newblock {Multi-PIE}.
\newblock In {\em {Automatic Face Gesture Recognition, 2008. FG ’08. 8th IEEE
  International Conference on}}, pages 1--8, {2008}.

\bibitem{grother2007performance}
P.~Grother and E.~Tabassi.
\newblock {Performance of Biometric Quality Measures}.
\newblock {\em {Pattern Analysis and Machine Intelligence, IEEE Transactions
  on}}, {29}({4}):531--543, {2007}.

\bibitem{guenther2012disparity}
M.~G{\"u}nther, D.~Haufe, and R.P. W{\"u}rtz.
\newblock Face recognition with disparity corrected {G}abor phase differences.
\newblock In A.E.P. Villa, W.~Duch, P.~{\'{E}}rdi, F.~Masulli, and G.~Palm,
  editors, {\em Artificial Neural Networks and Machine Learning}, volume 7552
  of {\em Lecture Notes in Computer Science}, pages 411--418. Springer Berlin,
  September 2012.

\bibitem{gunther2012open}
Manuel Gunther, Roy Wallace, and Sebastien Marcel.
\newblock {An Open Source Framework for Standardized Comparisons of Face
  Recognition Algorithms}.
\newblock {\em {IEEE International Workshop on Benchmarking Facial Image
  Analysis Technologies}}, {2012}.

\bibitem{mantasari2014calibration}
M.~I.~Mantasari, M.~G{\"{u}}nther, R.~Wallace, R.~Saedi, S.~Marcel, and
  D.~Van~Leeuwen.
\newblock Score calibration in face recognition.
\newblock {\em IET Biometrics}, 2014.

\bibitem{jain2007handbook}
Anil~K Jain, Patrick Flynn, and Arun~A Ross.
\newblock {\em Handbook of biometrics}.
\newblock Springer, 2007.

\bibitem{jammalamadaka2012algorithm}
Nataraj Jammalamadaka, Andrew Zisserman, Marcin Eichner, Vittorio Ferrari, and
  C.~V. Jawahar.
\newblock {Has My Algorithm Succeeded?: An Evaluator for Human Pose
  Estimators}.
\newblock In Andrew Fitzgibbon, Svetlana Lazebnik, Pietro Perona, Yoichi Sato,
  and Cordelia Schmid, editors, {\em {Computer Vision – ECCV 2012}}, pages
  114--128. {Springer Berlin Heidelberg}, {2012}.

\bibitem{jesorsky2001robust}
Oliver Jesorsky, Klaus~J Kirchberg, and Robert~W Frischholz.
\newblock Robust face detection using the hausdorff distance.
\newblock In {\em Audio-and video-based biometric person authentication}, pages
  90--95. Springer, 2001.

\bibitem{kazemi2014one}
Vahid Kazemi and Josephine Sullivan.
\newblock One millisecond face alignment with an ensemble of regression trees.
\newblock In {\em Computer Vision and Pattern Recognition (CVPR), 2014 IEEE
  Conference on}, pages 1867--1874. IEEE, 2014.

\bibitem{klare2012face}
Brendan~F. Klare and Anil~K. Jain.
\newblock {Face recognition: Impostor-based measures of uniqueness and
  quality}.
\newblock In {\em {Biometrics: Theory, Applications and Systems (BTAS), 2012
  IEEE Fifth International Conference on}}, pages 237--244, {2012}.

\bibitem{klontz2013case}
Joshua~C Klontz and Anil~K Jain.
\newblock A case study on unconstrained facial recognition using the boston
  marathon bombings suspects.
\newblock {\em Michigan State University, Tech. Rep}, 119:120, 2013.

\bibitem{li2005predicting}
Weiliang Li, Xiang Gao, and T.~E. Boult.
\newblock {Predicting biometric system failure}.
\newblock In {\em {Computational Intelligence for Homeland Security and
  Personal Safety, 2005. CIHSPS 2005. Proceedings of the 2005 IEEE
  International Conference on}}, pages 57--64, {2005}.

\bibitem{marques2000effects}
Joe Marques, Nicholas~M Orlans, and Alan~T Piszcz.
\newblock Effects of eye position on eigenface-based face recognition scoring.
\newblock {\em Image}, 8, 2000.

\bibitem{martinex2001pca}
A.M. Martinez and A.C. Kak.
\newblock {PCA} versus {LDA}.
\newblock {\em IEEE Transactions on Pattern Analysis and Machine Intelligence
  (TPAMI)}, 23(2):228--233, 2001.

\bibitem{min2005eye}
Jaesik Min, KevinW. Bowyer, and PatrickJ. Flynn.
\newblock Eye perturbation approach for robust recognition of inaccurately
  aligned faces.
\newblock In Takeo Kanade, Anil Jain, and NaliniK. Ratha, editors, {\em Audio-
  and Video-Based Biometric Person Authentication}, volume 3546 of {\em Lecture
  Notes in Computer Science}, pages 41--50. Springer Berlin Heidelberg, 2005.

\bibitem{verilook2011}
Neurotechnology.
\newblock {VeriLook C++ SDK Version 5.1}, {2011}.

\bibitem{nfiq2}
{NIST}.
\newblock {Development of NFIQ 2.0}, {2014}.

\bibitem{ozay2009improving}
N.~Ozay, Yan Tong, F.~W. Wheeler, and Xiaoming Liu.
\newblock {Improving face recognition with a quality-based probabilistic
  framework}.
\newblock In {\em {Computer Vision and Pattern Recognition Workshops, 2009.
  CVPR Workshops 2009. IEEE Computer Society Conference on}}, pages 134--141,
  {2009}.

\bibitem{paone2011difficult}
Jeffrey Paone, Soma Biswas, Gaurav Aggarwal, and Patrick Flynn.
\newblock Difficult imaging covariates or difficult subjects? - an empirical
  investigation.
\newblock In {\em Biometrics (IJCB), 2011 International Joint Conference on},
  pages 1--8, 2011.

\bibitem{park20073dmodel}
Unsang Park and Anil Jain.
\newblock {3D Model-Based Face Recognition in Video}.
\newblock In Seong-Whan Lee and Stan Li, editors, {\em {Advances in
  Biometrics}}, volume {4642}, chapter {Lecture Notes in Computer Science},
  pages 1085--1094. {Springer Berlin / Heidelberg}, {2007}.

\bibitem{paysan20093dface}
Pascal Paysan, Reinhard Knothe, Brian Amberg, Sami Romdhani, and Thomas Vetter.
\newblock {A 3D Face Model for Pose and Illumination Invariant Face
  Recognition}.
\newblock In {\em {Sixth IEEE International Conference on Advanced Video and
  Signal Based Surveillance, AVSS.}}, pages 296--301, {2009}.

\bibitem{pentland1994view}
A.~Pentland, B.~Moghaddam, and T.~Starner.
\newblock {View-based and modular eigenspaces for face recognition}.
\newblock In {\em {Computer Vision and Pattern Recognition, 1994. Proceedings
  CVPR ’94., 1994 IEEE Computer Society Conference on}}, pages 84--91, {jun}
  {1994}.

\bibitem{petersen2008matrix}
Kaare~Brandt Petersen and Michael~Syskind Pedersen.
\newblock The matrix cookbook.
\newblock {\em Technical University of Denmark}, pages 7--15, 2008.

\bibitem{phillips2009introduction}
P.~J. Phillips and J.~R. Beveridge.
\newblock {An introduction to biometric-completeness: The equivalence of
  matching and quality}.
\newblock In {\em {Biometrics: Theory, Applications, and Systems, 2009. BTAS
  ’09. IEEE 3rd International Conference on}}, pages 1--5, {2009}.

\bibitem{phillips2011introduction}
P.~J. Phillips, J.~R. Beveridge, B.~A. Draper, G.~Givens, A.~J. O’Toole,
  D.~S. Bolme, J.~Dunlop, Yui~Man Lui, H.~Sahibzada, and S.~Weimer.
\newblock {An introduction to the good, the bad, amp; the ugly face recognition
  challenge problem}.
\newblock In {\em {Automatic Face Gesture Recognition and Workshops (FG 2011),
  2011 IEEE International Conference on}}, pages 346--353.
  {\url{http://www.cs.colostate.edu/facerec/algorithms/baselines2011.php}},
  {2011}.

\bibitem{phillips2005overview}
P~Jonathon Phillips, Patrick~J Flynn, Todd Scruggs, Kevin~W Bowyer, Jin Chang,
  Kevin Hoffman, Joe Marques, Jaesik Min, and William Worek.
\newblock Overview of the face recognition grand challenge.
\newblock In {\em Computer vision and pattern recognition, 2005. CVPR 2005.
  IEEE computer society conference on}, volume~1, pages 947--954. IEEE, 2005.

\bibitem{phillips2011otherrace}
P.~Jonathon Phillips, Fang Jiang, Abhijit Narvekar, Julianne Ayyad, and
  Alice~J. O'Toole.
\newblock An other-race effect for face recognition algorithms.
\newblock {\em ACM Trans. Appl. Percept.}, 8(2):14:1--14:11, February 2011.

\bibitem{phillips2000feret}
P.~Jonathon Phillips, Hyeonjoon Moon, Syed~A. Rizvi, and Patrick~J. Rauss.
\newblock {The FERET evaluation methodology for face-recognition algorithms}.
\newblock {\em {Pattern Analysis and Machine Intelligence, IEEE Transactions
  on}}, {22}({10}):1090--1104, {oct} {2000}.

\bibitem{phillips2013existence}
P.J. Phillips, J.R. Beveridge, D.S. Bolme, B.A. Draper, G.H. Givens, Yui~Man
  Lui, Su~Cheng, M.N. Teli, and Hao Zhang.
\newblock On the existence of face quality measures.
\newblock In {\em Biometrics: Theory, Applications and Systems (BTAS), 2013
  IEEE Sixth International Conference on}, pages 1--8, Sept 2013.

\bibitem{riopka2003eyes}
Terry Riopka and Terrance Boult.
\newblock The eyes have it.
\newblock In {\em Proceedings of the 2003 ACM SIGMM workshop on Biometrics
  methods and applications}, WBMA '03, pages 9--16, New York, NY, USA, 2003.
  ACM.

\bibitem{rodriguez2006measuring}
Yann Rodriguez, Fabien Cardinaux, Samy Bengio, and Johnny Mari{\'{e}}thoz.
\newblock Measuring the performance of face localization systems.
\newblock {\em Image and Vision Computing}, 24(8):882 -- 893, 2006.

\bibitem{ross2009exploiting}
Arun Ross, Ajita Rattani, and Massimo Tistarelli.
\newblock Exploiting the “doddington zoo” effect in biometric fusion.
\newblock In {\em Biometrics: Theory, Applications, and Systems, 2009. BTAS'09.
  IEEE 3rd International Conference on}, pages 1--7. IEEE, 2009.

\bibitem{scheirer2011meta}
W.~J. Scheirer, A.~Rocha, R.~J. Micheals, and T.~E. Boult.
\newblock {Meta-Recognition: The Theory and Practice of Recognition Score
  Analysis}.
\newblock {\em {Pattern Analysis and Machine Intelligence, IEEE Transactions
  on}}, {33}({8}):1689--1695, {2011}.

\bibitem{shan2004curse}
Shiguang Shan, Yizheng Chang, Wen Gao, Bo~Cao, and Peng Yang.
\newblock Curse of mis-alignment in face recognition: problem and a novel
  mis-alignment learning solution.
\newblock In {\em IEEE International Conference on Automatic Face and Gesture
  Recognition (FG)}, pages 314--320. IEEE, 2004.

\bibitem{shi2008modeling}
Zhixin Shi, Frederick Kiefer, John Schneider, and Venu Govindaraju.
\newblock Modeling biometric systems using the general pareto distribution
  (gpd).
\newblock In {\em Proc. SPIE}, volume 6944, pages 69440O--69440O--11, 2008.

\bibitem{tabassi2005novel}
E.~Tabassi and C.~L. Wilson.
\newblock {A novel approach to fingerprint image quality}.
\newblock In {\em {IEEE International Conference on Image Processing, 2005.
  ICIP 2005.}}, {2005}.

\bibitem{tabassi2004fingerprint}
Elham Tabassi, Charles Wilson, and Craig~I. Watson.
\newblock {Fingerprint Image Qualitiy}.
\newblock {NIST Interagency/Internal Report (NISTIR) - 7151}, {National
  Institute of Standards and Technology (NIST)}, {April} {2004}.

\bibitem{turk1991eigenfaces}
M.~Turk and A.~Pentland.
\newblock Eigenfaces for recognition.
\newblock {\em Journal of Cognitive Neuroscience}, 3(1):71--86, 1991.

\bibitem{wagner2012towards}
A.~Wagner, J.~Wright, A.~Ganesh, Zihan Zhou, H.~Mobahi, and Yi~Ma.
\newblock Towards a practical face recognition system: Robust alignment and
  illumination by sparse representation.
\newblock {\em Pattern Analysis and Machine Intelligence, IEEE Transactions
  on}, 34(2):372--386, Feb 2012.

\bibitem{wallace2011intersession}
R.~Wallace, M.~McLaren, C.~McCool, and S.~Marcel.
\newblock Inter-session variability modelling and joint factor analysis for
  face authentication.
\newblock In {\em International Joint Conference on Biometrics (IJCB)}, pages
  1--8, 2011.

\bibitem{wallace2012ztnorm}
R.~Wallace, M.~McLaren, C.~McCool, and S.~Marcel.
\newblock Cross-pollination of normalisation techniques from speaker to face
  authentication using {G}aussian mixture models.
\newblock {\em IEEE Transactions on Information Forensics and Security},
  7(2):553--562, 2012.

\bibitem{wang2005sensitivity}
Haoshu Wang and Patrick~J. Flynn.
\newblock Sensitivity of face recognition performance to eye location accuracy.
\newblock {\em Biometric Technology for Human Identification II, Proc. SPIE
  5779}, pages 122--131, 2005.

\bibitem{wang2005automatic}
Peng Wang, M.B. Green, Qiang Ji, and J.~Wayman.
\newblock Automatic eye detection and its validation.
\newblock In {\em Computer Vision and Pattern Recognition - Workshops, 2005.
  CVPR Workshops. IEEE Computer Society Conference on}, pages 164--164, June
  2005.

\bibitem{wang2007modeling}
Peng Wang, Qiang Ji, and J.~L. Wayman.
\newblock {Modeling and Predicting Face Recognition System Performance Based on
  Analysis of Similarity Scores}.
\newblock {\em {Pattern Analysis and Machine Intelligence, IEEE Transactions
  on}}, {29}({4}):665--670, {2007}.

\bibitem{wein2005using}
Lawrence~M. Wein and Manas Baveja.
\newblock {Using fingerprint image quality to improve the identification
  performance of the U.S. Visitor and Immigrant Status Indicator Technology
  Program}.
\newblock {\em {Proceedings of the National Academy of Sciences of the United
  States of America}}, {102}({21}):7772--7775, {2005}.

\bibitem{wittman2006empirical}
M~Wittman, P~Davis, and PJ~Flynn.
\newblock Empirical studies of the existence of the biometric menagerie in the
  frgc 2.0 color image corpus.
\newblock In {\em Computer Vision and Pattern Recognition Workshop, 2006.
  CVPRW'06. Conference on}, pages 33--33. IEEE, 2006.

\bibitem{yager2010biometric}
Neil Yager and Ted Dunstone.
\newblock The biometric menagerie.
\newblock {\em Pattern Analysis and Machine Intelligence, IEEE Transactions
  on}, 32(2):220--230, 2010.

\bibitem{zhang2005local}
W.~Zhang, S.~Shan, W.~Gao, X.~Chen, and H.~Zhang.
\newblock Local {G}abor binary pattern histogram sequence ({LGBPHS}): a novel
  non-statistical model for face representation and recognition.
\newblock In {\em IEEE International Conference on Computer Vision (ICCV)},
  volume~1, pages 786--791 Vol. 1, 2005.

\bibitem{zhao2000sfs}
Wen~Yi Zhao and R.~Chellappa.
\newblock {SFS based view synthesis for robust face recognition}.
\newblock In {\em {Automatic Face and Gesture Recognition, 2000. Proceedings.
  Fourth IEEE International Conference on}}, pages 285--292, {2000}.

\bibitem{zuo2010adaptive}
Jinyu Zuo, F.~Nicolo, N.A. Schmid, and H.~Wechsler.
\newblock Adaptive biometric authentication using nonlinear mappings on quality
  measures and verification scores.
\newblock In {\em Image Processing (ICIP), 2010 17th IEEE International
  Conference on}, pages 4077--4080, Sept 2010.

\end{thebibliography}

\end{document}